\documentclass[review]{elsarticle}

\usepackage{hyperref}
\usepackage{siunitx}
\usepackage{xcolor}
\usepackage{amsmath}
\usepackage{amsfonts}
\usepackage{upgreek}
\usepackage{algorithm}
\usepackage[noend]{algpseudocode}
\usepackage{comment}
\usepackage{booktabs}
\usepackage{multirow}
\usepackage{siunitx}
\usepackage{float}
\usepackage{threeparttable}
\usepackage{rotating}
\usepackage[a4paper,bindingoffset=0.2in,%
            left=0.50in,right=0.75in,top=0.75in,bottom=0.75in,%
            footskip=.25in]{geometry}
\usepackage{graphicx}
\usepackage{caption}
\usepackage{subcaption}
\usepackage{tabularx}
\usepackage[nameinlink,capitalise]{cleveref}
\Crefname{figure}{Fig.}{Figs.}

\DeclareMathOperator*{\argmax}{argmax} % thin space, limits underneath in displays
\journal{TBD}

\newcommand{\calN}{\mathcal{N}}

\newcommand{\calX}{\mathcal{X}}
\newcommand{\R}{\mathbb{R}}
\newcommand{\xnew}{x_{\text{new}}}
\newcommand{\dC}[1]{{#1}^\text{o}\text{C}}
\newcommand{\GP}{\operatorname{GP}}
\newcommand{\bthetau}{\boldsymbol{\uptheta}_{u}}
\newcommand{\bthetag}{\boldsymbol{\uptheta}_{g}}
\newcommand{\bzero}{\mathbf{0}}
\newcommand{\bK}{\mathbf{K}}
\newcommand{\bKuu}{\mathbf{K}_{u,u}}

\newcommand{\C}{\mathbb{C}}

\makeatletter
\def\ps@pprintTitle{%
 \let\@oddhead\@empty
 \let\@evenhead\@empty
 \def\@oddfoot{}%
 \let\@evenfoot\@oddfoot}
 
\begin{document}

\begin{frontmatter}

\title{Learning Personalized Thermal Preferences via Bayesian Active Learning with Unimodality Constraints}

\author[label1]{Nimish Awalgaonkar}
\author[label1]{Ilias Bilionis\corref{cor1}}
\ead{ibilion@purdue.edu}
\author[label2,label3]{Xiaoqi Liu}\author[label2,label3]{Panagiota Karava}
\author[label2,label3]{Athanasios Tzempelikos}

    \address[label1]{School of Mechanical Engineering, Purdue University, 585 Purdue Mall, West Lafayette, IN, 47907, USA}

\address[label2]{Lyles School of Civil Engineering, Purdue University, 550 Stadium Mall Dr., West Lafayette, IN, 47907, USA}

\address[label3]{Center for High Performance Buildings, Ray W. Herrick Laboratories, Purdue University, 140 S. Martin Jischke Dr., West Lafayette, IN, 47907, USA}

\cortext[cor1]{corresponding author}

\begin{abstract}
Thermal preferences vary from person to person and may change over time. 
The main objective of this paper is to sequentially pose intelligent queries to occupants in order to optimally learn the indoor air temperature values which maximize their satisfaction. 
Our central hypothesis is that an occupant's preference relation over indoor air temperatures can be described using a scalar function of these temperatures, which we call the ``occupant's thermal utility function.'' 
Information about an occupant's preference over these temperatures is available to us through their response to thermal preference queries, ``prefer warmer'',``prefer cooler'' and ``satisfied '' which we interpret as statements about the derivative of their utility function, i.e. the utility function is ``increasing'',``decreasing,'' and ``constant'' respectively.
We model this hidden utility function using a Gaussian process prior with built-in unimodality constraint, i.e., the utility function has a unique maximum, and we train this model using Bayesian inference.
This permits an expected improvement based selection of next preference query to pose to the occupant, which takes into account both exploration (sampling from areas of high uncertainty) and exploitation (sampling from areas which are likely to offer an improvement over current best observation).
We use this framework to sequentially design experiments and illustrate its benefits by showing that it requires drastically fewer observations to infer the maximally preferred indoor air temperature values as compared to other methods.
This framework is an important step towards the development of intelligent HVAC systems which would be able to respond to occupants' personalized thermal comfort needs.
In order to encourage the use of our PE framework and ensure reproducibility in results, we publish an implementation of our work named GPPrefElicit\footnote{to be released upon acceptance of the paper} as an open-source package in the Python language .
\end{abstract}


\end{frontmatter}

%\linenumbers

\section{Introduction}\label{sec:intro}
Commercial buildings have significant impacts on humans and the environment. They consume more than 19\% of the total energy consumption in the US.
Heating, Ventilation and Air Conditioning (HVAC) systems account for 27\% of energy consumption and 45\% of peak electrical demand in commercial buildings and represent a substantial energy use reduction opportunity \cite{book2010us}.
The main purpose of HVAC systems in commercial buildings is to provide satisfactory thermal conditions and indoor air quality for occupants working inside these buildings. Providing satisfactory thermal conditions in indoor environments is critical since it directly affects occupants' comfort \cite{wagner2007thermal, frontczak2012quantitative}, health \cite{fisk1997estimates}, and productivity \cite{leaman1999productivity,wargocki1999perceived,wyon2004effects,tham2010room} and has a significant effect on energy performance of buildings \cite{korkas2016occupancy, yang2014thermal, jazizadeh2018personalized,liu2018model}.

To predict thermal comfort in indoor environments, researchers have developed empirical models with data collected from climate chamber experiments and field studies, using large populations \cite{wang2018individual}. 
These models have succeeded in predicting average comfort-related responses; consequently, they were adopted in standards for operation of heating, ventilation, and air-conditioning (HVAC) systems in office buildings \cite{en200715251}.
However, numerous field studies have shown that individual occupants prefer different thermal conditions and as such general thermal comfort models  cannot accurately predict individual thermal preferences \cite{ghahramani2015online, jazizadeh2013human, feldmeier2010personalized, lee2017bayesian}.
These approaches fail to capture that thermal preferences vary from person to person and may change over time \cite{ghahramani2015online}. 
As a result, typical HVAC systems operation cannot achieve high levels of satisfaction for all occupants.
Moreover, because of the conservative control settings based on general comfort standards, HVAC energy use in office buildings remains significant despite the higher equipment efficiency and the use of improved controls. 
Also, to implement general thermal comfort model, such as the most-widely used PMV model \cite{fanger1970thermal}, one needs very specific input features (e.g., metabolic rate, clothing insulation, air speed) that are difficult to obtain in a real-world setting and, therefore, they are often assumed or simplified \cite{erickson2012thermovote}.
Finally, the parameters of the PMV model (e.g., functions, coefficients) are trained using the original data set (i.e., Fanger's laboratory (chamber based) experimental data), and need to be updated to reflect the actual personalized thermal preferences of occupants in other settings.

Researchers have acknowledged these limitations and suggested solutions that incorporate building occupants in sensing and control frameworks and tune these systems based on personalized preferences to achieve optimal conditioning for improved occupant satisfaction and energy efficiency \cite{gao2013optimal, gao2013spot, jazizadeh2014user, liu2007neural, daum2011personalized, ghahramani2016infrared}. 
Results of initial studies have shown that tuning HVAC systems' operation based on occupants' feedback has the potential to increase their satisfaction and reduce energy consumption. For example, Ghahramani et al. \cite{ghahramani2016energy} showed that selecting the daily optimal setpoint temperature in the range of $22.5 \pm$\SI{1}{\celsius} , $22.5 \pm$\SI{2}{\celsius} and $22.5 \pm$\SI{3}{\celsius} resulted in average savings of 7.5\%, 12.7\% and 16.4\%, respectively, compared to the baseline setpoint temperature of \SI{22.5}{\celsius}.
Murakami et al. \cite{murakami2007field} developed an automatic control logic which maintained a balance between occupants’ satisfaction and energy consumption and resulted in 20\% energy savings without increasing the percentage of occupant dissatisfaction. 

\subsection{Current approaches for learning occupants' personalized thermal preferences}\label{subsec:personalized_thermal_preferences}
Thermal preferences are seen to vary from person to person and may change over time due to seasonal variations or acclimation \cite{ghahramani2015online}. Furthermore, what suits one group or a type of occupant may be uncomfortable for others \cite{wang2018individual, nakano2002differences}.
A wide array of field studies have shown individual differences in thermal comfort due to different age, weight, gender, historical thermal experiences etc. \cite{karjalainen2012thermal, indraganti2010effect, indraganti2015thermal}.
For example, Fountain et al. \cite{fountain1996expectations} argued that differences in individual occupant's response to comfort queries (on a 7-point thermal sensation scale) were frequently greater than 1 scale unit.
In another study, Humphreys and Nicol \cite{humphreys2002validity} analyzed ASHRAE database of field-studies and found that thermal comfort responses given by individual occupants have a standard deviation of around 1 scale unit (on the 7-point thermal sensation scale). 
One-unit difference in ASHRAE's 7-point scale corresponds to approximately $\dC{3}
$ difference in indoor air temperature values, further indicating that the individual 
differences between preferred indoor air temperature for occupants may be as large as $\dC{3}$.Due to these individual differences in thermal comfort, it is often very difficult to provide a thermal environment which is satisfactory for all of the occupants.
Personalized thermal comfort modeling approaches aim to address this challenge by building predictive models for inferring individual occupant's thermal satisfaction profile, instead of the average profile of a large population \cite{kim2018personal}.

The opportunities arising from development of personal comfort models has generated a lot of interest within the academic and industry communities.
%Researchers from industries are leveraging various cloud-based platforms and advanced data modeling schemes to deliver personalized heating and cooling in the office spaces (e.g. Google's Nest thermostat).
With the rapid development in sensing technology and increase in computational power, researchers are exploring new approaches to better predict individual occupant's thermal comfort in office buildings \cite{kim2018pp}.
Many studies have focussed on development of personalized models which estimate the model parameters to describe individual occupant's thermal comfort status (expressed as preference, satisfaction, or sensation) based on data collected from the field.
These models inferred personalized thermal comfort of occupants by establishing correlations between environmental measurements and occupants’ response to comfort queries (obtained via surveys).
They employed different forms of surrogate models such as neural networks \cite{liu2007neural}, support vector machines \cite{rana2013feasibility, jiang2016modelling}, Gaussian processes \cite{cheung2017longitudinal, guenther2018feature}, fuzzy rules \cite{jazizadeh2014user}, and Bayesian networks \cite{lee2017bayesian, lee2019inference} to approximate individual occupant’s thermal preferences, resulting in improved data representation and predictive accuracy.
The results obtained using these models show significant improvement in performance (20-30\% gain in terms of predictive accuracy) compared to conventional comfort models (PMV) \cite{kim2018pp}.  
For example, Ghahramani et al. \cite{ghahramani2015online} introduced an online Bayesian network-based learning algorithm for modeling personalized thermal preferences.
They tested their framework on 33 different subjects by collecting their responses to comfort queries over a course of approximately 3 months. The results showed an average accuracy of 70.14\% (as compared to 56.1\%   accuracy obtained by PMV model on the same dataset).
In \cite{daum2011personalized}, Daum et al. proposed a learning method based on multinomial logistic regression and showed that personalized thermal satisfaction profiles were needed in order to predict occupant’s thermal sensation properly.
In this study, authors collected data from 28 different occupants during a period of three years.
They found that the individual satisfaction profiles evolved as new information (data) was collected and used, and that congruency of the proposed model increased from 0.66 to 0.87 after 90 votes were added.
They concluded that, for obtaining converged satisfaction profiles, more than 90 (indoor air temperature based) data points were needed.
In another study, Feldmeier and Paradiso \cite{feldmeier2010personalized} developed an occupant-centric personalized HVAC control system which inferred individual occupant’s preferences using linear discriminant analysis.
They conducted this study from May through August of 2009.

As seen from all these works, without sufficient survey data, the proposed personal comfort models would largely fail to infer individual occupant-specific thermal preference needs.
However, securing data collection through surveys is difficult due to the cost associated with conducting the experiments, potential fatigue and eventual decay in participation \cite{kim2018personal, kim2018pp}. Having occupants responding
to surveys for a long time is certainly not a practical solution that can be implemented in real buildings. 
A potential solution to address these problems is to deploy novel active learning methods, that are able to lean fast and efficiently using a small amount of data. 
Therefore, in this paper, an alternative robust online strategy/framework is proposed to actively collect the ``most informative'' data in order to describe individual occupants' thermal preferences and infer their maximally preferred indoor air temperatures.
\subsection{Research contributions}\label{subsec:research_contri}
When it comes to development of smart thermostat in offices, our overarching goal is to develop a recommender system that makes temperature set-point recommendations with the aim of maximizing some occupant's satisfaction in the light of inherent uncertainty over occupants and indoor environment. Traditionally, big companies like Amazon, Netflix, Google, Facebook etc. have had deployed recommender systems to assist users when they search in large information spaces such as collection of products (books, movies, music), documents (news articles, Wikipedia articles) etc \cite{smith2017two, gomez2016netflix, ricci2015recommender}. 
These recommender systems assist in saving a user's efforts in searching for products by interacting with users in order to reduce uncertainty in their preferences.
Preference elicitation (PE) is a crucial component of recommender systems that aims to make optimal recommendations to the users by actively asking them questions about their preferences \cite{guo2010gaussian, guo2010real, guo2011bayesian}.
An important requirement for PE framework is that they should be able to make optimal or near optimal recommendations based only on a small number of survey questions.

The objective of this paper is to present, for the first time (in the context of smart building applications), a PE framework which sequentially poses intelligent queries to a new occupant in order to optimally learn the indoor air temperature values which maximize his/her satisfaction.
Our central hypothesis is that an occupant's preference relation over indoor air temperatures can be described using a scalar function of these temperatures, which we call the ``occupant's thermal utility function.'' 
The mathematical characteristics of our PE framework are that:
(1) It maintains a flexible representation of an occupant's thermal utility function using Gaussian processes (GP);
(2) By adhering to a fully Bayesian paradigm, it handles uncertainty in a principled manner;
(3) It sequentially selects queries that yield the maximum expected improvement in utility;
and (4) It incorporates prior knowledge from different sources \cite{guo2011bayesian}, e.g., that there exists a unique maximally preferred indoor air temperature.
From a thermal comfort perspective, our framework comes under the paradigm of design of experiments for training personalized thermal preference models \cite{kim2018personal, kim2018pp}.
The key practical characteristics of our PE framework are that:
(1) It takes an individual occupant as the unit of analysis rather than population or groups of people;
(2) It uses direct feedback from occupants to train and suggests new temperature to query next;
(3) It prioritizes cost-effective and easily-obtainable data (we only use indoor air temperature as a predictor which affects occupant’s satisfaction);
(4) It has the capacity to adapt as new data is introduced into the modeling framework.
A detailed comparison of our approach with respect to existing personalized thermal comfort models is shown in the Table~\ref{table:comfort studies}. 
This PE framework is an important step towards the development of intelligent HVAC systems which would be able to respond to individual occupants’ personalized thermal comfort needs.
In order to encourage the use of our PE framework and ensure the reproducibility of our results, we publish an implementation of our work named GPPrefElicit\footnote{https://github.com/nawalgao/GPPrefElicit} in the Python language .

To summarize, this paper starts with an introduction of the PE framework used for inferring the maximally preferred indoor air temperature values in Section \ref{sec:methodology}.
In Section \ref{sec:proof_of_concept}, the performance of this framework is evaluated (by showing how this framework would work in case of synthetic occupants).
In Section \ref{sec:DOE},  the results of an experimental study (using our newly developed PE framework) for inferring maximally preferred indoor air temperatures is presented. 
We then discuss some of the limitations of this PE framework in Section \ref{sec:discussion}.
Section \ref{sec:conclusion} summarizes our findings and concludes the paper.

\begin{sidewaystable}[htbp]
\caption{Summary of major studies related to personalized comfort models \cite{kim2018pp}}
\label{table:comfort studies}
%\tiny
\centering\small\setlength\tabcolsep{2pt}
\resizebox{\textwidth}{!}{
\begin{threeparttable}
\begin{tabular}
{ccccccc}
\toprule
\multirow{2}{*}{Model} &
\multirow{2}{*}{Data} &
\multicolumn{2}{c}{Model inputs} &
\multirow{2}{*}{Model output} &
\multirow{2}{*}{Methodology} &
\multirow{2}{*}{Model evaluation} \\
\cline{3-4}
& & {Occupant's response} &
{Measurements\tnote{a}} & & & \\
\midrule
\cite{cheung2017longitudinal} & 
\begin{tabular}{c} Field data \\ from 15 subjects \end{tabular} &
\begin{tabular}{c} Continuous thermal acceptability scale \\ with 4 labels (clearly acceptable/just acceptable/ \\just unacceptable/clearly unacceptable)  \end{tabular}&
Ta, RH, CO$_{2}$ &
\begin{tabular}{c}Continuous thermal acceptability scale \\ with 4 labels (clearly acceptable/just acceptable/ \\just unacceptable/clearly unacceptable)  \end{tabular}&
\begin{tabular}{c} Gaussian \\ processes \end{tabular}&
\begin{tabular}{c} $R^2$ between predicted and actual votes = 0.18 and 0.26 for 2 subjects resp. \\ Approx. 190 datapoints were used for training.  \\ Datapoints were randomly collected \\ (all at once, no active learning framework setup). \end{tabular} \\
\hline{}
\cite{lee2017bayesian} & 
\begin{tabular}{c} Field data \\ from ASHRAE \\ RP-884 database \end{tabular} &
\begin{tabular}{c} 3-point thermal preference \\ (warmer/satisfied / cooler)\end{tabular} &
Ta, RH, MRT, Va &
\begin{tabular}{c} 3-point thermal preference \\ (warmer/satisfied / cooler) \\ Probability of occupant \\ belonging to a specific cluster.\end{tabular} &
\begin{tabular}{c} Bayesian inference, \\ Logistic regression, \\ Infinite Dirichlet processes \\ based clustering \end{tabular} &
\begin{tabular}{c} Maximized the model evidence with optimal number of clusters.
\\ Model validated using 9 synthetic occupants. \\
Once trained on ASHRAE dataset,
approx. more than \\10 number of datapoints were
further collected for\\ training and inferring 
the satisfaction profiles of new occupants.\\
Datapoints were randomly collected \\ (all at once, no active learning framework setup).
\end{tabular} \\
\hline{}
\cite{ghahramani2015online} &
\begin{tabular}{c} Field data \\ from 33 subjects \end{tabular} &
\begin{tabular}{c} 11-point thermal preference scale \\ with 3 labels (warmer/satisfied / cooler)\end{tabular} &
Ta &
\begin{tabular}{c} 3 comfort conditions \\ (uncomfortably warm/ \\ comfortable / \\ uncomfortably cool) \end{tabular} &
\begin{tabular}{c} Probabilistic graphical model, \\ online learning \end{tabular}&
\begin{tabular}{c} Mean accuracy:
proposed model = 70.1\%  ;
PMV model (on same data) = 56.1.\% \\
Minimum number of datapoints used for training the model = 19. \\
Datapoints were randomly queried (in an active learning setup). \\
Kolmogorov-Smirnov test was used to remove \\ statistically irrelevant data points as new votes are added.
\end{tabular} \\
\hline{}
\cite{jazizadeh2013human} &
\begin{tabular}{c} Field data \\ from 4 subjects \end{tabular} &
\begin{tabular}{c}Continuous thermal preference scale \\
with 2 labels (cooler and warmer) at the extreme ends \end{tabular} &
Ta &
\begin{tabular}{c}
Indoor air temperatures associated with \\
5-level thermal sensation index \\
(very cold/ cold/ neutral/ warm/ very warm).
\end{tabular} &
Fuzzy rules &
\begin{tabular}{c}
Mean error in actual vs. predicted associated air temp. = $\dC{1.17}$ \\
Model validated using 7 different synthetic personas. \\
Field experiment was conducted over \\a period of 2 months
to collect sufficient data. \\
A total number of 114, 77, 76 and 61 datapoints \\
were obtained from subjects 1-4 for training the model. \\
Datapoints were  randomly collected \\ (all at once, no active learning framework setup).\end{tabular} \\
\hline{}
\cite{rana2013feasibility} & 
\begin{tabular}{c} Field data \\ from 6 subjects \end{tabular} &
\begin{tabular}{c}
ASHRAE 7-point thermal sensation scale
\end{tabular} &
Ta, RH &
ASHRAE 7-point thermal sensation scale &
\begin{tabular}{c}
Support vector machine, \\
humidex
\end{tabular} &
Overall accuracy using campus dataset = 80\%\\
\hline{}
\cite{daum2011personalized} &
\begin{tabular}{c} Field data \\ from 6 subjects \end{tabular} &
ASHRAE 7-point thermal sensation scale &
Ta &
\begin{tabular}{c}
Probability of 3 comfort conditions \\(too hot/comfortable/too cold)
\end{tabular} &
Logistic regression &
\begin{tabular}{c}
Congruency (convergence to true satisfaction profile) \\
increased from 0.66 to 0.87 after 90 votes were added for one subject. \\
Datapoints were  randomly collected \\ (all at once, no active learning framework setup). \\
In an active learning setup, older (than 30 days) votes \\
in the vicinity of Ta$\pm \dC{0.25}$ were replaced with new entry. \\
Comfort distance was then recalculated with new data entry.
\end{tabular} \\
\hline{}
\begin{tabular}{c}
\textbf{Our} \\ \textbf{Model} 
\end{tabular} &
\begin{tabular}{c}
Field data \\ from 6 subjects 
\end{tabular} &
\begin{tabular}{c}
3-point thermal preference \\
(prefer warmer/ satisfied / prefer colder)
\end{tabular} &
Ta &
\begin{tabular}{c}
Maximally preferred \\ indoor air temperature. \\
Utility function values.
\end{tabular} &
\begin{tabular}{c}
Bayesian inference, \\
Bayesian active learning, \\
Gaussian processes with unimodality constraints, \\
Expected improvement in utility
\end{tabular} &
\begin{tabular}{c}
Model validated using 3 synthetic occupants \\
(with different utility profiles). \\
Datapoints needed to infer the max. preferred temp. values: \\
minimum number of datapoints : 5 \\
maximum number of datapoints : 11 \\
Thermal comfort queries are sequentially (intelligently) collected \\
(one by one, active learning setup).
\end{tabular}
\\
\bottomrule
\end{tabular}
\begin{tablenotes}\footnotesize
\item [a] Note: Ta = indoor air temperature, RH = relative humidity,
MRT = metabolic rate, Va = air velocity, CO$_{2}$ = carbon dioxide level
\end{tablenotes}
\end{threeparttable}
}
\end{sidewaystable}

\section{Methodology}\label{sec:methodology}

\subsection{Problem formulation}\label{subsec:problem_formulation}
We assume that each occupant has a well-defined preference relation over the set of all  possible indoor thermal environments.
If this preference relation satisfies a set of reasonable preference axioms \cite{morgenstern1953theory}, then we can encode it as a scalar function of the thermal environment $x$ taking values in a set $\calX$.
That is, for a given occupant there exists a function $u:\calX\rightarrow\R$ such that the statement ``the occupant prefers the thermal environment $x$ to the thermal environment $x'$'' is equivalent to the mathematical statement $u(x) > u(x')$ \cite{furnkranz2010preference}. 
For example, since we associate the thermal environment of a room with its air temperature in degrees C, $u$ is a real function of indoor air temperature and a preference of $\dC{25}$ over $\dC{20}$ is equivalent to $u(\dC{25}) > u(\dC{20})$.
We call $u$, the \textit{thermal utility function} of the occupant.
The maximally preferred indoor air temperature $x^{*}$ is then given as:
\begin{equation}\label{eqn:x_pref_abs}
x^{*} = \argmax_{x \in \calX} u(x),
\end{equation}
where the maximization takes place over the set of all possible indoor air temperatures $\calX = [\dC{20}, \dC{28}]$.

The problem presented in \Cref{eqn:x_pref_abs} is hard because information about $u$ is only indirectly available to us \cite{gonzalez2017preferential}.
One way around this limitation is to ask the occupant to compare between two different thermal environments of the room \cite{sui2018advancements, busa2018preference, houlsby2012collaborative, dewancker2018sequential}.
Indeed, \cite{xiong2018inferring} and \cite{awalgaonkar2018design} followed exactly this approach to infer visual preferences of occupants in private office spaces. 
However, our pilot studies revealed that this approach is not suitable for eliciting thermal preferences, reason being that, as the time required to change the steady state indoor air temperature is of the order of half an hour (in our experimental setup), it becomes very likely that the occupant forgets the initial temperature and pairwise comparisons between two consecutive thermal environments of the room is not viable in such a scenario. 
For this reason, we have resorted to instantaneous thermal preference queries of the form ``prefer warmer,'' ``prefer cooler,'' ``satisfied '' which we interpret as statements about the derivative of the utility function, i.e., they mean that $u(x)$ is ``increasing,'' ``decreasing,'' ``constant'' respectively.
An addendum is that such queries have low cognitive load \cite{chajewska2000making}. 

\subsection{Outline of Preference Elicitation framework}\label{subsec:outline_pe}
The proposed PE framework operates as follows:
\begin{enumerate}
    \item 	An occupant working in an office space is exposed to indoor air temperature $x$ in $\calX$.
    After 30 minutes, we ask the occupant about his/her thermal preference in the form of a query, ``How satisfied are you with current thermal conditions?''
    \item 	If the occupant responds ``I'm satisfied with current condition'', we record discrete variable response $y$ as $0$.
    If the occupant responds ``I prefer warmer'' we record variable $y$ to be equal to $1$. Lastly, if the occupant responds ``I prefer cooler'' we record variable $y$ to be equal to $-1$.
    \item Using all data collected so far, we update our beliefs about the occupant's utility function $u$, see Sec.~\ref{subsec:utility}.
    \item If the convergence conditions (stopping criterion) we set for elicitation framework are satisfied, we STOP, see Sec.~\ref{sec:proof_of_concept} and Sec.~\ref{sec:DOE}.
    \item We identify the indoor air temperature $\xnew$ in $\calX$ that maximizes the expected improvement in the occupant's utility, see  Sec.~\ref{sec:next_point_selection}.
    \item We implement the thermal condition $\xnew$ and we GOTO 1. 
\end{enumerate}

\subsection{Learning an occupant’s thermal utility function}\label{subsec:utility}
Assume that we have observed $n$ number of thermal preference queries so far, which are given as:
\begin{equation}\label{eqn:data}
D_n = (X_n, \mathbf{y}_n),
\end{equation}
where $X_n = (x_{1}, \dots x_n)$  are the observed indoor air temperature and $\mathbf{y}_n = (y_1, \dots, y_n)$ is the occupant's response to the queries. 
In this section, we discuss how to make use of thermal preference data $D_{n}$ to update our beliefs about the utility function $u$.
For notational simplicity, we skip the subscript $n$ and we write $X$ instead of $X_n$, $\mathbf{y}$ instead of $\mathbf{y}_n$, and $D$ instead of $D_n$.
We follow a Bayesian approach where we put a suitable prior on the space of $u$'s, condition it on $D_n$ and sample from the resulting posterior using a variant of Markov chain Monte Carlo.
The technical details are outlined below.

From previous research, we know that an occupant's thermal utility function is smooth, it has a unique maximum, and that ``extreme'' indoor air temperatures are rarely preferred by individual occupants \cite{lee2017bayesian, gunay2018modelling, gunay2018development}.
Mathematically, we assume that $u$ is infinitely differentiable, and that it is unimodal.
The latter means that there exists a temperature $c_{0}$ in $\calX$ such that $u$ is monotonically increasing for temperatures smaller than $c_{0}$ and monotonically decreasing for temperatures greater than $c_{0}$.
Of course, this implies that the derivative of $u$ is positive for temperatures less than $c_0$ and negative for temperatures greater than $c_0$.
We present examples of such utility functions in Sec.~\ref{sec:proof_of_concept}.
%In this study, we are going to incorporate prior belief that $u(\cdot)$ is unimodal with a single maximum into our GP-PE modeling framework.
To encode this prior information, we construct a hierarchical Bayesian model for unimodal functions based on Gaussian processes \cite{andersen2017bayesian}.
We encode the prior belief about the unimodality of the utility function via virtual derivative observations on $u$ that would satisfy the above defined inequality constraints \cite{riihimaki2010gaussian}. 

\subsubsection{Encoding our beliefs about the smoothness of the thermal utility}
\label{subsubsec:smoothness}
Our model assigns a zero-mean GP prior on $u$'s, i.e.,
\begin{equation}
    \label{eqn:gp}
    u | \boldsymbol{\uptheta}_{u} \sim \GP(0,k(\cdot,\cdot;\boldsymbol{\uptheta}_{u})),
\end{equation}
where $k(\cdot,\cdot;\boldsymbol{\uptheta}_{u})$ is a positive definite function known as the covariance function, and $\boldsymbol{\uptheta}_{u}$ are its parameters.
The smoothness of function samples from a GP prior is directly connected to the smoothness of the covariance function \cite{rasmussen2006gaussian, banerjee2003smoothness}.
Therefore, to enforce that sampled $u$'s are infinitely differentiable, we use the squared exponential covariance function:
\begin{equation}\label{eqn:sqrexp}
k(x, x';\boldsymbol{\uptheta}_{u}) = \nu_{u} \exp\left\{-\frac{(x - x')^2}{2\rho_u^2} \right\},
\end{equation}
where $\bthetau = (\nu_{u},\rho_{u})$: the signal variance $\nu_{u}>0$ and lengthscale $\rho_{u}>0$.
Following a fully Bayesian modeling approach, we need to place priors over these hyperparameters $\bthetau$ such that they reflect our beliefs about them i.e., what do we believe about the length of the ‘wiggles’ in utility function? How sensitive is the utility function to change in any of the input features? How much farther away is the true utility function from its mean function values?
Since we are quite uncertain about answers to these questions, we reflect this uncertainty by assigning uninformative Gamma prior distributions over signal variance $\nu_{u}$ and lengthscale $\rho_{u}$. 
We assume that $\nu_u$ and $\rho_u$ are a priori independent, i.e., $p(\boldsymbol{\uptheta}_{u})=p(\nu_{u})p(\rho_u)$ and we choose:
\begin{equation}\label{eqn:hyperprior1}
\begin{array}{ccc}
p(\nu_{u}) &=& \mathcal{G}(\nu_{u}|\alpha_1, \beta_1),\\
p(\rho_{u}) &=& \mathcal{G}(\rho_{u}|\alpha_2, \beta_2),
\end{array}
\end{equation}
where $\mathcal{G}(\cdot|\alpha, \beta)$ is probability density of a Gamma r.v. (random variable) with shape parameter $\alpha$ and scale parameter $\beta$ \cite{thom1958note}.
We set hyperparameters $\alpha_{1} = 1$, $\beta_{1} = 1$, $\alpha_{2} = 1$ and $\beta_{2} = 1$ reflecting our uncertain beliefs.

A GP defines a probability measure on the space of utility functions such that any finite collection of function values follows a multivariate Gaussian distribution \cite{rasmussen2006gaussian}.
For example, by assigning a GP prior on the space of $u$'s, the joint probability density over the utility function values at all the observed indoor air temperature $X = (x_{1}, ..., x_{n})$, denoted by vector $\mathbf{u} =(u(x_{1}),\dots,u(x_n))$, is the multivariate zero-mean Gaussian $p(\mathbf{u} |X, \bthetau) = \calN\left(\mathbf{u}|\bzero, \bKuu\right)$,
where the  $N\times N$ covariance matrix $\bKuu$, by definition, has elements:
\begin{equation}\label{eqn:cov_bt_u}
    (\bKuu)_{ij} := \C[u(x_i), u(x_j)|\bthetau] := k(x_j,x_j; \bthetau),
\end{equation}
where $\C[\cdot,\cdot|\bthetau]$ is the covariance operator conditional on $\bthetau$.
In \ref{appen_subsec: GPR with and without monotonicity} and \ref{appen_subsec: GPR with and without unimodality}, we show how GP would work in a conventional regression problem (see \cref{fig:appen_mon_dec_gp_fit} and \cref{fig:appen_unimodal_normal_gp_fit}).
%As we can see from Eq. (\ref{eq3}), GPs are fully specified by covariance function $k(\cdot, \cdot)$.
%The covariance between utility function values at room temperatures $x_{i}$ and $x_{j}$ can therefore be written as $\text{Cov}[u(x_{i}), u(x_{j})] = k(x_{i}, x_{j})$.

Furthermore, the derivative of $u$ is also a GP since differentiation is a linear operator \cite{rasmussen2006gaussian}.
This makes it possible to include derivative observations in the GP model, or to compute predictions related to it \cite{wu2017bayesian, wu2017exploiting, eriksson2018scaling}. 
More importantly, $(u,u')$, seen as a two-output function of indoor air temperature, is a two-output Gaussian process, i.e., all finite dimensional joint distributions of $u$ and $u'$ values are multivariate Gaussian.
The covariance function of the $(u,u')$ process is given as follows (see \cite{riihimaki2010gaussian}):
\begin{eqnarray}
\label{eqn:cov_u_uder}
\C[u'(x), u(x')|\bthetau] &=& \frac{\partial k(x,x';\bthetau)}{\partial x} = 
k(x,x';\bthetau)\frac{x - x'}{\rho_u^2},\\
\label{eqn:cov_uder_uder}
\C[u'(x), u'(x')|\bthetau] &=& \frac{\partial^2 k(x,x';\bthetau)}{\partial x \partial x'} = k(x,x';\bthetau)\rho_u^{-2}\left(1-\frac{(x - x')^{-2}}{\rho_u^2}\right).
\end{eqnarray}
For example, with these definitions, the joint distribution of function values $\mathbf{u}$ and function derivatives $\mathbf{u'} = (u'(x_1),\dots, u'(x_N))$ is the multivariate zero-mean Gaussian:
\begin{equation}\label{eqn:gp_u_uder}
    p(\mathbf{u}, \mathbf{u'} |X, \bthetau) = \calN\left(
    \left(\begin{array}{c}\mathbf{u}\\ \mathbf{u'}\end{array}\right)
    \middle|\bzero, \bK_u\right),
\end{equation}
where $\bK_u$ is the $(2N)\times(2N)$ matrix defined by:
\begin{equation}\label{eqn:cov_u_all}
\bK_u = 
\left(
\begin{matrix}
\bKuu & \bK_{u',u}^T \\
\bK_{u',u} & \bK_{u',u'}
\end{matrix}
\right),
\end{equation}
with $\bKuu$ as in Eq.~(\ref{eqn:cov_bt_u}), $\bK_{u',u}$ computed using Eq. (\ref{eqn:cov_u_uder}) and $\bK_{u',u'}$ computed using Eq.~(\ref{eqn:cov_uder_uder}).
This result generalizes to any collection of utility and utility derivatives values.

\subsubsection{Encoding our beliefs about the unimodality of the utility}
\label{sec:unimodality}
Assume that $u$ is a unimodal function of indoor air temperature and let $c_0$ in $\calX$ be the unique maximally preferred indoor air temperature.
Then, since $u$ is continuously differentiable, it holds that $u'(x) > 0$ for $x < c_{0}$ and $u'(x) < 0$ for $x > c_{0}$.
In other words, as we move through the interval $\calX$, the utility derivative $u'$ must flip signs from positive to negative only once.
It is not trivial to model this constraint directly since, apart from the single sign flipping, $u'$ is arbitrary. In \ref{appen_subsec: GPR with and without monotonicity}, we show how a unimodal Gaussian process model (UGP) \cite{andersen2017bayesian} would work in a conventional regression setting.
The idea is twofold.
First, we introduce an easy-to-model latent GP $g$ that satisfies the same sign flips as $u'$, and second we force the signs of $g$ and $u'$ to agree.
We discuss these two points sequentially.

Let $g$ be a zero-mean latent GP with a squared exponential covariance function, hyperparameters $\bthetag$, and hyperparameter prior $p(\bthetag)$. We assign the same uninformative Gamma prior distribution over $p(\bthetag)$ as in Eq.(\ref{eqn:hyperprior1}).
We want to force $g$ to flip signs at most once in $\calX$.
The easiest way to encode this information is using a monotonicity constraint (see \ref{appen_subsec: GPR with and without monotonicity}).
In particular, it suffices to force $g$ to be monotonically decreasing, which is equivalent to forcing the derivative $g'$ to be negative throughout $\calX$.
A common approach is to enforce the monotonicity constraint using a set of virtual derivative observations.
Basically, instead of using ``$g'$ is negative throughout $\calX$'', we use the weaker, but more practical, constraint ``$g'$ is negative on a finite set of indoor air temperatures $\tilde{X} = (\tilde{x}_1,\dots,\tilde{x}_j)$ in $\calX$.''
The set $\tilde{X}$ should cover as much of $\calX$ as possible.
In this work, we take $\tilde{X} = (20, 20.5, 21, ..., 28)$.
Let $\mathbf{\tilde{g}}$ and $\mathbf{\tilde{g}}'$ be the $J$-dimensional vectors of values and derivatives of $g$ at the set points $\tilde{X}$.
The conditional prior joint distribution of $g$ and $g'$ is:
\begin{equation}
    \label{eqn:g_joint}
    p(\tilde{\mathbf{g}}, \tilde{\mathbf{g}}' | \tilde{X}, \bthetag) \propto \calN\left(
    \left(\begin{array}{c}
         \tilde{\mathbf{g}}  \\
         \tilde{\mathbf{g}}' 
    \end{array}\right)
    \middle|
    \mathbf{0},
    \tilde{\mathbf{K}}_g
    \right)\cdot
    \left(\prod_{j=1}^J \Phi\left(-\nu_g g'(\tilde{x}_j)\right)\right),
\end{equation}
where the first part of the right hand side is the equivalent of Eq.~(\ref{eqn:gp_u_uder}).
The second part of the right hand side of Eq.~(\ref{eqn:g_joint}) puts higher probability to negative $\tilde{\mathbf{g}}'$.
Here, $\Phi$ is the cumulative distribution function of the standard normal distribution, also called probit link function and is given as: $\Phi(z) = \int_{-\infty}^{z} \calN(\gamma| 0,1)d\gamma$.
The positive parameter $\nu_g$ controls how fast the probit transitions from zero to one.
We take $\nu_g = 10^6$, which practically turns $\Phi(-\nu_g g'(\tilde{x}_j))$ into a step function.

Second, we force the signs of $g$ and $u'$ to agree.
Similarly to the previous paragraph, we weaken the constraint ``the signs of $g$ and $u'$ agree throughout $\calX$ to the more practical ``the signs of $g$ and $u'$ agree on a finite set of points $\tilde{X}$.''
To achieve this, we simply introduce the set of virtual observations $\tilde{\mathbf{y}} = \left(\text{sign}(u'(\tilde{x}_1)),\dots, \text{sign}(u'(\tilde{x}_j))\right)$.
We connect $\tilde{y}$ to the sign of $\tilde{g}$ through the generative model:
\begin{equation}
    p(\tilde{\mathbf{y}} | \tilde{\mathbf{g}}) \propto \prod_{j=1}^J\text{Ber}(\tilde{y}_j| \Phi(\nu_{\tilde{y}}\tilde{g}_j)),
\end{equation}
where $\text{Ber}(\cdot| \lambda)$ is the probability mass density of a Bernoulli r.v. with probability of success $\lambda$, and $\nu_{\tilde{y}}$ is a positive parameter.
The parameter $\nu_{\tilde{y}}$ controls the strength of the connection between the $\tilde{\mathbf{y}}$ and the sign of $\tilde{\mathbf{g}}$.
In the limit of $\nu_{\tilde{y}}\rightarrow \infty$, the two match perfectly.
In the limit of $\nu_{\tilde{y}}\rightarrow 0$, there is no connection between the two, i.e., $\mathbf{\tilde{y}}$ is completely random.
For any finite, nonzero $\nu_{\tilde{y}}$ the strength of the connection between the signs depends on the magnitude of $g$.
The two signs match well when the absolute value of $g$ is large.
As expected, the largest uncertainty about the sign is at the point where $g$ crosses zero, i.e., at the maximally preferred temperature.
Note that $\nu_{\tilde{y}}$ can be absorbed by the signal variance $\eta_g$ of $g$.
Thus, without loss of generality, we take $\nu_{\tilde{y}} = 1$.

Finally, we condition the original GP to satisfy the virtual observations $\tilde{y}$.
This yields a prior model for $u$ that satisfies the unimodality constraint.
Unfortunately, because of the nonlinear nature of the sign operator, the resulting process is not a GP.
Nevertheless, the joint prior of the utility and utility derivative values at arbitrary points is a modified multivariate Gaussian of the form.
This is enough for characterizing the posterior through sampling.
In particular, the joint prior of $\tilde{\mathbf{u}}$, $\tilde{\mathbf{u}}'$, and $\mathbf{u}'$ is 
\begin{equation}\label{eqn:modified_prior}
    p(\mathbf{\tilde{u}}, \mathbf{u}', \tilde{\mathbf{u}}' | X, \tilde{X}, \mathbf{\tilde{y}}, \bthetau) \propto p(\mathbf{\tilde{u}}, \mathbf{u}', \tilde{\mathbf{u}}' | X, \tilde{X}, \bthetau)\cdot \left(\prod_{j=1}^J\Phi(\nu_v \tilde{y}_j \tilde{u}_j')\right).
\end{equation}
The first part of the right hand side is a trivial extension of Eq.~(\ref{eqn:gp_u_uder}).
The second part of the right hand side of Eq.~(\ref{eqn:modified_prior}) acts as an indicator function.
It give higher probability when the signs of $\mathbf{\tilde{y}}$ and $\tilde{\mathbf{u}}$ match and lower when they do not.
We use $\nu_v = 10^6$.

\subsubsection{The likelihood of the data}
\label{sec:like}
The response to a preference query gives us information about the sign of derivative of utility function at the indoor air temperature where we ask these queries.
%For example, if we ask the comfort query when the room temperature is \SI{20}{\celsius} and occupant responds “I prefer warmer”, then that means that derivative of utility function at temperature \SI{20}{\celsius}, i.e. $\text{sign}[u'(20^{\text{o}}\text{C})] = 1$, where 1 represents that the derivative is positive. Similarly, if the occupant responds “I prefer cooler temperature at \SI{27}{\celsius}, then $\text{sign}[u'(27^{\text{o}}\text{C})] = -1$, where -1 represents that the derivative is negative.
%If the occupant has no preference (i.e., he likes the current temperature) at \SI{24}{\celsius}, then $\text{sign}[u'(24^{\text{o}}\text{C})] = 0$, where 0 represents that the derivative is 0.
Essentially, the observed data $\mathbf{y}$ are noisy observations of the sign of derivatives $\mathbf{u}' = \left(u'(x_1),\dots, u'(x_N)\right)$.
The likelihood function is a model of the measurement process and it establishes the connection between $\mathbf{y}$ and $\mathbf{u}$.
We define:
\begin{equation}\label{eqn:likelihood_all}
    p(\mathbf{y}| \mathbf{u}', X) \propto \prod_{i = 1}^{N}\Phi(\nu_ly_iu_i'),
\end{equation}
where $\nu_l$ is a positive parameter controlling the level of the noise.
The proposed likehood encodes the following intuitive characteristics.
First, the likelihood is high when $y_i$ and $u_i'$ have the same sign.
Second, the likelihood is low when $y_i$ and $u_i'$ have opposite signs.
Third, as $u_i'$ gets close to zero, any answer is equally probable.
Without loss of generality, we take $\nu_l = 1$ a the signal variance $\eta_u$ of $u$ can absorb it.

\subsubsection{The posterior state of knowledge}
\label{sec:post}
The posterior probability density of all model parameters is given by Bayes rule:
\begin{equation}\label{eqn:posterior}
    p(\mathbf{\tilde{u}}, \mathbf{u}', \tilde{\mathbf{u}}', \mathbf{\tilde{g}}, \mathbf{\tilde{g}}', \mathbf{\tilde{y}}, \bthetau, \bthetag | \tilde{X}, D) \propto
    p(\mathbf{y}| \mathbf{u}', X)
    p(\mathbf{\tilde{u}}, \mathbf{u}', \tilde{\mathbf{u}}' | X, \tilde{X}, \mathbf{\tilde{y}}, \bthetau)
    p(\tilde{\mathbf{y}} | \tilde{\mathbf{g}})
    p(\tilde{\mathbf{g}}, \tilde{\mathbf{g}}' | \tilde{X}, \bthetag)
    p(\bthetau)
    p(\bthetag),
\end{equation}
where all the terms of the right hand side have been already introduced in the previous sections.
Note, that it is possible to analytically marginalize over the virtual observations $\mathbf{\tilde{y}}$ getting:
\begin{equation}\label{eqn:marginalized_posterior}
    \begin{array}{ccc}
    p(\mathbf{\tilde{u}}, \mathbf{u}', \tilde{\mathbf{u}}', \mathbf{\tilde{g}}, \mathbf{\tilde{g}}', \bthetau, \bthetag | \tilde{X}, D) &=& \sum_{\tilde{\mathbf{y}}} p(\mathbf{\tilde{u}}, \mathbf{u}', \tilde{\mathbf{u}}', \mathbf{\tilde{g}}, \mathbf{\tilde{g}}', \mathbf{\tilde{y}}, \bthetau, \bthetag | \tilde{X}, D)\\
    &\propto& 
    p(\mathbf{y}| \mathbf{u}', X)p(\tilde{\mathbf{g}}, \tilde{\mathbf{g}}' | \tilde{X}, \bthetag)
    p(\bthetau)
    p(\bthetag)\cdot\\
    && 
    \prod_{j=1}^J\left[\Phi(-\nu_{u}\tilde{u}_j')\Phi(-\nu_{\tilde{y}}g_j)
    +\Phi(\nu_{u}\tilde{u}_j')\Phi(\nu_{\tilde{y}}g_j)
    \right].
    \end{array}
\end{equation}
To get the marginalized posterior of any variable, one just needs to integrate out all other variables.
In particular, the joint posterior probability density of the utility values $\mathbf{\tilde{u}}$ at the virtual indoor air temperature values $\tilde{X}$ and the hyperparameters $\bthetau$ is given by:
\begin{equation}\label{eqn:predictive}
    p(\mathbf{\tilde{u}}, \bthetau | \tilde{X},D) = \int p(\mathbf{\tilde{u}},\bthetau,\mathbf{u}', \tilde{\mathbf{u}}', \mathbf{\tilde{g}}, \mathbf{\tilde{g}}', \bthetau, \bthetag | D, \tilde{X})d\mathbf{u}'d\mathbf{\tilde{u}'}d\mathbf{\tilde{g}}d\mathbf{\tilde{g}} d\bthetag.
\end{equation}
%Using the sum rule of probability, the marginal posterior distribution over utility function values $\tilde{u}}$, its derivatives $\tilde{\mathbf{u}}'$ and hyperparameters $\boldsymbol{\uptheta}_{u}$ is then given as:
%\begin{equation}\label{eqn:post_u_u_der}
%p(\mathbf{\tilde{u}}, \tilde{\mathbf{u}}',\bthetau| X, \mathbf{y}) =
%\int p(\mathbf{\tilde{u}}, \mathbf{u}', \tilde{\mathbf{u}}', \mathbf{\tilde{g}}, \mathbf{\tilde{g}}', \mathbf{y}, \bthetau, \bthetag | X, \tilde{X})
%d\mathbf{\tilde{g}}d\mathbf{\tilde{g}}'d\bthetag.
%\end{equation}
%This marginal posterior distribution can then be interpreted as a Gaussian process subjected to unimodality constraints.
%For a fully Bayesian treatment of our model, we also need to characterize our updated (posterior) beliefs about the hyperparameters $\boldsymbol{\uptheta} = [\boldsymbol{\uptheta}_{u}, \boldsymbol{\uptheta}_{g}]$.
%Using sum rule, the posterior distribution over hyperparameters $\boldsymbol{\uptheta}$ is given as:
%\begin{equation}\label{eqn:post_theta}
%p(\boldsymbol{\uptheta}_{g},\boldsymbol{\uptheta}_{u}|\mathbf{y},X,\hat{X}, \tilde{X}) =
%\int \int \int \int p(\mathbf{u}, \mathbf{u'},\mathbf{g}, \mathbf{g'}, \boldsymbol{\uptheta}_{g}, %\boldsymbol{\uptheta}_{u}|\mathbf{y},X,\hat{X}, \tilde{X}) d\mathbf{g}d\mathbf{g'}d\mathbf{u}d\mathbf{u'}.
%\end{equation}
%It gets cumbersome to explicitly show the conditioning of posterior probability densities on $\tilde{X}$.

\subsubsection{Making predictions at unobserved indoor air temperatures}
\label{sec:predict}
Now that we have quantified our posterior beliefs regarding the utility function values at virtual indoor air temperatures $\tilde{X}$ (see Eq.(\ref{eqn:predictive})), 
we shift our attention to prediction of utility function at an arbitrary indoor air temperature value $x^{*}$ in $\calX$.
%We are interested in answering the question, ``Given that we have observed thermal preference data $D_{N}$, what is the value of utility function $u$ at new temperature $x^{*}$?''
This posterior predictive distribution over utility function values $u^* = u(x^{*})$ is:
\begin{equation}\label{eqn:post_pred}
\begin{array}{ccc}
p(u^{*}|x^{*}, \tilde{X}, D)
&=&
\int p(u^*, \tilde{\mathbf{u}}, \bthetau | x^*, \tilde{X}, D) d\tilde{\mathbf{u}}d\bthetau\\
&=& \int p(u^* | x^*, \tilde{X}, \tilde{\mathbf{u}}, \bthetau, D)p(\mathbf{\tilde{u}}, \bthetau |  \tilde{X}, D)d\mathbf{\tilde{u}}d\bthetau,
\end{array}
\end{equation}
where, going from the first to the second line, we used Bayes rule.
The first term inside the integral of the second line of Eq.~(\ref{eqn:post_pred}) is the posterior probability density of $u^*$ conditioned on the observations $D=(X,\mathbf{y})$, the virtual observations $(\tilde{X},\tilde{\mathbf{u}})$, and the hyperparameters $\bthetau$, while the second term is given in Eq.~(\ref{eqn:predictive}).
Since the virtual points $\tilde{X}$ cover the input space $\calX$ densely, the information contained in $(\tilde{X}, \mathbf{\tilde{u}})$ overshadows the information in $D$.
In other words, to a good approximation, the effect of the observations $D$ can be dropped.
Mathematically, when we know that $\tilde{\mathbf{u}}$ and $\bthetau$ are jointly distributed as specified in Eq.~(\ref{eqn:predictive}), then
$$
p(u^* | x^*, \tilde{X}, \tilde{\mathbf{u}}, \bthetau, D)\approx p(u^* | x^*, \tilde{X}, \tilde{\mathbf{u}}, \bthetau),
$$
and, thus, Eq.~(\ref{eqn:post_pred}) becomes:
\begin{equation}\label{eqn:post_pred_approx}
p(u^{*}|x^{*}, \tilde{X}, D) \approx \int p(u^{*}|x^{*},\tilde{X}, \mathbf{\tilde{u}}, \bthetau)
p(\mathbf{\tilde{u}}, \bthetau |\tilde{X}, D)d\mathbf{\tilde{u}}d\bthetau.
\end{equation}
Finally, $p(u^* | x^*, \tilde{X}, \tilde{\mathbf{u}}, \bthetau)$ is analytically available:
\begin{equation}\label{eqn:conditional}
p(u^* | x^*, \tilde{X}, \tilde{\mathbf{u}}, \bthetau) = 
\calN\left(u^{*} \middle| m_{u}(x^*; \tilde{X}, \tilde{\mathbf{u}}, \bthetau), \sigma_u^2(x^*; \tilde{X}, \tilde{\mathbf{u}}, \bthetau)\right),
\end{equation}
where the mean is:
\begin{equation}\label{eqn:u_mean}
    m_u(x^*; \tilde{X}, \tilde{\mathbf{u}}, \bthetau) = \bK_{u^{*}, \tilde{u}}\bK_{\tilde{u},\tilde{u}}^{-1}\tilde{\mathbf{u}},
\end{equation}
and the variance is:
\begin{equation}\label{eqn:u_var}
      \sigma_u^2(x^*; \tilde{X}, \tilde{\mathbf{u}}, \bthetau) = K_{u^{*},u^{*}} - \bK_{u^{*},\tilde{u}}\bK^{-1}_{\tilde{u},\tilde{u}}\bK^{T}_{u^{*},\tilde{u}}.
\end{equation}
All the $K$-quantities in Eqs.~(\ref{eqn:u_mean}) and~(\ref{eqn:u_var}) involve the prior covariance of $u$ with hyperparameters $\bthetau$.
In particular, the $1 \times J$ matrix $\bK_{u^*,\tilde{u}}$ is the cross-covariance matrix between $u^*$ and $\tilde{\mathbf{u}}$, the $J \times J$ matrix $\bK_{\tilde{u},\tilde{u}}$ is the covariance matrix of $\mathbf{\tilde{u}}$, and the scalar $K_{u^*,u^*}$ is the a priori variance of $u^*$, i.e., $K_{u^*, u^*} = k(x^*,x^*;\bthetau)$.

\subsubsection{Characterizing our state of knowledge about the maximally preferred indoor air temperature}
\label{sec:maximally_preferred}
In this section we summarize our state of knowledge about the maximally preferred indoor air temperature values.
To this end, we define the random variable representing the location of the maximally preferred indoor air temperature:
\begin{equation}\label{eqn:x_best}
    X_{\text{best}} \equiv  X_{\text{best}}[u] := \argmax_{x\in \calX} u(x).
\end{equation}
The uncertainty in $X_{\text{best}}$ is, of course, induced by our uncertainty about the utility $u$.
Our state of knowledge is neatly captured by the probability density $p(x_{\text{best}} | D)$ which is formally given by
\begin{equation}
    p(x_{\text{best}} | \tilde{X}, D) = \mathbb{E}\left[\delta\left(X_{\text{best}}[u]-x_{\text{best}}\right)\middle| \tilde{X}, D\right],
\end{equation}
where $\mathbb{E}[\cdot|D]$ is the conditional expectation over the stochastic process $u$ conditioned on $D$.
In what follows, we derive an efficient approximation of this probability density so that it can be characterized by sampling.

Instead of working directly with the random variable $X_{\text{best}}$, we define a discretized version of it that considers only the utility values at the virtual temperatures:
\begin{equation}
    \tilde{X}_{\text{best}} \equiv \tilde{X}_{\text{best}}[\mathbf{\tilde{u}}] := \arg\max_{\tilde{x}\in\tilde{X}}u(\tilde{x}).
\end{equation}
This suffices, because as, as the number of virtual temperatures $J\rightarrow \infty$, $\tilde{X}_{\text{best}}$ converges to $X_{\text{best}}$ almost surely.
Subsequently, instead of probability density function $p(x_{\text{best}} |\tilde{X}, D)$, we characterize the probability mass function $p(\tilde{x}_{\text{best}} |\tilde{X}, D)$ defined by:
\begin{equation}\label{eqn:x_best}
    \begin{array}{ccc}
    p(\tilde{x}_{\text{best}} |\tilde{X}, D) &=& \mathbb{E}\left[\delta\left(\tilde{X}_{\text{best}}[\mathbf{\tilde{u}}]-x_{\text{best}}\right)\middle| \tilde{X}, D\right]\\
    &=& \int \delta\left(\tilde{X}_{\text{best}}[\mathbf{\tilde{u}}]-x_{\text{best}}\right)p(\tilde{\mathbf{u}} | \tilde{X}, D)d\tilde{\mathbf{u}}.
    \end{array}
\end{equation}
We approximate the second line of Eq.~(\ref{eqn:x_best}) by building a histogram from samples from the marginalized posterior $p(\tilde{\mathbf{u}} | \tilde{X}, D)$.

Eq.~(\ref{eqn:x_best}) can be used to decide when to stop the preference elicitation process.
In this work, we monitor the 95\% credible interval, i.e., the range of indoor air temperature $[a,b]\subset\calX$ for which
\begin{equation}
    \mathbb{P}(a\le \tilde{X}_{\text{best}}\le b | \tilde{X}, D) := \int_a^bp(\tilde{x}_{\text{best}} |\tilde{X}, D) = 0.95,
\end{equation}
and one of the criterion for stopping the elicitation process is when $b-a$ becomes smaller than a specific threshold. For the purpose of this paper, we have set this temperature threshold to be $\dC{1}$.

\subsubsection{Sampling from the posterior and posterior predictive distributions}
\label{sec:sampling}
%The non-Gaussian nature of conditional likelihood (Eq. (\ref{eqn:likelihood_all})), monotonic latent function $g(\cdot)$ (Eq. (\ref{eqn:p_g_g_der_z})) and unimodal utility function $u(\cdot)$ (Eq. (\ref{eqn:p_u_u_der_z})) make the above posterior and posterior predictive integrals to be analytically intractable.
%Furthermore, the Bayesian hierarchical structure of unimodal GP makes the hyperparameters and latent functions to be strongly coupled a posteriori. 
%We propose to tackle this intractability concerned with Bayesian inference in unimodal GP model through the use of stochastic simulations based on Hamiltonian Monte Carlo (HMC) methods \cite{duane1987hybrid}.
The posterior probability density is analytically intractable and it can only be characterized via sampling. 
We use Hamiltonian Monte Carlo (HMC) \cite{duane1987hybrid} to sample from Eq.~(\ref{eqn:marginalized_posterior}).
%We make use of HMC sampler to sample from the joint posterior distribution over hyperparameters and latent functions.
We implemented the model using the GPflow 0.4.0 package \cite{matthews2017gpflow}, a GP library that uses TensorFlow for its core computations and Python for its front end.
After analyzing the traces, autocorrelation, and the effective sample size (ESS) \cite{gilks1995markov} of all quantities,
we decided to burn the first 5,000 samples, and gather a total of $S=3,000$ samples by keeping one sample out of every 3 samples \cite{gelman2013bayesian}.
In what follows, we denote HMC samples from the joint posterior conditioned on $D_n = (X_n, \mathbf{y}_n)$ by $\left(\mathbf{\tilde{u}}^{(n,s)}, \mathbf{u}'^{(n,s)}, \tilde{\mathbf{u}}'^{(n, s)}, \mathbf{\tilde{g}}^{(n, s)}, \mathbf{\tilde{g}}'^{(n, s)}, \bthetau^{(n, s)}, \bthetag^{(n, s)} \right)$ for $s=1,\dots,S$.
We use these samples to carry out any integration over the posterior using sampling averages.

\subsection{Selecting the next thermal preference query}
\label{sec:next_point_selection}
In this paper, we develop a PE framework that uses the previously seen thermal preference data $D_n$ to select the next query $x_{n+1}$ to pose to the occupant by maximizing the expected improvement in utility.
%For addressing this problem, we have to come up with a
%new mathematical function which dictates the value of asking a new question at temperature $x_{q}$ .
%In traditional Bayesian Global optimization literature, these functions are called ``acquisition functions'' \cite{brochu2010tutorial}.
%We select the next indoor temperature to query by maximizing this acquisition function.
Symbolically, we select the next indoor air temperature $x_{n+1}$ to query by solving
\begin{equation}\label{eqn:pe_nxt_point_select}
x_{n+1} = \argmax_{x^* \in \mathcal{X}_n} \text{EUI}\left(x^*; D_n\right),
\end{equation}
where $\mathcal{X}_n \subset \calX$ denotes the set of all the available indoor air temperature values reachable from the current indoor air temperature $x_n$,
$\text{EUI}\left(x^*; D_n\right)$ denotes the expected utility improvement of a hypothetical query at $x^*$ given the data $D_n$ (to be defined below).
In practice, the set of reachable indoor air temperatures $\mathcal{X}_n$ is finite and, thus, the optimization problem of Eq.~(\ref{eqn:pe_nxt_point_select}) can be solved by a simple search algorithm.
In our studies, $\mathcal{X}_n$ is a grid of $\dC{0.5}$-separated indoor air temperatures reaching $\pm \dC{3}$ below and above the current indoor air of the room $x_n$.
See Sub-section \ref{subsec:framework_in_action} for examples.

We define EUI by generalizing the expected improvement (EI) acquisition function (for a detailed discussion of EI see the Bayesian global optimization literature \cite{snoek2012practical}).
%In this section, we define mathematically IUEI and we discuss how it can be approximated via sampling averaging.
Traditionally, EI is defined to be the expected marginal gain with respect to the current best observed quantity of interest.
However, the traditional definition of EI is not directly applicable because the quantity of interest, the utility function, is not directly observed in our problem.
%However, we do not have direct access to this utility (since it is not observed).
%Our only way of evaluating EI is to calculate in expectation, the marginal gain over inferred posterior predictive utility function values (see Eq.(\ref{eqn:sampling_post_pred}) and Eq.(\ref{eqn:post_pred}).
%The EI acquisition function estimates the expected marginal gain with respect to the current best expected utility function value of a hypothetical observation $x^*$.
%The current best utility is given by:
%We denote this current best utility function value by $u_{\text{best}}$ and is given as:

Therefore, in order to tackle this problem, we develop a context-specific EI that exploits the semi-analytical tractability of the approximation to the point-predictive posterior we derived in Eq.~(\ref{eqn:post_pred_approx}).
We start by conditioning on the virtual observations $(\tilde{X}, \mathbf{\tilde{u}})$ and the hyperparameters $\bthetau$, and we proceed to integrate them out using samples from their posterior, see Eq.~(\ref{eqn:post_pred}) and Sec.~\ref{sec:sampling}.
Our current best (conditional) estimate of the utility over the observed indoor air temperatures $X_n$ is given by:
\begin{equation}\label{eqn:pe_ubest}
\hat{u}_{\text{best}}(X_n, \tilde{X}, \mathbf{\tilde{u}}, \bthetau) := \max_{x\in X_n} \mathbb{E}\left[u(x_i) | \tilde{X}, \mathbf{\tilde{u}}, \bthetau\right] =
              \max_{x\in X_n} m_u(x; \tilde{X}, \mathbf{\tilde{u}}, \bthetau),
\end{equation}
where $\mathbb{E}[\cdot|\cdot]$ is the conditional expectation operator which we evaluated using Eq.~(\ref{eqn:conditional}).
Thus, the (conditional) improvement in utility of a hypothetical observation $x^*$ in $\calX$ is
\begin{equation}\label{eqn:improvement}
    \text{UI}(x^*; X_n, \tilde{X}, \mathbf{\tilde{u}}, \bthetau) := \max\{u(x^*) - \hat{u}_{\text{best}}(X_n, \tilde{X}, \mathbf{\tilde{u}}, \bthetau), 0\}.
\end{equation}
The expected (conditional) improvement in utility is then:
\begin{equation}\label{eqn:pe_ei}
\begin{array}{ccc}
\text{EUI}(x_*; X_n, \tilde{X}, \mathbf{\tilde{u}}, \bthetau) &:=& \mathbb{E}\left[\text{UI}(x^*; X_n, \tilde{X}, \mathbf{\tilde{u}}, \bthetau) \middle| \tilde{X}, \mathbf{\tilde{u}}, \bthetau \right]\\
&=& 
\sigma_u(x^*; \tilde{X}, \mathbf{\tilde{u}}, \bthetau)\cdot \\
&& \left[\gamma(x^*; X_n, \tilde{X}, \mathbf{\tilde{u}}, \bthetau)\Phi\left(\gamma(x^*; X_n, \tilde{X}, \mathbf{\tilde{u}}, \bthetau)\right)+ \phi\left(\gamma(x^*; X_n, \tilde{X}, \mathbf{\tilde{u}}, \bthetau)\right)\right],
\end{array}
\end{equation}
where
\begin{equation}\label{eqn:gamma}
    \gamma(x^*; X_n, \tilde{X}, \mathbf{\tilde{u}}, \bthetau) := \frac{\hat{u}_{\text{best}}(X_n, \tilde{X}, \mathbf{\tilde{u}}, \bthetau) - m_u(x^*; \tilde{X}, \mathbf{\tilde{u}}, \bthetau)}{\sigma_u(x^*;\tilde{X}, \mathbf{\tilde{u}}, \bthetau)},
\end{equation}
and $\Phi(\cdot)$ is the probability density function of the standard normal.
Finally, we integrate over the joint posterior of $\mathbf{\tilde{u}}$ and $\bthetau$ to get:
\begin{equation}\label{eqn:pe_integrated_ei}
\begin{array}{ccc}
\text{EUI}\left(x^*; D_n\right) &:=& \mathbb{E}\left[\text{EUI}(x_*; X_n, \tilde{X}, \mathbf{\tilde{u}}, \bthetau) \middle| \tilde{X}, D_n\right]\\
&=& \int \text{EUI}(x_*; \tilde{X}, \mathbf{\tilde{u}}, \bthetau) p(\mathbf{\tilde{u}}, \bthetau | X_n, \mathbf{y}_n, \tilde{X})d\mathbf{\tilde{u}}d\bthetau\\
&\approx& \frac{1}{S}\sum_{s=1}^S\text{EUI}\left(x_*; X_n, \tilde{X}, \mathbf{\tilde{u}}^{(n, s)}, \bthetau^{(n, s)}\right),
\end{array}
\end{equation}
where in the last step we used a sampling average approximation of the integral, see Sec.~\ref{sec:sampling}.

We present a concise algorithmic representation of the proposed PE scheme in Algorithm \ref{alg:PE}.

\begin{algorithm}[H]
\caption{Preference Elicitation algorithm}\label{alg:PE}
\begin{algorithmic}[1]
\State \textbf{Input:} Maximum number of queries (available budget) $N$.
\State \textbf{Initialize:} Query the occupant, collect the first preference query $x_1$ and receive first observation $y_1$ constituting the initial dataset $D_1$.
\For{$n = 2$ to $N$}
\State Sample from the posterior, Eq.(\ref{eqn:marginalized_posterior}), using the procedure of Sec.~\ref{sec:sampling}.
\State For all $x^*$ in $\calX_n$, compute the expected improvement in utility $\text{EUI}\left(x^{*}; D_{n-1}\right)$ (see Eq.(\ref{eqn:pe_integrated_ei})).
\State Identify the next indoor air temperature to query: $x_{n} = \argmax_{x^* \in \mathcal{X}_n} \text{EUI}\left(x^*; D_{n-1}\right)$ (see Eq.(\ref{eqn:pe_nxt_point_select})).
\State Query the occupant at $x_n$ and obtain $y_{n}$.
\State 	Augment previous preference data with this new query:$X_n = (X_{n-1}, x_n),\mathbf{y}_n=(\mathbf{y}_{n-1}, y_n)), D_n = (X_n, \mathbf{y}_n)$.
\EndFor
\State Sample from the posterior, Eq.(\ref{eqn:marginalized_posterior}), using the procedure of Sec.~\ref{sec:sampling}.
\State \textbf{Report:} Maximally preferred indoor air temperature value.
\end{algorithmic}
\end{algorithm}

%where $\alpha_{\text{EI}}(\cdot)$ depends on observed comfort data and hyperparameters $\boldsymbol{\uptheta}$.
%The above defined expectation in Eq. (\ref{eqn:pe_integrated_ei}) represents the correct generalization to account for uncertainties in hyperparameters
%This expectation over hyperparameters enables us to balance the trade-off between exploration and exploitation.
%When the framework is exploring, it will account for uncertainties in hyperparameters and choose new temperatures where the uncertainty in utility function predictions is large.
%When the framework is exploiting, it will choose new temperatures where the mean of utility function value is large.
%Since the integral in Eq. (\ref{eqn:pe_integrated_ei}) is intractable, we use the Monte Carlo estimate of IEUI which is given as:
%\begin{equation}\label{eqn:pe_integrated_ei_sampling}
%\hat{\alpha}(x_*| D_N, \boldsymbol{\uptheta}) \approx
%\frac{1}{S} \sum_{s = 1}^S  \alpha_{\text{EI}}\left(x_*| D_N, \boldsymbol{\uptheta}^{(s)}\right).
%\end{equation}

\section{Validation of the methodology}\label{sec:proof_of_concept}
\subsection{Synthetic occupant data generation}
\begin{figure}[H]
  \centering
  \begin{subfigure}[b]{0.32\linewidth}
    \includegraphics[width=\linewidth]{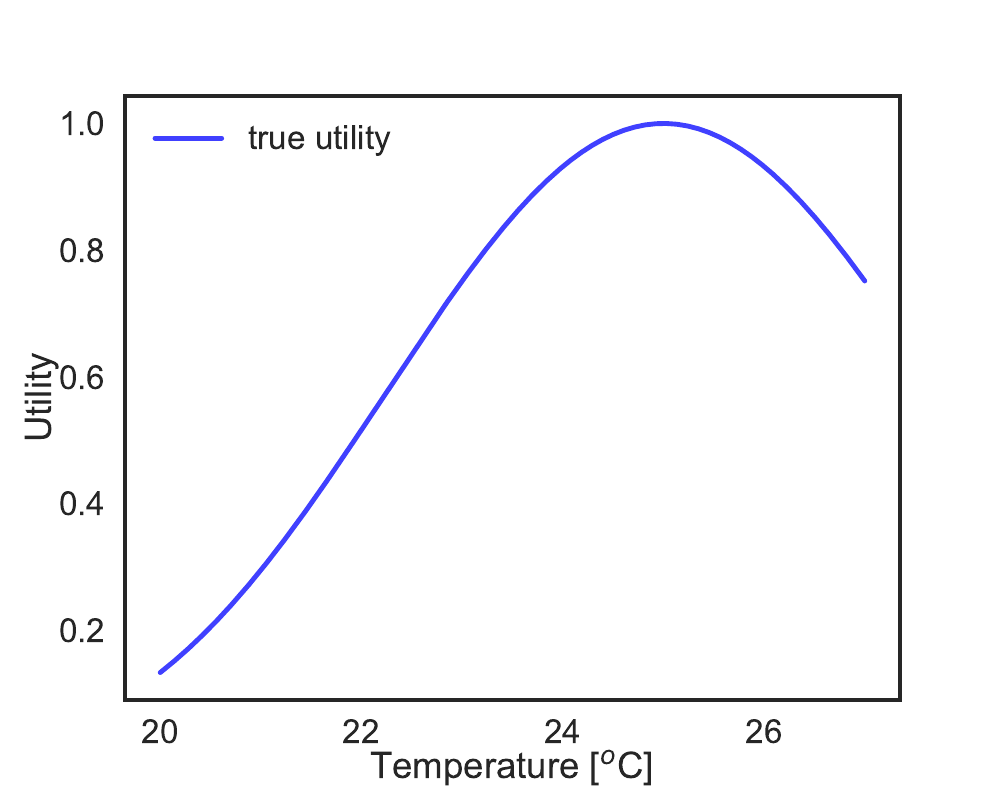}
     \caption{Synthetic Occupant 1}
  \end{subfigure}
  \begin{subfigure}[b]{0.32\linewidth}
    \includegraphics[width=\linewidth]{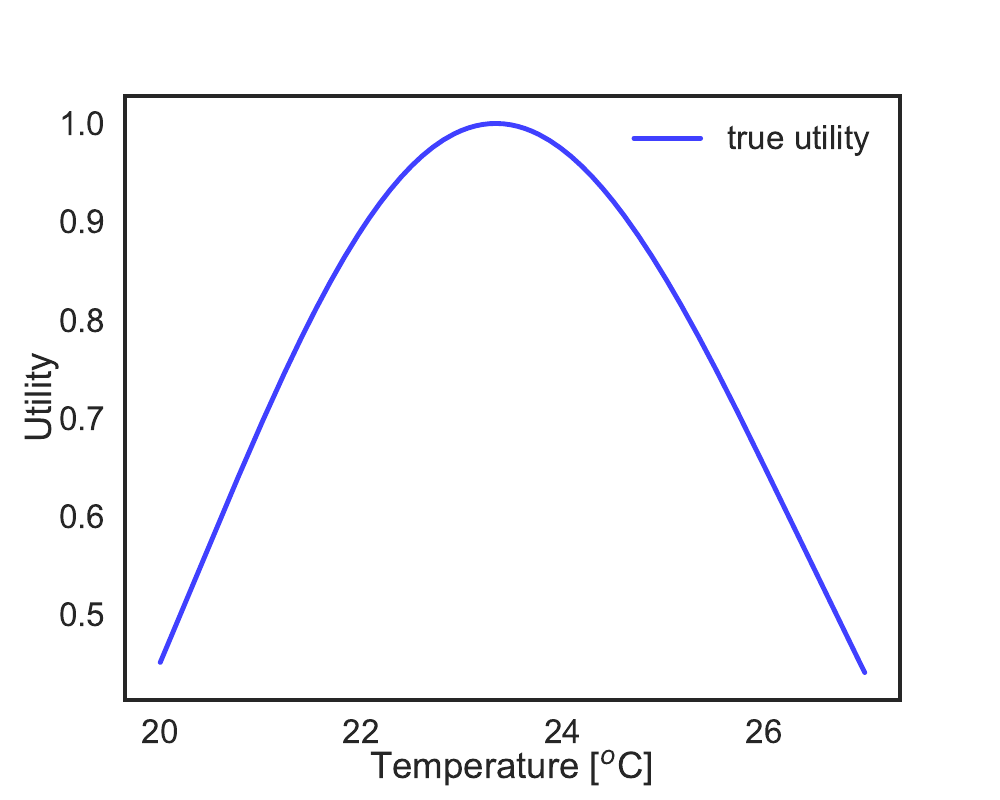}
    \caption{Synthetic Occupant 2}
  \end{subfigure}
  \begin{subfigure}[b]{0.32\linewidth}
    \includegraphics[width=\linewidth]{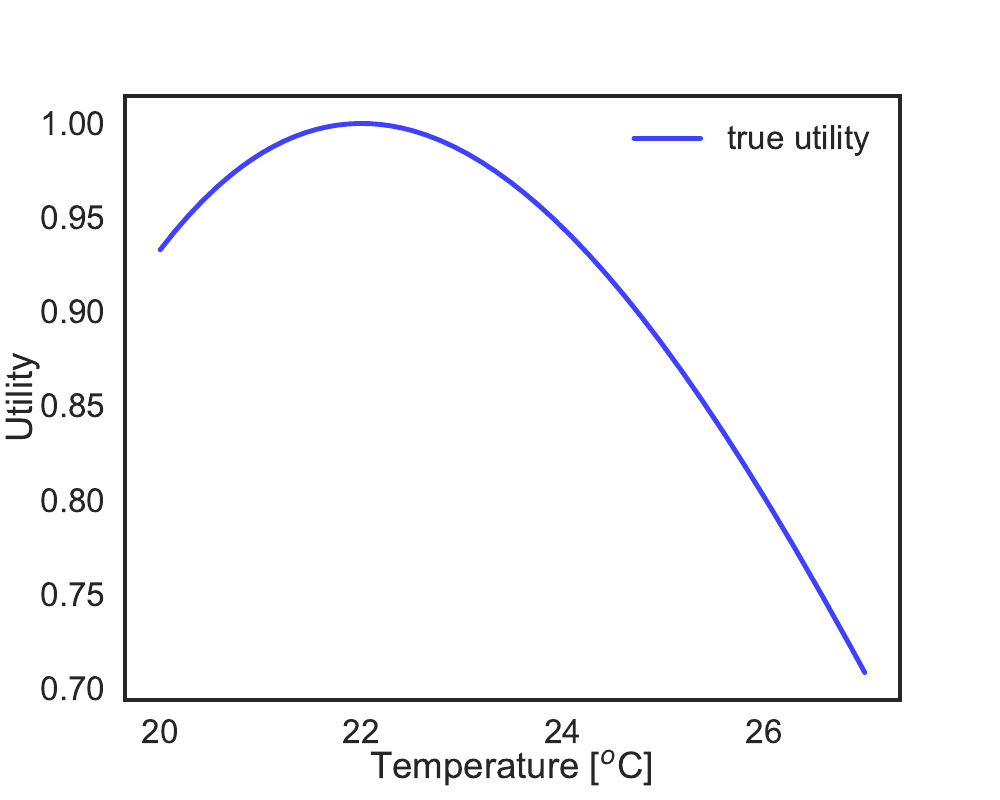}
    \caption{Synthetic Occupant 3}
  \end{subfigure}
  \caption{Utility functions governing the preferences of synthetic occupants}
  \label{fig:syn_occupant_utilities}
\end{figure}

In order to validate that our PE framework can indeed discover the maximally preferred indoor air temperature value, we first run our PE framework using three synthetic occupants' preference data.
For generating a thermal preference dataset of synthetic occupants,we take the following approach:
\begin{enumerate}
    %\item Assume that the thermal state of the room is completely defined by the air temperature feature (1D case).
    \item  We assume that we have synthetic occupants working in private offices whose utilities for different indoor air temperatures are as shown in the \Cref{fig:syn_occupant_utilities}.
    We want to learn the indoor air temperatures which would provide maximum thermal satisfaction to these occupants. 
    For synthetic occupant 1, the maximally preferred indoor air temperature is $\dC{25}$ ;
    for synthetic occupant 2, the maximally preferred indoor air temperature is $\dC{23.34}$ and
    for synthetic occupant 3, the maximally preferred indoor air temperature is $\dC{22.1}$.
    \item  We assume that we can only query thermal preference responses (``How satisfied are you with current thermal condition?'') from these occupants.
    \item  The outcome (response of the synthetic occupant) for a given indoor air temperature is generated as described in sub-sub-section \ref{sec:like}, i.e., the outcome is drawn from the distribution $p(y|x, u(\cdot)) \propto \Phi(\nu_lyu'(x))$, where the response $y$ takes values $-1$ (prefer cooler), $0$ (satisfied) and $-1$ (prefer warmer).
    We start with an initial indoor air temperature of $\dC{21}$ and elicit thermal query related responses for all the three occupants. 
    At this thermal indoor air temperature, each of the three synthetic occupants prefer warmer temperatures $(y =1)$.
\end{enumerate}

\subsection{The framework in action}\label{subsec:framework_in_action}
\begin{figure}[H]
\centering
\begin{tabular}{ccc}

\includegraphics[width=50mm]{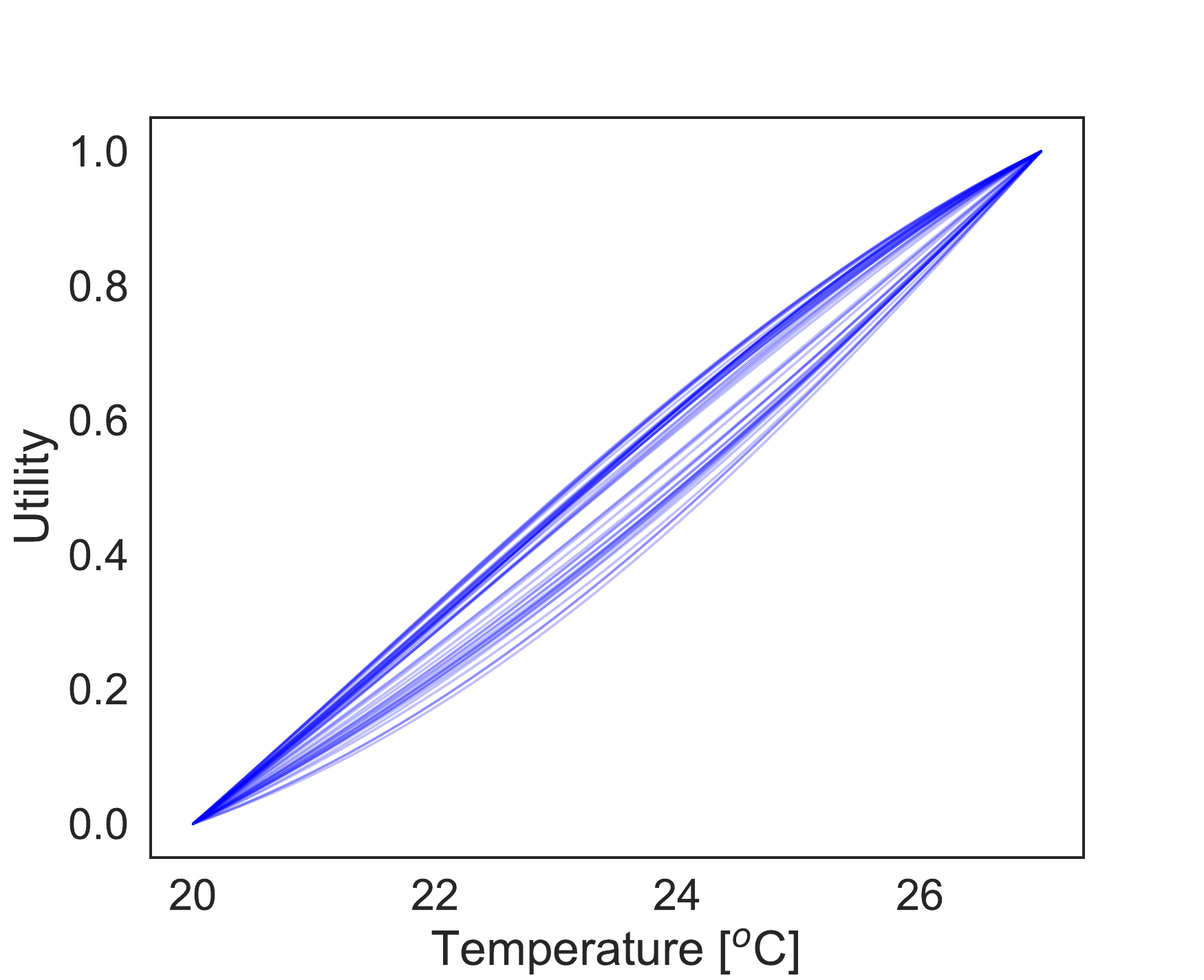} &  
\includegraphics[width=50mm]{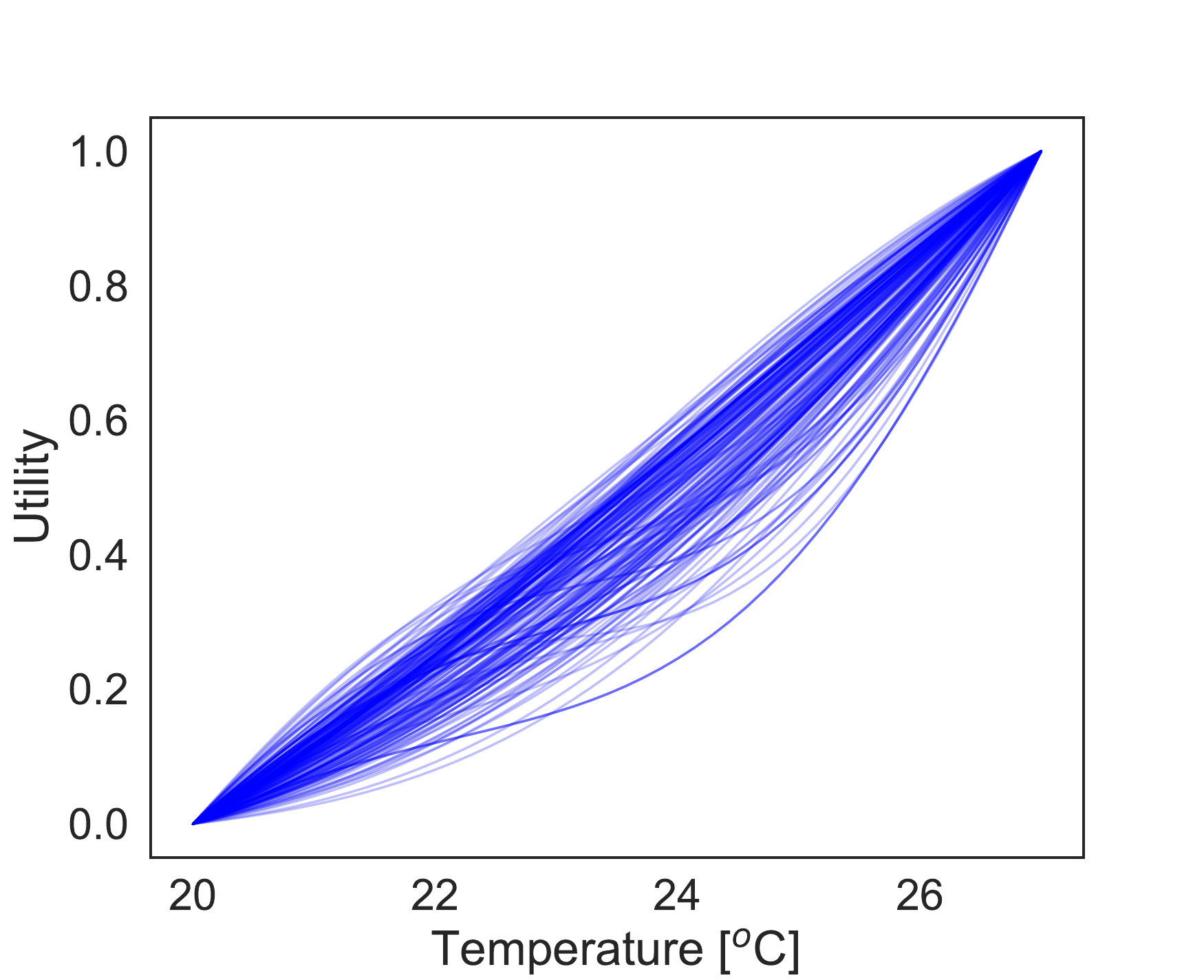} \\
(a) utility samples ($N = 2$) &
(b) utility samples ($N = 2$) \\
\includegraphics[width=50mm]{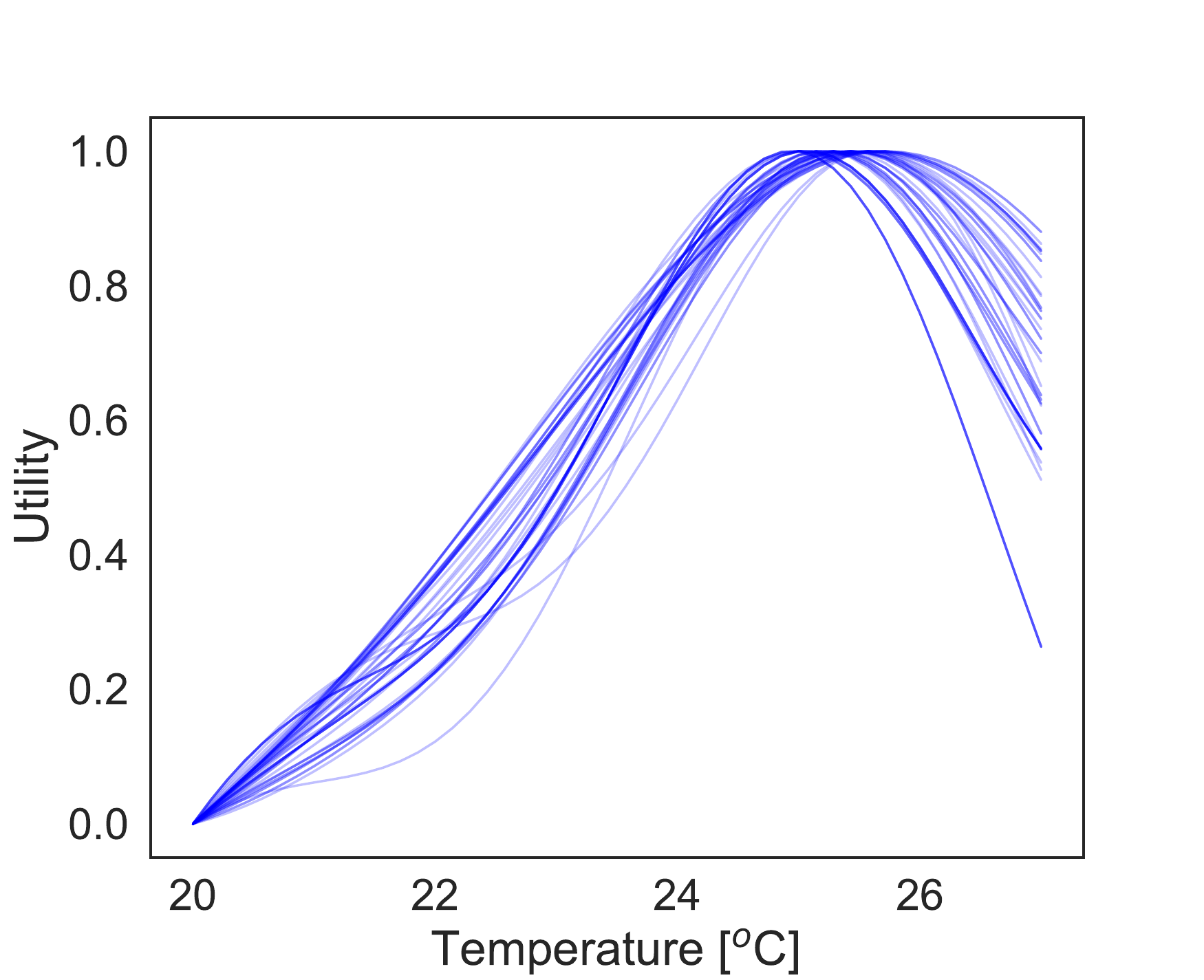} &  
\includegraphics[width=50mm]{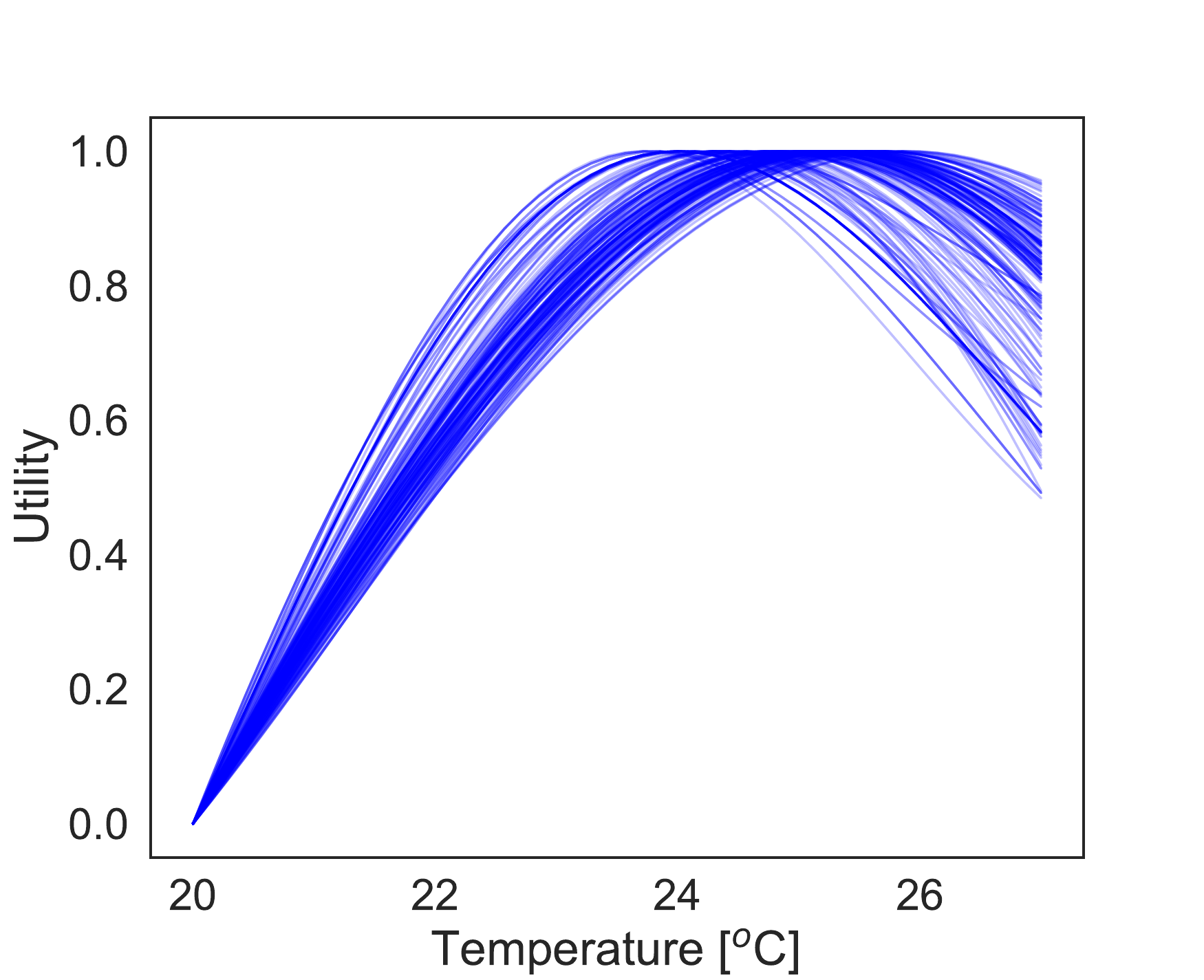} &
\includegraphics[width=50mm]{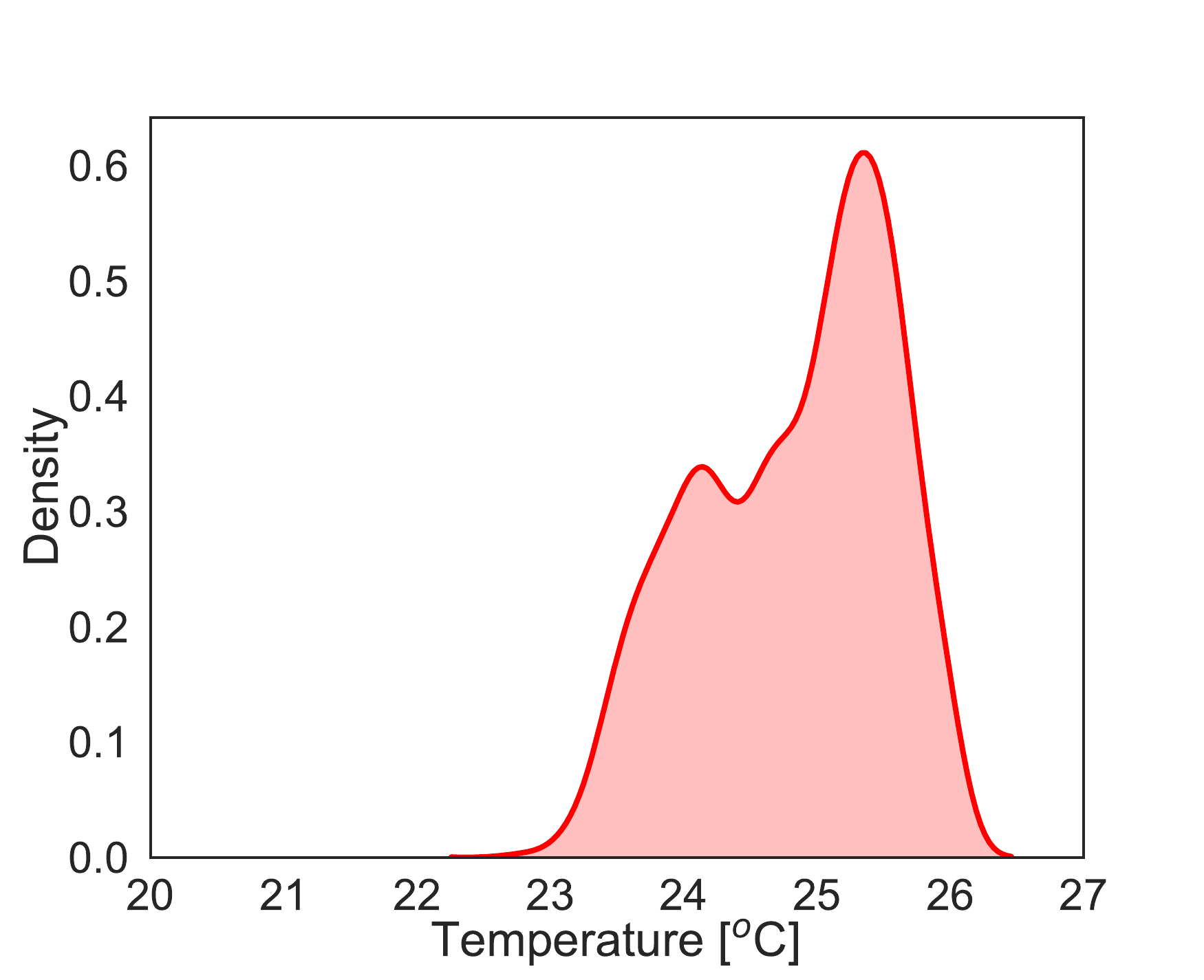} \\
(c) utility samples ($N = 3$) &
(d) utility samples ($N = 3$) &
(e) max. preferred temp. ($N = 3$)\\

\includegraphics[width=50mm]{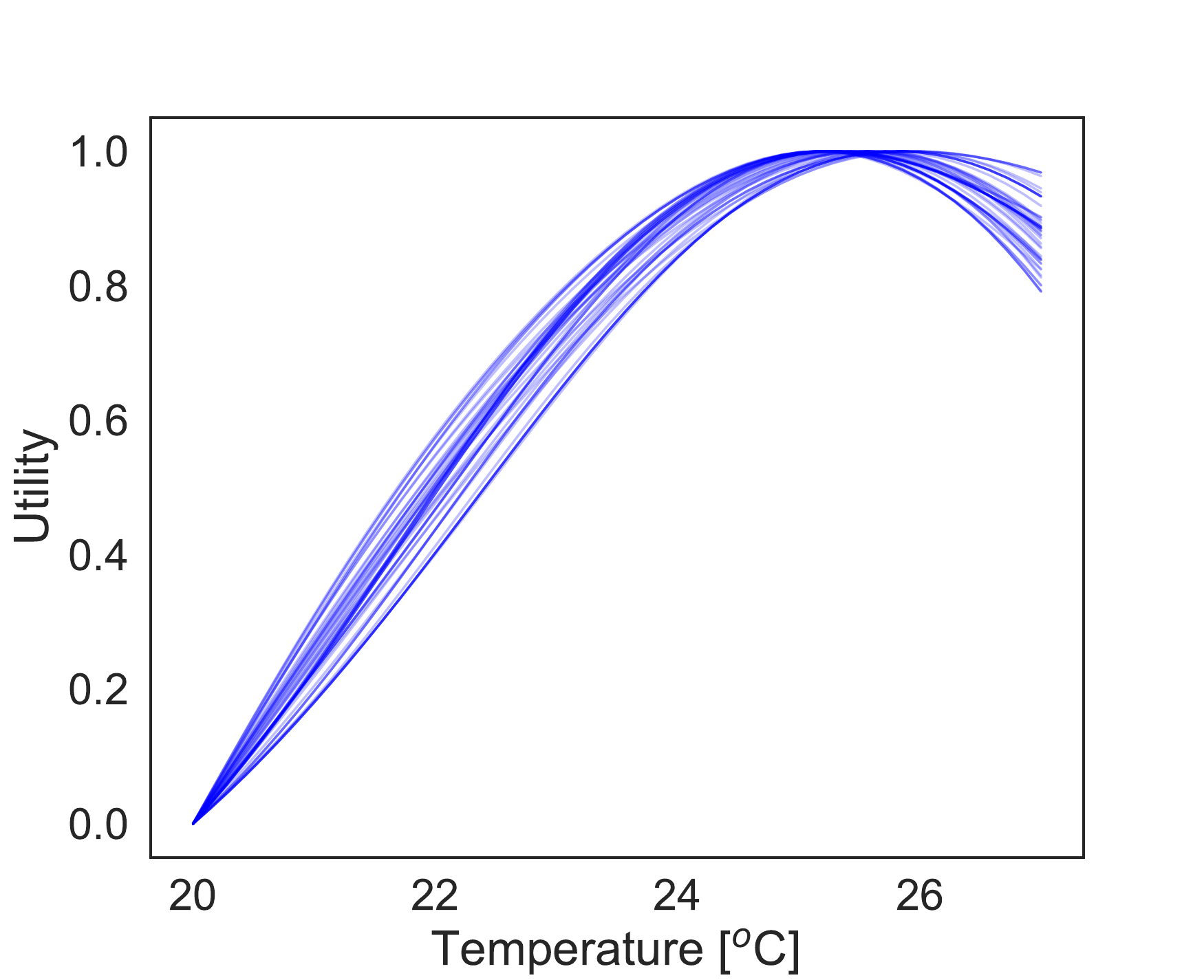} &  
\includegraphics[width=50mm]{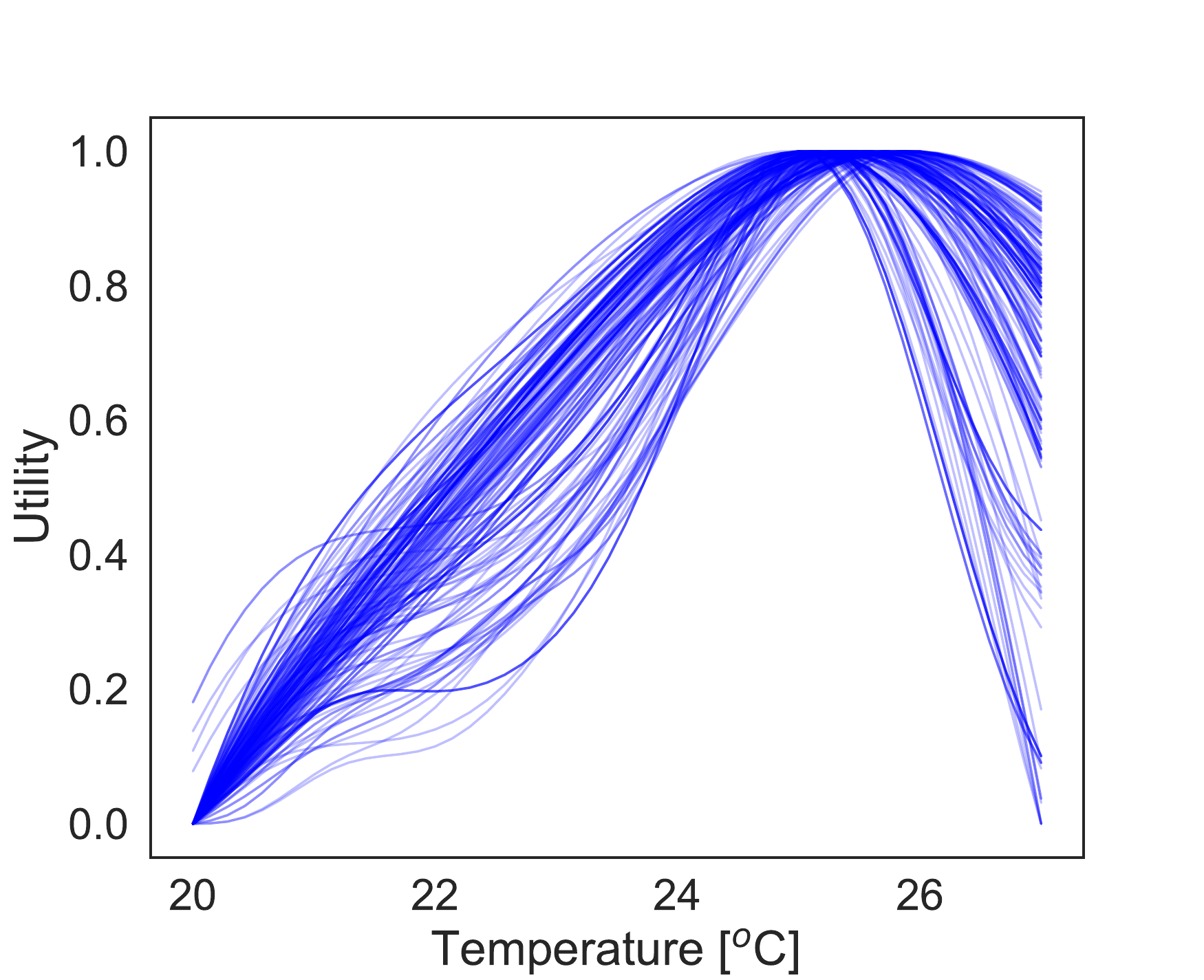} &
\includegraphics[width=50mm]{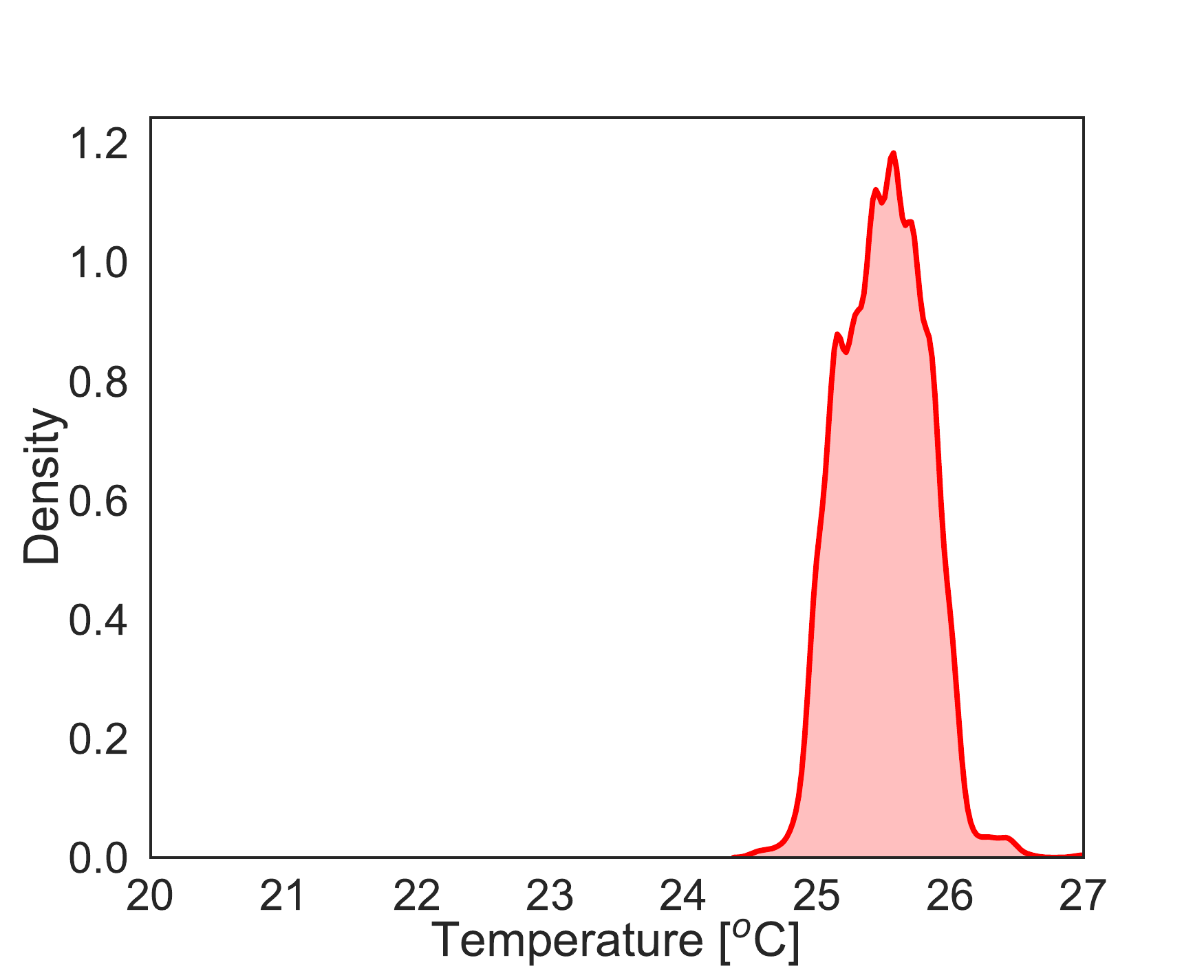} \\
(f) utility samples ($N = 4$) &
(g) utility samples ($N = 4$) &
(h) max. preferred temp. ($N = 4$) \\
\includegraphics[width=50mm]{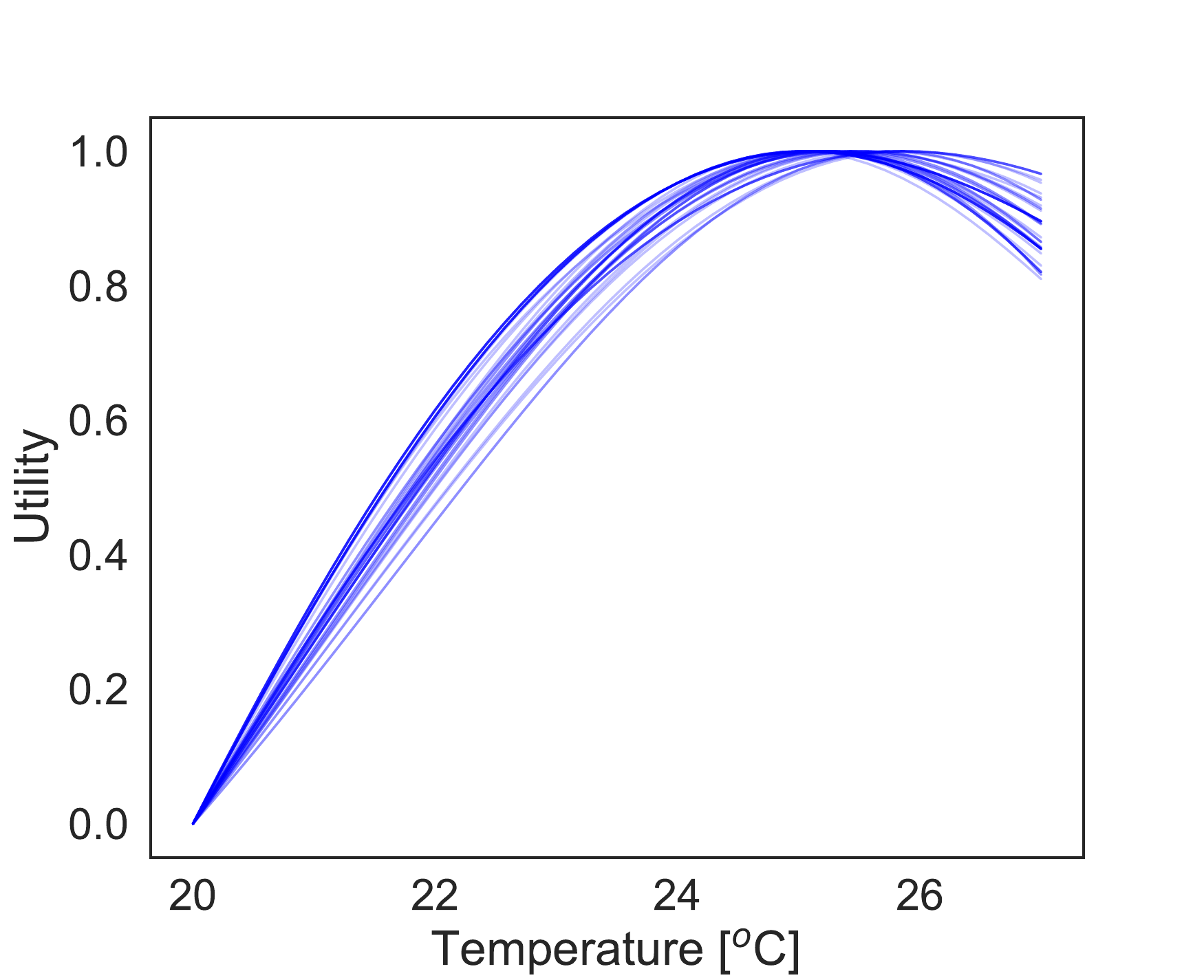} &  
\includegraphics[width=50mm]{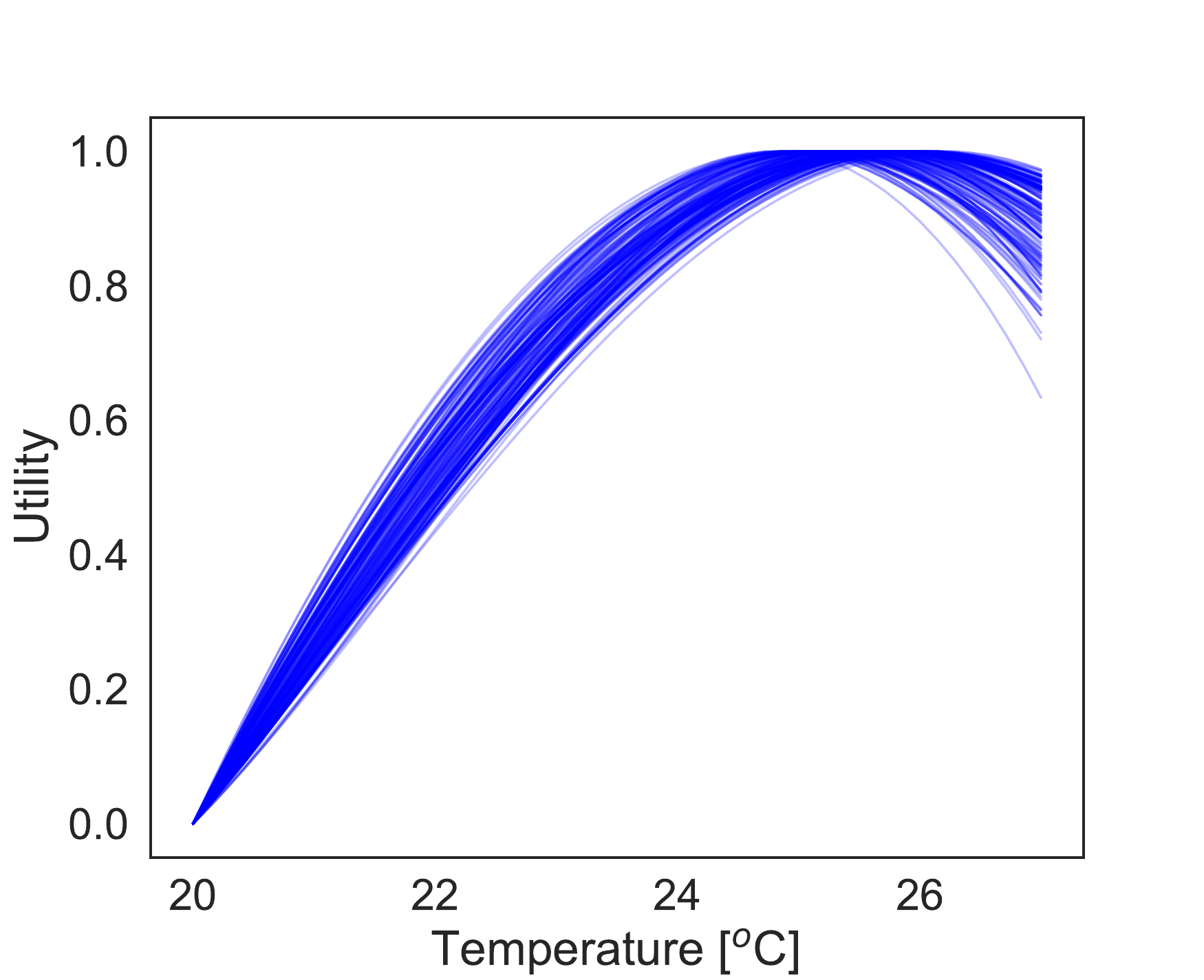} &
\includegraphics[width=50mm]{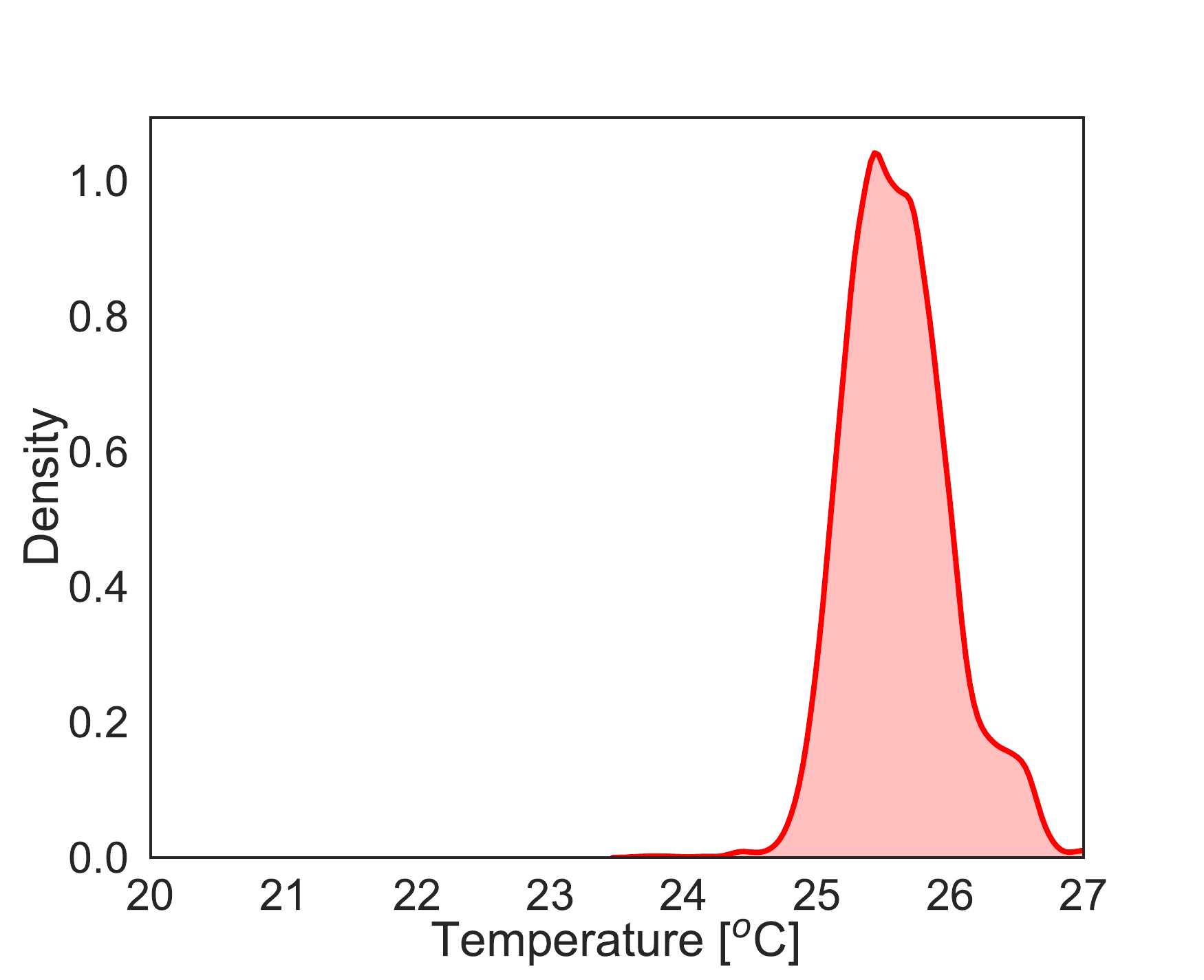} \\
(i) utility samples ($N = 5$) &
(j) utility samples ($N = 5$) &
(k) max. preferred temp. ($N = 5$) \\
\includegraphics[width=50mm]{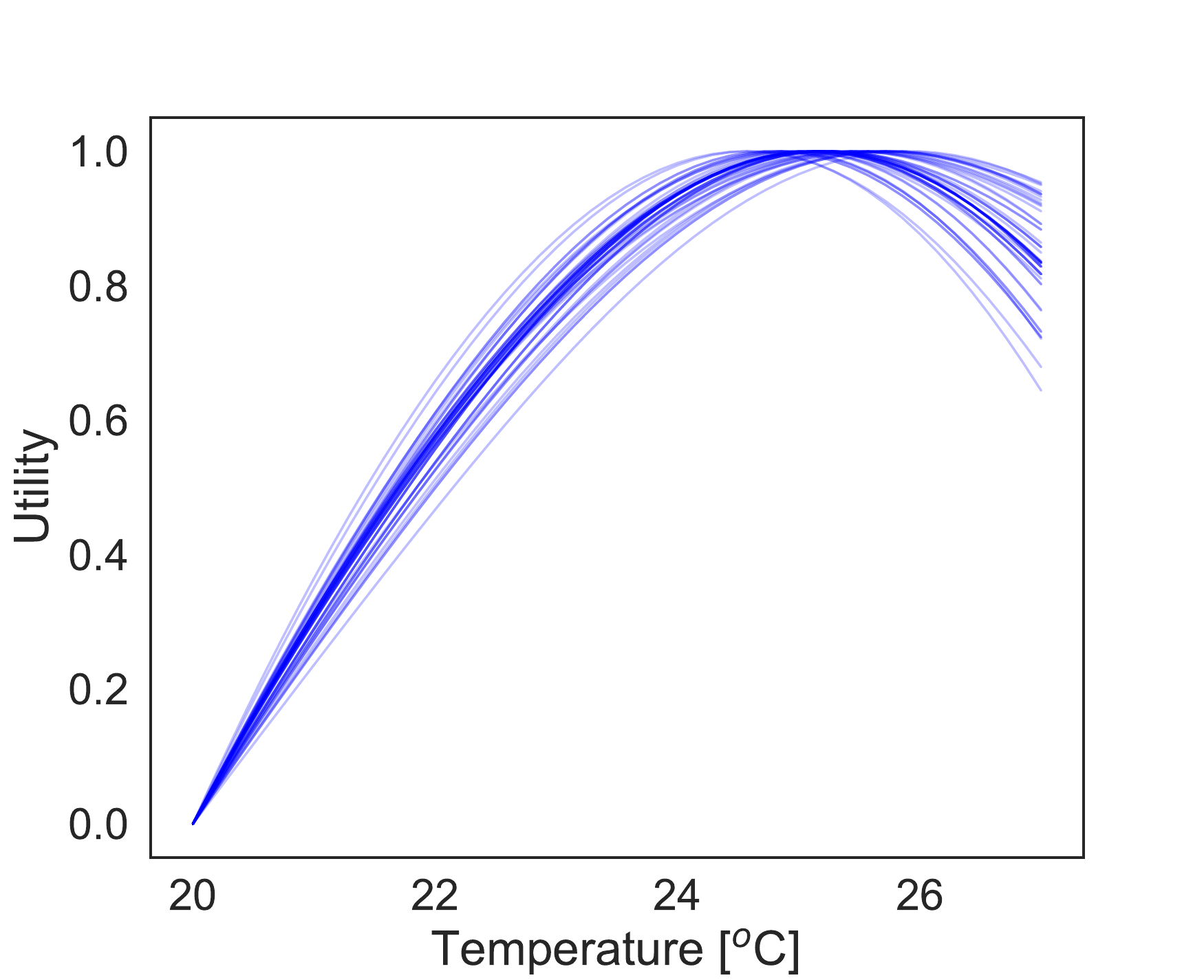} &  
\includegraphics[width=50mm]{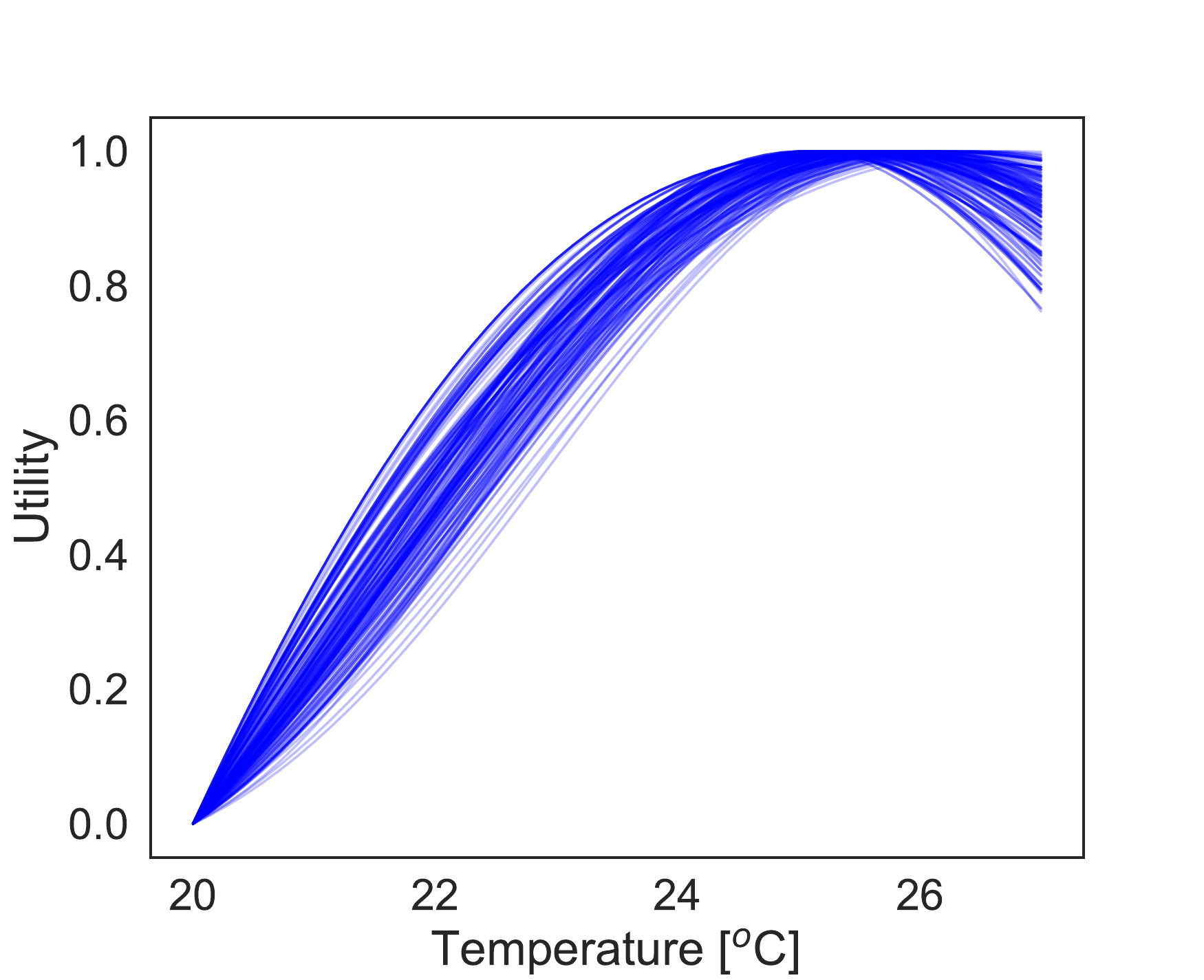} &
\includegraphics[width=50mm]{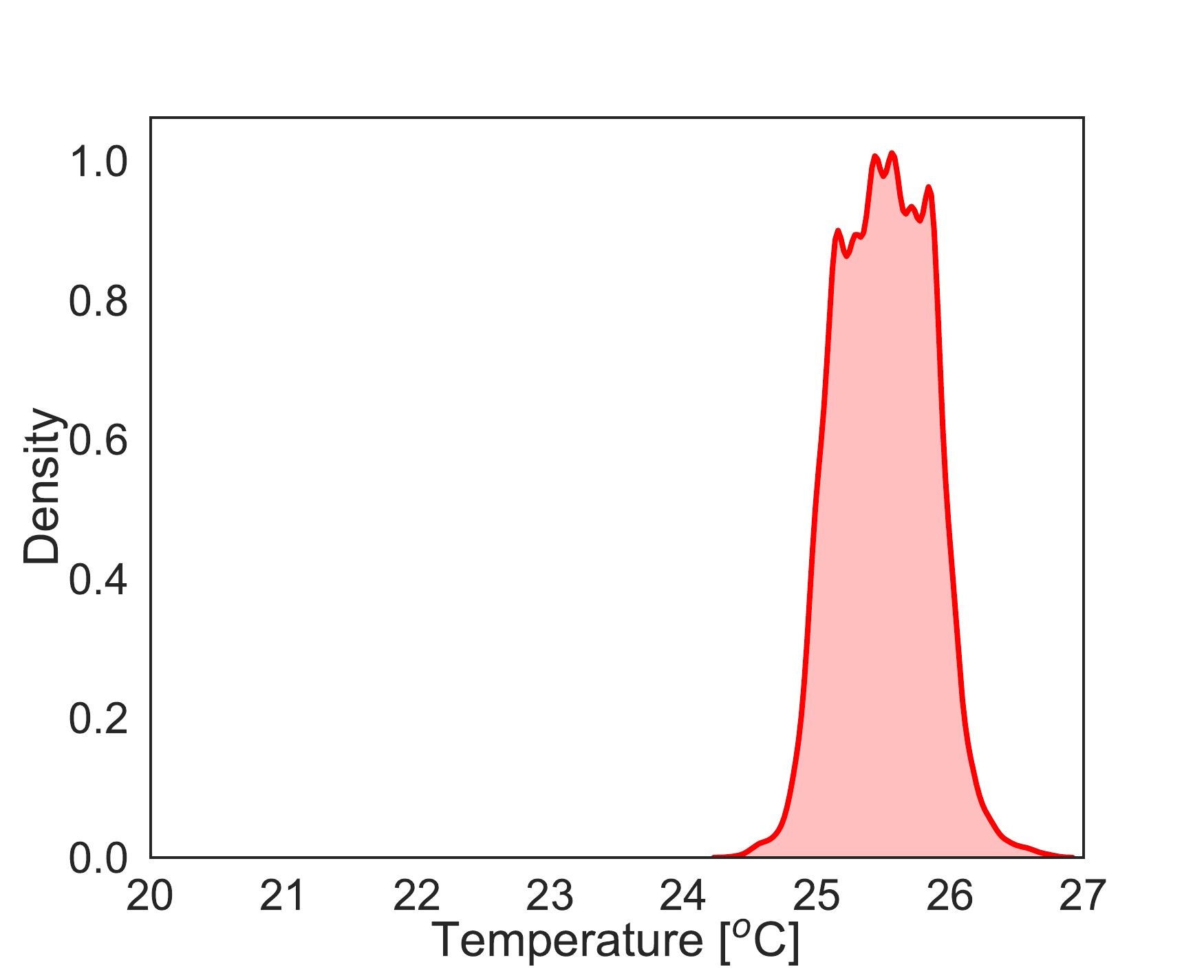} \\
(l) utility samples ($N = 6$) &
(m) utility samples ($N = 6$) &
(n) max. preferred temp. ($N = 6$) \\
\end{tabular}
\caption{Posterior predictions for synthetic occupant 1. \textbf{First two columns:} Samples from posterior predictive distribution over normalized utility function. \textbf{Last column:}.
Predictive distribution over maximally preferred indoor air temperature values.}
\label{fig:synthetic_occupant1_pp}
\end{figure}

\begin{figure}[H]
\centering
\begin{tabular}{ccc}
\includegraphics[width=50mm]{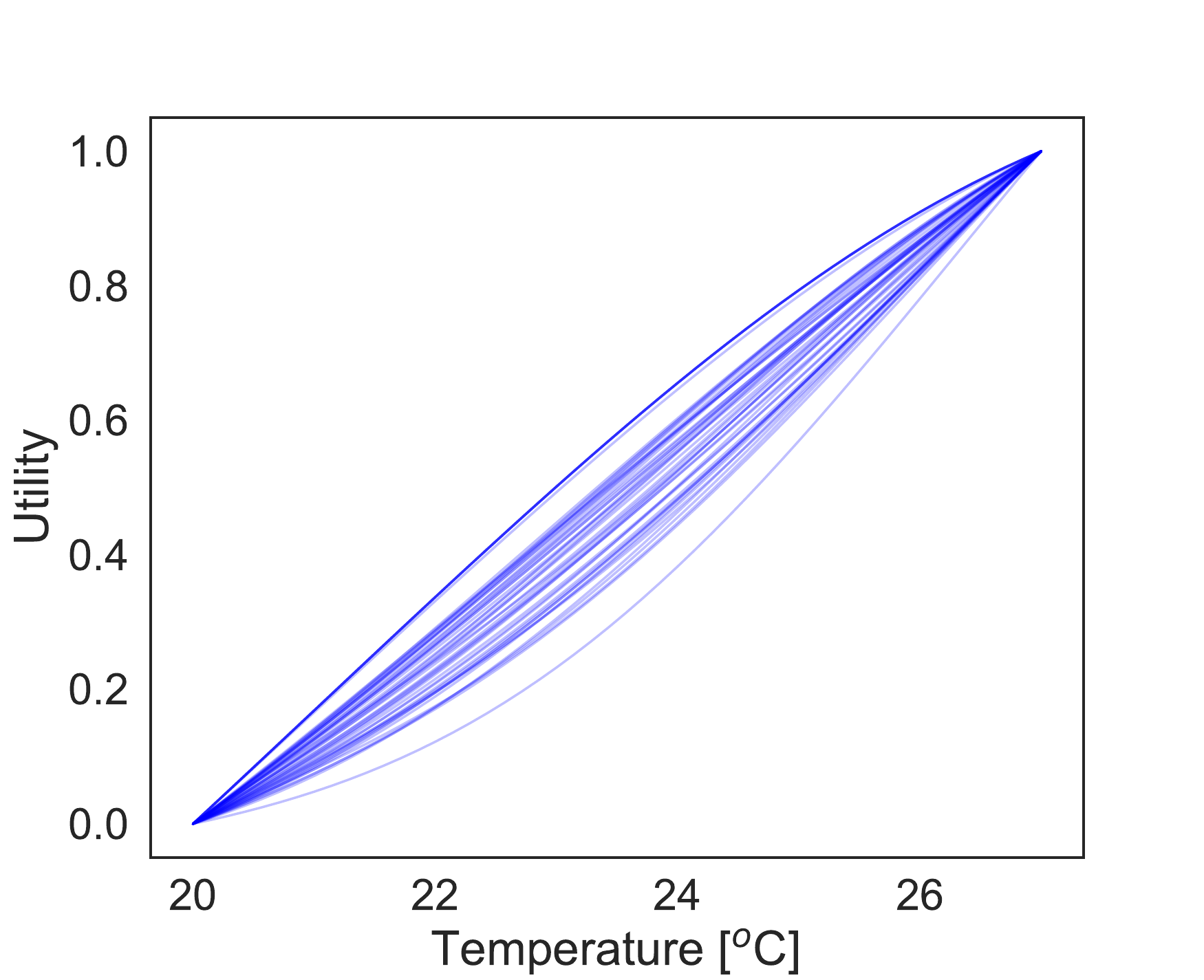} &  
\includegraphics[width=50mm]{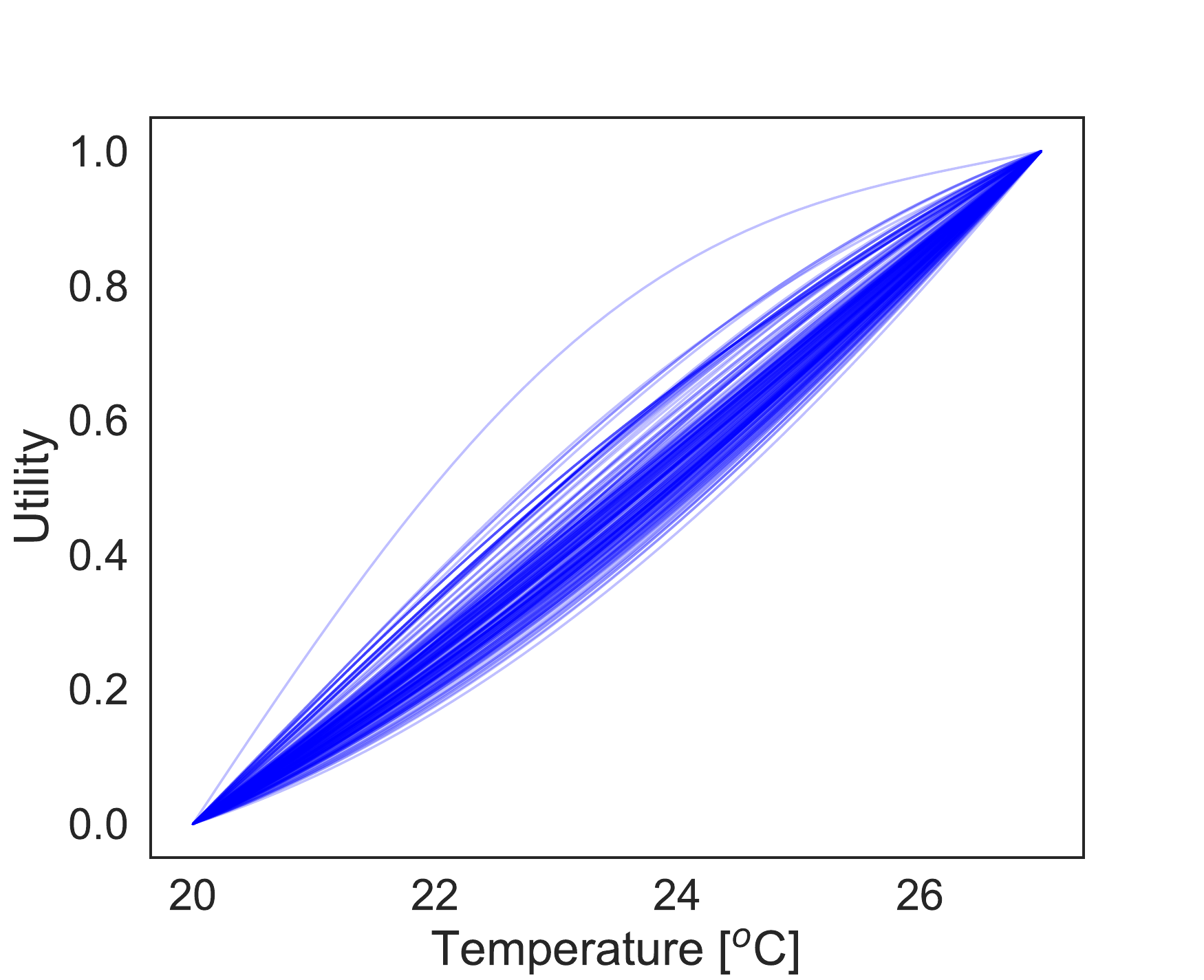} \\
(a) utility samples ($N = 1$) &
(b) utility samples ($N = 1$) \\
\includegraphics[width=50mm]{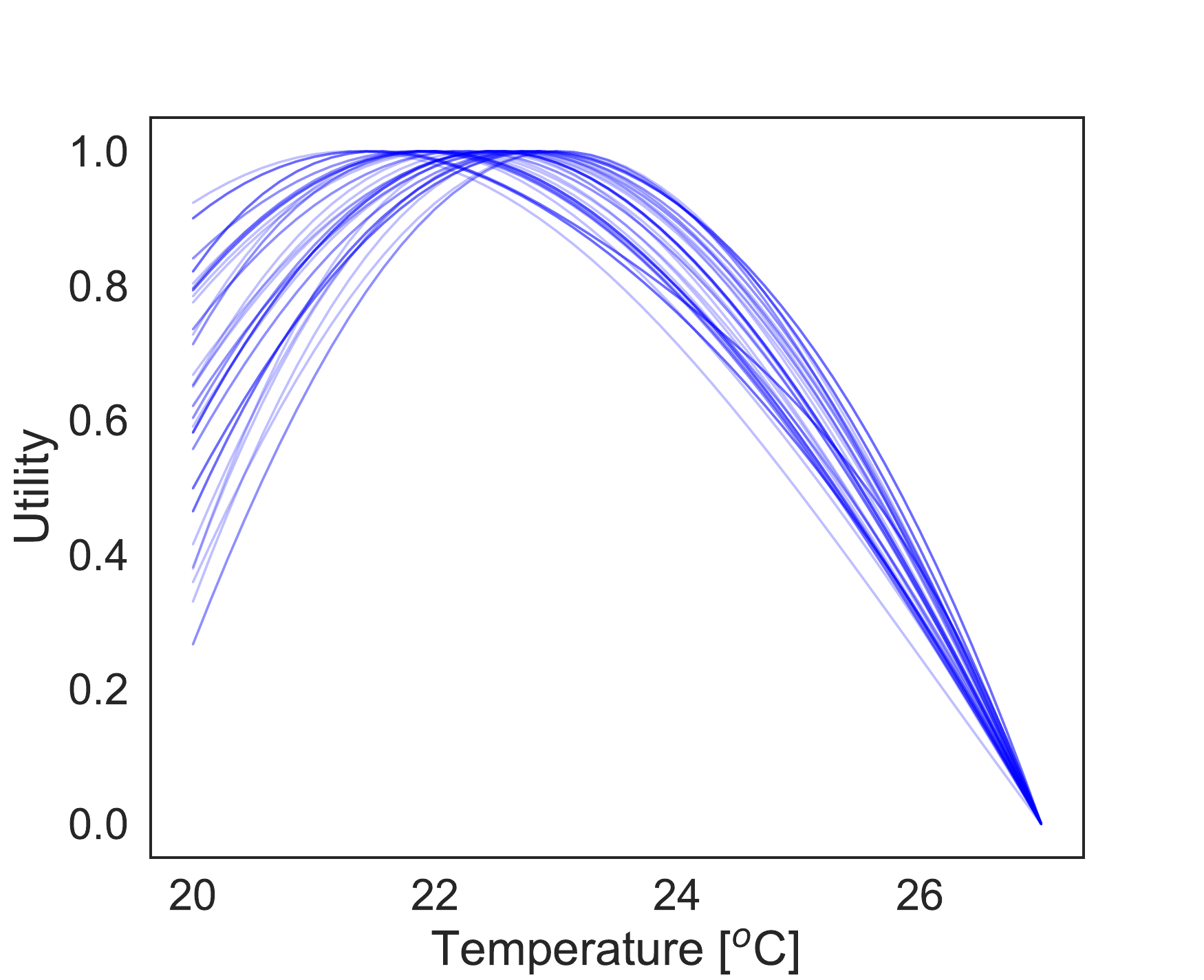} &  
\includegraphics[width=50mm]{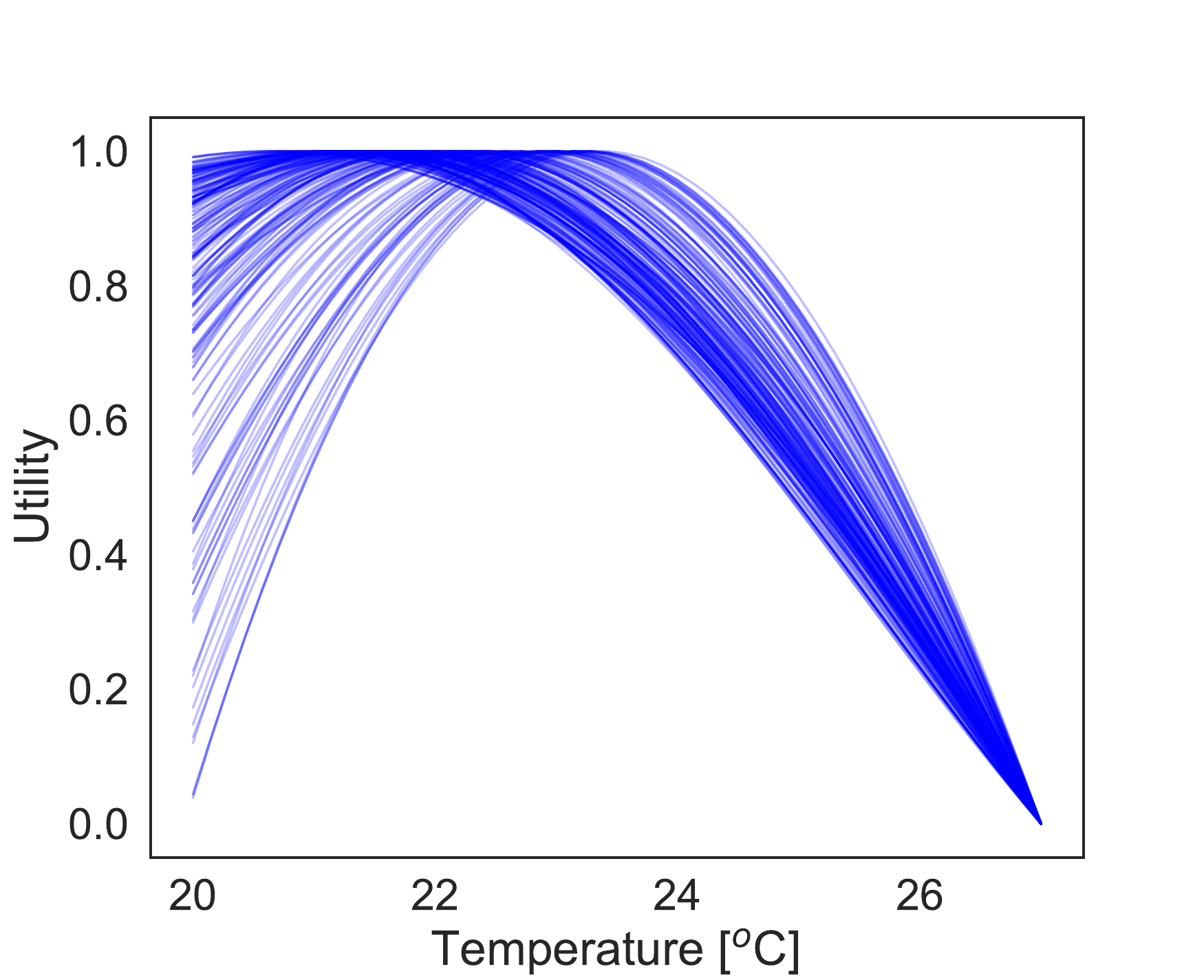} &
\includegraphics[width=50mm]{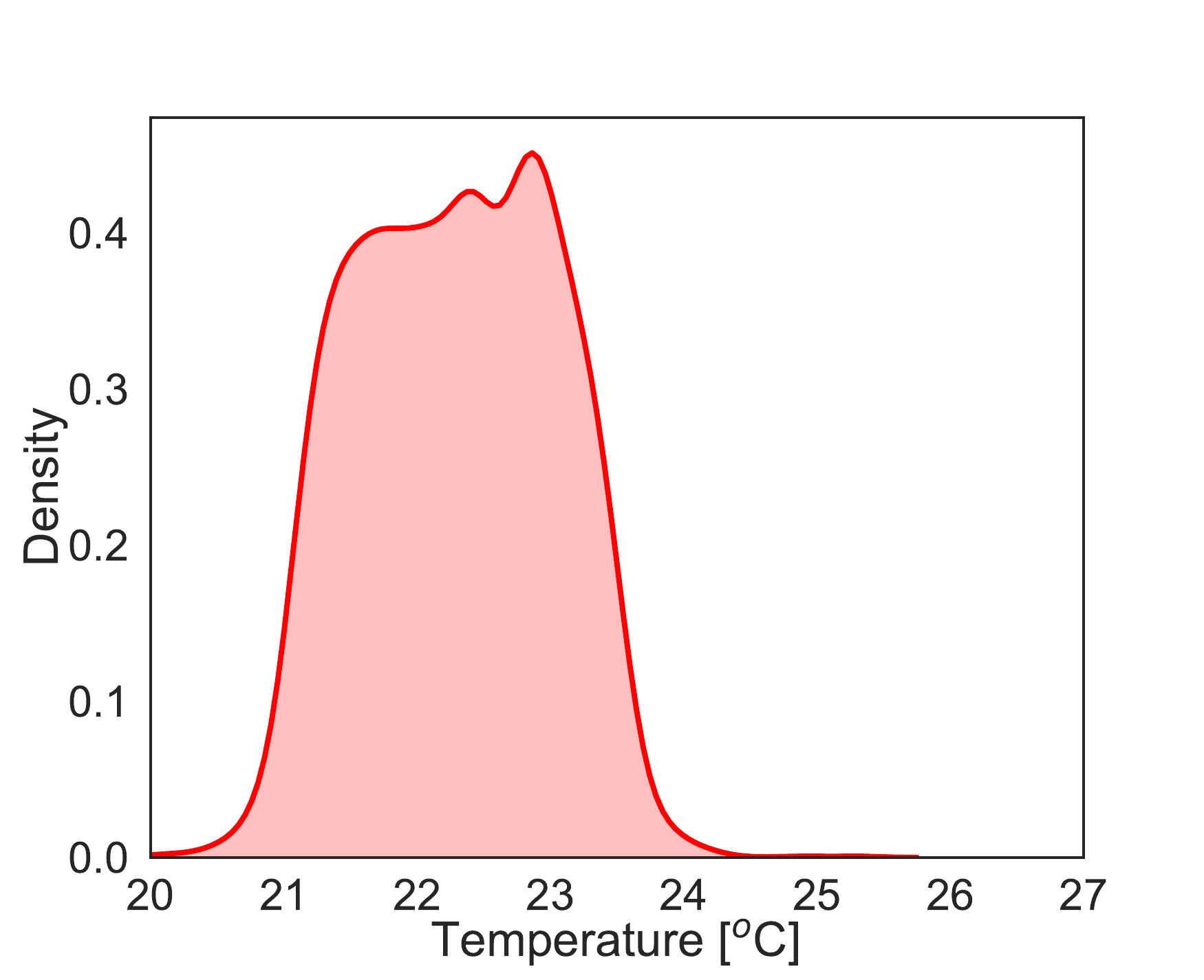} \\
(c) utility samples ($N = 2$) &
(d) utility samples ($N = 2$) &
(e) maximally preferred temp. ($N = 2$) \\
\includegraphics[width=50mm]{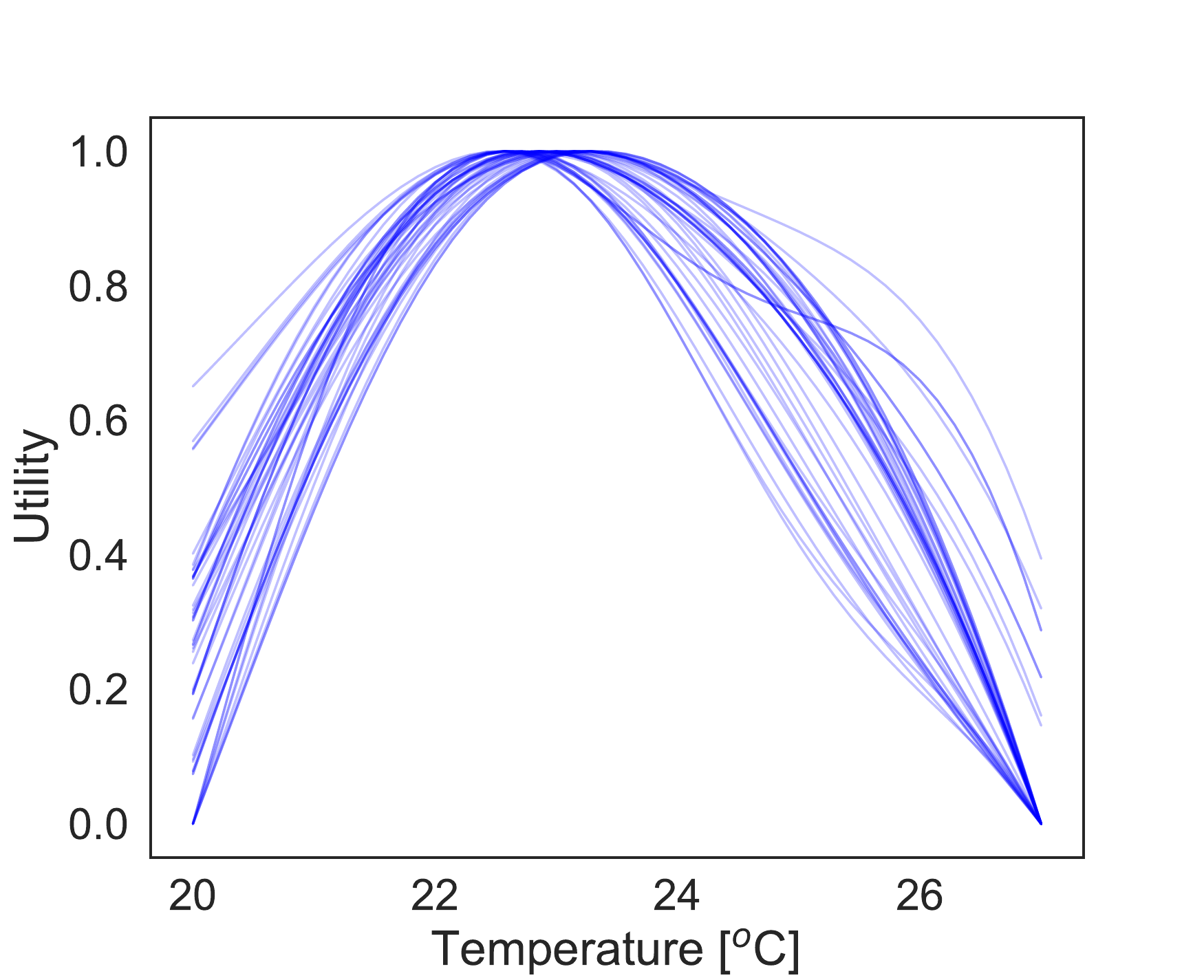} &  
\includegraphics[width=50mm]{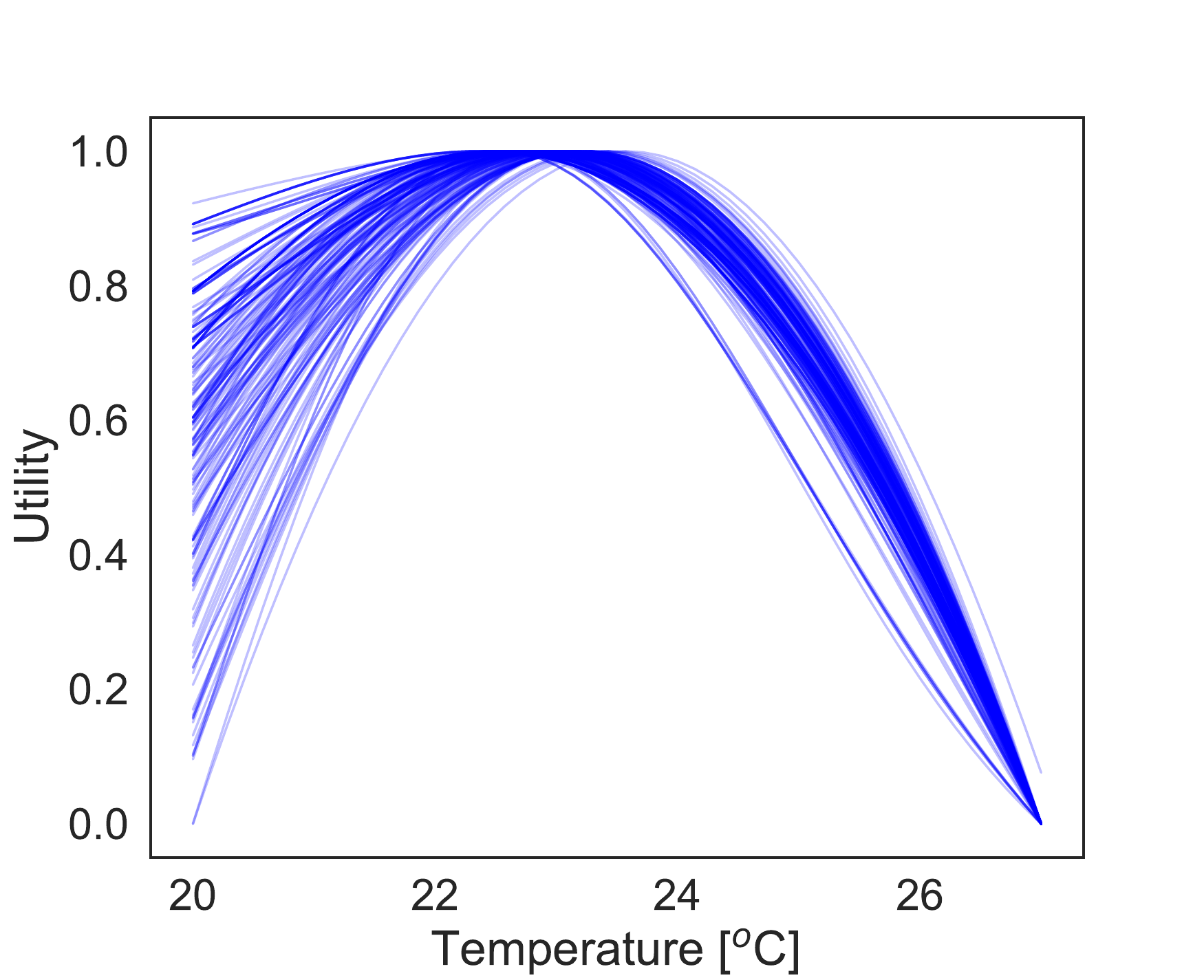} &
\includegraphics[width=50mm]{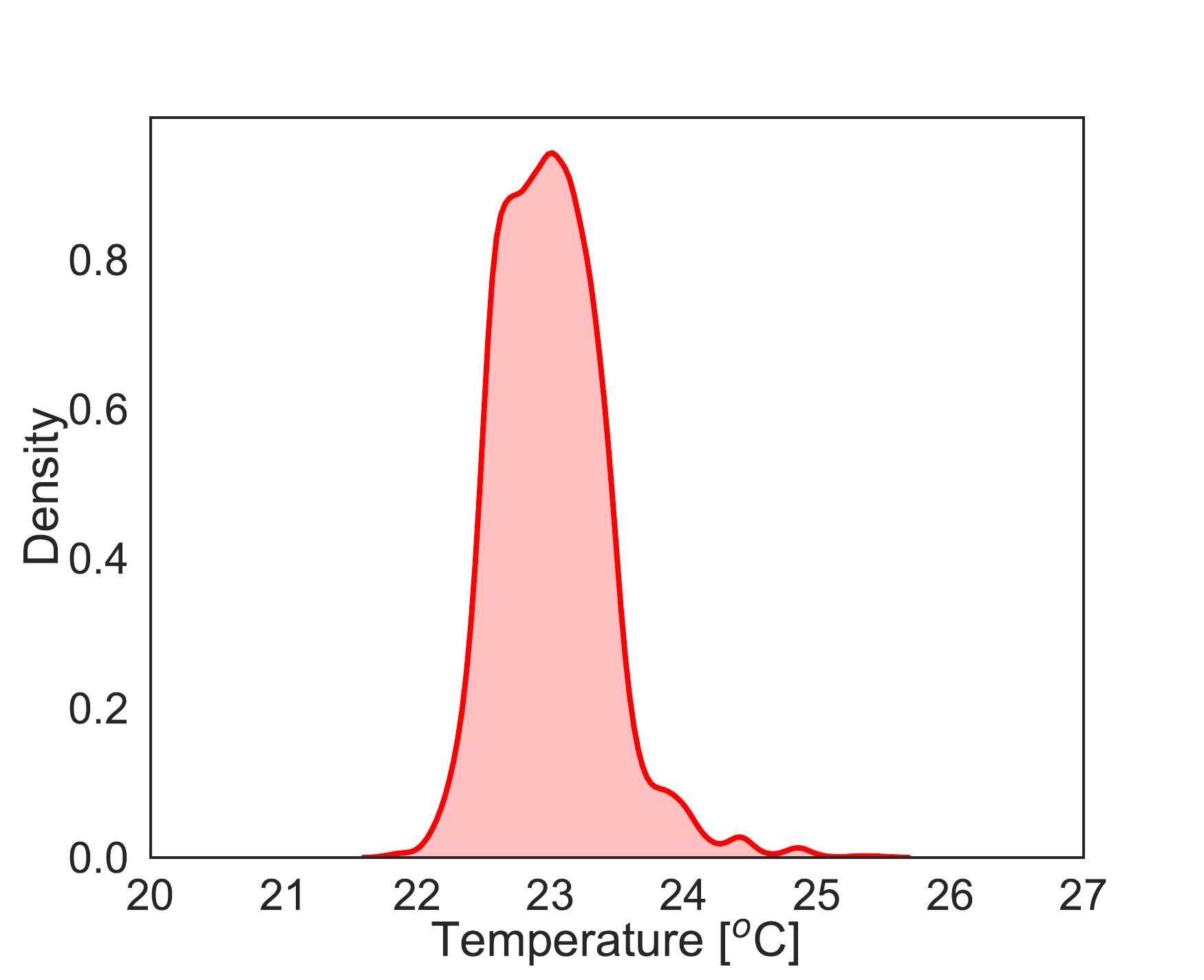} \\
(f) utility samples ($N = 3$) &
(g) utility samples ($N = 3$) &
(h) maximally preferred temp. ($N = 3$) \\
\includegraphics[width=50mm]{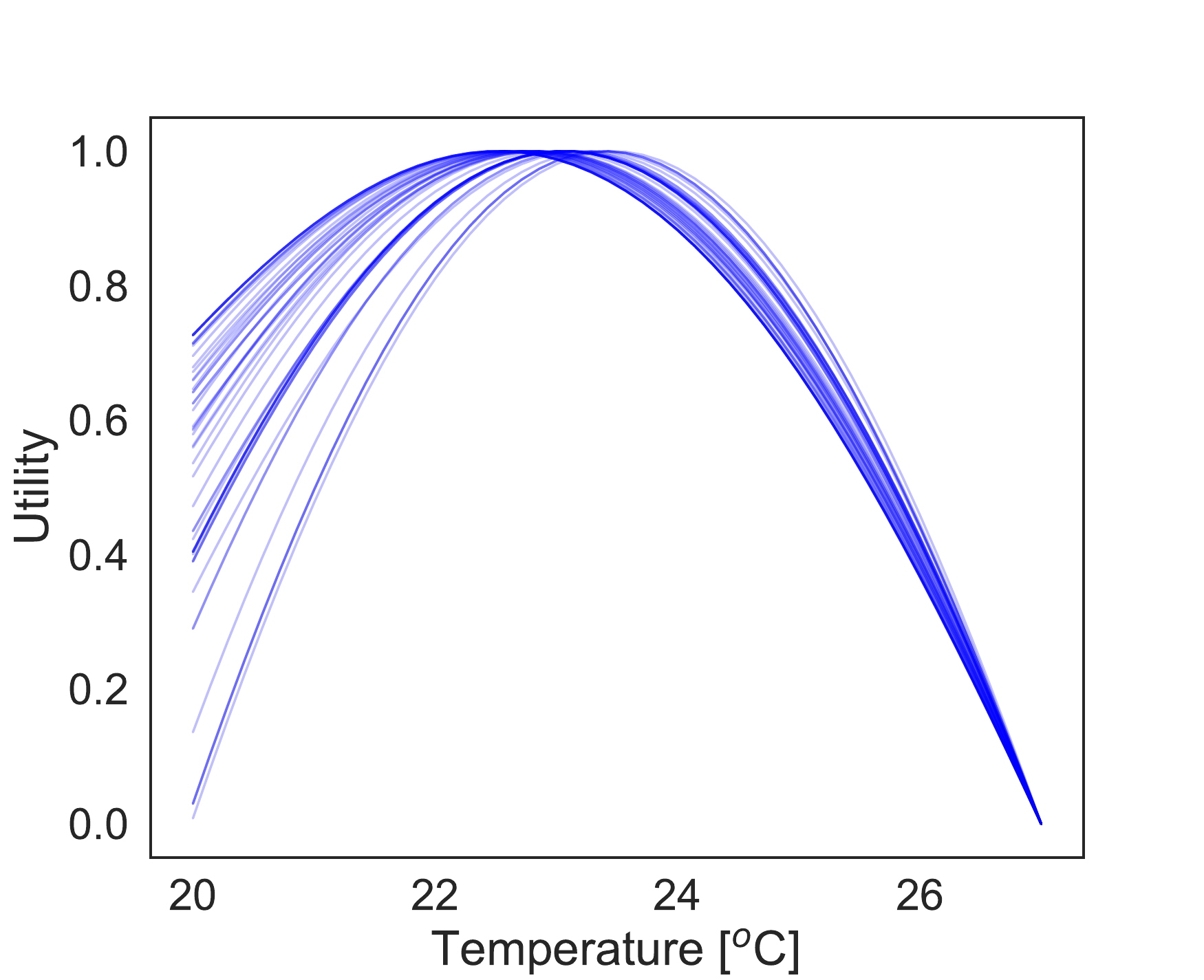} &  
\includegraphics[width=50mm]{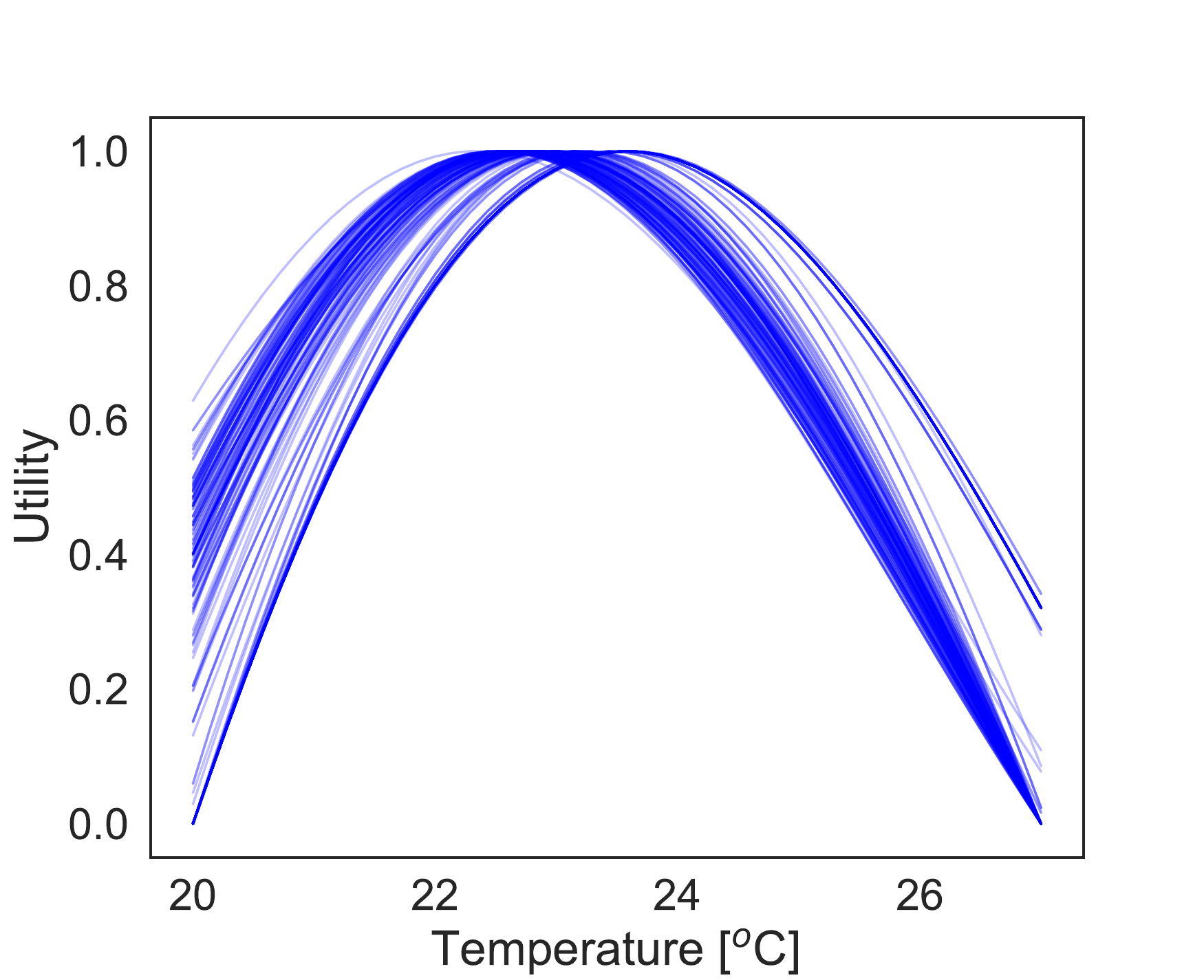} &
\includegraphics[width=50mm]{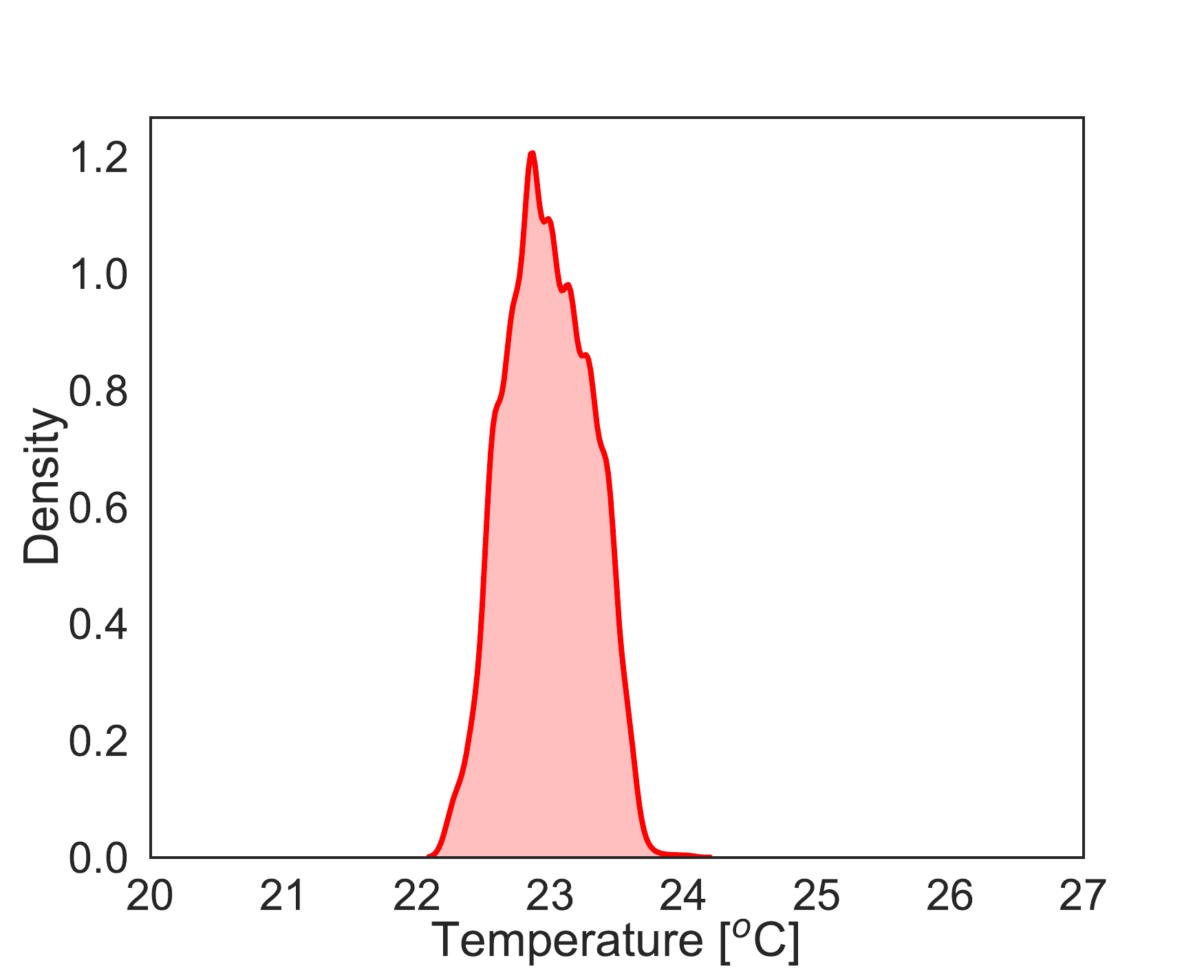} \\
(i) utility samples ($N = 4$) &
(j) utility samples ($N = 4$) &
(k) maximally preferred temp. ($N = 4$) \\
\end{tabular}
\caption{Posterior predictions for syntheic occupant 2. \textbf{First two columns:} Samples from posterior predictive distribution over normalized utility function. \textbf{Last column:} Posterior predictive distribution over maximally preferred indoor air temperature values.}
\label{fig:synthetic_occupant2_pp}
\end{figure}

\begin{figure}[H]
\centering
\begin{tabular}{ccc}
\includegraphics[width=50mm]{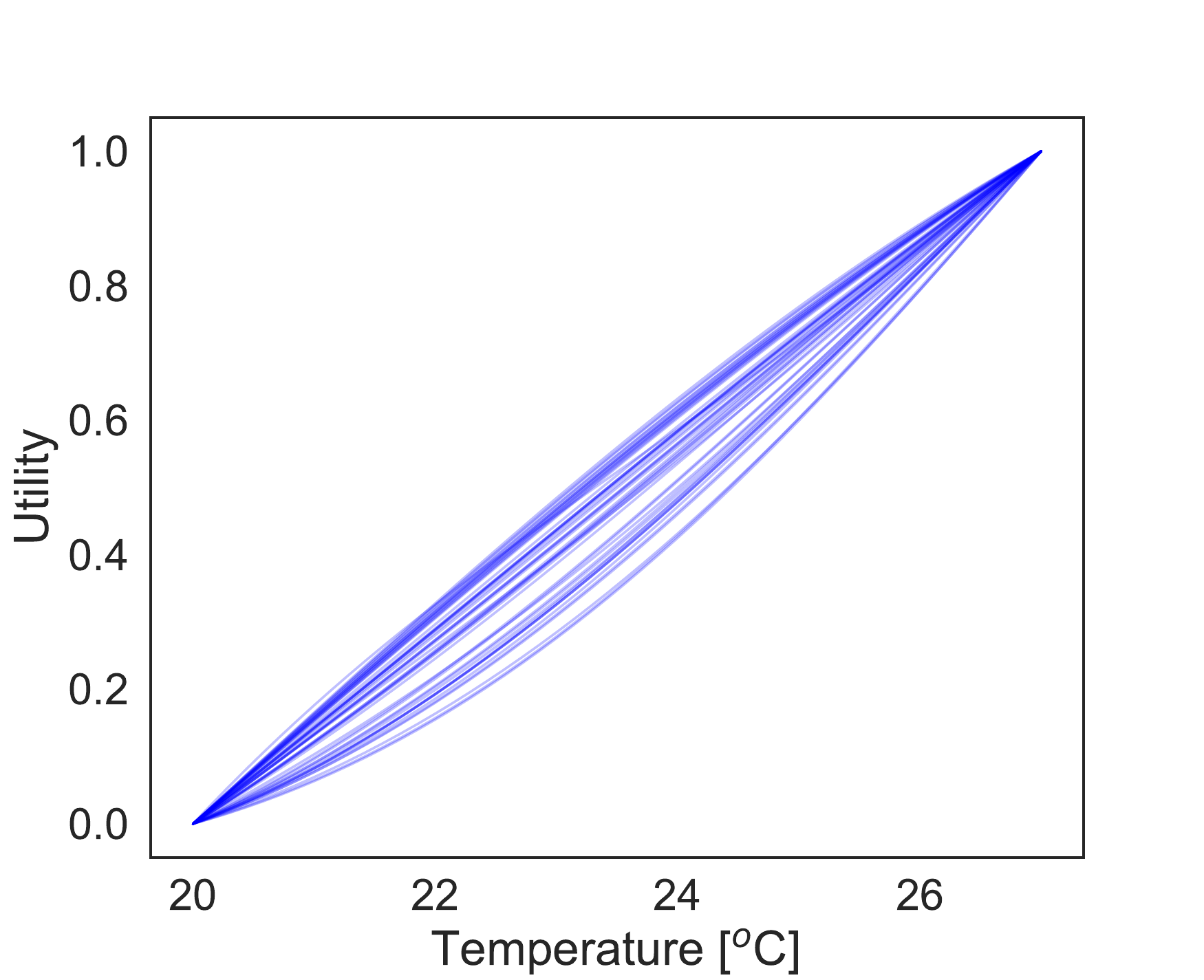} &  
\includegraphics[width=50mm]{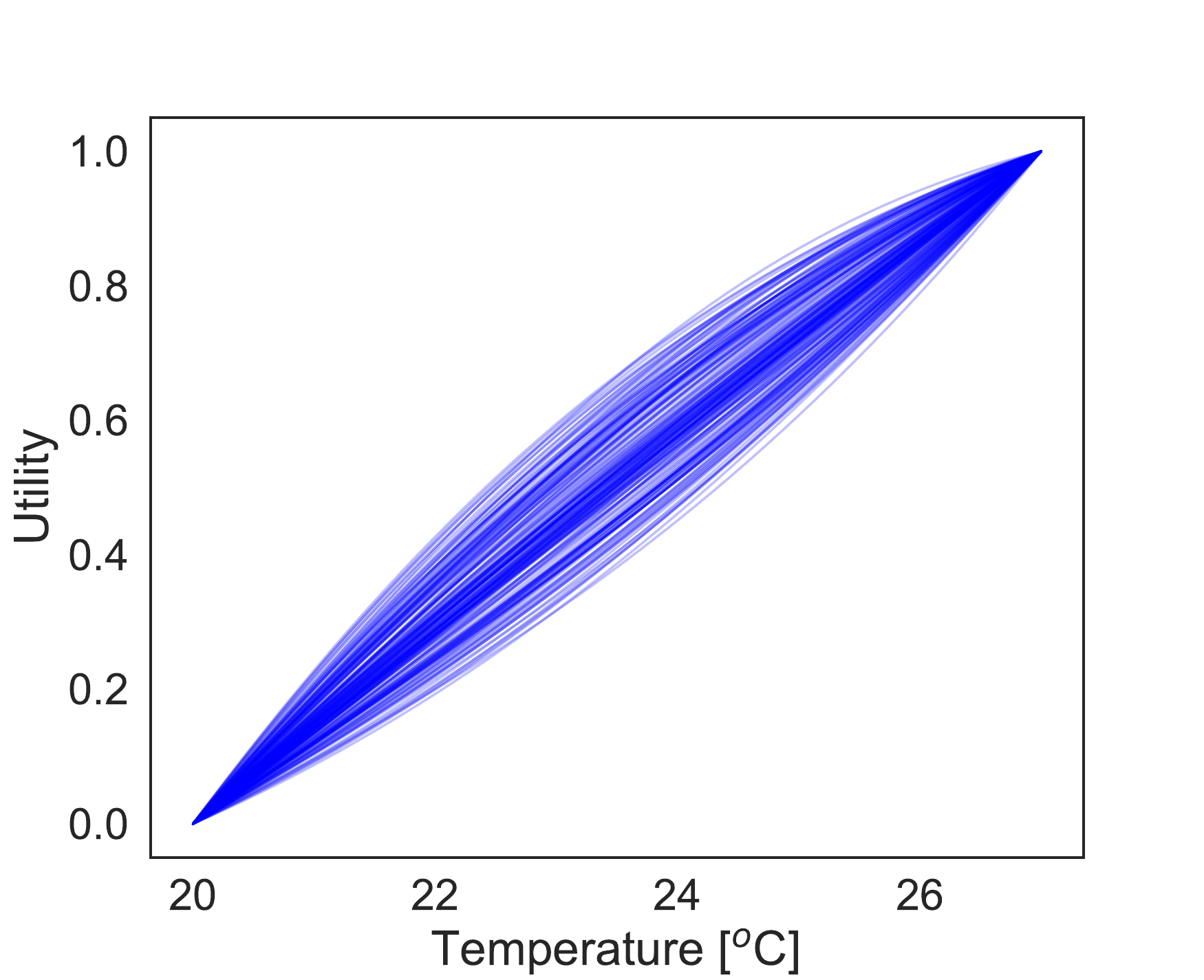} \\
(a) utility samples ($N = 1$) &
(b) utility samples ($N = 1$) \\
\includegraphics[width=50mm]{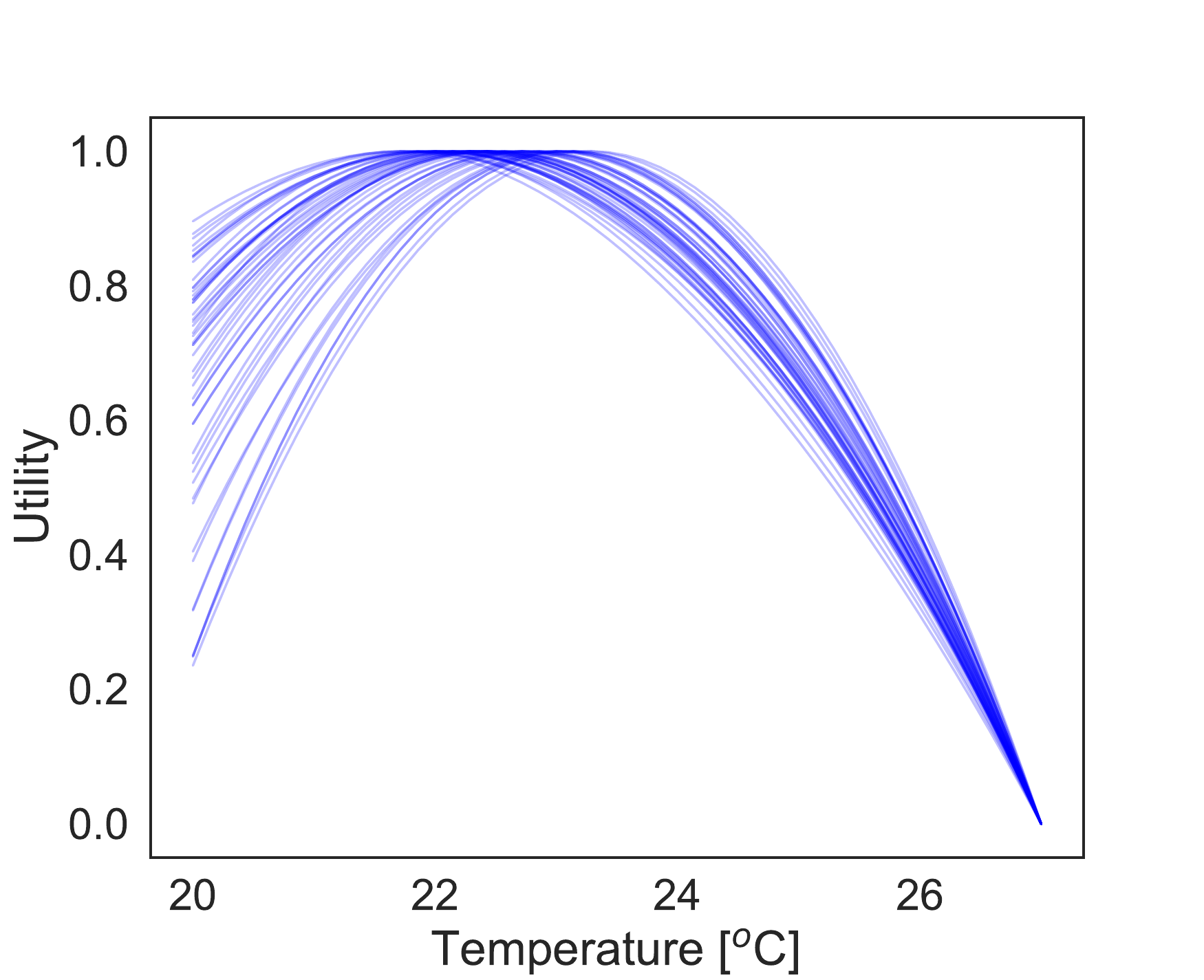} &  
\includegraphics[width=50mm]{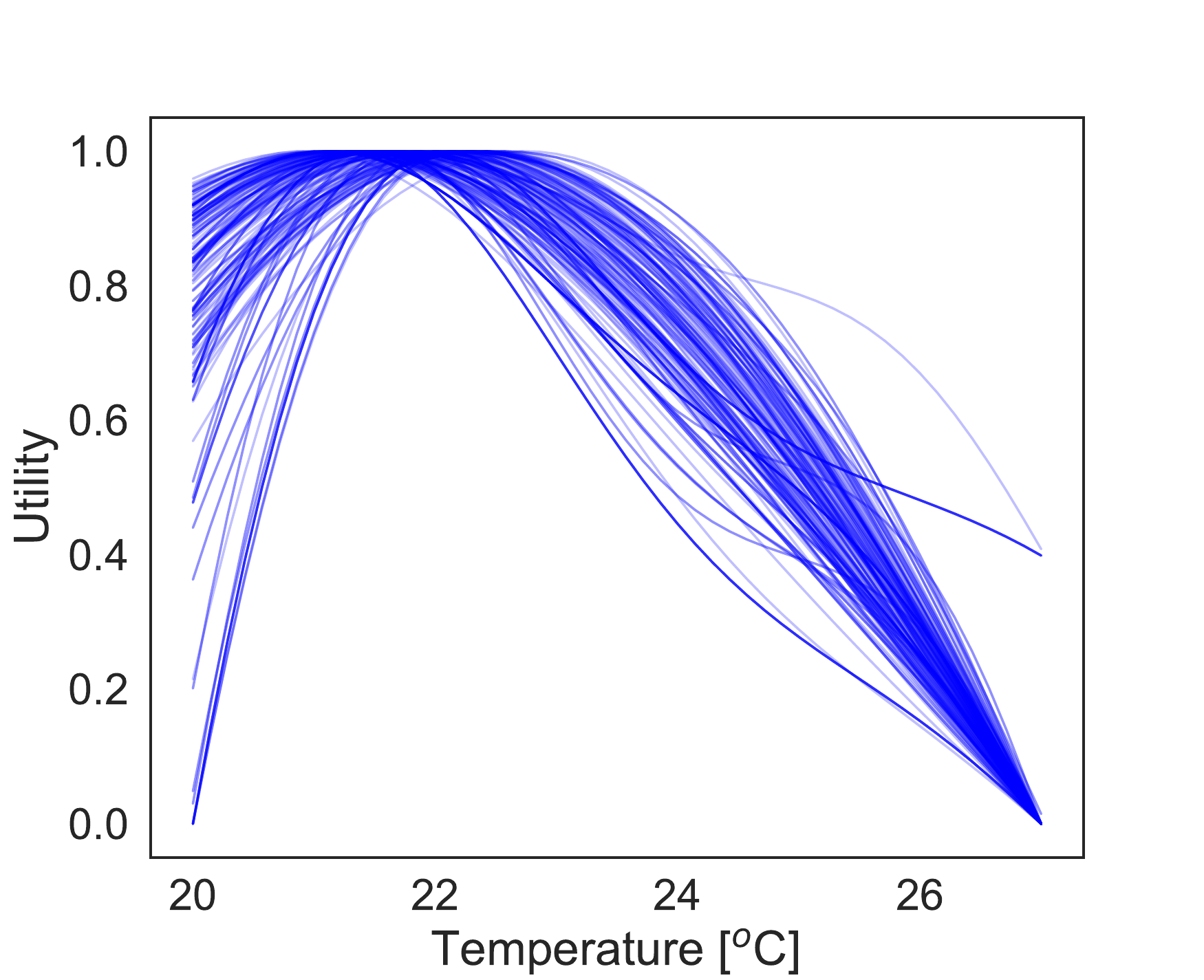} &
\includegraphics[width=50mm]{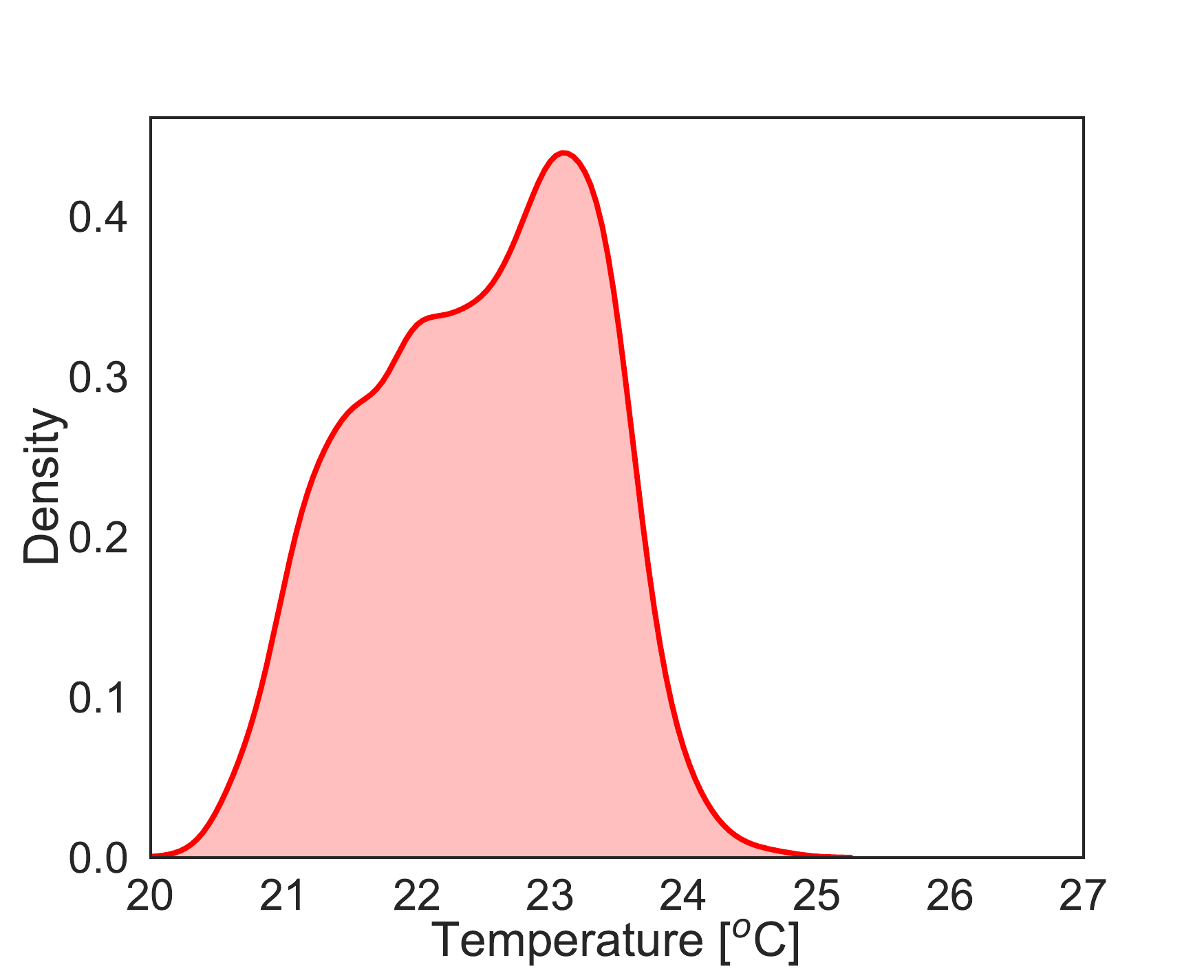} \\
(c) utility samples ($N = 2$) &
(d) utility samples ($N = 2$) &
(e) maximally preferred temp. ($N = 2$) \\
\includegraphics[width=50mm]{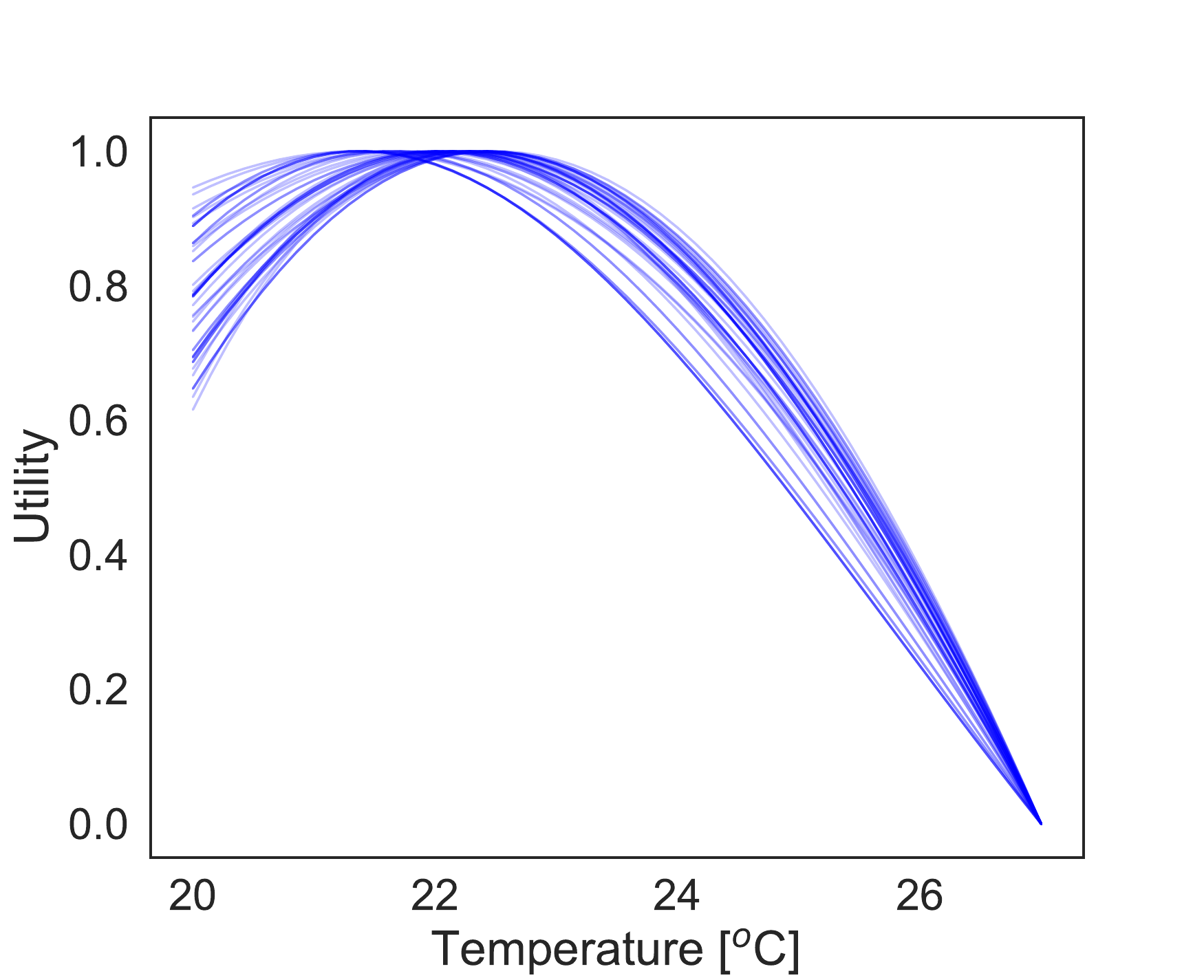} &  
\includegraphics[width=50mm]{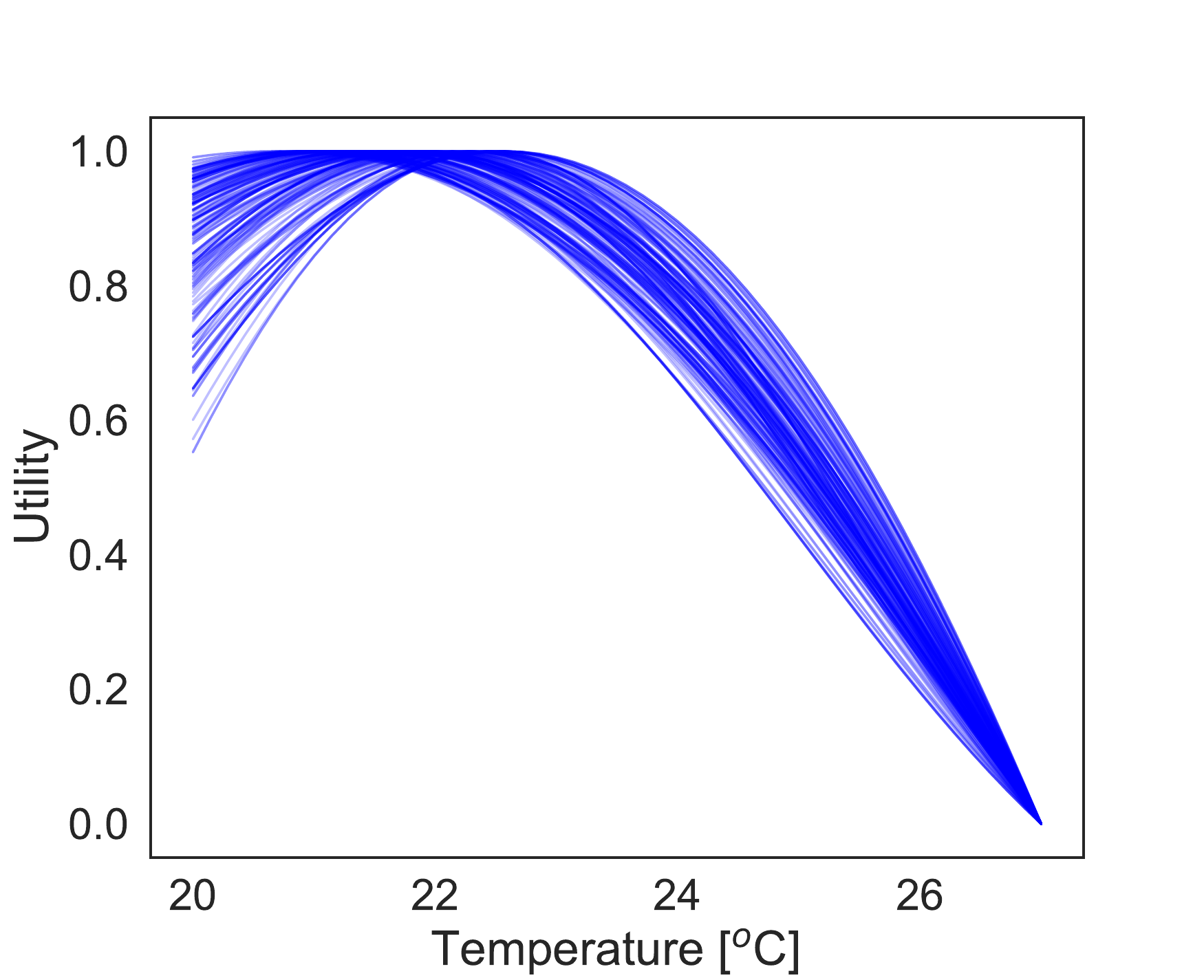} &
\includegraphics[width=50mm]{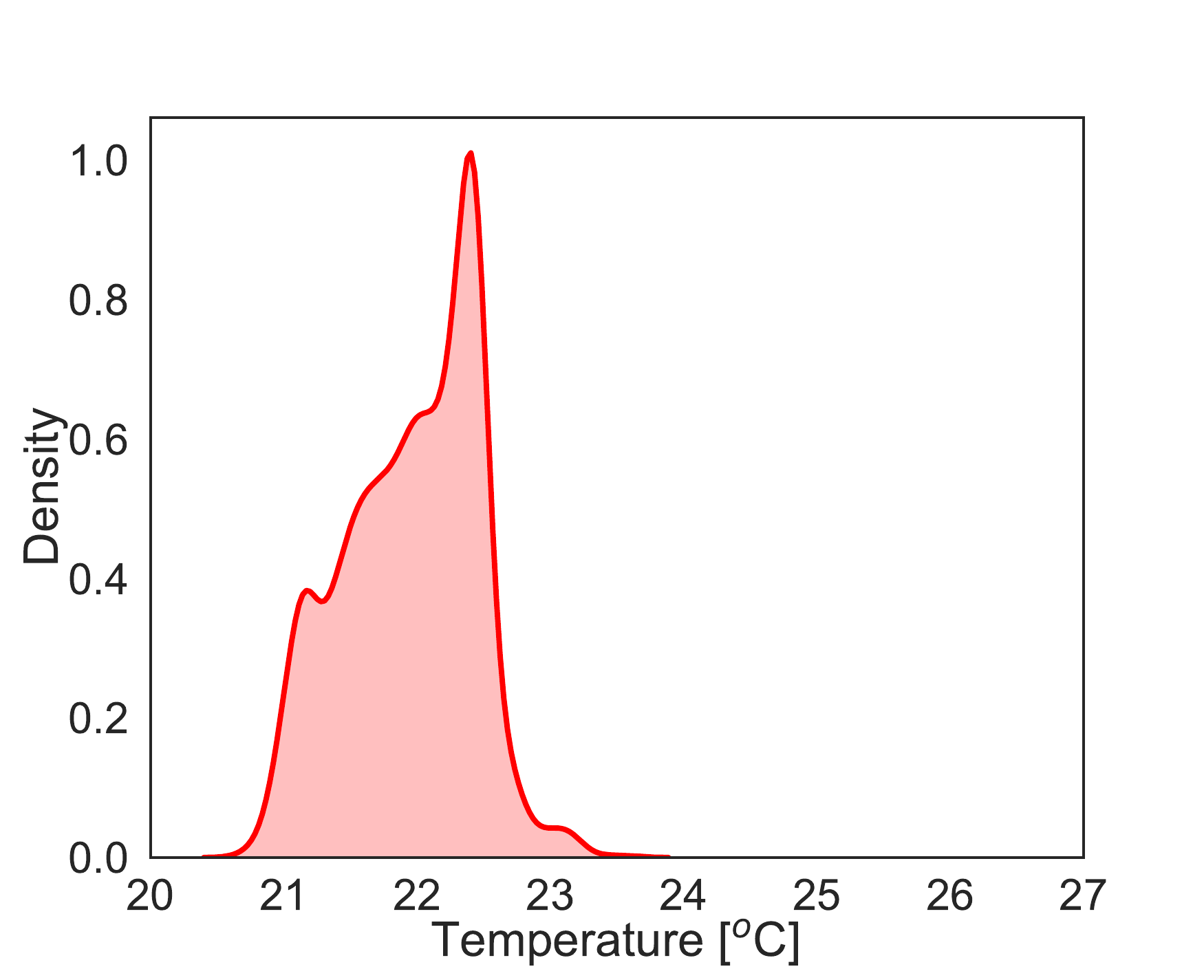} \\
(f) utility samples ($N = 3$) &
(g) utility samples ($N = 3$) &
(h) maximally preferred temp. ($N = 3$) \\
\includegraphics[width=50mm]{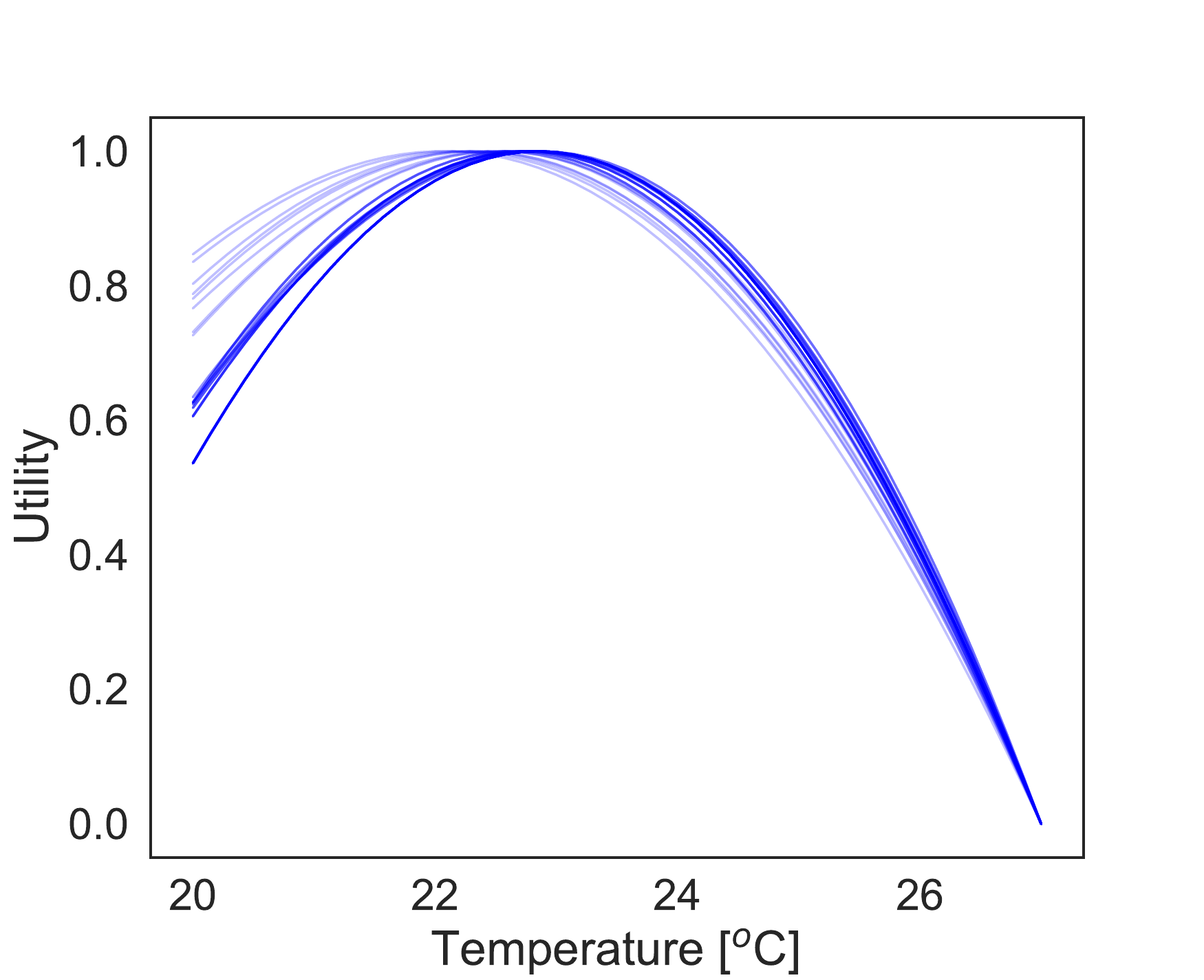} &  
\includegraphics[width=50mm]{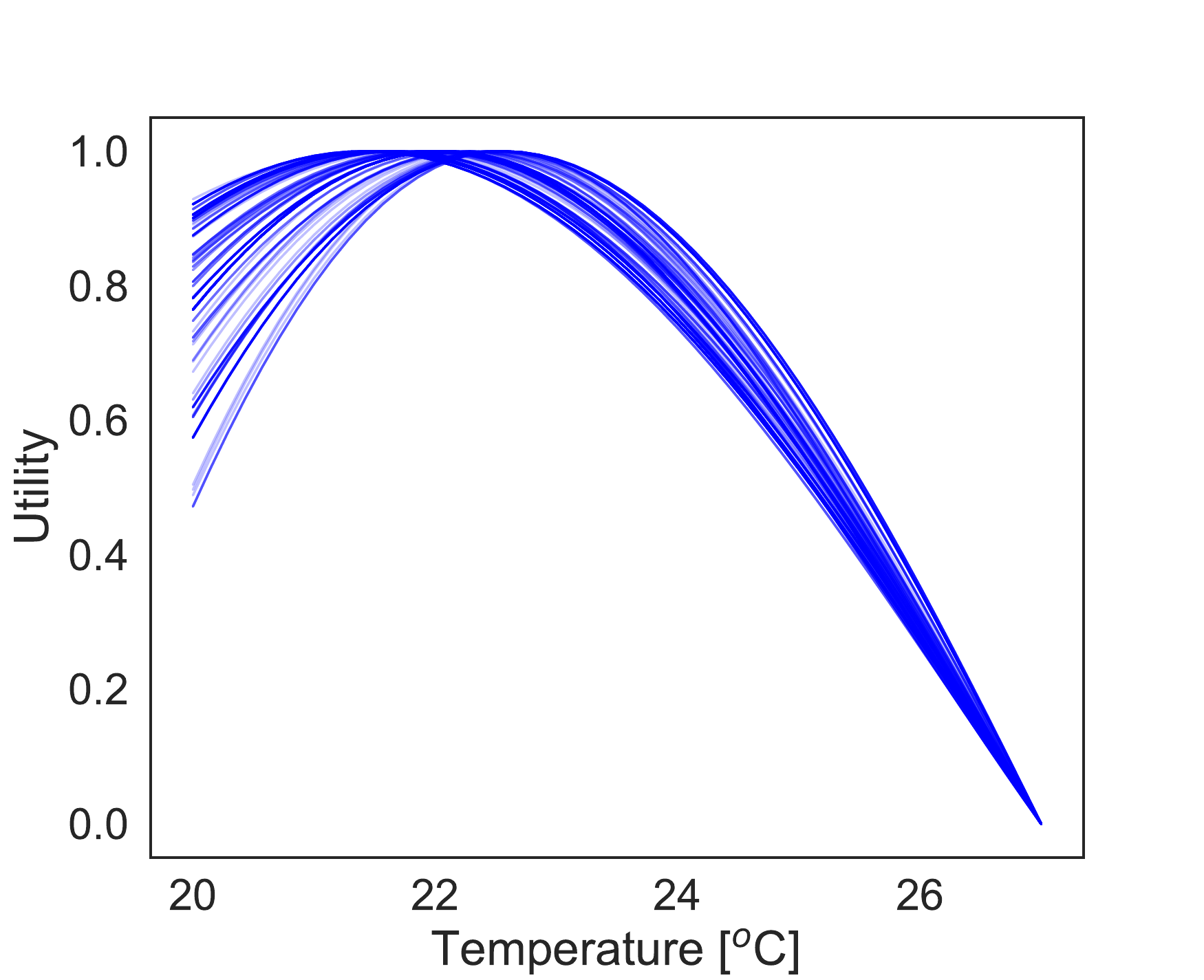} &
\includegraphics[width=50mm]{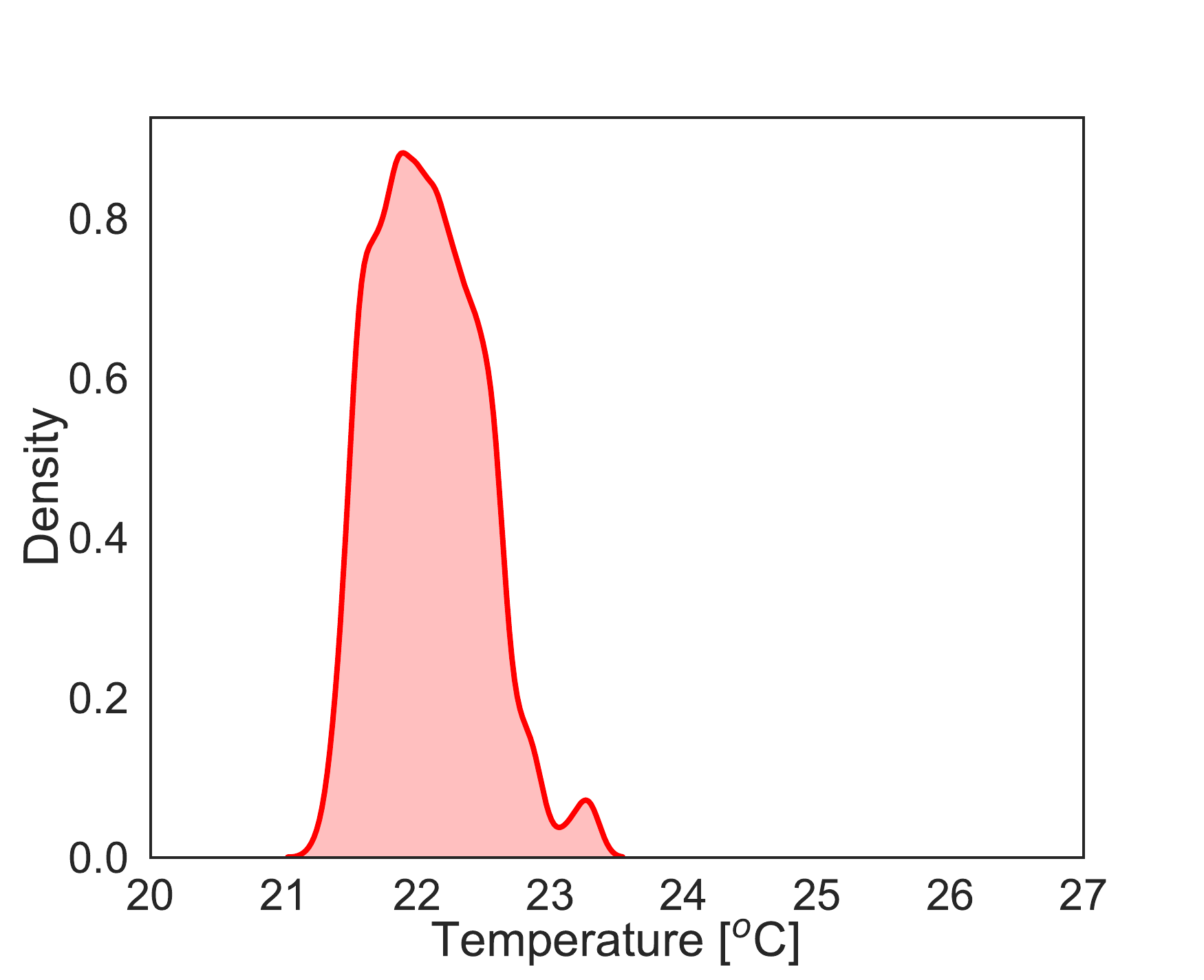} \\
(i) utility samples ($N = 4$) &
(j) utility samples ($N = 4$) &
(k) maximally preferred temp. ($N = 4$) \\
\includegraphics[width=50mm]{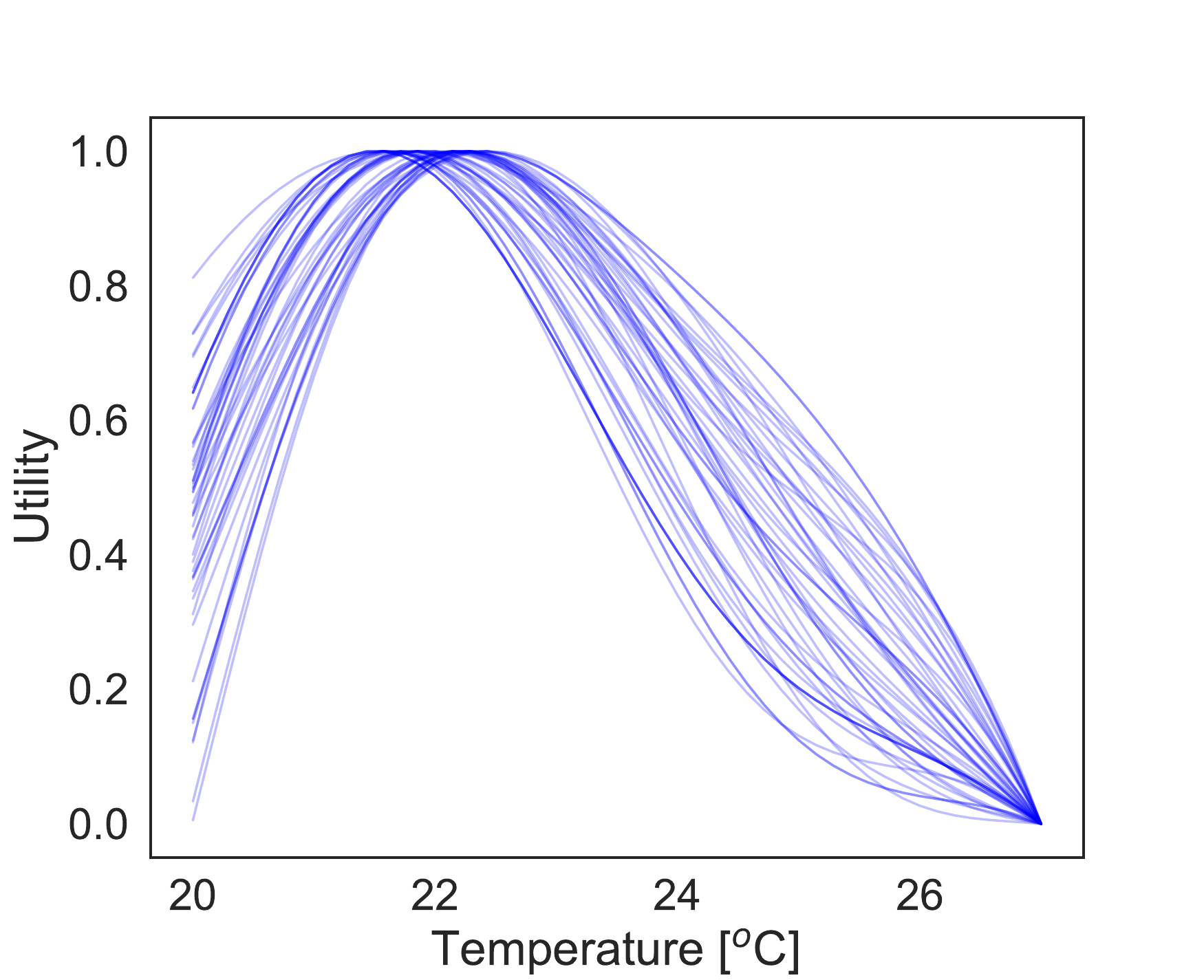} &  
\includegraphics[width=50mm]{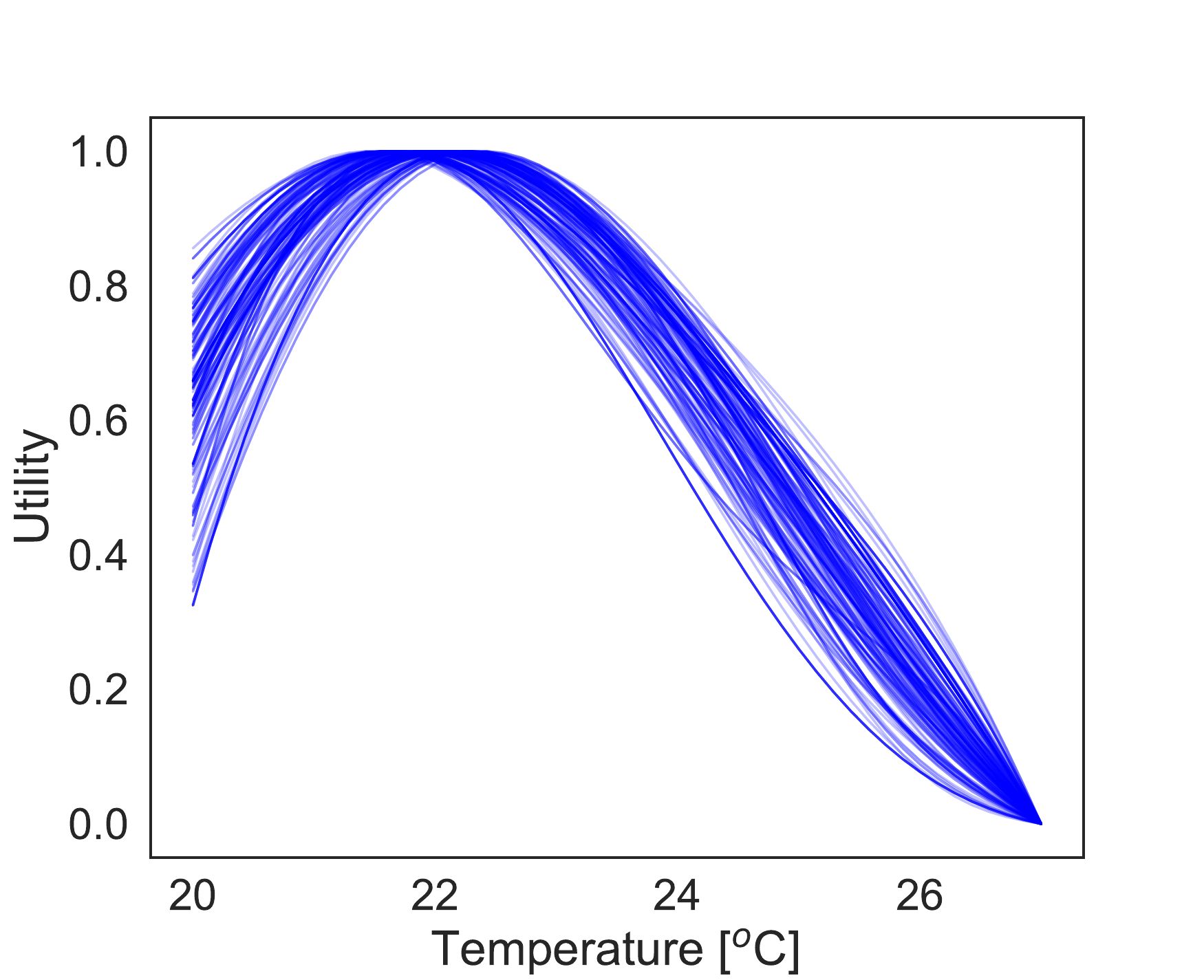} &
\includegraphics[width=50mm]{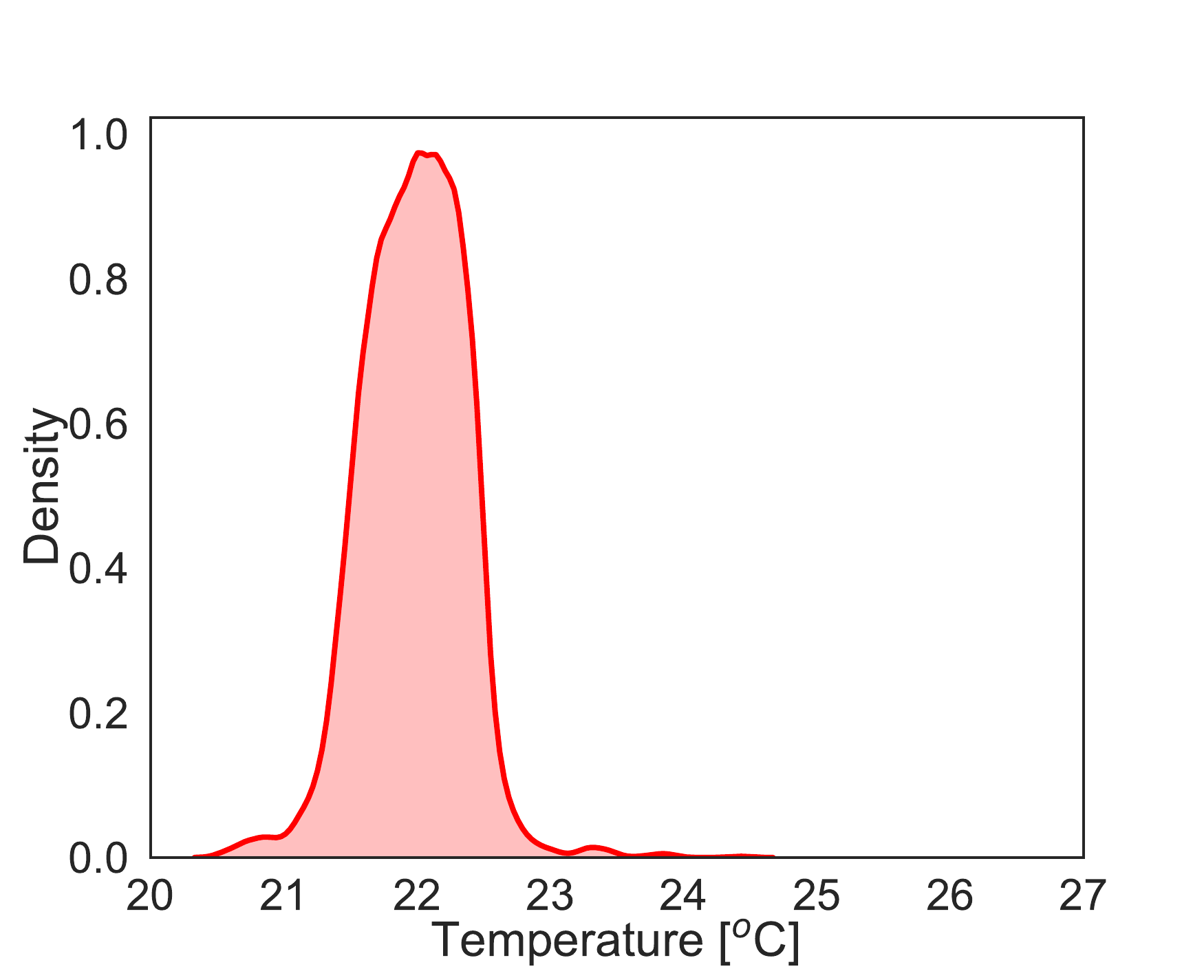} \\
(l) utility samples ($N = 5$) &
(m) utility samples ($N = 5$) &
(n) maximally preferred temp. ($N = 5$) \\
\end{tabular}
\caption{Posterior predictions for synthetic occupant 3. \textbf{First two columns:} Samples from posterior predictive distribution over normalized utility function. \textbf{Last column:} Posterior predictive distribution over maximally preferred indoor air temperature values.}
\label{fig:synthetic_occupant3_pp}
\end{figure}

For selecting the next indoor air temperature to query, the set-up is as follows:
\begin{enumerate}
    \item 
    As mentioned earlier, to find the next query we search over a grid of $\dC{0.5}$- separated indoor air temperatures reaching $\pm \dC{3}$ below and above the current indoor air temperature of the room.
    %The search for the next room temperature to query is performed in a grid of $\dC{0.5}$ from current %room temperature values.
    %We also put one more constraint in the search, i.e. we only look for room temperature values $\pm\dC{3}$ from the current room temperature values.
    %This is done in order to restrict the framework from making temperature changes that are too large and potentially uncomfortable for the occupants (such as from $\dC{20}$ to $\dC{28}$ etc.).
    For example, if the current indoor air temperature value is $x_n=\dC{24}$, then our framework would look for next temperature to query from the set of available indoor air temperatures $\calX_n = \{21, 21.5, 22,22.5,23,23.5, 24.5, 25, 25.5, 26,26.5,27\}$.
    \item We start each elicitation process with a query at $x_1=\dC{21}$.
    We continue to run this elicitation process until we converge towards the maximally preferred indoor air temperature, see the end of the Sec.~\ref{sec:maximally_preferred}.
    All of the indoor air temperatures queried along with their associated thermal preference responses, EUI values, ratio of EUI value at previous iteration and EUI value at current iteration (EUI ratio) for synthetic occupant 1, occupant 2 and occupant 3 are shown in \Cref{table:syn_occupant}.
    The posterior predictive distributions over utility function $u$ and maximally preferred indoor air temperature values are shown in \Crefrange{fig:synthetic_occupant1_pp}{fig:synthetic_occupant3_pp}.
\end{enumerate}

\begin{table}[H]
\centering
\resizebox{\textwidth}{!}{
\begin{tabular}{ccccccccccccc}
\toprule
\multirow{2}{*}{Query \#} &
\multicolumn{4}{c}{Synthetic occupant 1} &
\multicolumn{4}{c}{Synthetic occupant 2} &
\multicolumn{4}{c}{Synthetic occupant 3} \\
\cmidrule(lr){2-5} 
\cmidrule(lr){6-9}
\cmidrule(lr){10-13}
& {Temp. ($\dC{}$)} & {Response} & {EUI} & {EUI Ratio} & {Temp. ($\dC{}$)} & {Response} & {EUI} & {EUI Ratio} &
{Temp. ($\dC{}$)} & {Response} & {EUI} & {EUI Ratio}\\
\midrule

1  & 21  & 1 & 0.465 & - &
21 &  1 & 0.534 & - &
21 & 1 & 0.537 & -\\
2  & 23.5 & 1 & 0.285 & 1.631 &
23.5 & -1 & 0.1963 & 8.30 &
23.5 & -1 & 0.238 & 2.22\\
3 & 26 & -1 & 0.01 & 28.5 &
23 & 1 & 0.295 & 0.665 & 
22.5 & -1 & 0.009 & 26.4\\
4 & 24.5 & 1 & 0.016 & 0.625 &
23.5 & -1 & 0.210 & 1.40 &
21.5 & 1 & 0.010 & 0.9\\
5 & 25.5 & -1 & 0.009 & 2 &
23 & 1 & 0.234 & 0.897 &
22 & 1 & 0.005 & 2\\
6 & 25 & 0 & 0.009 & 1 &
23 & 1 & 0.221 & 1.05 & 
22 & 1 & 0.005 & 1\\
7 & 25 & 0 & 0.006 & 1.5 & 
23.5 & -1 & 0.231 & 0.956 &
22.5 & -1 & 0.005 & 1\\
8 & 25 & 0 & 0.007 & 0.857 &
23.5 & -1 & 0.008 & 28.875 &
22 & 1 & 0.005 & 1\\
9 & 25 & 0 & 0.008 & 0.875 &
23.5 & -1 & 0.005 & 1.6 &
22.5 & -1 & 0.005 & 1\\
10 & 25 & 0 & - & - &
23.5 & -1 & - & - 
& 22 & 1 & - & -\\
\bottomrule
\end{tabular}
}
\caption{Synthetic occupants' preference data collected using the newly developed PE framework.}
\label{table:syn_occupant}
\end{table}

As we can see from \Crefrange{fig:synthetic_occupant1_pp}{fig:synthetic_occupant3_pp}, our PE framework searches through all of the available indoor air temperature values to converge towards the temperature that maximizes the thermal satisfaction/utility of the occupants.
For synthetic occupant 1, the maximum preferred indoor air temperature value was $\dC{25}$ and our framework was able to infer it in six queries  (\Cref{fig:synthetic_occupant1_pp}). 
In case of synthetic occupant 2, due to the constraints we have put on the
framework, i.e., to only search for temperatures values with minimum temperature difference of 0.5, our framework converges to the maximally preferred indoor air temperature value of $\dC{23.5}$ instead of actual maxima at $\dC{23.34}$.
This happens in six queries to the occupant (\Cref{fig:synthetic_occupant2_pp}).
Next, the process converges to indoor air temperature value of $\dC{22}$  for synthetic occupant 3  (actual maxima is at $\dC{22.2}$) after collecting five active thermal preference queries (\Cref{fig:synthetic_occupant3_pp}).

As shown in \Crefrange{fig:synthetic_occupant1_pp}{fig:synthetic_occupant3_pp}, all of the inferred posterior predictive utility samples are unimodal.
Because of this strict unimodality constraint, the EUI acquisition function governing the search for maximally preferred indoor air temperature value corrects the boundary over-exploration effect which is typically seen in conventional Bayesian global optimization problems \cite{siivola2017correcting}.
Intuitively, once we have observed that a given occupant responds ``prefers warmer'' at \SI{21}{\celsius}, the model becomes sufficiently confident that this occupant is going to respond ``prefers warmer'' at $\{\SI{20}{\celsius}, \SI{19.5}{\celsius}, \SI{19}{\celsius}\}$ temperature values as well.
Similarly, if the occupant responds that he ``prefers cooler'' at \SI{26}{\celsius}, then our model becomes confident that this occupant is also going to respond ``prefers cooler'' at $\{\SI{27}{\celsius}, \SI{27.5}{\celsius}, \SI{28}{\celsius}\}$ temperature values.
Because of the unimodality constraint, in this case, the model would place high probability of the maximally preferred indoor air temperature of being in between \SI{21}{\celsius} and \SI{26}{\celsius}.
%However, if this unimodality assumption is not valid in another application, this constraint can be relaxed and elicitation using simple GP model can be implemented (all of the steps illustrated in Algorithm.\ref{alg:PE} would be the same, except that now you would
%use conventional GP model instead of unimodal GP).
%In this section we have shown the robustness of our framework when concerned with inferring maximally preferred indoor air temperature values in case of different synthetic occupants.

\subsection{Performance of Expected Improvement based Preference Elicitation Framework}\label{subsec:performance of expected improvement}
In this sub-section, we analyze the performance of our newly developed PE framework and illustrate some of its key properties. Put simply, we seek to answer the following questions:
\begin{enumerate}
    \item How long does it take for PE framework to converge towards the maximally preferred indoor air temperature values?
    \item How does changing the initial preference query posed to the occupant affect the convergence of PE framework?
    \item How is the EUI based PE framework performance compared to just randomly searching the input space? Does it converge faster towards the maximally preferred indoor air temperature values?
\end{enumerate}

\subsubsection{Convergence towards maximally preferred indoor air temperature values}
\begin{figure}[H]
  \centering
  \begin{subfigure}[b]{0.32\linewidth}
    \includegraphics[width=\linewidth]{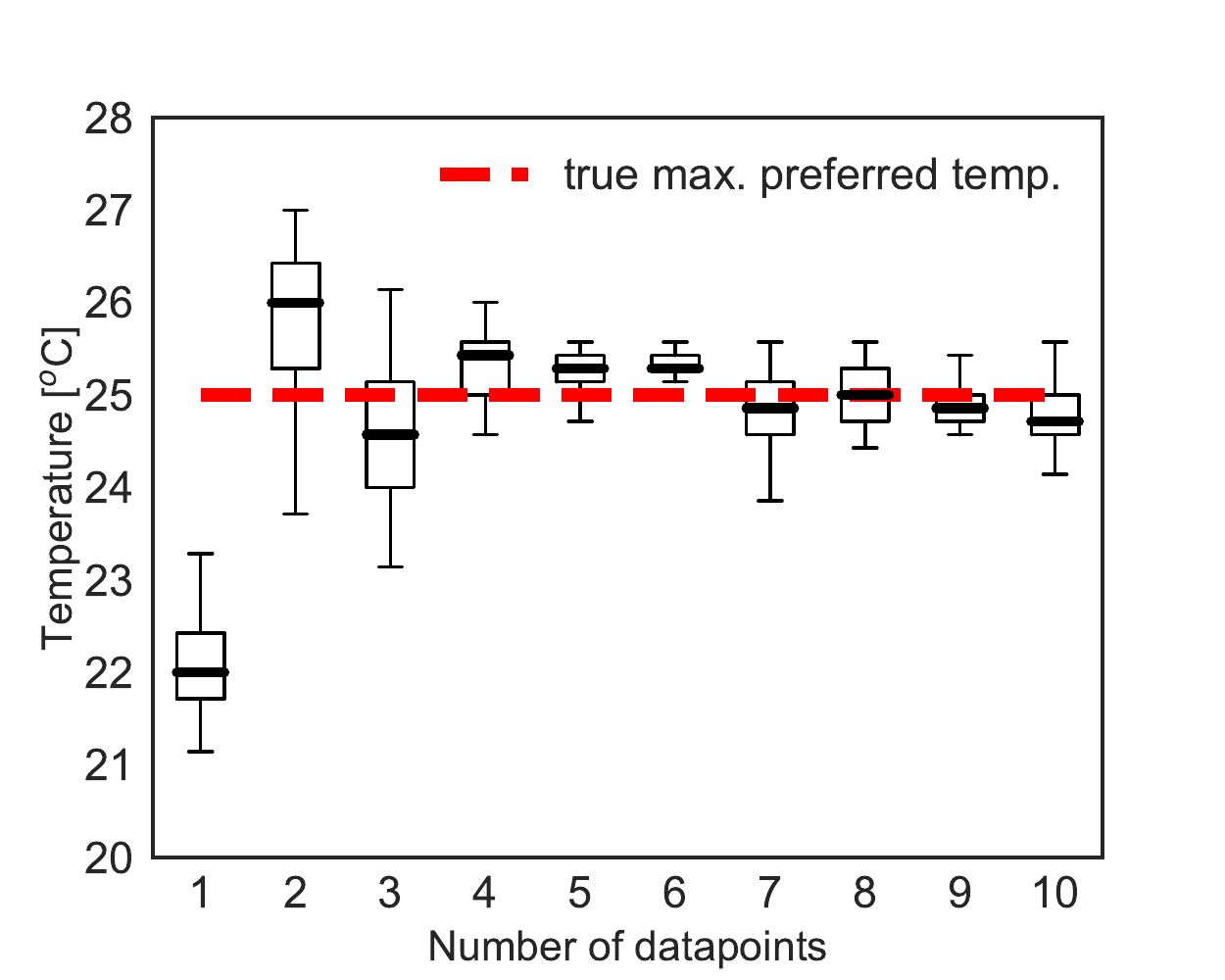}
     \caption{Synthetic Occupant 1}
     \label{fig:conv_o1}
  \end{subfigure}
  \begin{subfigure}[b]{0.32\linewidth}
    \includegraphics[width=\linewidth]{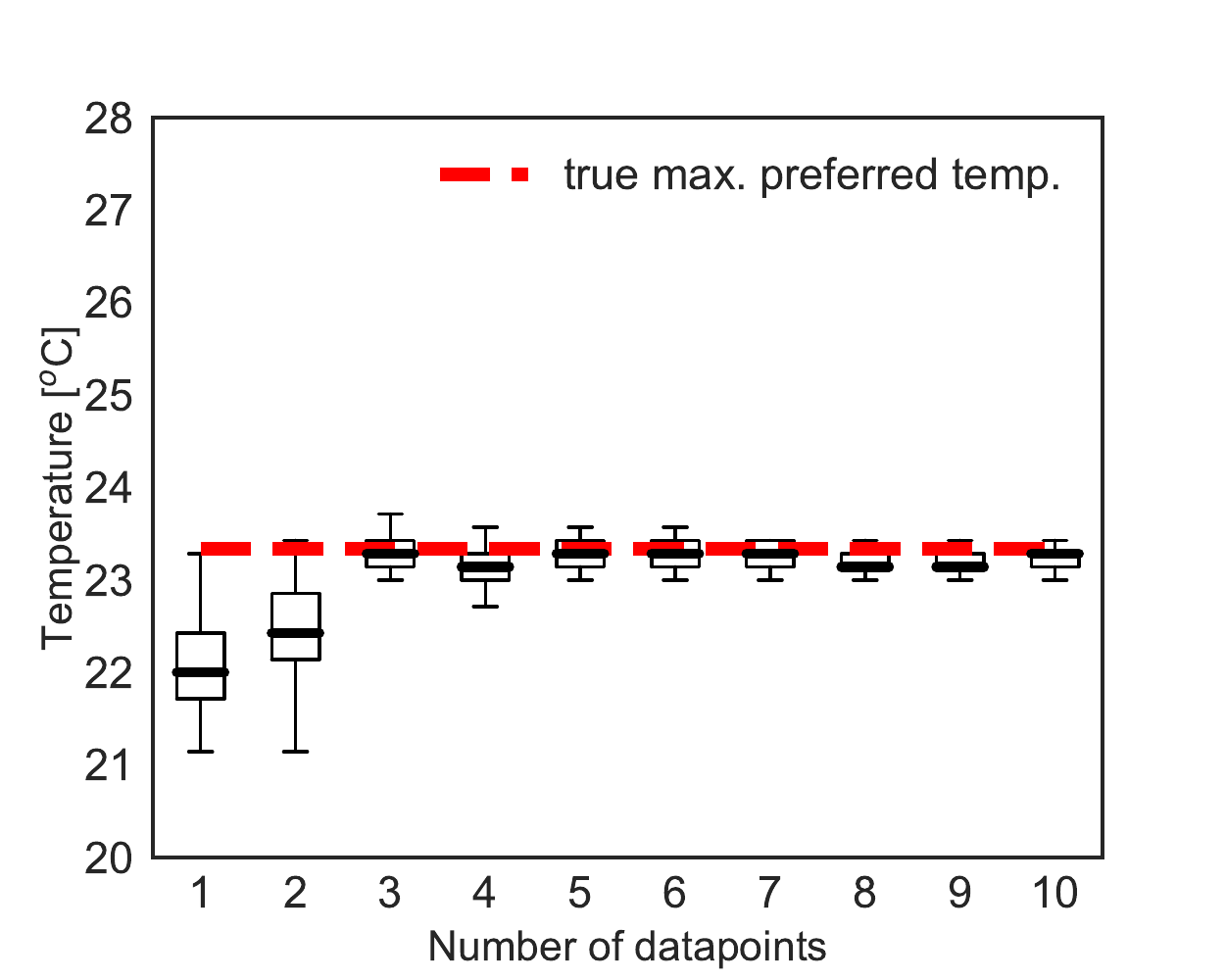}
    \caption{Synthetic Occupant 2}
    \label{fig:conv_o2}
  \end{subfigure}
  \begin{subfigure}[b]{0.32\linewidth}
    \includegraphics[width=\linewidth]{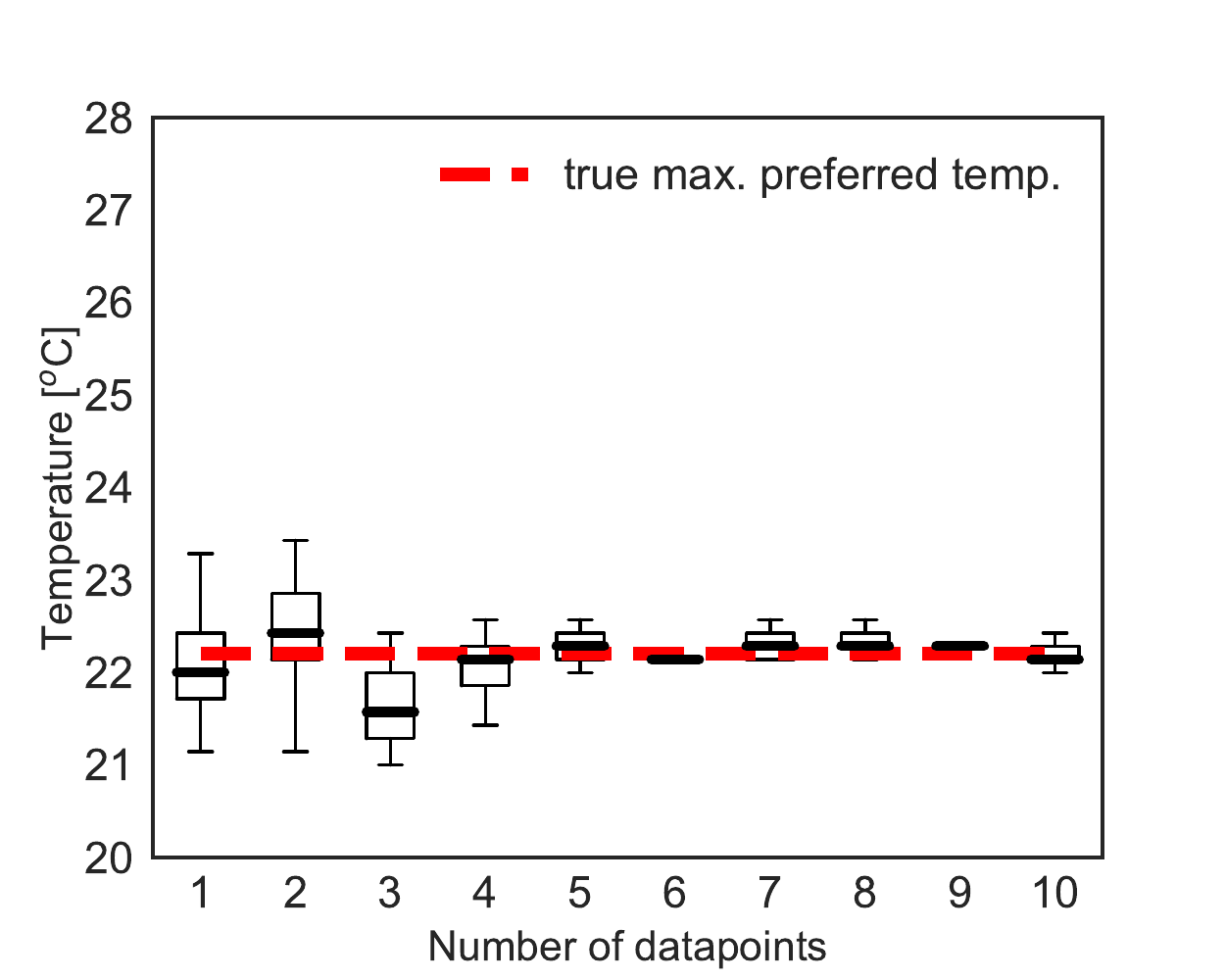}
    \caption{Synthetic Occupant 3}
    \label{fig:conv_o3}
  \end{subfigure}
\caption{Convergence of EUI based PE framework to maximally preferred indoor air temp. values}
\label{fig:max_pref_convergence}
\end{figure}

As discussed in the previous sub-section (see \cref{subsec:framework_in_action}), each of the PE run starts with one initial thermal preference query to the occupants at $\dC{21}$ and a total budget of 10 queries are posed to the occupants.
We report the posterior distribution over maximally preferred indoor air temperature values at each step of the PE runs. 
The boxplots of these inferred posterior distribution over maximally preferred indoor air temperature values (for each of the synthetic occupants) are illustrated in \cref{fig:max_pref_convergence}. Minimum and maximum values of inferred preferred indoor air temperature values are used to define the whiskers of the boxplots.
The interquantile range (IQR) for visualization of boxplots is calculated as the difference between the first and third quantiles and 1.5 times IQR is used to define the outliers.
As seen from these sub-figures, our PE framework converges to the maximally preferred indoor air temperature values (\cref{fig:conv_o1} shows convergence of PE runs to $\dC{25}$ in case of synthetic occupant 1 ; \cref{fig:conv_o2} shows convergence of PE runs to $\dC{23.34}$ in case of synthetic occupant 2 and \cref{fig:conv_o3} shows convergence of PE runs to $\dC{22.1}$ in case of synthetic occupant 3).

\subsubsection{Convergence analysis }
\begin{figure}[H]
  \centering
  \begin{subfigure}[b]{0.32\linewidth}
    \includegraphics[width=\linewidth]{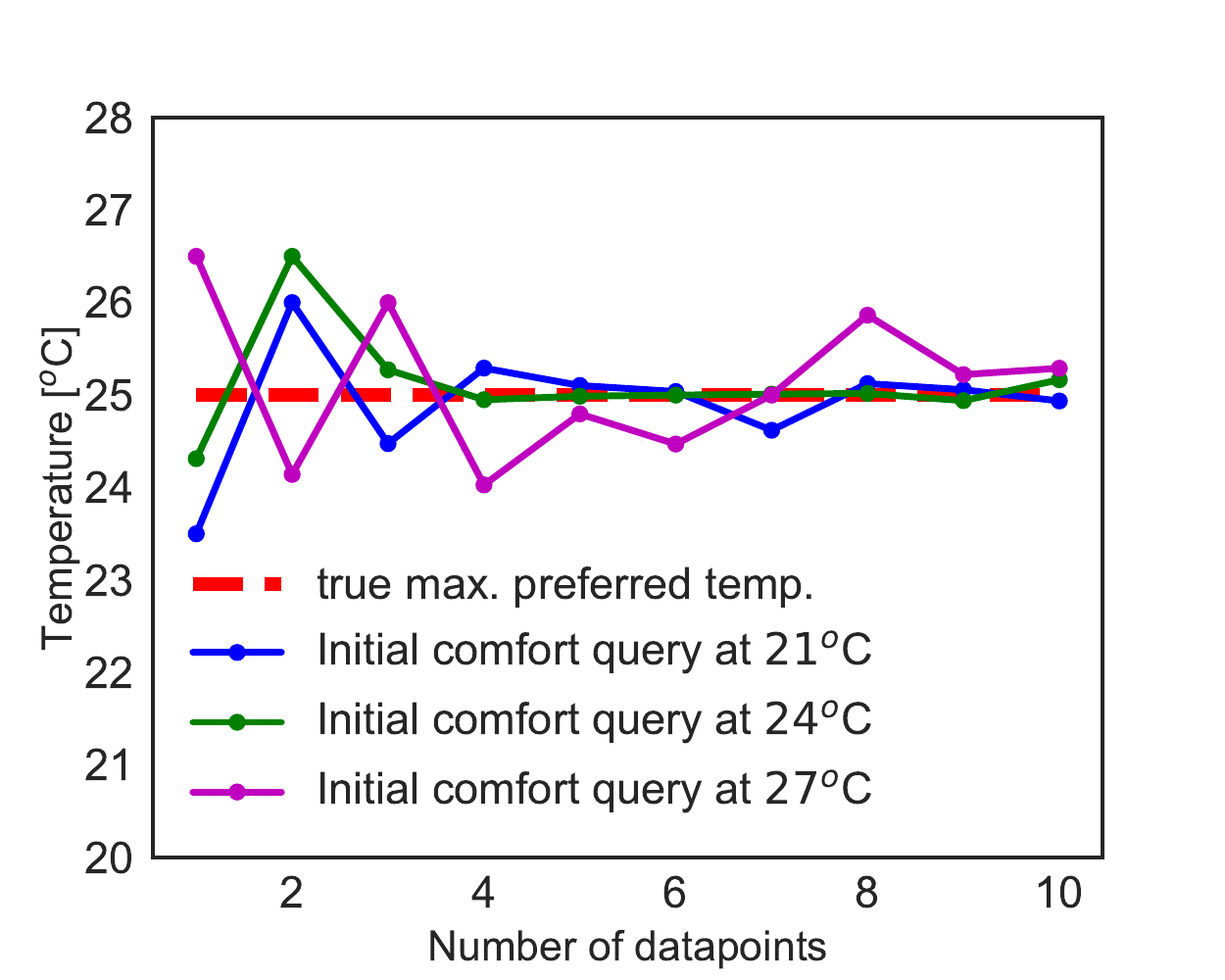}
     \caption{Synthetic Occupant 1}
     \label{fig:conv_random_init_o1}
  \end{subfigure}
  \begin{subfigure}[b]{0.32\linewidth}
    \includegraphics[width=\linewidth]{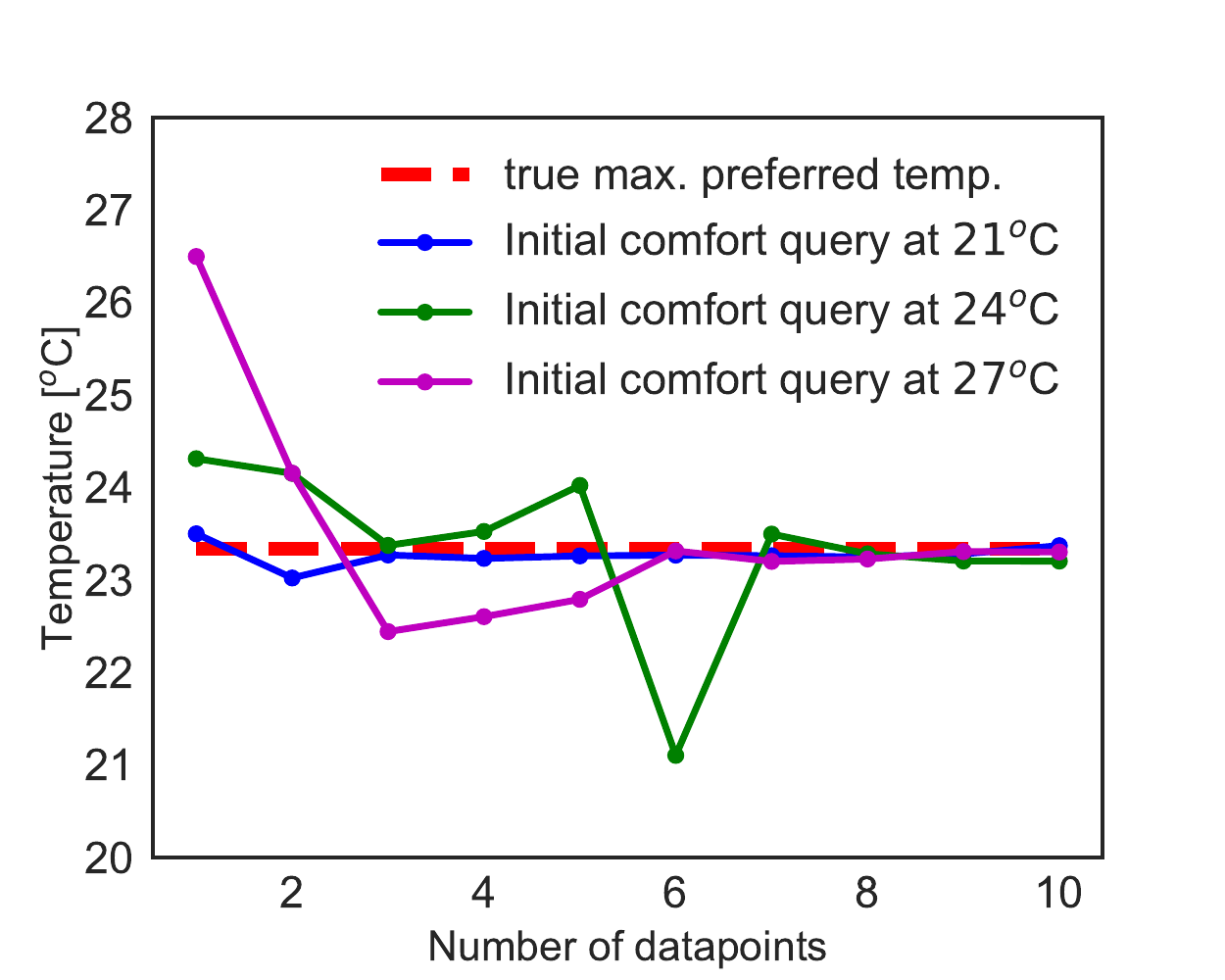}
    \caption{Synthetic Occupant 2}
    \label{fig:conv_random_init_o2}
  \end{subfigure}
  \begin{subfigure}[b]{0.32\linewidth}
    \includegraphics[width=\linewidth]{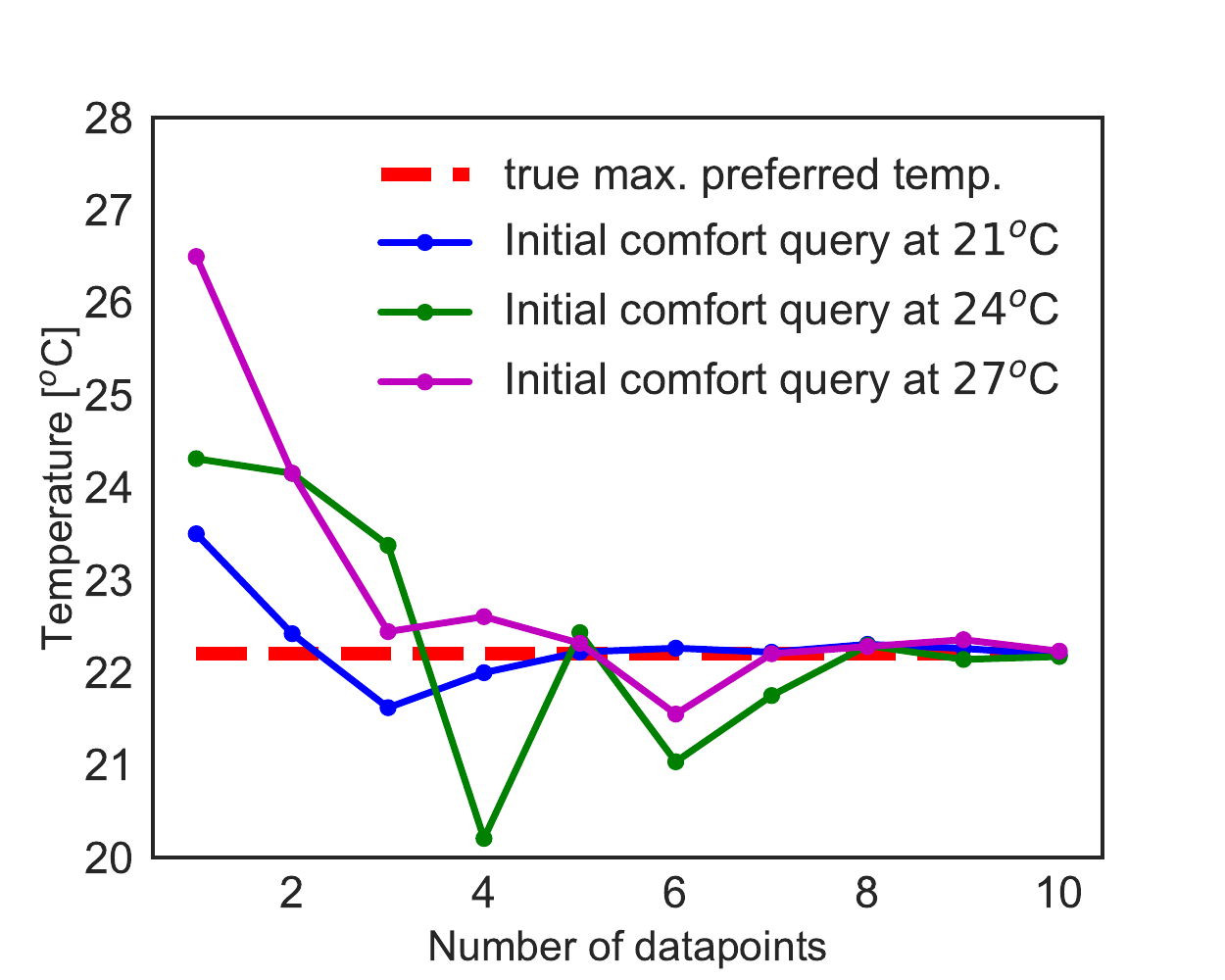}
    \caption{Synthetic Occupant 3}
    \label{fig:conv_random_init_o3}
  \end{subfigure}
\caption{Convergence of EUI based PE framework (from random initial comfort query points)}
\label{fig:conv_random_init}
\end{figure}

In this sub-sub-section, we show the robustness of our PE framework to converge towards the maximally preferred indoor air temperature values (independent of the initial query posed to the occupants).
That is, no matter the initial thermal preference query posed to the occupant (whether it be at $\dC{21}$ or $\dC{24}$ or $\dC{27}$), in a perfectly ideal experimental setting, it will converge towards the maximally preferred indoor air temperatures (in less than 10 queries) to occupants. 
We run the PE scheme (for all the occupants) three times, starting from initial comfort query at $\dC{21}$, $\dC{24}$ and $\dC{27}$. 
We report the mean of inferred posterior maximally preferred indoor air temperature values at each step of the PE runs (for all the three occupants, with different initial preference query)  in \cref{fig:conv_random_init}. As seen from these sub-figures, no matter from where we start the PE runs, our algorithm will eventually converge towards the maximally preferred indoor air temperature values (in less than 10 queries).

\subsubsection{Expected Improvement in Utility (EUI) vs. Random Search (RS)}
In this sub-sub-section, we address the question of ``how fast does the EUI based PE framework converge towards the maximally preferred indoor air temperature values (as compared to just randomly searching (RS) over the indoor air temperature values)?'' 
For evaluating this comparison between EUI vs. RS, we introduce the concept of averaged Euclidean distance (AED) performance metric.

\begin{figure}[H]
  \centering
  \begin{subfigure}[b]{0.32\linewidth}
    \includegraphics[width=\linewidth]{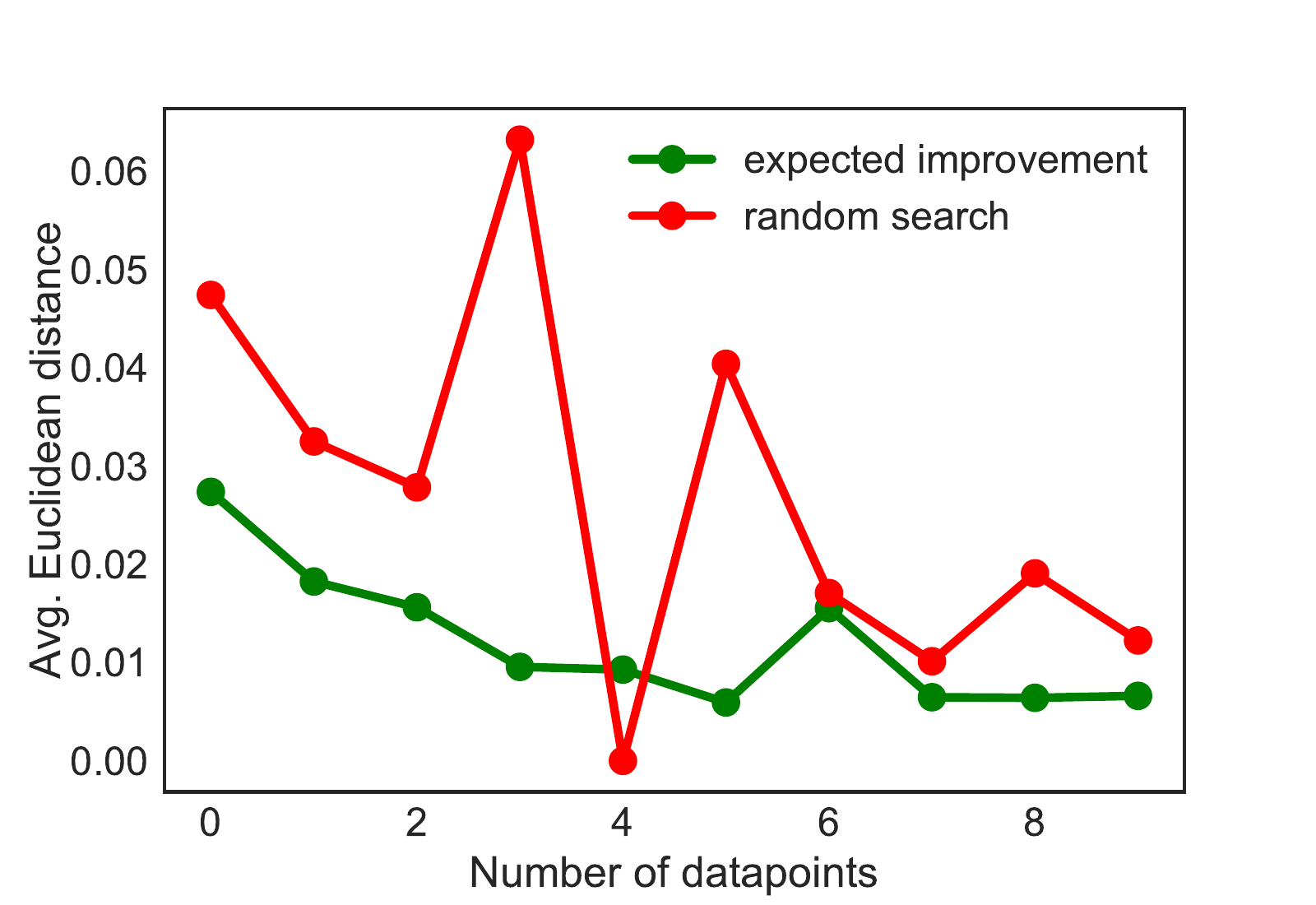}
     \caption{Synthetic Occupant 1}
     \label{fig:conv_euc_o1}
  \end{subfigure}
  \begin{subfigure}[b]{0.32\linewidth}
    \includegraphics[width=\linewidth]{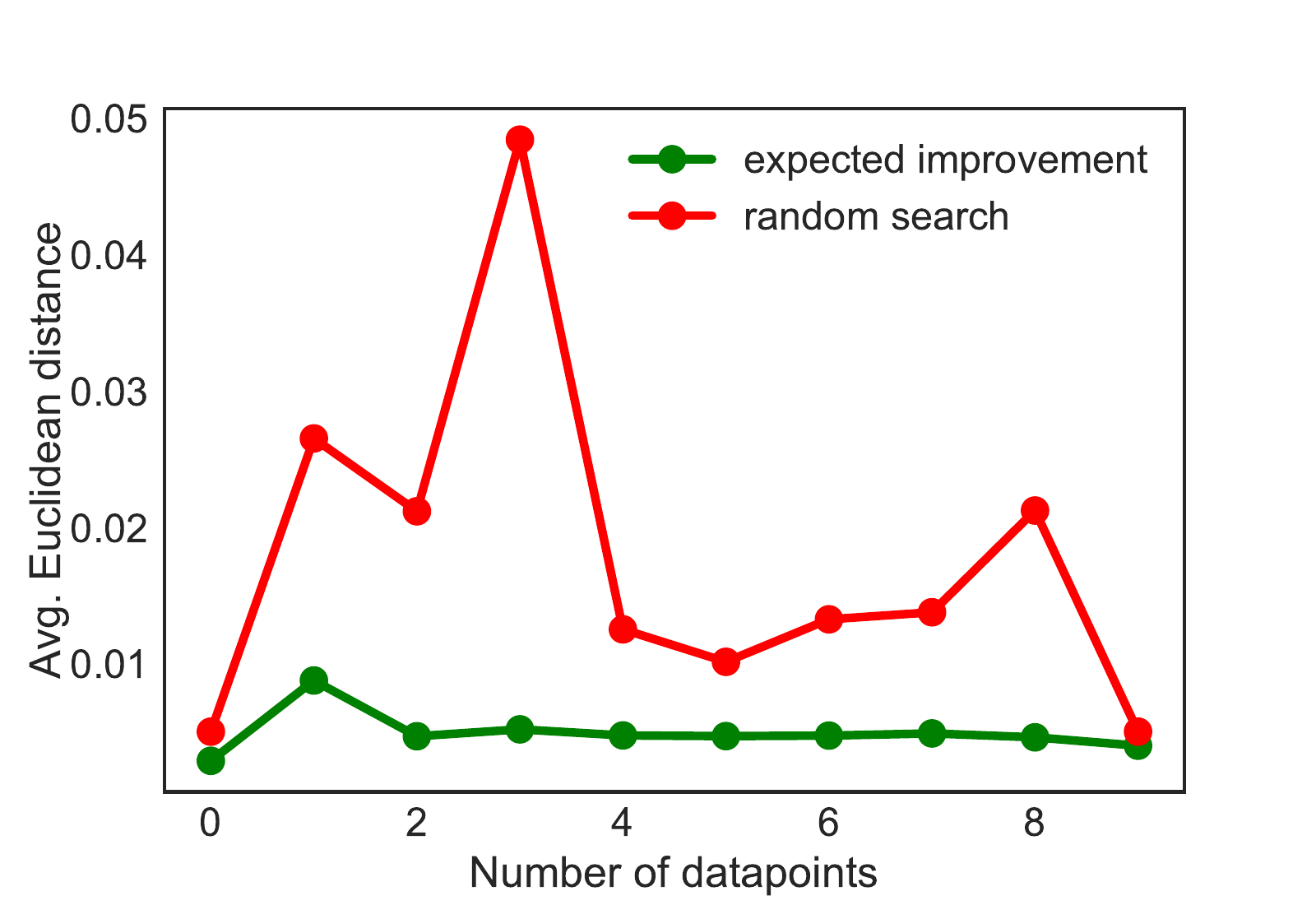}
    \caption{Synthetic Occupant 2}
    \label{fig:conv_euc_o2}
  \end{subfigure}
  \begin{subfigure}[b]{0.32\linewidth}
    \includegraphics[width=\linewidth]{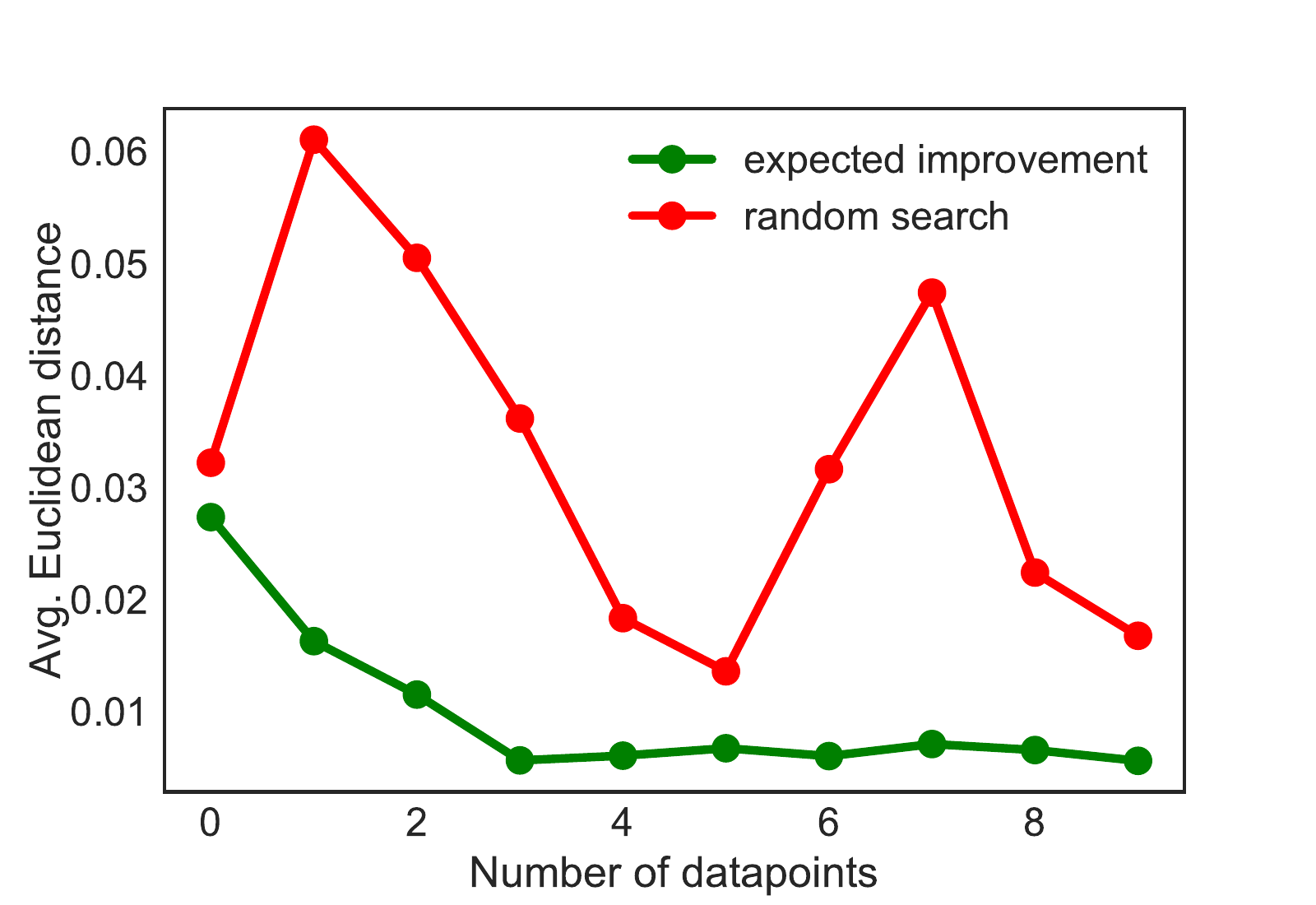}
    \caption{Synthetic Occupant 3}
    \label{fig:conv_euc_o3}
  \end{subfigure}
\caption{Convergence of EUI based PE framework vs. Random Search}
\label{fig:ei_vs_rs}
\end{figure}

AED between two vectors gives a measure of ``similarity'' between the two vectors.
In this problem, we calculate the averaged Euclidean distance between the inferred maximally preferred indoor air temperature (at each step of the elicitation runs) and true maximally preferred indoor air temperature values. 
We run both EUI and RS based elicitation 3 times (trials) with different initial thermal preference query posed to the occupants (same for both EUI and RS based exploration) and we report the average performance across all these trials. The lower the value of AED metric, the better the given framework and vice versa. It is given as \cite{awalgaonkar2018design} :
\begin{equation}\label{eqn:euc_dist_metric}
%p(\tilde{x}_{\text{best}} |\tilde{X}, D)
\begin{array}{ccc}

\text{AED}(\tilde{x}_{\text{best}}, x_{\text{true}} |\tilde{X}, D_n) &\approx& 
\frac{1}{S}\left(\sum_{s=1}^S (\tilde{x}^{(s)}_{\text{best}} - x_{\text{true}})^2\right)^{1/2},
\end{array}
\end{equation}
where $\tilde{x}^{(s)}_{\text{best}}$ are HMC samples approximating posterior distribution over maximally preferred indoor air temperature values and $x_{\text{true}}$ is the true maximally preferred indoor air temperature ($\dC{25}$ in case of occupant 1 ; $\dC{23.34}$ in case of occupant 2 and $\dC{22.1}$ in case of occupant 3). 
\cref{fig:ei_vs_rs} shows the performance of EUI approach against randomized data collection (as we add more and more thermal comfort queries).
The results are consistent across all the three occupants, that is, EUI exploration approach has consistently been proven better than randomized data collection (since it converges in less number of thermal preference queries to the occupants). 
For more in-depth analysis of the convergence properties of expected improvement acquisition function, we refer readers to the work done by \cite{vazquez2010convergence}.

\subsection{Performance of Unimodal Gaussian Process Preference Learning Model}\label{subsec:performance of expected improvement}
As we have iterated numerous times in the previous sections, the objective of this paper is to sequentially pose intelligent queries to occupants in order to optimally learn the indoor air temperature which would maximize their satisfaction.
However, after analyzing the posterior distribution over utilities and inferred maximally preferred indoor air temperature values, a question which naturally arises in our minds is ``can the Unimodal-GP based preference learning model be used in for classifying prefer warmer/ prefer cooler conditions?'' Indeed, due to the strict unimodality constraints we have imposed on the structure of utility function, one positive/favorable side-effect of running our UEI based PE scheme is that the trained Unimodal-GP can be further used to classify prefer warmer vs. prefer cooler conditions. 

In order to validate if our systematic approach provides higher accuracy for classifying prefer warmer vs. prefer cooler conditions, we compared the performance of our proposed Unimodal-GP preference learning model against two well-known classification techniques : logistic regression and support vector machine (SVM) with linear basis function. Each of the above-mentioned classification models has one or several hyperparameters that need to be properly tuned for providing their best performance during classification. We do not discuss the process of hyperparameter tuning in this paper. Instead, for learning more about hyperparameter optimization of machine learning models, we refer readers to work done by \cite{bishop2006pattern}. In order to achieve fair comparison between these models, we provided two sets of training data: (1) 5 datapoints based on EUI search (first 5 rows of \cref{table:syn_occupant}) and (2) 5 randomly selected (RS) thermal preference queries to the models, trained them and then checked the classification results on a testing test (different from the training data). Specifically, we used hit-rate accuracy (HRA) as the measure of performance of different classification models (same as in \cite{ghahramani2015online}). 
HRA is used as a statistical measure of how well a binary classification model correctly identifies or excludes a condition. 
Put simply, HRA is defined as the average chance of correct prediction.
Higher the value of HRA, the better the performance of model is. In \cref{table:hit_rate_accuracy}, we report the 3-fold validation HRA across all the above mentioned models. As it can be seen in \cref{table:hit_rate_accuracy}, hit rate accuracy of the proposed Unimodal-GP (trained on data collected using EUI based PE framework) was relatively higher than the other classification models. The relatively higher accuracy of the proposed model is due to the strict unimodality prior constraints imposed on the structure of utility functions as well as due to the collection of most informative thermal preference query data (as elicited by our PE framework). However, as promising as this may be, we also like to note that one should be cautious while interpreting \cref{table:hit_rate_accuracy}, as the HRA metrics values were obtained using synthetic occupants' (ideal) preference data whose true utilities were unimodal, favoring our unimodal GP model and this might not always be the case when concerned with real occupants. 

\begin{table}[H]
\centering
\resizebox{\textwidth}{!}{
\begin{tabular}{ccccccc}
\toprule
\multirow{3}{*}{Model} &
\multicolumn{6}{c}{Hit rate accuracy (\%)} \\ 
\cmidrule(lr){2-7}
& 
\multicolumn{2}{c}{Synthetic occupant 1} &
\multicolumn{2}{c}{Synthetic occupant 2} &
\multicolumn{2}{c}{Synthetic occupant 3} \\
\cmidrule(lr){2-3} 
\cmidrule(lr){4-5}
\cmidrule(lr){6-7}
& {Average} & {Standard deviation} &
{Average} & {Standard deviation} &
{Average} & {Standard deviation} \\
\midrule
Unimodal GP (trained on EUI data) & 100\% & 0\% & 97.33\% & 2.5\% & 100\% & 0\%\\
Logistic Regression (trained on EUI data) & 100\% & 0\% & 81.33\% & 6.6\% & 74.66\% & 5.24\%\\
SVM (trained on EUI data) & 93.33\% & 4.10\% & 46\% & 8.48\% & 56.66\% & 5.73\%\\
Unimodal GP (trained on RS data) & 70\% & 5.65\% & 68.66\% & 13.19\% & 72.66\% & 4.10\%\\
Logistic Regression (trained on RS data) & 64.66\% & 11.46\% & 73.33\% & 6.6\% & 69.33\% & 6.8\%\\
SVM (trained on RS data) & 64.66\% & 11.46\% & 78\% & 3.26\% & 69.33\% & 6.8\%\\

\bottomrule
\end{tabular}
}
\caption{Hit Rate Accuracy of Preference Learning Models}
\label{table:hit_rate_accuracy}
\end{table}

\section{Experimental results}\label{sec:DOE}
\subsection{Experiment Setup}\label{subsec:experimental_setup}
\begin{figure}[H]
        \centering
        \begin{minipage}{0.3\linewidth}
        \begin{subfigure}{\linewidth}
        \includegraphics[width=\linewidth]{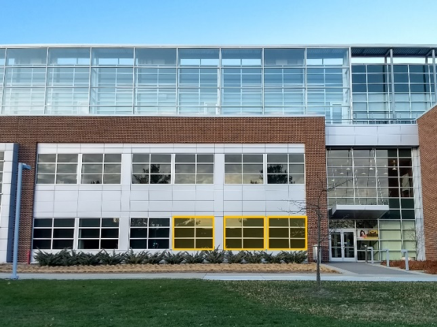}
        \caption{Three offices used in the study}
        \label{fig:herrick_office}
        \end{subfigure}
        \begin{subfigure}{\textwidth}
        \includegraphics[width=\linewidth]{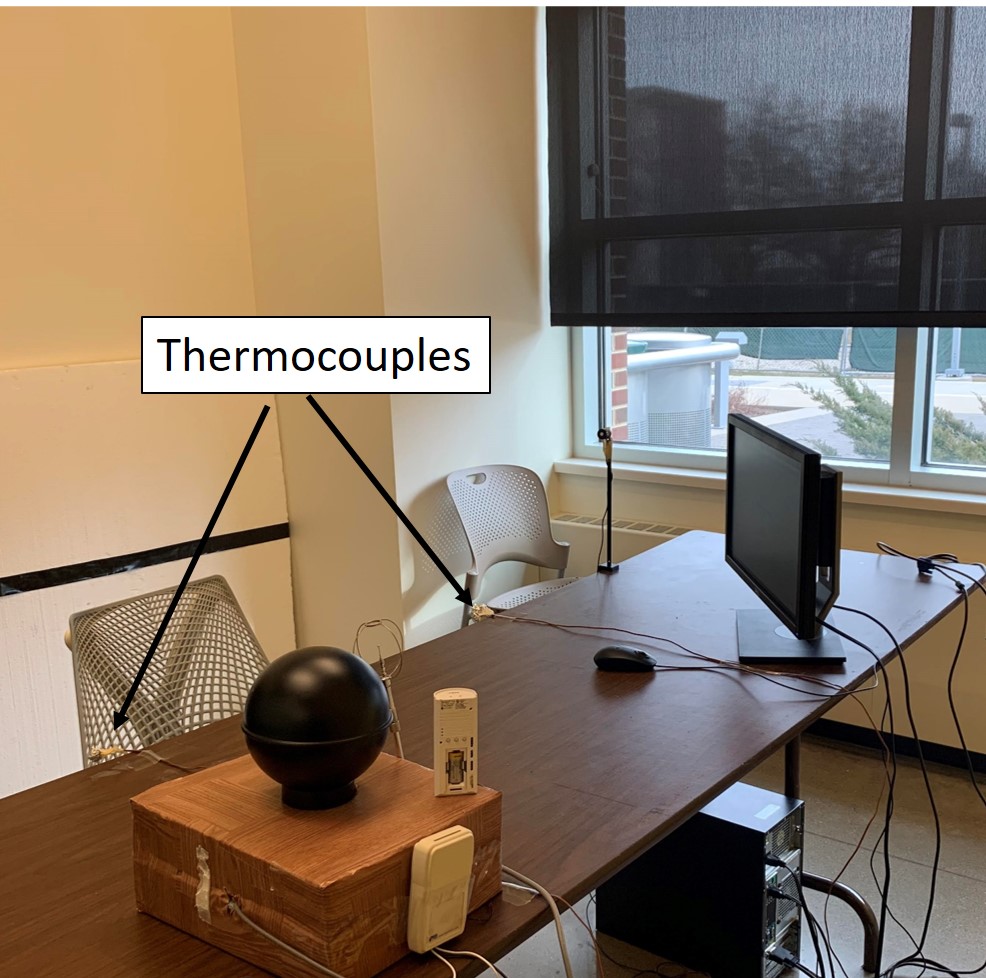}
        \caption{Sensor location and office layout}
        \label{fig:sensor_loc}
        \end{subfigure}
        \end{minipage}
        \hfil
        \begin{minipage}{0.4\linewidth}
        \begin{subfigure}{\linewidth}
        \includegraphics[width=\linewidth]{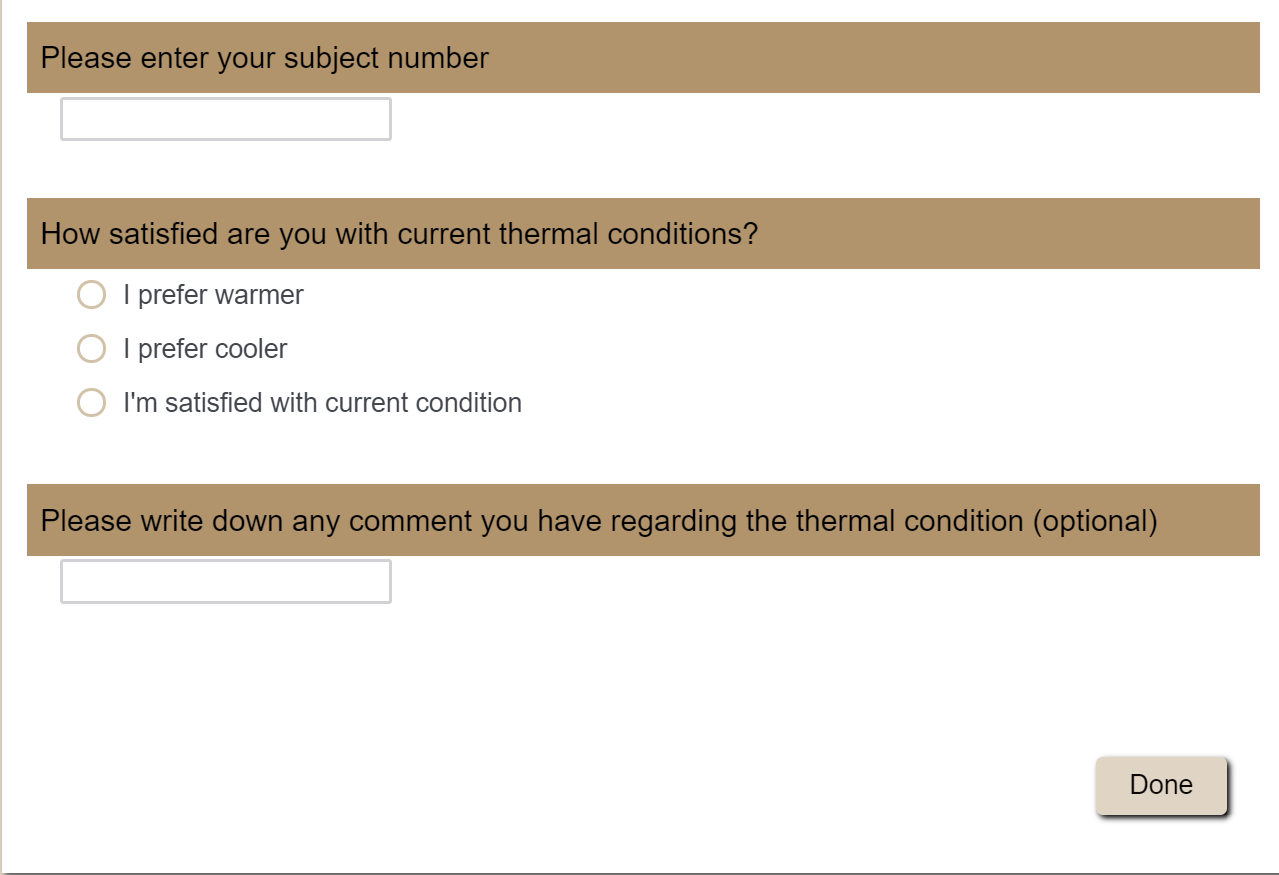}
        \caption{Survey questionnaire}
        \label{fig:survey}
        \end{subfigure}
        \end{minipage}
    \caption{Experimental setup}
    \end{figure}

We conducted real-time experiments in three identical south-facing private offices (3.3m x 3.7m x 3.2m high) in a building located in West Lafayette, Indiana. 
\Cref{fig:herrick_office} shows a general view of this building and the monitored offices.
The offices have one exterior curtain wall facade with 54\% window-to-wall ratio and high-performance glazing units with selective low-emissivity coating (normal visible transmittance: 70\%, normal solar transmittance: 33\%).
Dark-colored motorized interior roller shades are installed in the offices with a total normal visible transmittance of 2.53\% and an openness factor of 2.18\%. Each office has two electric lighting fixtures with two 32-W T5 fluorescent lamps (total of 128 W).  
The heating and cooling delivery to the spaces is achieved through a Variable Air Volume (VAV) system with a central air handling unit (AHU), which supplies conditioned air to the offices at a constant temperature, but variable flow rate.
Each office has a VAV box with zone damper that can modulate the supply air flow rate in the cooling mode and a reheat coil to increase the supply air temperature as needed to achieve the setpoint temperatures. The cooling and heating source (chilled water and steam) are provided from the campus plant. 
Furthermore, a Building Management System (BMS) is operated through the installed Tridium JACE controllers and Niagara/AX software framework, which in addition to a variety of internet-enabled features provides the ability to monitor, control and automate all of the building systems. 
During the course of this study, the air flow rate, supply air temperature, reheat coil mode, occupancy,  setpoint temperature, shading position, and electric light level were monitored. The sensors/actuators for these variables were connected with the BMS using BACnet protocol. 
The indoor air temperature (J-type thermocouples, resolution: $\dC{0.01}$, accuracy: 0.4\%) was measured at seating height and on two sides of the occupant's regular work position (see \cref{fig:sensor_loc}). They were placed in a closed proximity (less than 1 m) to the test subjects.
The average reading of two was used to reduce the influence of spatial temperature distribution. 
In addition, we monitored, the relative humidity (BAPI-Stat4, resolution: 1\%, accuracy: 2\%), air velocity (SENSOR anemometer, resolution: 0.01 m/s, accuracy: 1\%) and globe temperature ($\mu$-SMART black globe temperature sensor, resolution: $\dC{0.1}$, accuracy: 0.5\%) near the desks, transmitted solar irradiance (LI-COR 200-SL pyranometer, resolution: 0.1 W/m2, accuracy: 3\%) on the facade.
The sensors for these variables were connected with the National Instrument wireless data acquisition system, which communicated with the BMS using Modbus protocol.
The instruments used were installed in configurations that would not interfere with occupants’ positions or tasks.
In this study, we control the dimming level of the electric lights to obtain 300 lux on the workplane and fix the shade position at 25\% to eliminate the potential visual environment impacts on the thermal preference of test-subjects.
During the experiment, we deployed an automated control setup with no manual overrides allowed. This study was approved by the Institutional Review Board (IRB Protocol \#: 1503015873).

We conducted this study over the span of 23 days in October 2018. 
Overall, six test-subjects (20 - 35 years old; three males and three females), not familiar with this research, participated in this field study.
Each participant occupied one office everyday between 9:00 AM to 4:00 PM.
Upon arrival in the morning (9:00 AM), we set the indoor air temperature to the minimum temperature value that was achievable during that particular day (for most of the occupants, this temperature was $x_1 = \dC{21}$).
We asked all the occupants to perform their usual work routine (computer-related work, reading, writing etc.) during the day.
Also, we advised them to wear clothes they would normally wear while working inside offices (to marginalize out the impact of clothing on their thermal preferences).
They were free to take breaks to have lunch, use washrooms, etc., that is, breaks one would normally take while working in offices.
This setup was used to create realistic occupancy dynamics and allows for marginalizing out its impact on thermal preferences.
During the experiments, the setpoint indoor air temperature of the offices were automatically adjusted by the research team without giving override access to the participants. 
After 30 minutes, when the participants adopted to the steady state thermal conditions (also supported in findings by \cite{ghahramani2015online}), we asked them to answer a short web-based questionnaire (see \cref{fig:survey}), and based on their responses, the PE algorithm determined the next indoor air temperature to test for the participants. 

Before conducting the actual field studies with real occupants, we analyzed the impact of temperature difference constraints, temperature range to query over, the time difference between consecutive queries, etc.,  by running pilot experiments using expert occupants (authors of this paper and other researchers from our group). After conducting these preliminary pilot studies, the indoor air temperatures we decided to test over were from $\dC{20}$ to $\dC{28}$ (supported in previous studies by \cite{guenther2019feature,lee2017bayesian}), with the minimum perceivable temperature difference set to be $\dC{0.5}$ (supported in a previous study by \cite{wang2018individual}). 
That is, we are going to test $\calX_n  \subset \{20, 20.5, 21, 21.5 ... . ,28\}$ for all queries $n=1,\dots,N$.
Also, to avoid drastic difference in indoor air temperature values (i.e. to avoid going from $\dC{20}$ to $\dC{28}$ and vice versa), we constrained the search to temperatures $\pm$\SI{3}{\celsius} from the current value.

%\subsection{Limitations of our PE framework}\label{subsec:exp_pe_limitations}
%One of the limitations of our framework is the failure to capture evolution of occupants’ preferences over time/seasons.
%We have acknowledged this issue and when it comes to addressing such dynamic variations in preferences (across seasons), we suggest readers to follow a brute force strategy in which to run the PE framework for each of the seasons and quantify thermal preferences for each season separately.
%In this paper, we have focused on the development of a simple, robust, low-cost, easy to compute and easy to implement personalized PE framework which focuses on room temperature as its most important feature and as such have not included other latent features affecting preferences (time/seasonal factors such as metabolic rate, clothing level etc.).
%Our future work involves addressing these limitations by extending this PE framework to incorporate temporal effects \cite{liu2015modeling, gao2017collaborative} into our GP-PE based latent factor model.

\subsection{Practical Elicitation Stopping Criteria for Experiment Design}
Collecting sufficient data that expresses occupants' satisfaction for different thermal environment is critical. 
In this work, we sequentially collected thermal preference query (survey) data (based on the next query selected using PE) to infer personalized maximally preferred indoor air temperatures for different occupants. 
In this regard, one question which naturally arises in our minds is: ``When should we stop our PE run? How do we ensure that we have collected sufficient amount of data to infer the above mentioned quantity of interest?'' An ideal way to answer these questions would be to pose such questions from a Bayesian decision theoretic perspective i.e. ``what is the expected loss we would incur if we were to stop the PE run at iteration $i$?'' \cite{gelman2013bayesian}. In this way, in an ideal setting (when we are able to perfectly quantify the losses), we would stop the run when the expected loss becomes less than a specific threshold value.

However, when it comes to thermal preferences of building occupants, it is difficult to objectively quantify losses. Therefore, in this work, we follow more of a heuristic based approach to stop our PE run (collection of data). In a practical experimental setting, we will stop collecting the query data at iteration $i$, when either of the below mentioned conditions are met:
\begin{enumerate}
    \item EUI values obtained at $i$, $i-1$ and $i-2$ iterations is less than $0.01$ ; elicited temperature value has converged to a specific temperature and thermal preference query responses $y_{i}$ and $y_{i-1}$ are $0$ (i.e. there are satisfied with the current indoor air temperature).
    \item EUI ratio values obtained at $i$, $i-1$ were less than $1$ ;  elicited temperature value has converged to a specific temperature and query responses $y_{i}$ and $y_{i-1}$ are $0$ (i.e. there are satisfied with the current indoor air temperature).
    \item Elicitation budget ($N = 10$) is exhausted.
\end{enumerate}

\subsection{Analysis of the results}
\label{subsec:analysis of results}
Following the same design of experiments strategy discussed in the sub-section \ref{subsec:outline_pe}, we elicit the maximally preferred  indoor air temperature values for six new real occupants by actively asking them questions regarding their thermal preferences.
We summarize the obtained thermal preference dataset (after running the PE scheme) in Table~\ref{table:real_occupant1} and Table~\ref{table:real_occupant2}.
The posterior predictive distributions over utility functions and the location of maximum preferred indoor air temperature values (after PE scheme has completed its run) in case of real occupants 1-6 are
depicted in \Cref{fig:real_occupants}.
In case of real occupant 1, the maximum preferred indoor air temperature was  inferred to be $\dC{24}$ (see \Cref{fig:r1_m}).
Samples from the posterior predictive distribution over utility function for this occupant (conditioned on the five thermal preference queries collected after running PE scheme) are also shown in \Cref{fig:r1_u2} and \cref{fig:r1_u3}.
Each of the posterior utility samples represents the collected thermal preference data and as seen, due to the unimodality constraints imposed in the GP prior, the derivative of utility function changes sign from positive to negative at $\dC{24}$ (in case of real occupant 1).
Similarly, \Crefrange{fig:r2_u2}{fig:r6_m} represent posterior utility samples and the maximally preferred indoor air temperature values in case of real occupants 2-6, respectively.

The 95\% credible intervals \cite{gelman2013bayesian} of the maximally preferred indoor air temperature values (for each of the real occupants) after the completion of elicitation run are shown in \Cref{table:real_occupant_credible}.
From \Cref{table:real_occupant_credible} and \Crefrange{fig:r1_u2}{fig:r1_m}, we can say with 95\% probability that the maximum preferred indoor air temperature value for occupant 1 would lie in the range of $23.4-\dC{24.7}$.
Since this occupant was quite sensitive to change in indoor air temperature, we were able to infer this maximally preferred indoor air temperature value  in only five thermal preference queries to the occupant.
Similarly, with 95\% probability, the maximally preferred indoor air temperature value of occupant 2 was inferred to lie in between $21.9-\dC{23.1}$.
This occupant was also quite sensitive to change in indoor air temperature values and, therefore, we only needed six thermal preference queries to get sufficiently close to the maximally preferred indoor air temperature values.
In case of occupant 3, the 95\% credible interval of the posterior distribution over maximally preferred indoor air temperature was inferred to be $22-\dC{24}$.
This occupant was less sensitive to temperature changes as compared to occupants 1\&2.
Similarly, with 95\% probability, the maximally preferred indoor air temperature value of occupant 4 lay in $22.6-\dC{24}$.
Finally, we inferred that occupants 5\&6 were the least sensitive to temperature changes.
They exhibited no specific maximally preferred indoor air temperature.
PE framework inferred that occupant 5 prefers any indoor air temperature in $20-\dC{23.5}$ and occupant 6 any value in $22.8-\dC{26.8}$.

\begin{table}[H]
\centering
\resizebox{\textwidth}{!}{
\begin{tabular}{ccccccccccccc}
\toprule
\multirow{2}{*}{Query \#} &
\multicolumn{4}{c}{Real occupant 1} &
\multicolumn{4}{c}{Real occupant 2} &
\multicolumn{4}{c}{Real occupant 3} \\
\cmidrule(lr){2-5} 
\cmidrule(lr){6-9}
\cmidrule(lr){10-13}
& {Temp. ($\dC{}$)} & {Response} & {EUI} & {EUI Ratio} & {Temp. ($\dC{}$)} & {Response} & {EUI} & {EUI Ratio} &
{Temp. ($\dC{}$)} & {Response} & {EUI} & {EUI Ratio}\\
\midrule

1  & 21  & 1 & 0.299 & - &
21.5 &  1 & 0.276 & - &
22 & 1 & 0.296 & -\\

2  & 23.5 & 1 & 0.012 & 24.91 &
24 & -1 & 0.228 & 1.19 &
24.5 & -1 & 0.255 & 1.16\\

3 & 26 & -1 & 0.014 & 0.86 &
23 & -1 & 0.009 & 25.33 & 
23.5 & 0 & 0.229 & 1.11\\

4 & 24.5 & -1 & 0.006 & 2.19 &
22.5 & 0 & 0.008 & 1.12 &
23 & 0 & 0.253 & 0.90\\

5 & 24 & 0$^*$ & 0.006 & 1 &
22 & 1 & 0.008 & 1 &
22.5 & 0 & 0.220 & 1.15\\

6 & 24 & 0$^*$ & 0.006 & 1 &
22.5 & 0$^*$ & 0.008 & 1 & 
23.5 & -1 & 0.210 & 1.04\\

7 & 24 & 0 & 0.006 & 1 & 
22.5 & 0 & 0.008 & 1 &
23 & 0$^*$ & 0.279 & 0.752\\

8 & - & - & - & - &
- & - & - & - &
23 & 0$^*$ & 0.284 & 0.982\\
\midrule[\heavyrulewidth]
\multicolumn{13}{l}{\footnotesize$^*$ additional information is provided by the occupant (see \cref{table:additional_comments_1})} \\
\bottomrule
\end{tabular}
}
\caption{Real occupants' preference data collected using the newly developed PE framework.}
\label{table:real_occupant1}
\end{table}

\begin{table}[H]
\centering
\resizebox{\textwidth}{!}{
\begin{tabular}{ccccccccccccc}
\toprule
\multirow{2}{*}{Query \#} &
\multicolumn{4}{c}{Real occupant 4} &
\multicolumn{4}{c}{Real occupant 5} &
\multicolumn{4}{c}{Real occupant 6} \\
\cmidrule(lr){2-5} 
\cmidrule(lr){6-9}
\cmidrule(lr){10-13}
& {Temp. ($\dC{}$)} & {Response} & {EUI} & {EUI Ratio} & {Temp. ($\dC{}$)} & {Response} & {EUI} & {EUI Ratio} &
{Temp. ($\dC{}$)} & {Response} & {EUI} & {EUI Ratio}\\
\midrule

1  & 22  & 0$^*$ & 0.269 & - &
21 &  0 & 0.409 & - &
21.5 & 1 & 0.320 & -\\

2  & 24 & -1 & 0.250 & 1.076 &
23.5 & -1 & 0.233 & 1.74 &
23.5 & 1$^*$ & 0.335 & 0.95\\

3 & 21 & 1 & 0.215 & 1.16 &
23 & 0 & 0.295 & 0.79 & 
26 & 0$^*$ & 0.225 & 1.49\\

4 & 23 & 0 & 0.253 & 0.849 &
21 & 0 & 0.235 & 1.25 &
26.5 & 0 & 0.257 & 0.87\\

5 & 21.5 & 1 & 0.239 & 1.05 &
20 & 1 & 0.216 & 1.08 &
28 & -1$^*$ & 0.213 & 1.20\\

6 & 22.5 & 1 & 0.008 & 29.87 &
22 & 0$^*$ & 0.229 & 0.94 & 
26 & 0 & 0.235 & 0.90\\

7 & 23 & 0$^*$ & 0.008 & 1 & 
22.5 & 0 & 0.233 & 0.98 &
27 & -1 & 0.20 & 1.17\\

8 & 23.5 & 0$^*$ & 0.007 & 1.14 &
21.5 & 0$^*$ & 0.214 & 1.08 &
26.5 & 0$^*$ & 0.196 & 1.02\\

9 & 23.5 & 0 & 0.008 & 0.87 &
22.5 & 0$^*$ & 0.237 & 0.90 &
24.5 & 0$^*$ & 0.30 & 0.65\\

10 & 23 & 0$^*$ & 0.006 & 1.33 &
22.5 & 0 & 0.241 & 0.98 &
25 & 0$^*$ & 0.248 & 1.20\\

\midrule[\heavyrulewidth]
\multicolumn{13}{l}{\footnotesize$^*$ additional information is provided by the occupant (see \cref{table:additional_comments_2})} \\
\bottomrule
\end{tabular}
}
\caption{Real occupants' preference data collected using the newly developed PE framework.}
\label{table:real_occupant2}
\end{table}

% Real occupants data table 3
\begin{table}[H]
    \centering
    \begin{tabular}{cccc}
    \toprule
    \multirow{2}{*}{Real occupant} &
    \multirow{2}{*}{Number of queries} &
    \multicolumn{2}{c}{Maximum preferred indoor air temp. ($\dC{}$)} \\\cmidrule{3-4}
    & & {Median} & {95\% credible interval} \\
    \midrule
        1 & 5 & 24.0 & 23.4 - 24.7 \\
        2 & 6 & 22.5 & 21.9 - 23.1 \\
        3 & 7 & 23.1 & 22.0 - 24.0 \\
        4 & 8 & 23.3 & 22.6 - 24.0 \\
        5 & 10 & 22.3 & 20.2 - 23.4 \\
        6 & 10 & 25.0 & 22.8 - 26.8 \\
    \end{tabular}
    \caption{Posterior maximally preferred indoor air temperatures for real occupants.}
    \label{table:real_occupant_credible}
\end{table}

\begin{figure}[H]
  \centering
  \begin{subfigure}[b]{0.32\linewidth}
    \includegraphics[width=\linewidth]{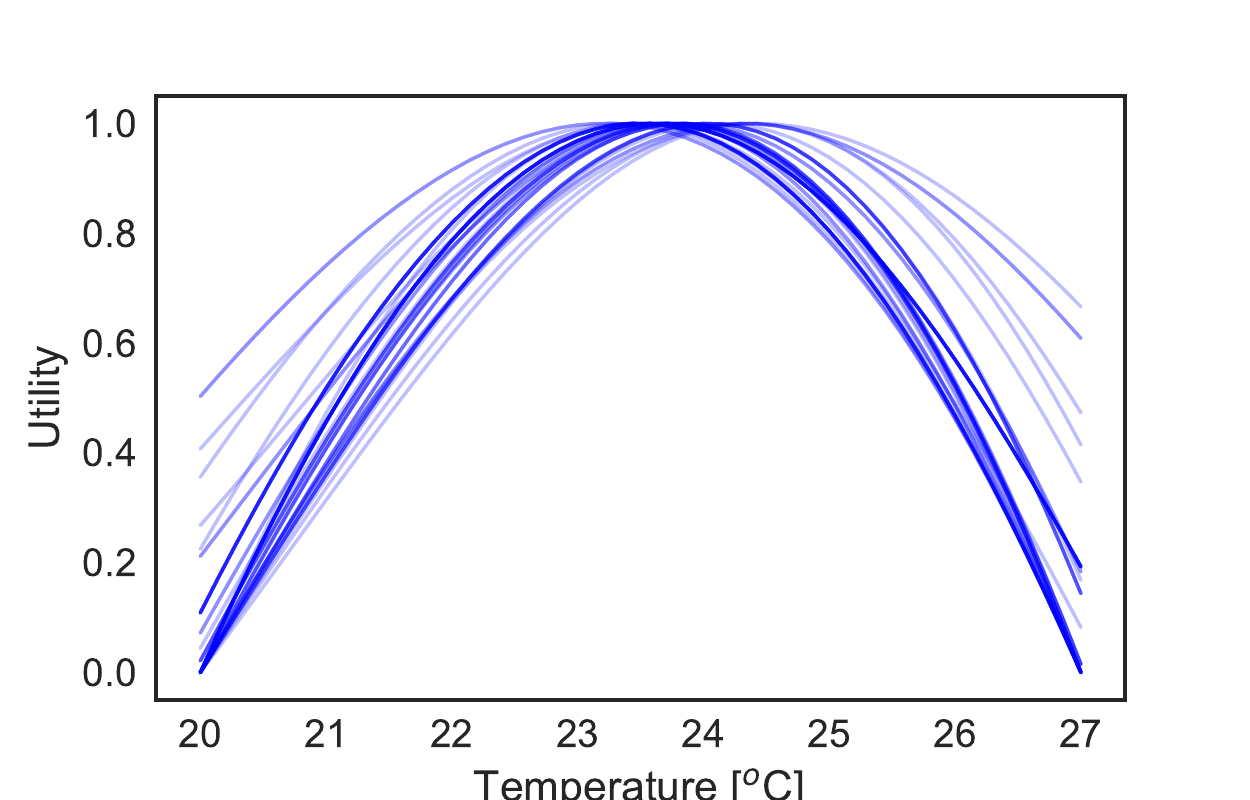}
     \caption{utility samples (occupant 1)}
     \label{fig:r1_u2}
  \end{subfigure}
  \begin{subfigure}[b]{0.32\linewidth}
    \includegraphics[width=\linewidth]{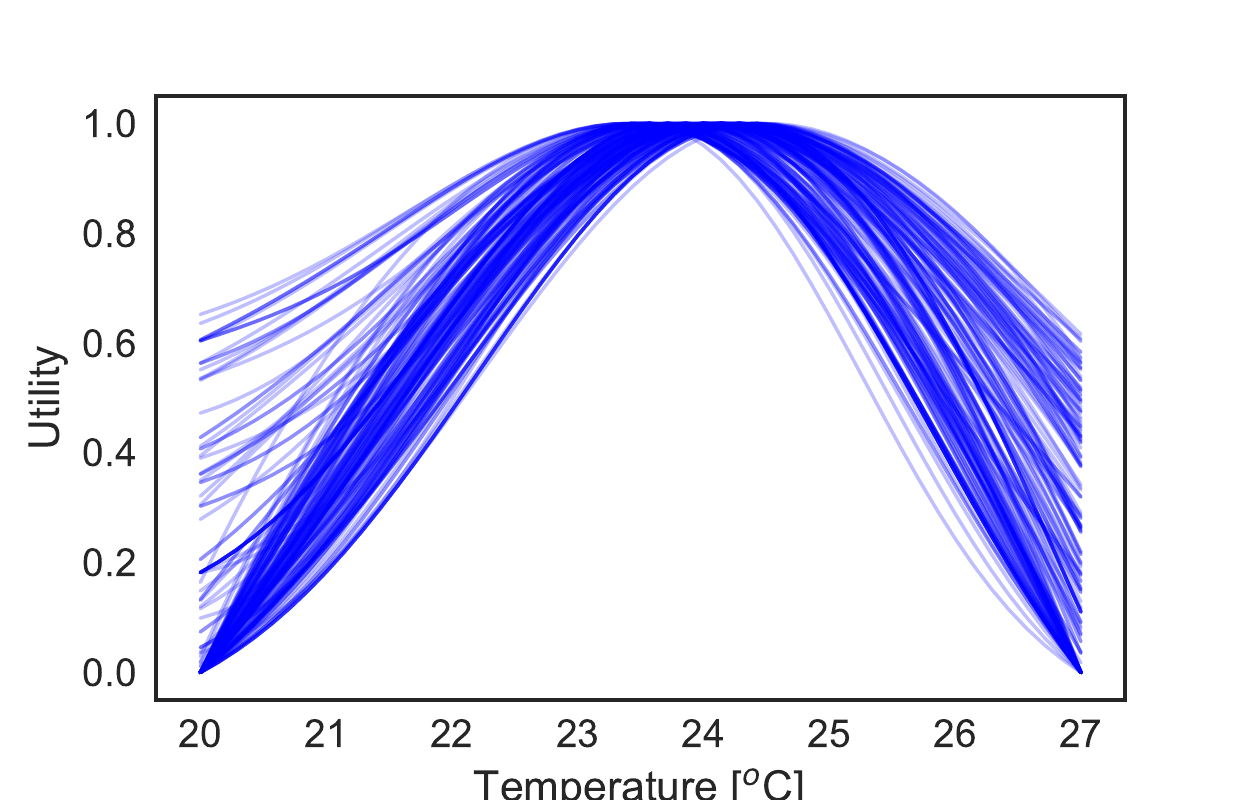}
    \caption{utility samples (occupant 1)}
    \label{fig:r1_u3}
  \end{subfigure}
  \begin{subfigure}[b]{0.32\linewidth}
    \includegraphics[width=\linewidth]{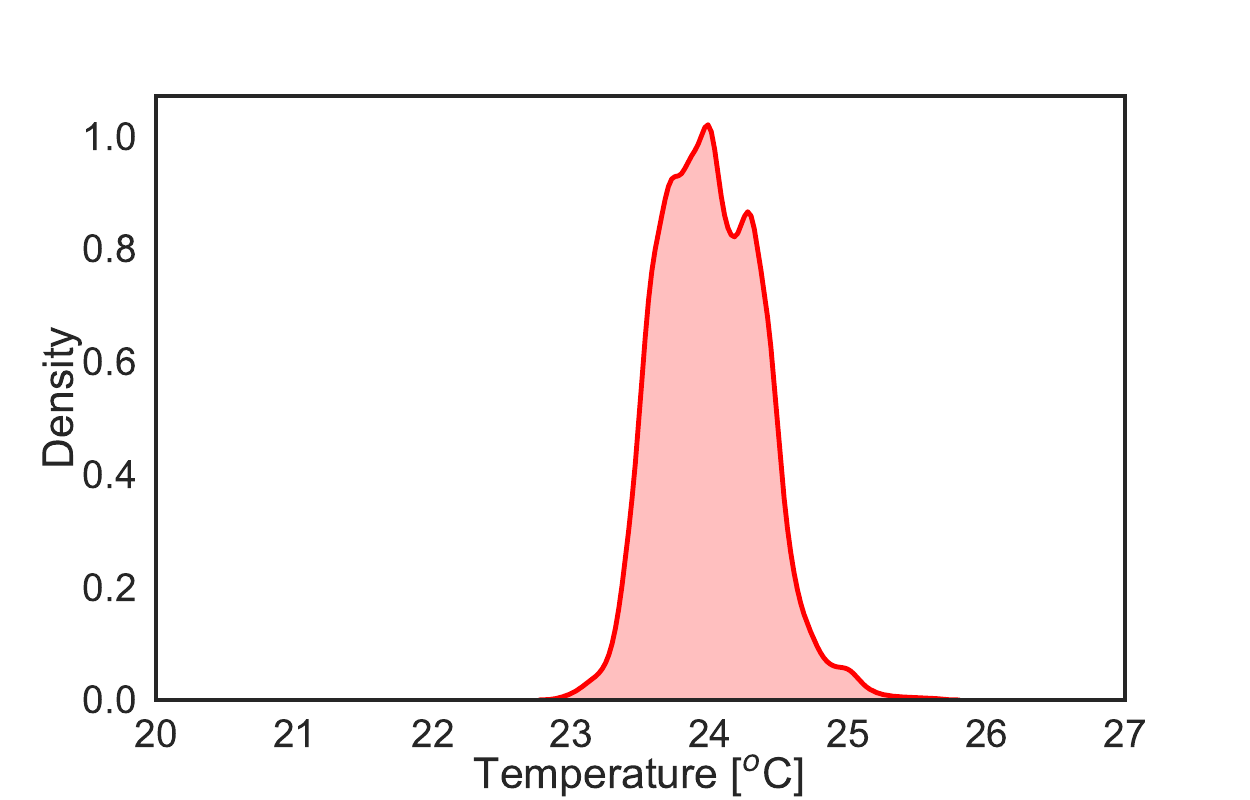}
    \caption{max. preferred temp. (occupant 1)}
    \label{fig:r1_m}
  \end{subfigure}
  
 \begin{subfigure}[b]{0.32\linewidth}
    \includegraphics[width=\linewidth]{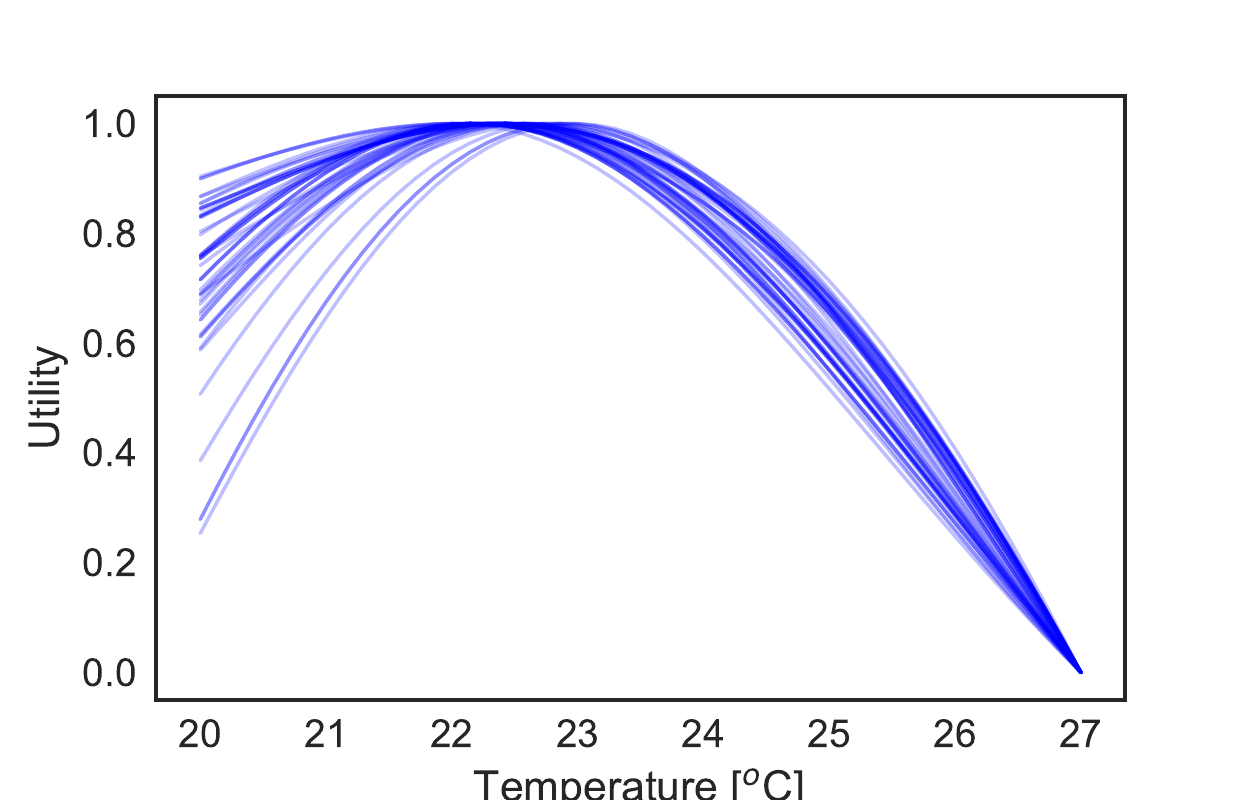}
    \caption{utility samples (occupant 2)}
    \label{fig:r2_u2}
  \end{subfigure}
  \begin{subfigure}[b]{0.32\linewidth}
    \includegraphics[width=\linewidth]{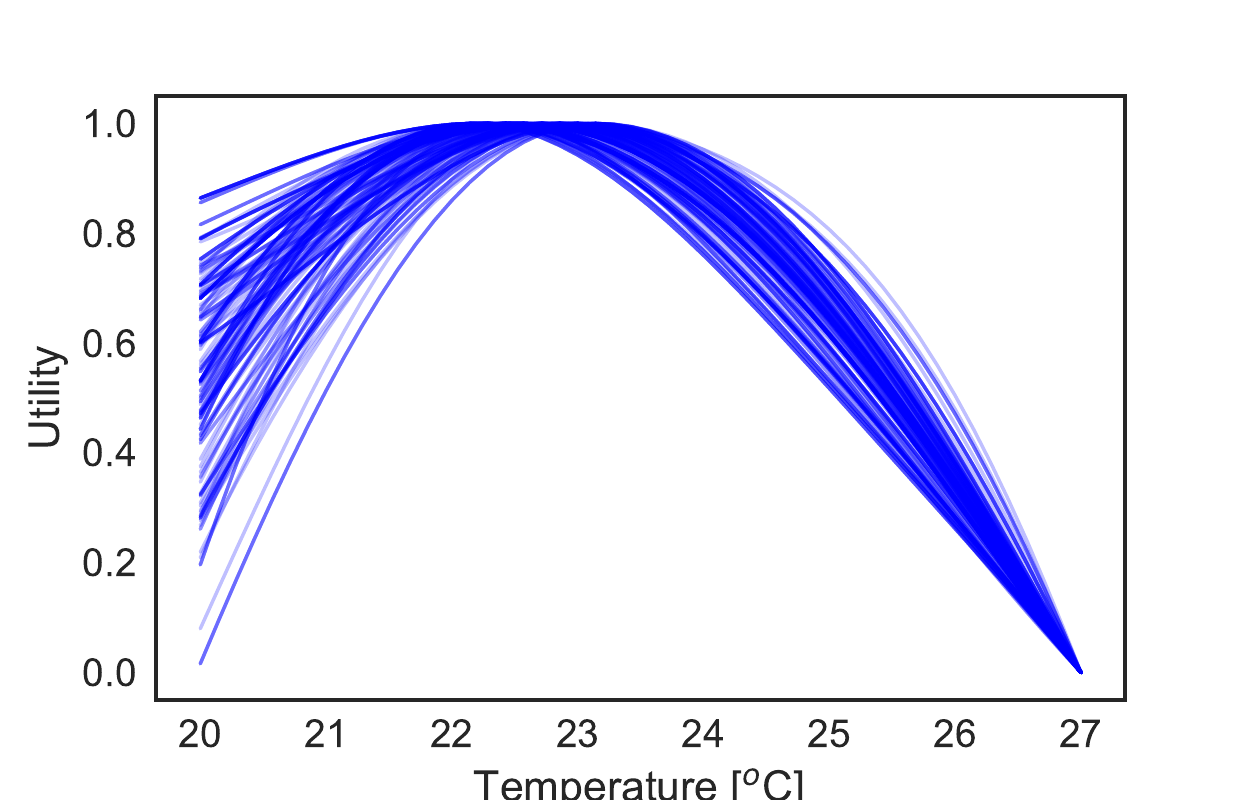}
    \caption{utility samples (occupant 2)}
    \label{fig:r2_u3}
  \end{subfigure}
  \begin{subfigure}[b]{0.32\linewidth}
    \includegraphics[width=\linewidth]{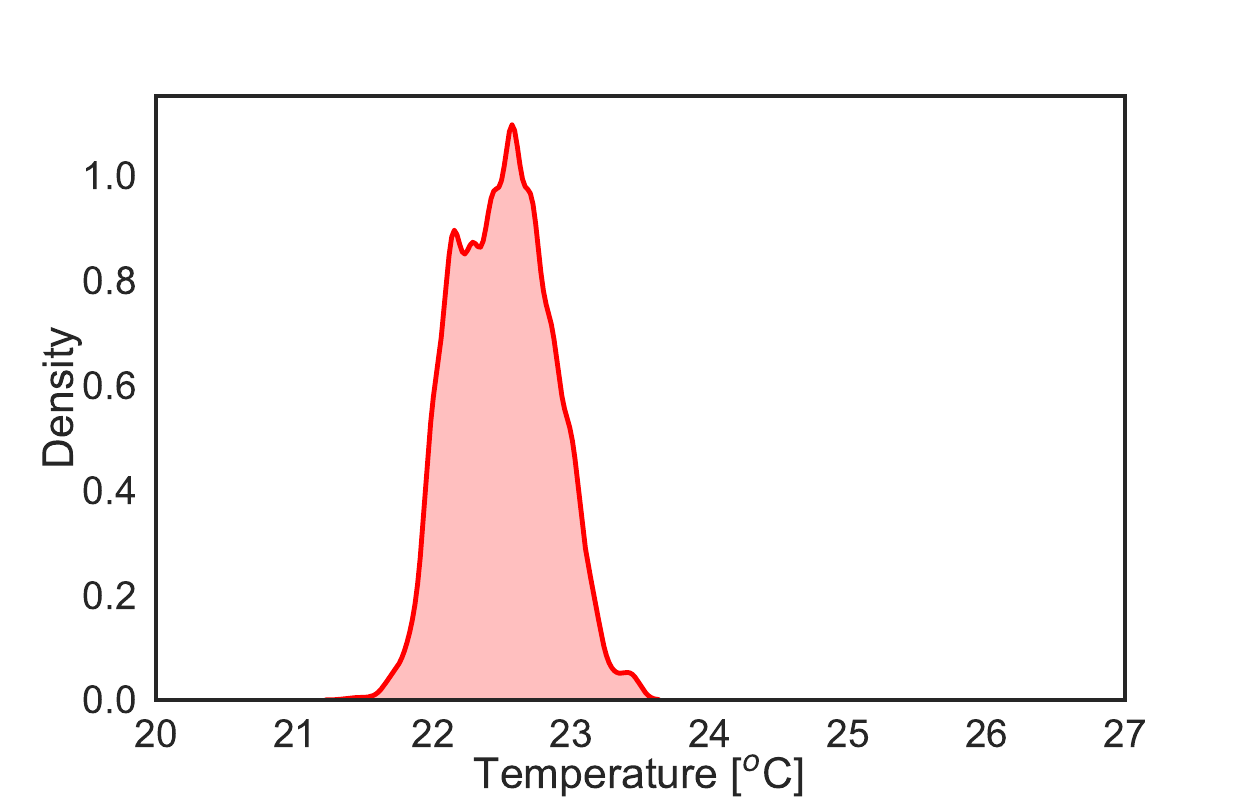}
    \caption{max. preferred temp. (occupant 2)}
    \label{fig:r2_m}
  \end{subfigure}
  \end{figure}

\begin{figure}[H]
 \ContinuedFloat
  \centering
  \begin{subfigure}[b]{0.32\linewidth}
    \includegraphics[width=\linewidth]{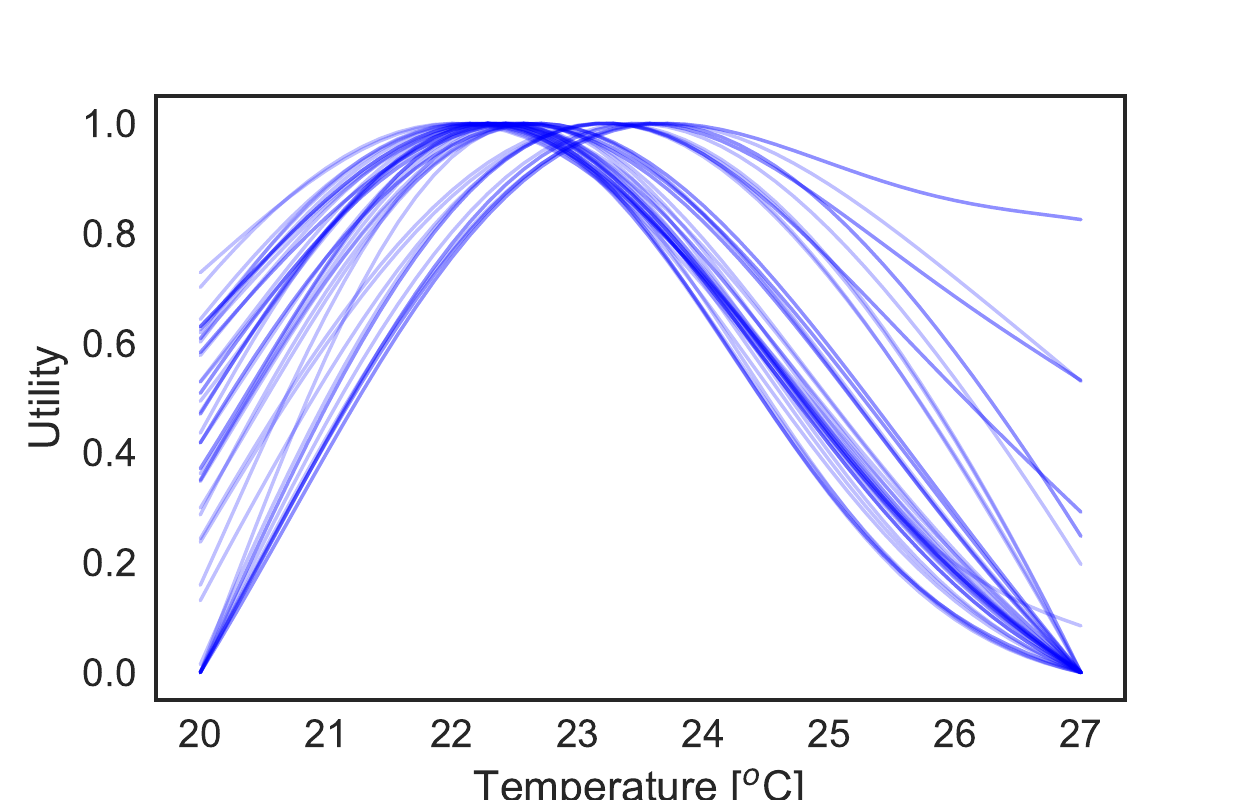}
    \caption{utility samples (occupant 3)}
    \label{fig:r3_u2}
  \end{subfigure}
  \begin{subfigure}[b]{0.32\linewidth}
    \includegraphics[width=\linewidth]{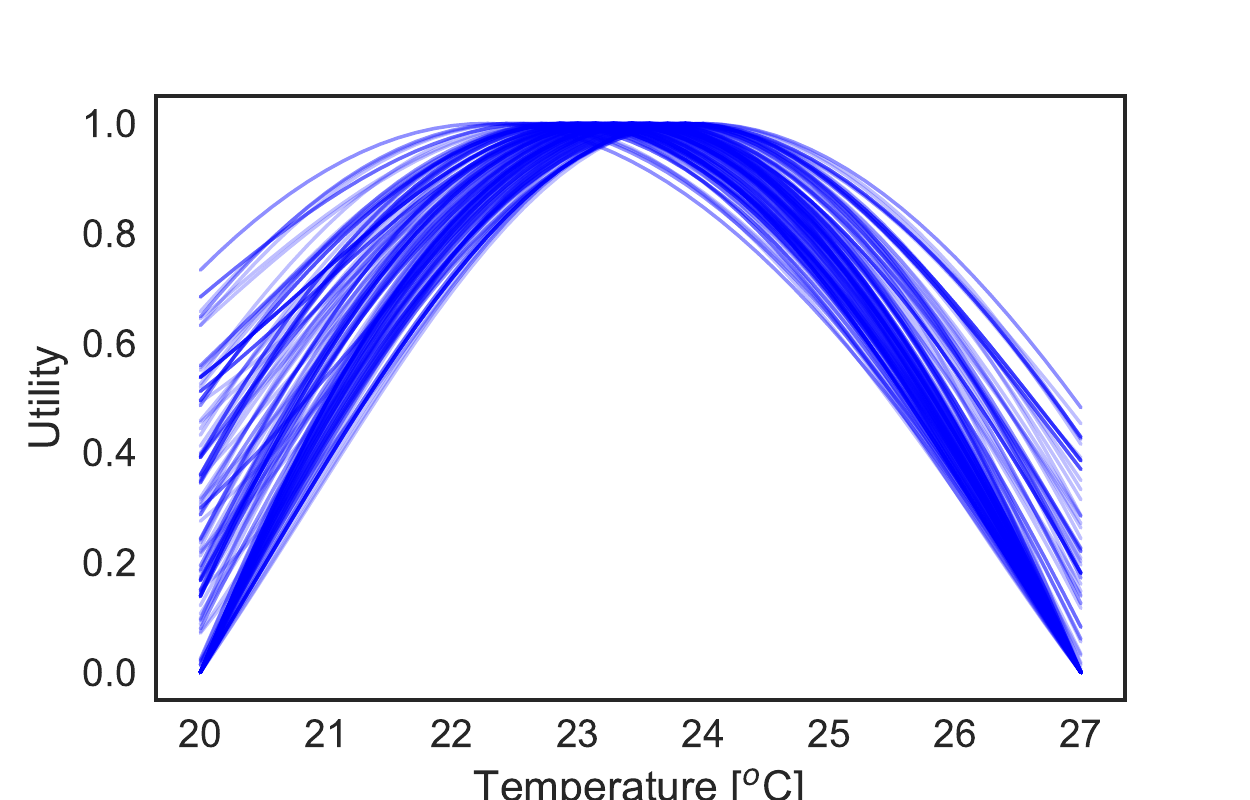}
    \caption{utility samples (occupant 3)}
    \label{fig:r3_u3}
  \end{subfigure}
  \begin{subfigure}[b]{0.32\linewidth}
    \includegraphics[width=\linewidth]{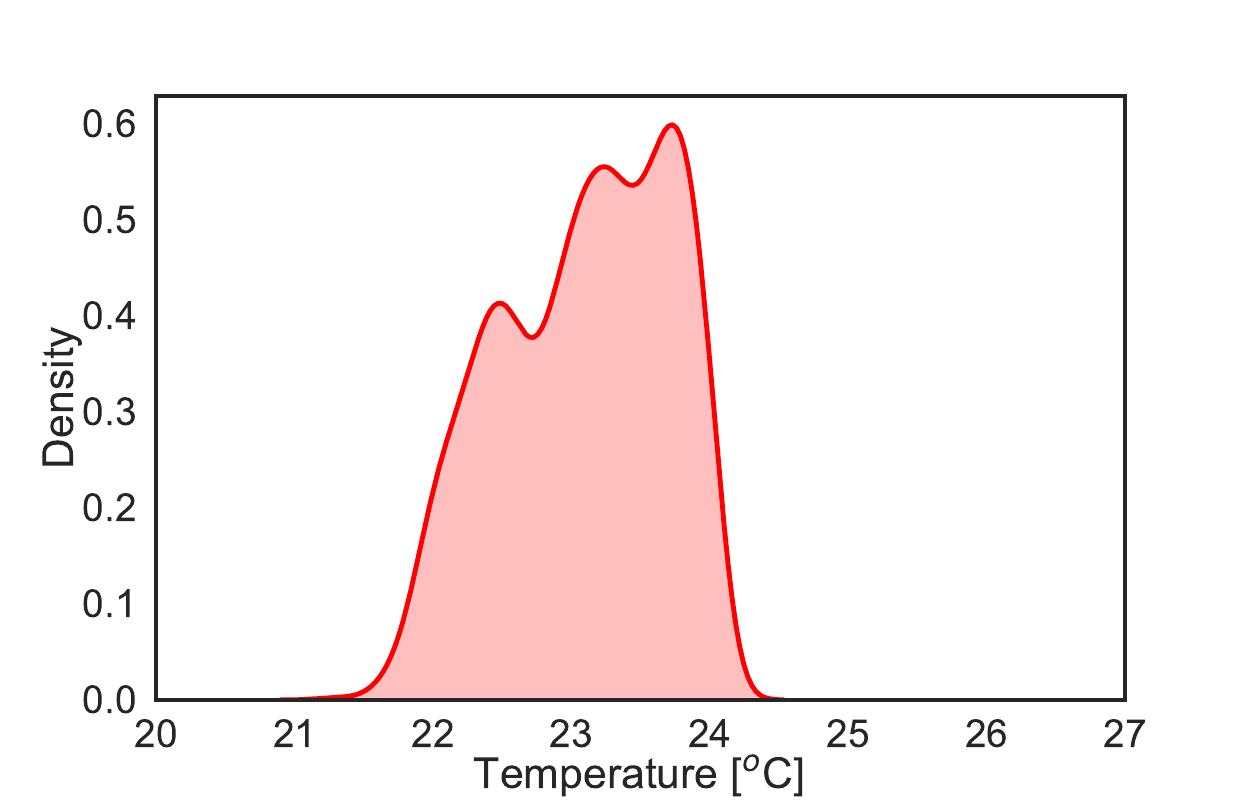}
    \caption{max. preferred temp. (occupant 3)}
    \label{fig:r3_m}
  \end{subfigure}
  
  \begin{subfigure}[b]{0.32\linewidth}
    \includegraphics[width=\linewidth]{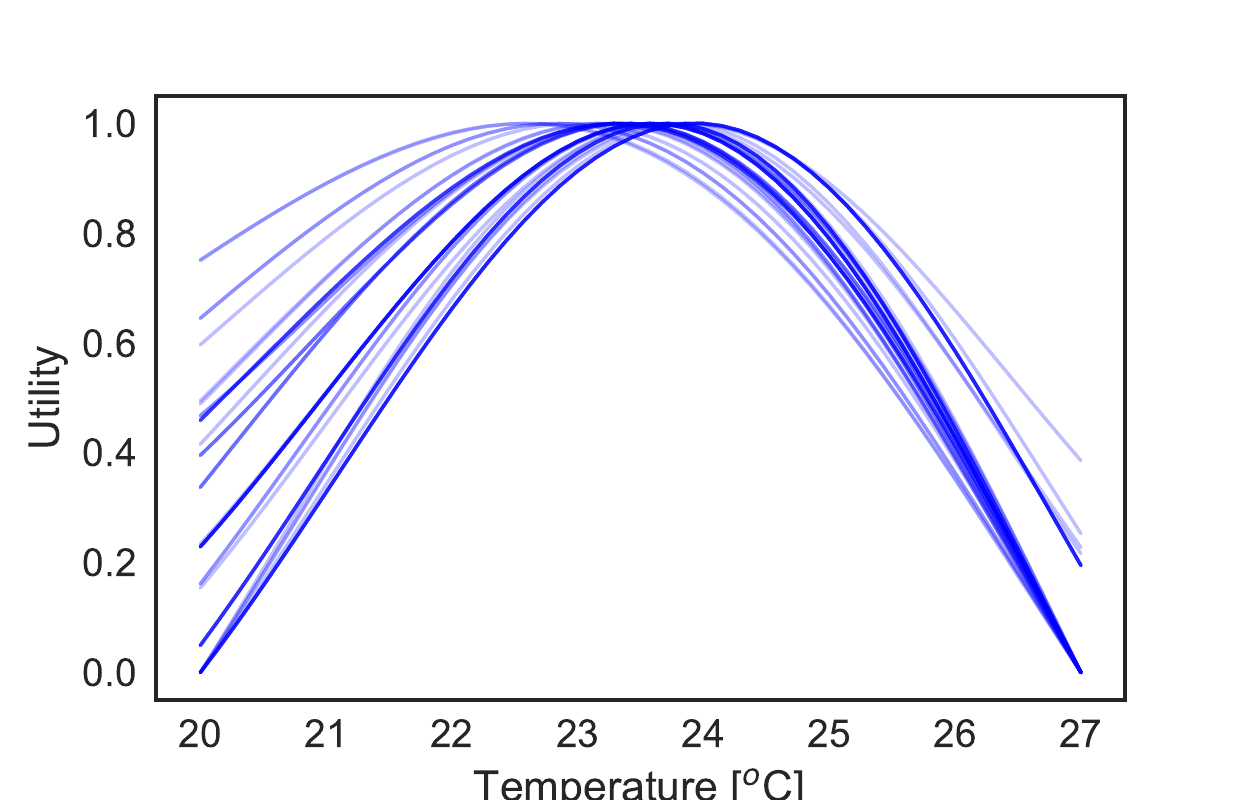}
    \caption{utility samples (occupant 4)}
    \label{fig:r4_u2}
  \end{subfigure}
  \begin{subfigure}[b]{0.32\linewidth}
    \includegraphics[width=\linewidth]{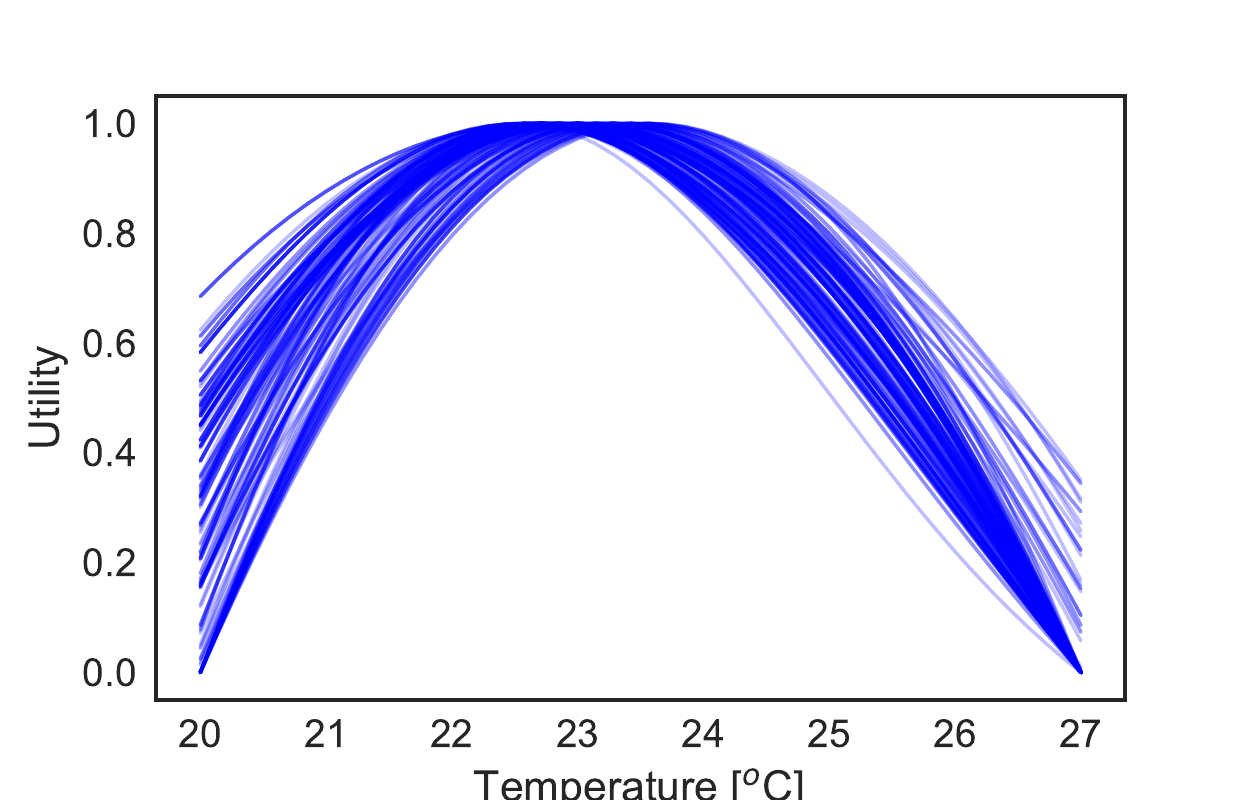}
    \caption{utility samples (occupant 4)}
    \label{fig:r4_u3}
  \end{subfigure}
  \begin{subfigure}[b]{0.32\linewidth}
    \includegraphics[width=\linewidth]{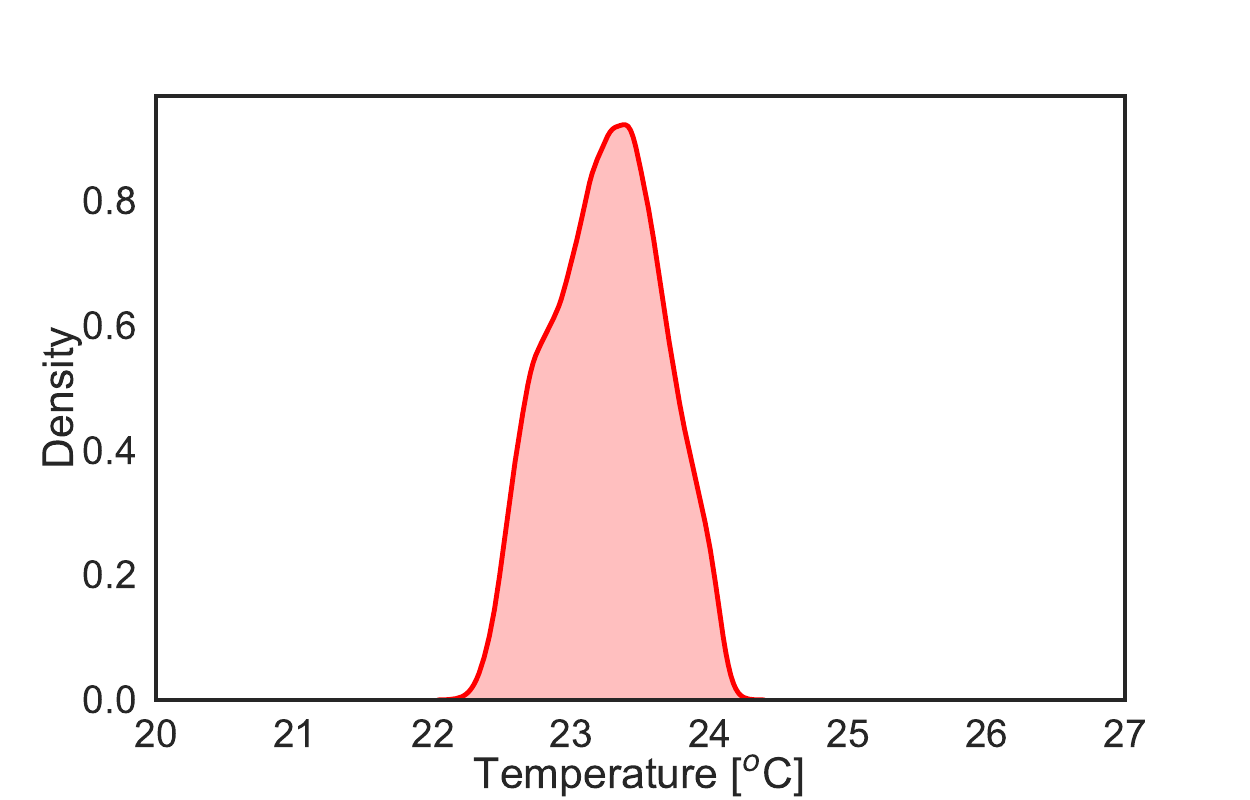}
    \caption{max. preferred temp. (occupant 4)}
    \label{fig:r4_m}
  \end{subfigure}
  \begin{subfigure}[b]{0.32\linewidth}
    \includegraphics[width=\linewidth]{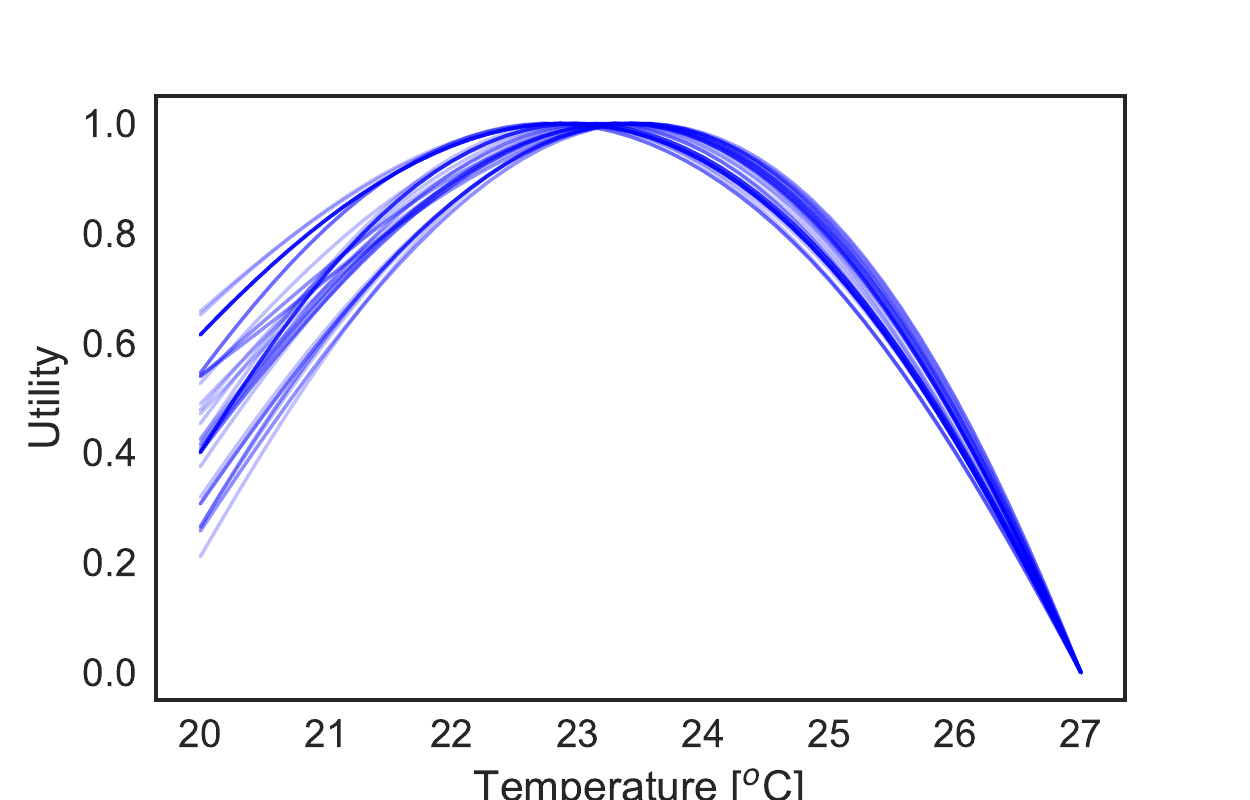}
    \caption{utility samples (occupant 5)}
    \label{fig:r5_u2}
  \end{subfigure}
  \begin{subfigure}[b]{0.32\linewidth}
    \includegraphics[width=\linewidth]{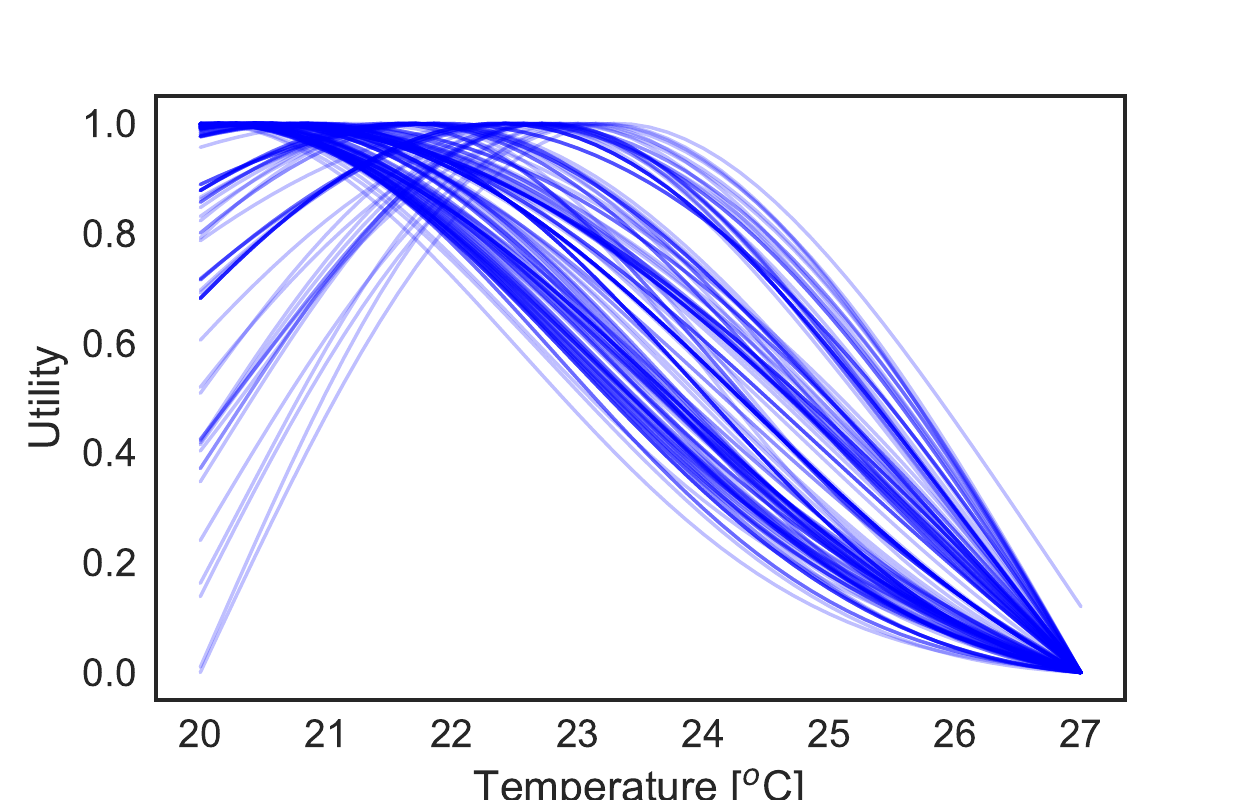}
    \caption{utility samples (occupant 5)}
    \label{fig:r5_u3}
  \end{subfigure}
  \begin{subfigure}[b]{0.32\linewidth}
    \includegraphics[width=\linewidth]{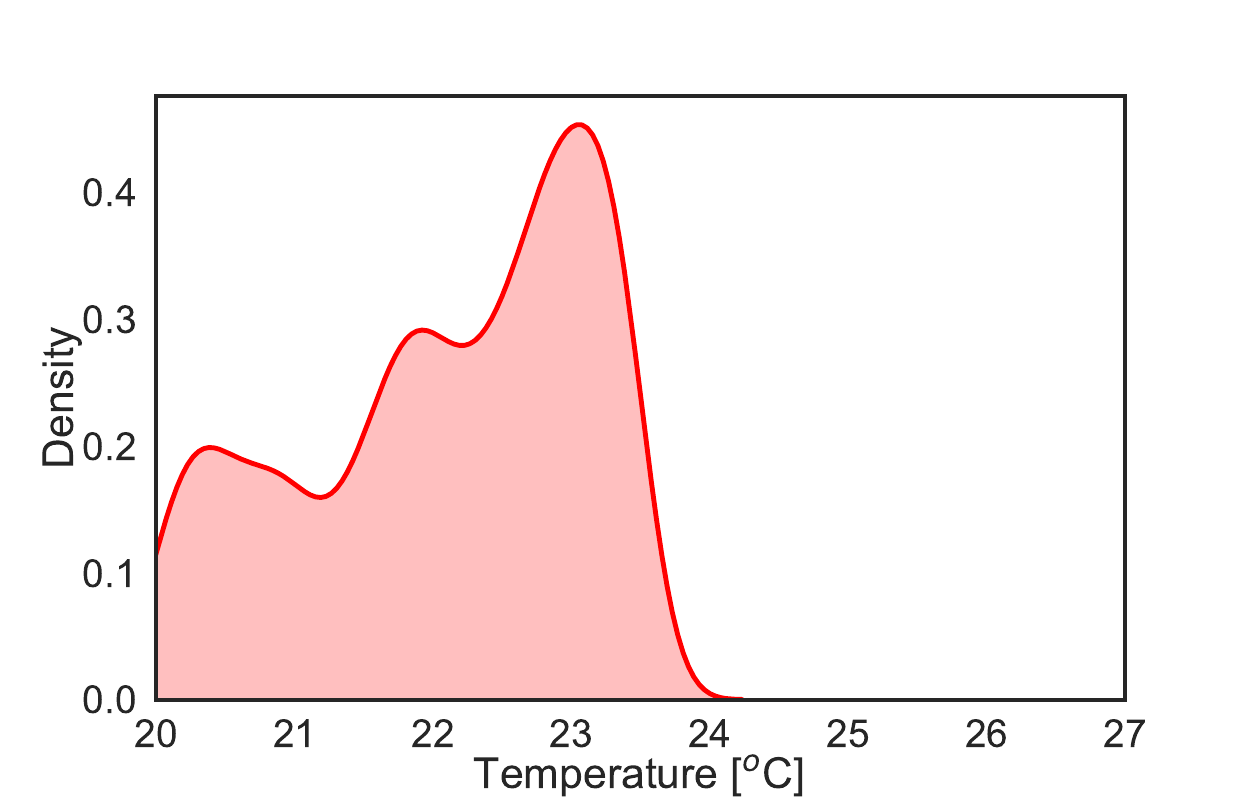}
    \caption{max. preferred temp. (occupant 5)}
    \label{fig:r5_m}
  \end{subfigure}
  
   \begin{subfigure}[b]{0.32\linewidth}
    \includegraphics[width=\linewidth]{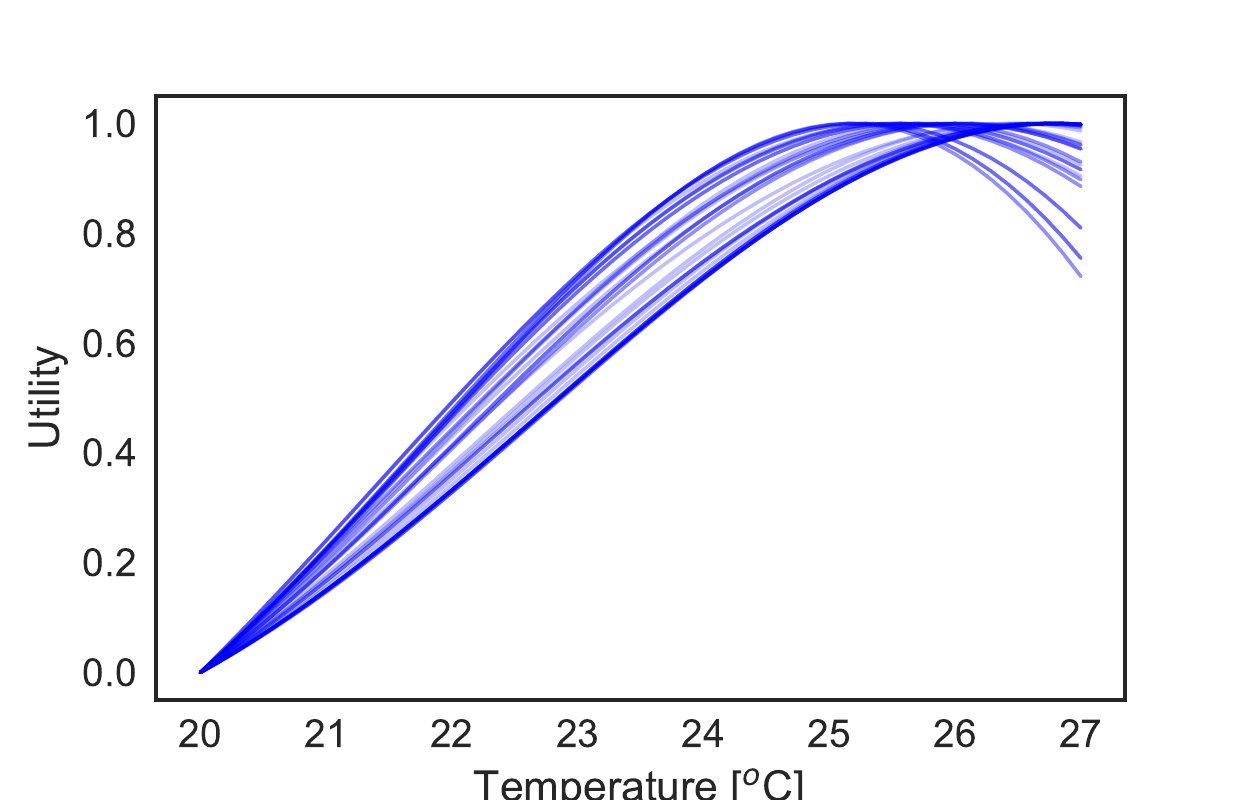}
    \caption{utility samples (occupant 6)}
    \label{fig:r6_u2}
  \end{subfigure}
  \begin{subfigure}[b]{0.32\linewidth}
    \includegraphics[width=\linewidth]{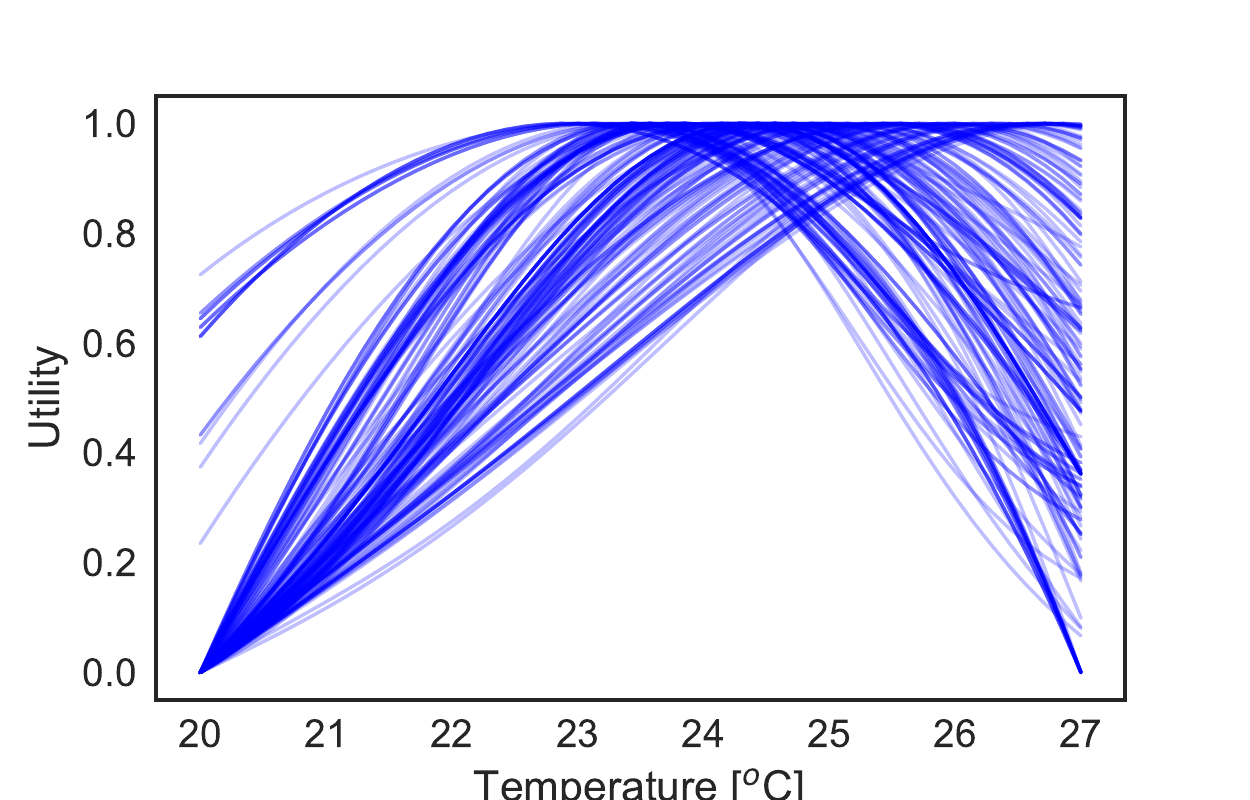}
    \caption{utility samples (occupant 6)}
    \label{fig:r6_u3}
  \end{subfigure}
  \begin{subfigure}[b]{0.32\linewidth}\label{fig:r6_m}
    \includegraphics[width=\linewidth]{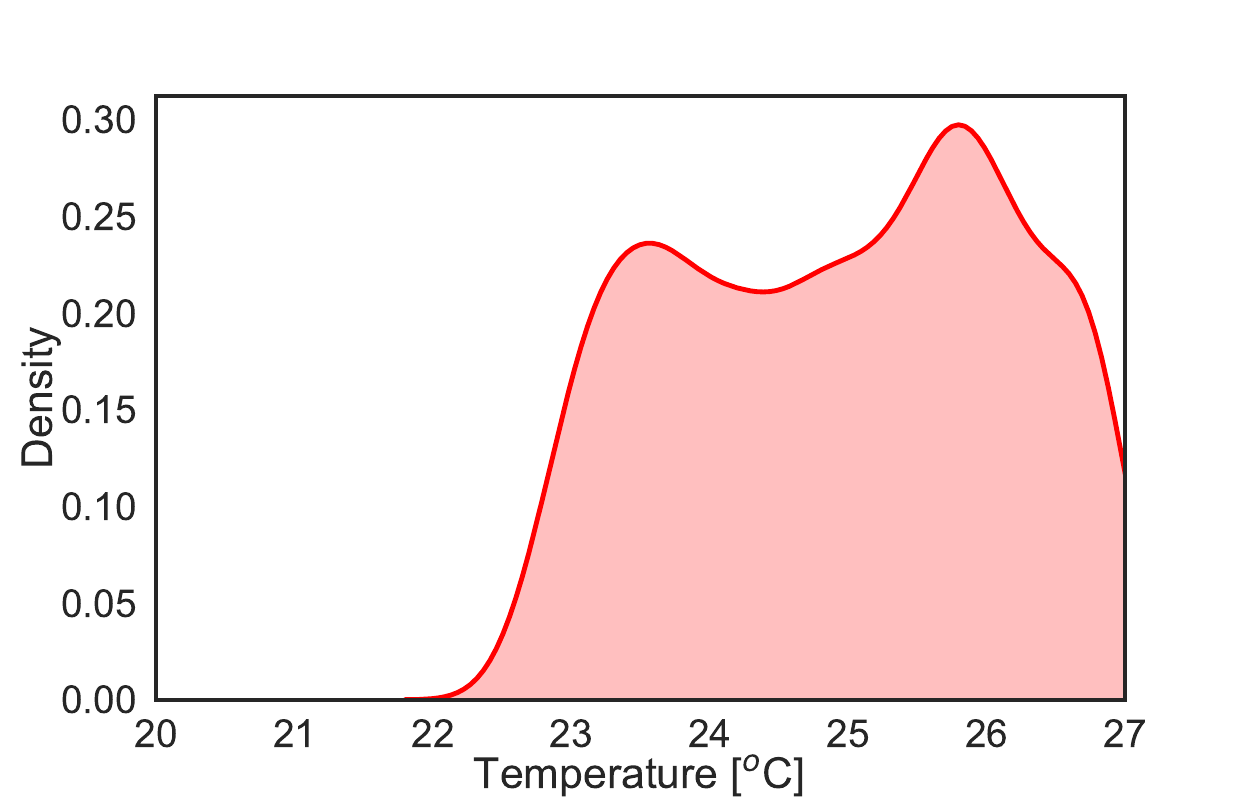}
    \caption{max. preferred temp. (occupant 6)}
    \label{fig:r6_m}
  \end{subfigure}
  
  \caption{Posterior predictions for real occupants. \textbf{First two columns:} Samples from posterior predictive distribution over normalized utility function. \textbf{Last column:} Predictive distribution over maximally preferred indoor air temperature values.}
  \label{fig:real_occupants}
\end{figure}

\section{Discussion}\label{sec:discussion}
This section first presents an analysis of the convergence of PE runs used to infer the maximally preferred indoor air temperature values (see \cref{subsec:pe_run_conv}). Then, in \cref{subsec:limitations}, the limitations of the proposed methodology and possible ways to address these limitations are discussed.

\subsection{Assessing the convergence of PE runs}\label{subsec:pe_run_conv}
\begin{figure}[H]
  \centering
  \begin{subfigure}[b]{0.32\linewidth}
    \includegraphics[width=\linewidth]{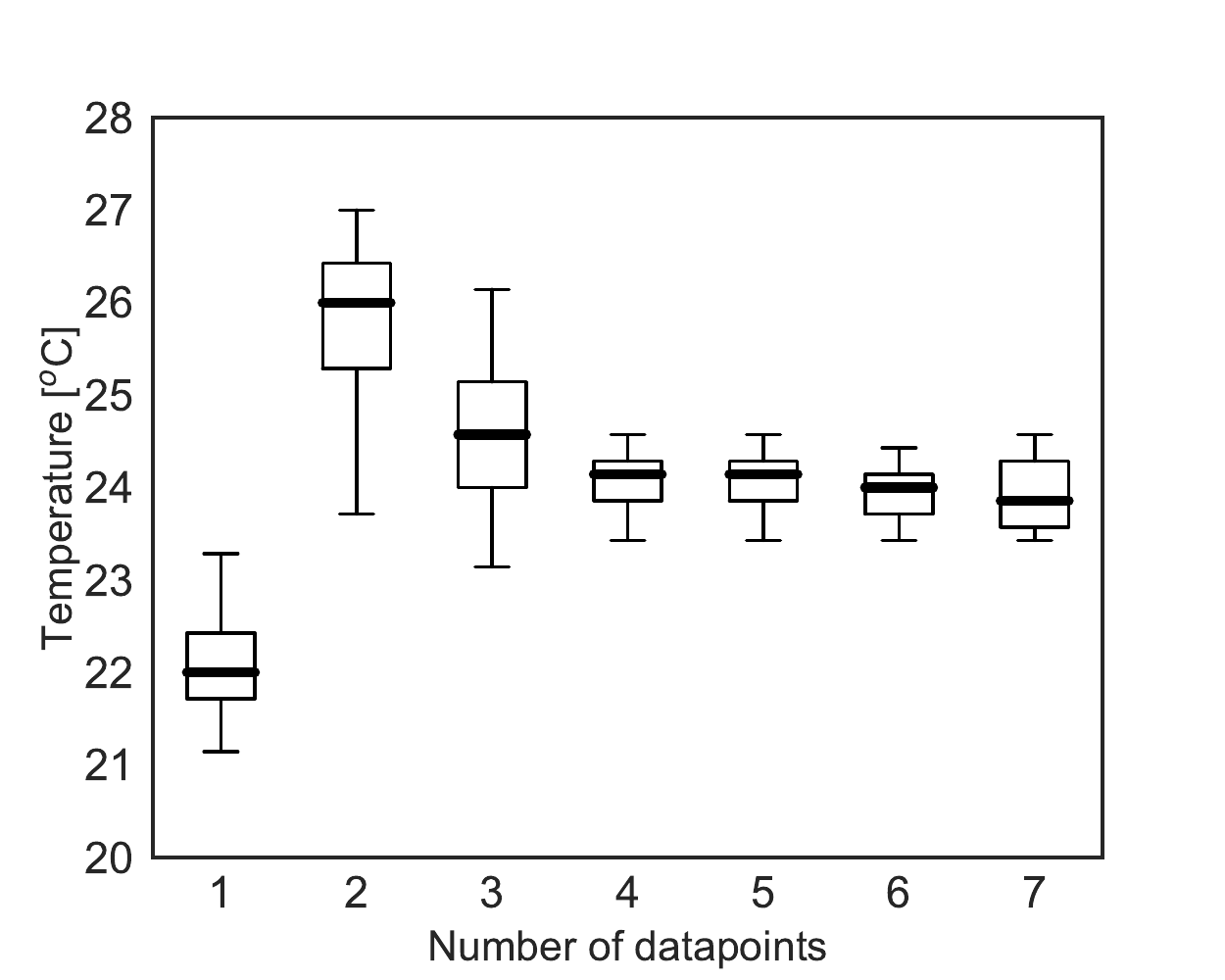}
     \caption{Real Occupant 1}
     \label{fig:conv_realo1}
  \end{subfigure}
  \begin{subfigure}[b]{0.32\linewidth}
    \includegraphics[width=\linewidth]{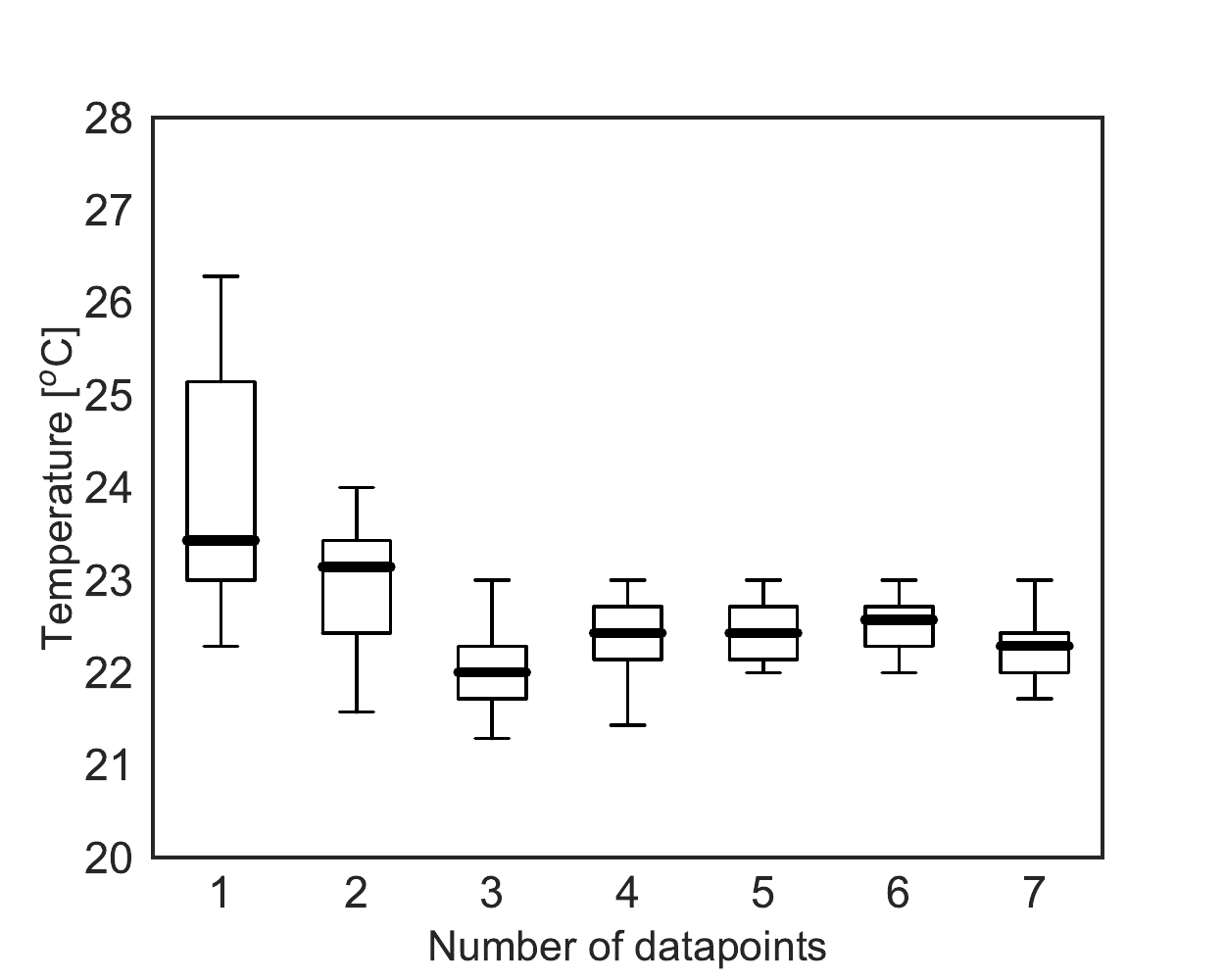}
    \caption{Real Occupant 2}
    \label{fig:conv_realo2}
  \end{subfigure}
  \begin{subfigure}[b]{0.32\linewidth}
    \includegraphics[width=\linewidth]{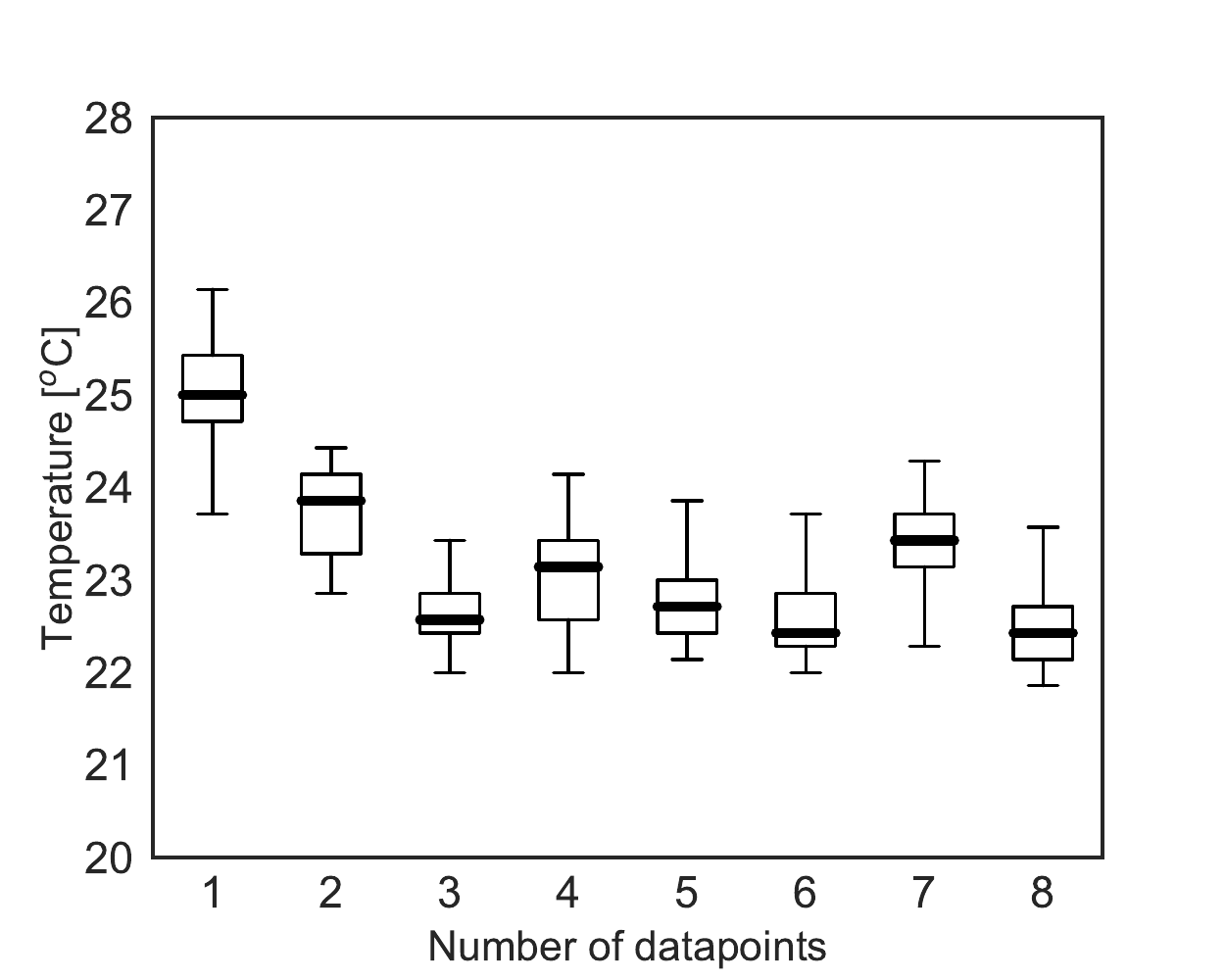}
    \caption{Real Occupant 3}
    \label{fig:conv_realo3}
  \end{subfigure}
  
  \begin{subfigure}[b]{0.32\linewidth}
    \includegraphics[width=\linewidth]{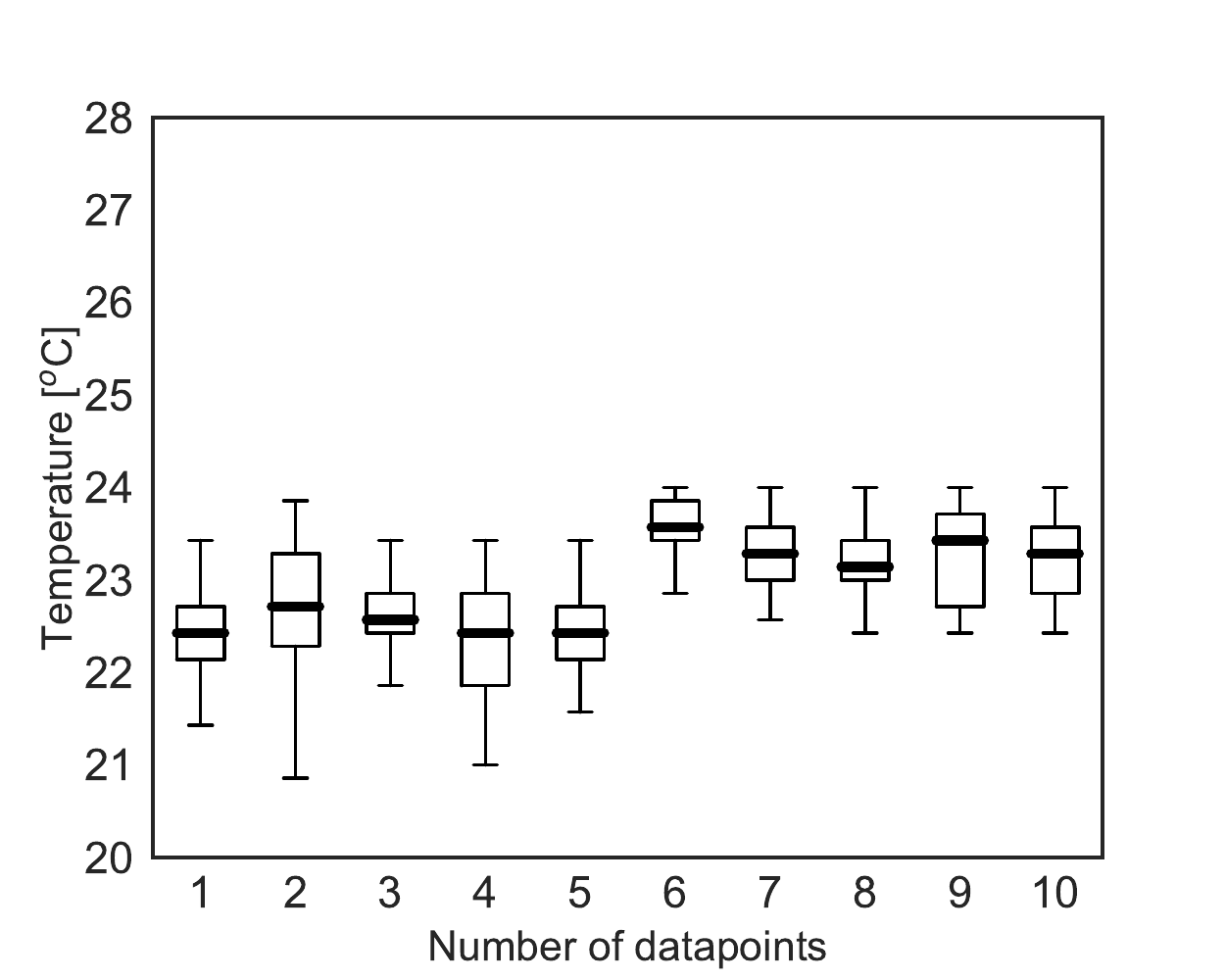}
     \caption{Real Occupant 4}
     \label{fig:conv_realo4}
  \end{subfigure}
  \begin{subfigure}[b]{0.32\linewidth}
    \includegraphics[width=\linewidth]{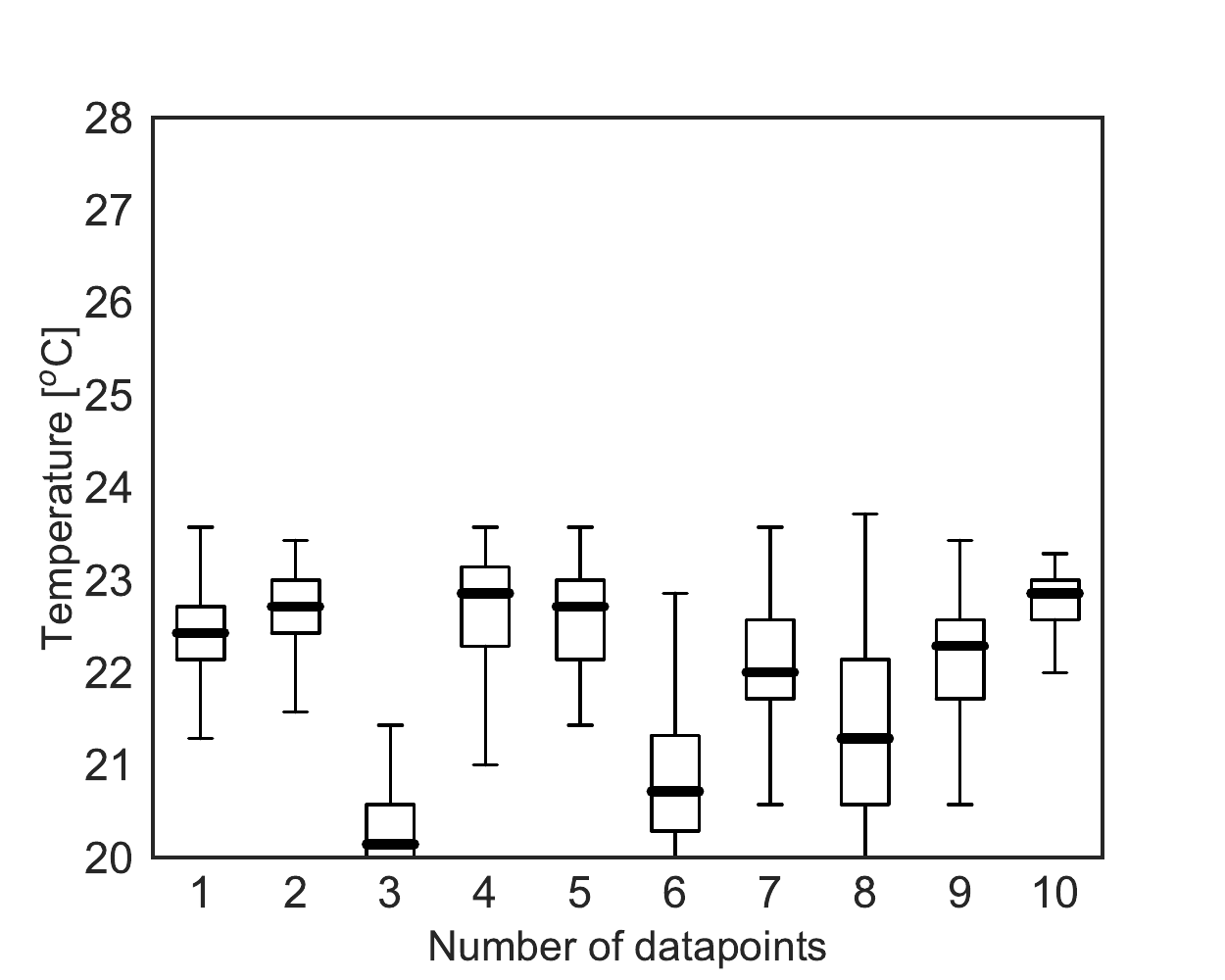}
    \caption{Real Occupant 5}
    \label{fig:conv_realo5}
  \end{subfigure}
  \begin{subfigure}[b]{0.32\linewidth}
    \includegraphics[width=\linewidth]{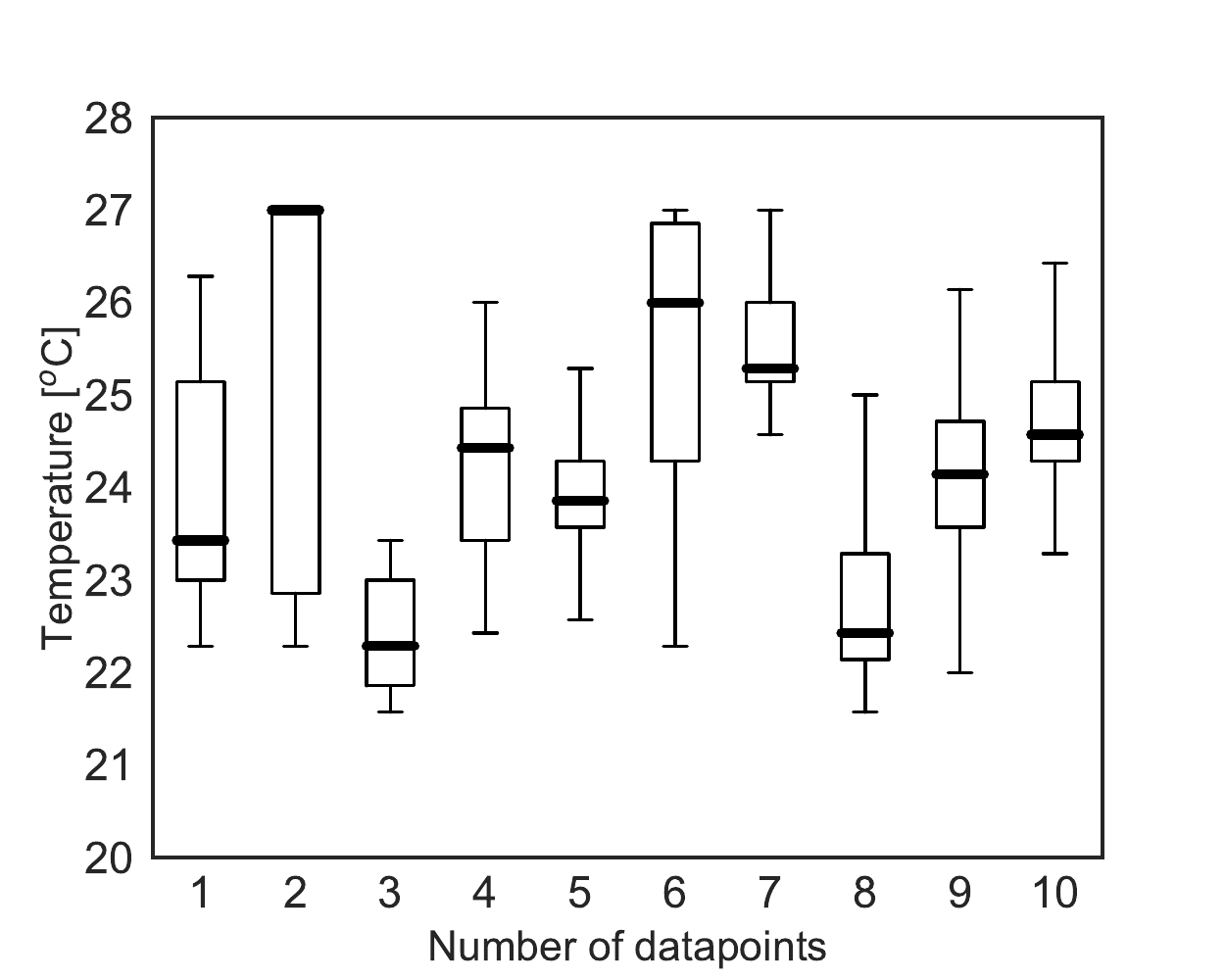}
    \caption{Real Occupant 6}
    \label{fig:conv_realo6}
  \end{subfigure}
\caption{Convergence of EUI based PE framework to maximally preferred indoor air temp. values}
\label{fig:max_pref_convergence_real_occupants}
\end{figure}

\begin{figure}[H]
  \centering
  \begin{subfigure}[b]{0.32\linewidth}
    \includegraphics[width=\linewidth]{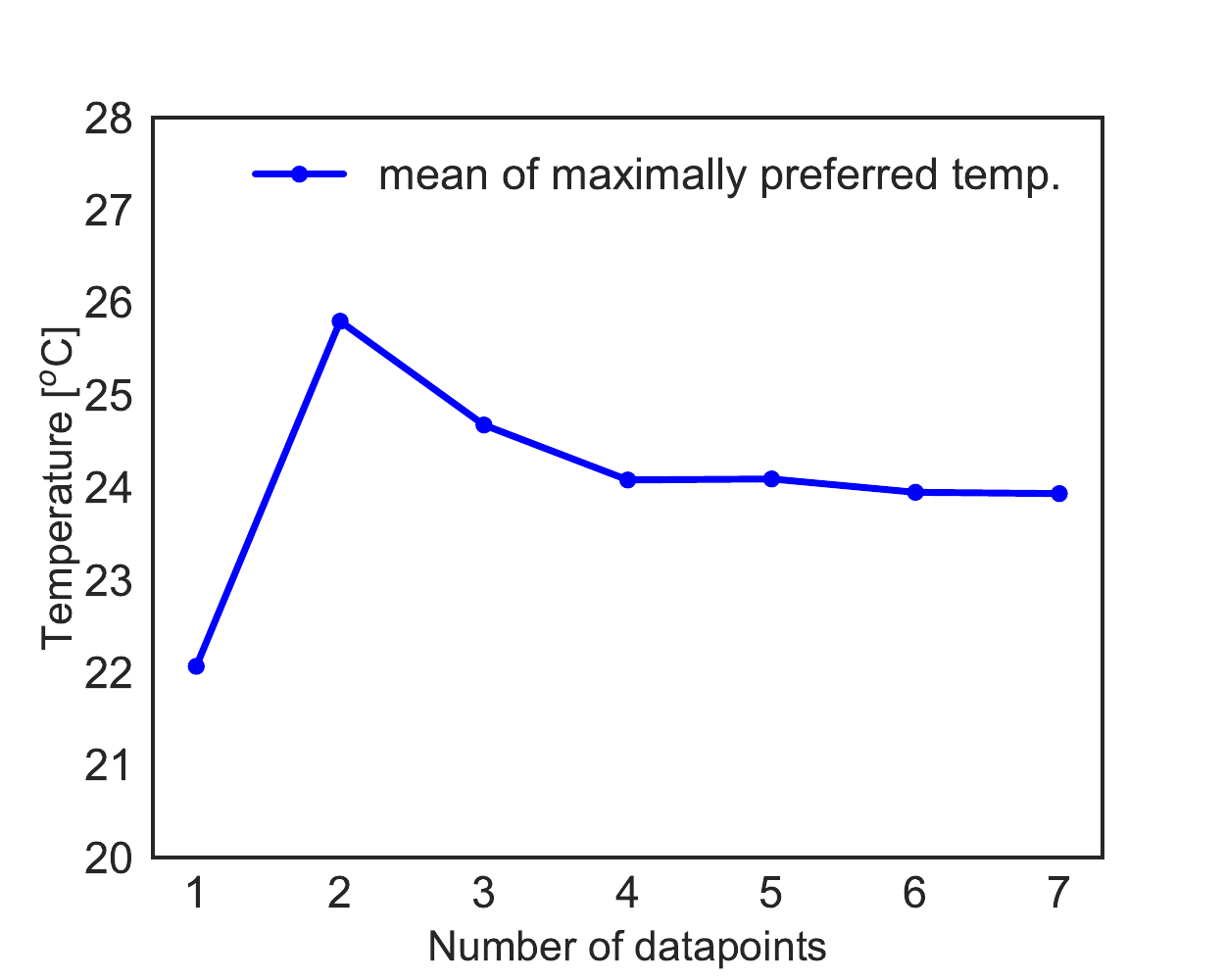}
     \caption{Real Occupant 1}
     \label{fig:conv_mean_ro1}
  \end{subfigure}
  \begin{subfigure}[b]{0.32\linewidth}
    \includegraphics[width=\linewidth]{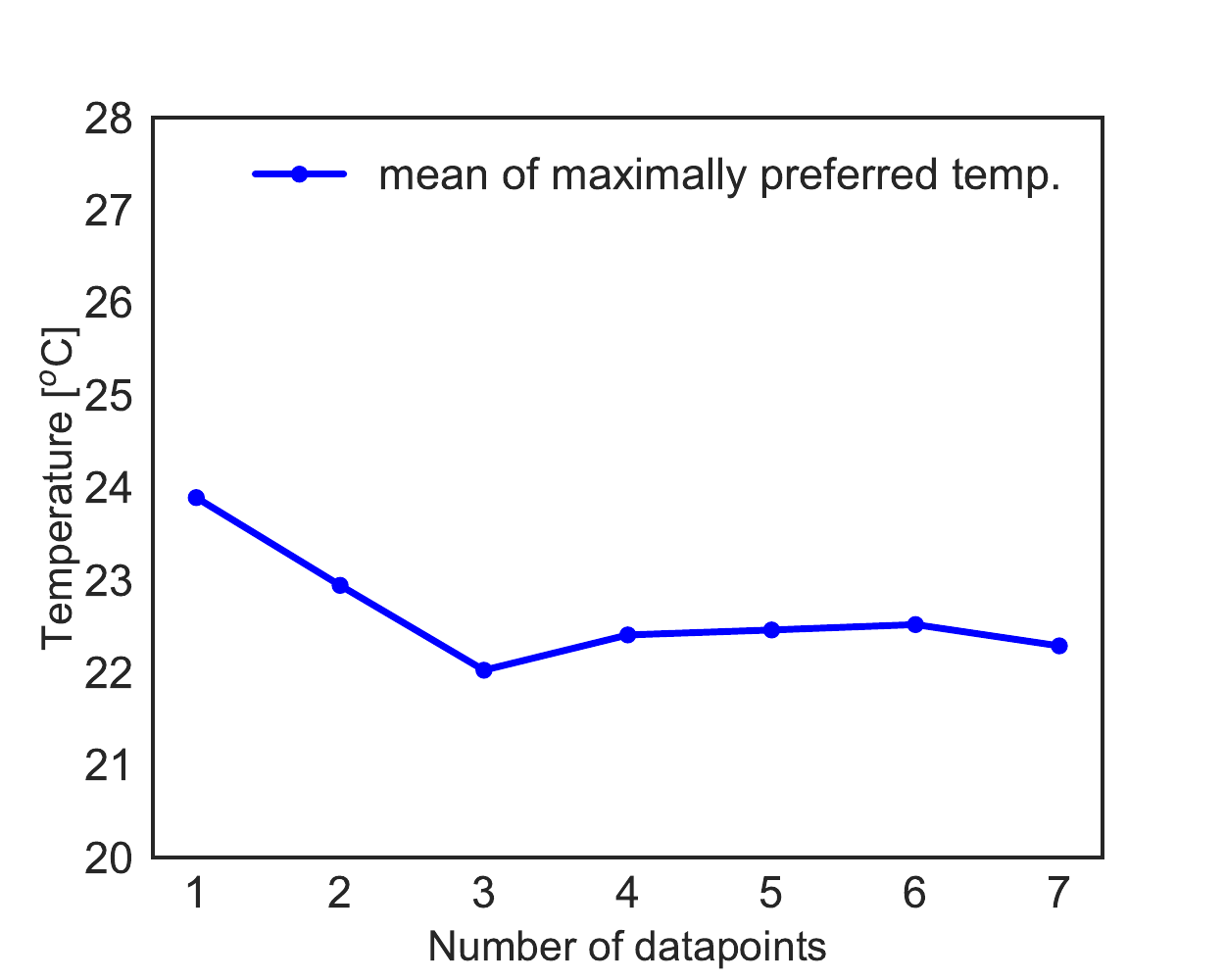}
    \caption{Real Occupant 2}
    \label{fig:conv_mean_ro2}
  \end{subfigure}
  \begin{subfigure}[b]{0.32\linewidth}
    \includegraphics[width=\linewidth]{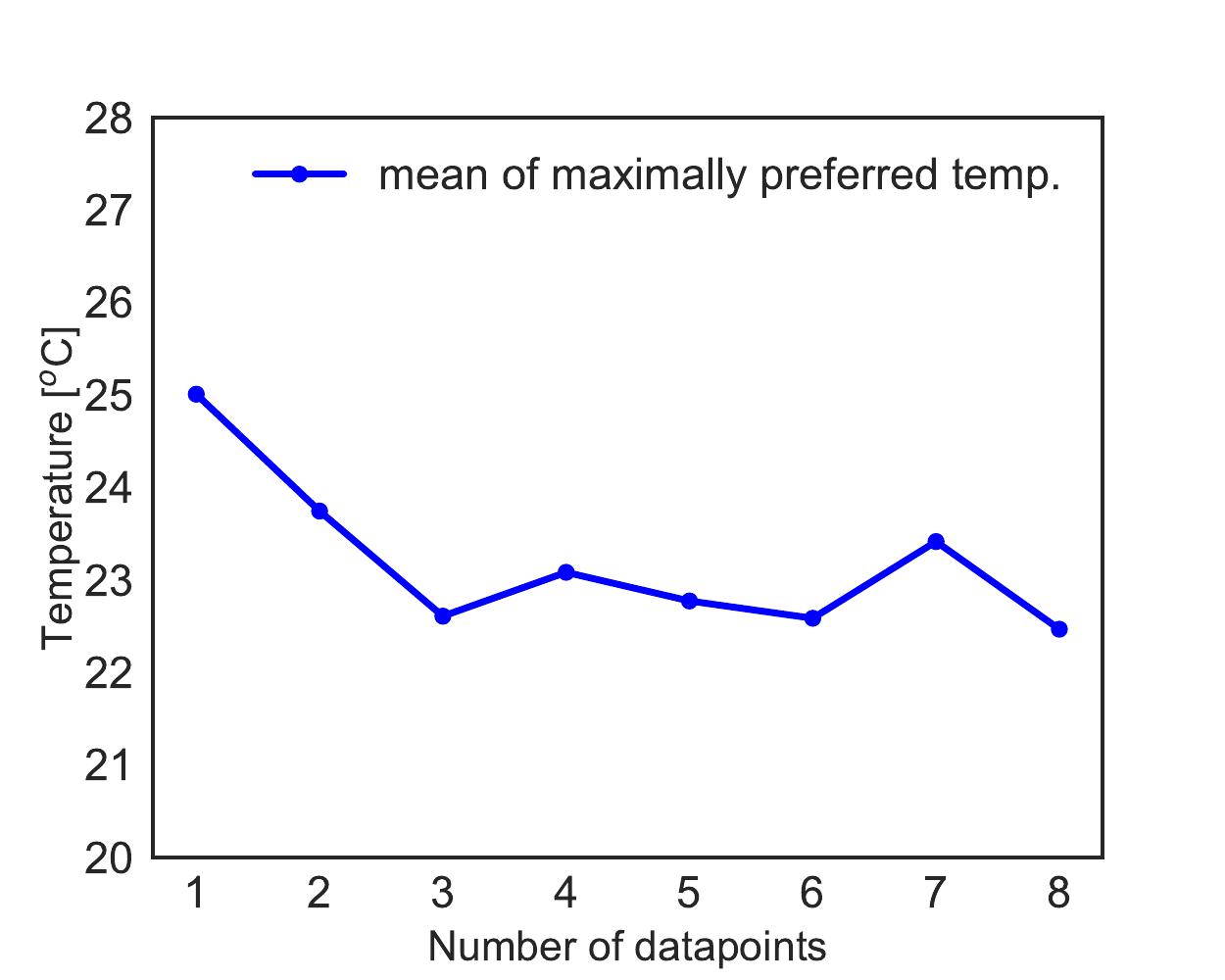}
    \caption{Real Occupant 3}
    \label{fig:conv_mean_ro3}
  \end{subfigure}
  
  \begin{subfigure}[b]{0.32\linewidth}
    \includegraphics[width=\linewidth]{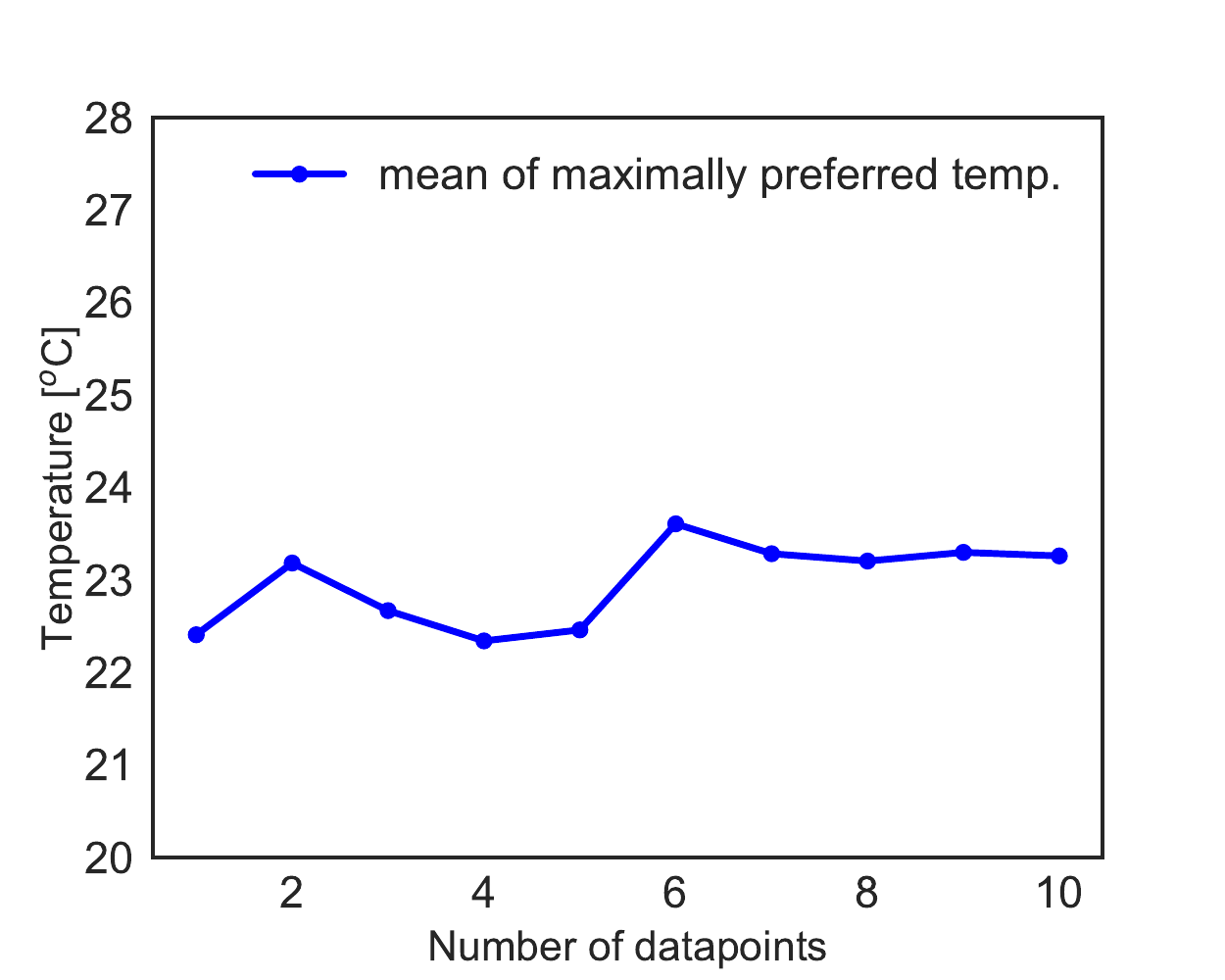}
     \caption{Real Occupant 4}
     \label{fig:conv_mean_ro4}
  \end{subfigure}
  \begin{subfigure}[b]{0.32\linewidth}
    \includegraphics[width=\linewidth]{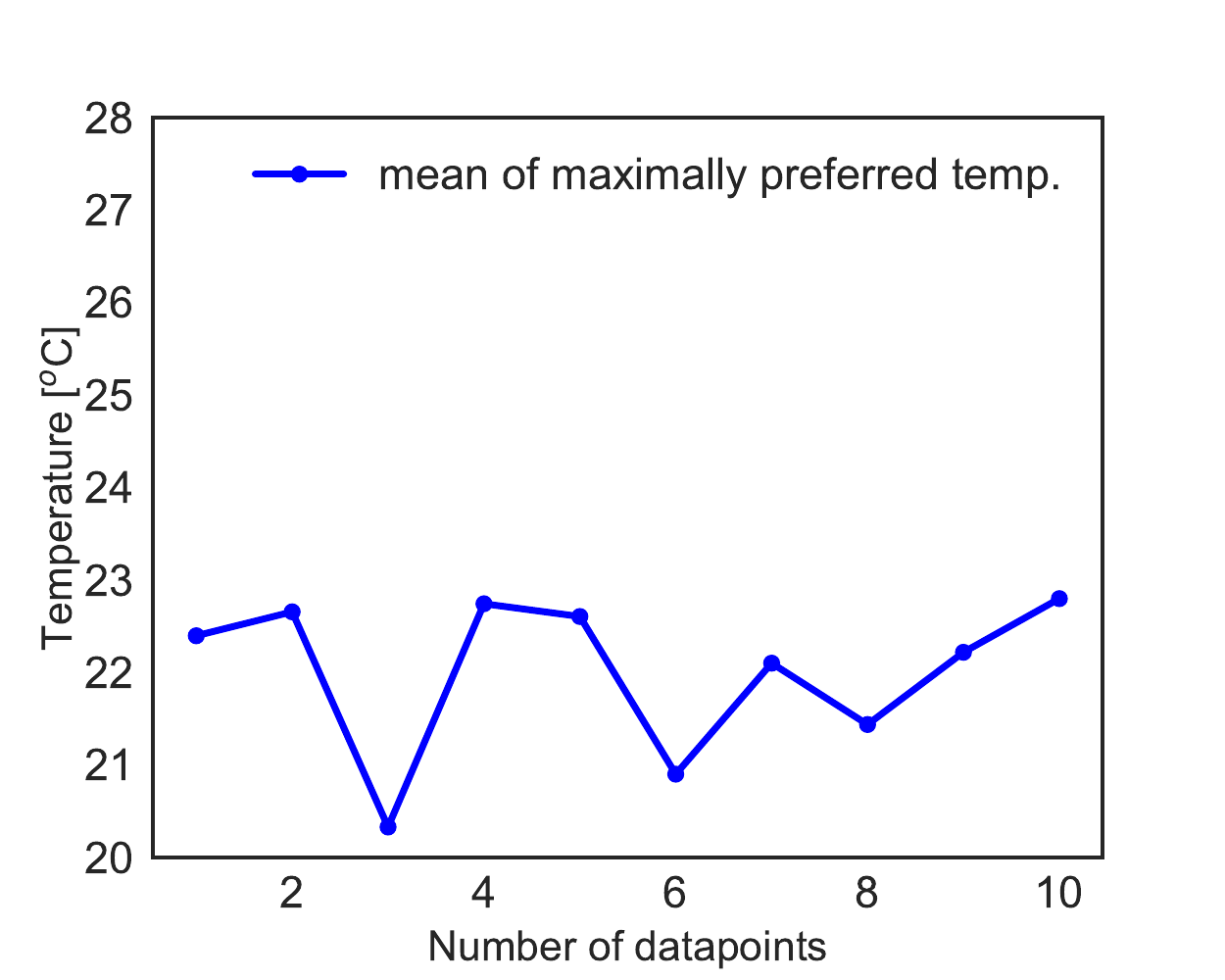}
    \caption{Real Occupant 5}
    \label{fig:conv_mean_ro5}
  \end{subfigure}
  \begin{subfigure}[b]{0.32\linewidth}
    \includegraphics[width=\linewidth]{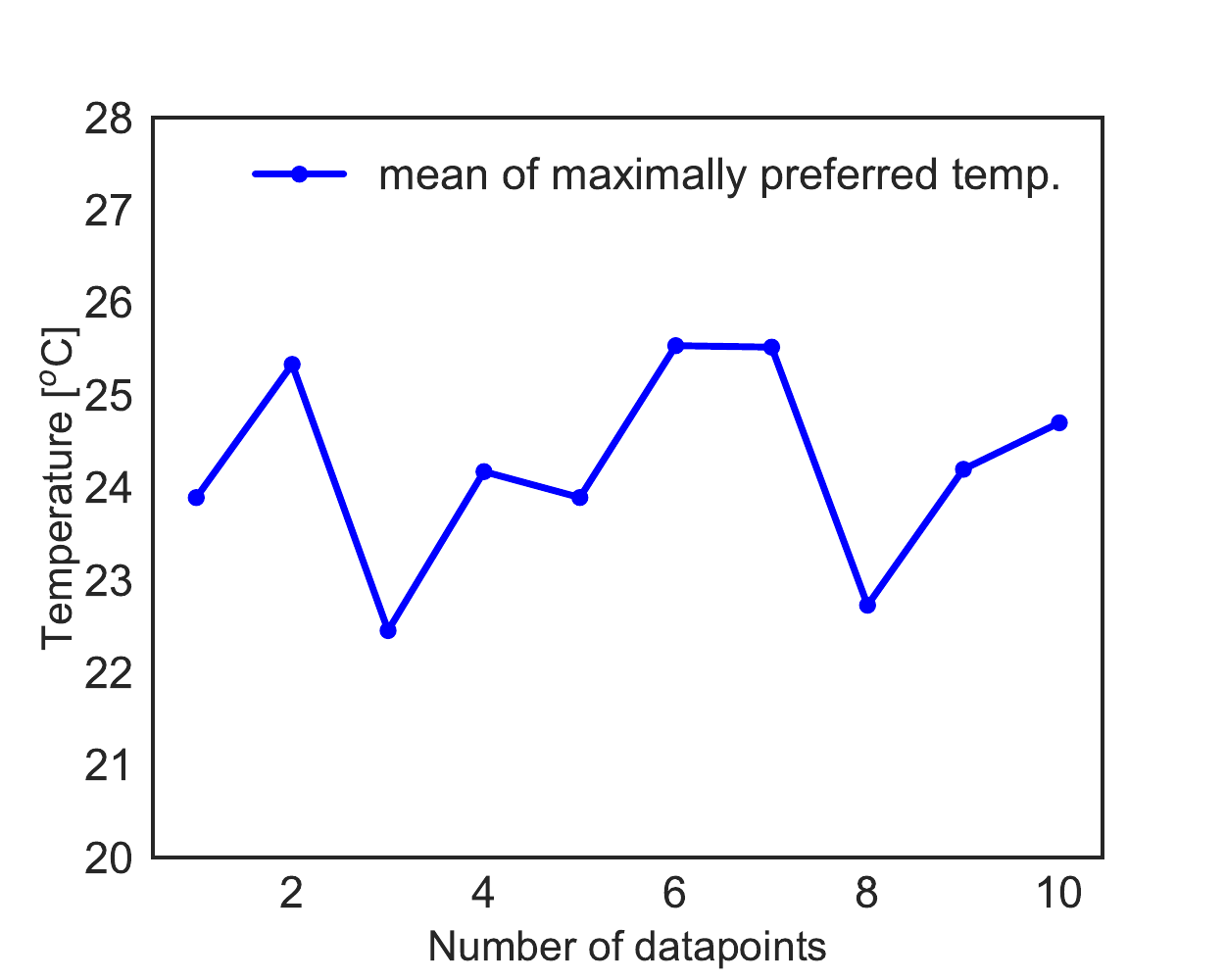}
    \caption{Real Occupant 6}
    \label{fig:conv_mean_ro6}
  \end{subfigure}
\caption{Convergence of EUI based PE framework to maximally preferred indoor air temp. values}
\label{fig:mean_max_pref_convergence_real_occupants}
\end{figure}

The inferred posterior distributions over maximally preferred indoor air temperature values (at each step of the PE runs) are shown in the \cref{fig:max_pref_convergence_real_occupants}. 
The mean of the inferred maximally preferred indoor air temperature values (at each step of the PE runs) are shown in the \cref{fig:mean_max_pref_convergence_real_occupants}.
It can be seen from \cref{fig:mean_max_pref_convergence_real_occupants} that as we run our PE chains and keep on collecting more and more informative query data, we converge towards the maximally preferred indoor air temperature values for each occupant. 

This convergence is further validated by the additional comments (information) given by the occupants to the optional question posed in the survey ``Please write down any comment you have regarding the thermal condition (optional)'' (see \cref{fig:survey}). 
The answers to this optional survey question, given by each of the occupants are shown in the \Crefrange{table:additional_comments_1}{table:additional_comments_2}. As it can be seen from these \Crefrange{table:additional_comments_1}{table:additional_comments_2}, as the PE runs converges to the maximally preferred indoor air temperatures, occupants are likely to comment favorably to reflect their preferences for the current indoor air temperature values they are exposed to. For example, occupant 1 filled out this additional section of the survey (at $i = 5$ of the PE run), commenting ``I would prefer these conditions'' when the air temperature of the room was set to $\dC{24}$. In fact, this is exactly what we expected from the occupants. Due to the nature of these PE experiments (sequentially changing indoor air temperatures and collecting thermal preference responses), occupants expected the indoor air temperature to change (in future) and in order to avoid this temperature change (and possible discomfort), occupants responded to this additional optional question to let us know about their preference for the current indoor air temperature values.   

\begin{table}[H]
\centering
\resizebox{\textwidth}{!}{
\begin{tabular}{ccccccc}
\toprule
\multirow{2}{*}{Query \#} &
\multicolumn{2}{c}{Real occupant 1} &
\multicolumn{2}{c}{Real occupant 2} &
\multicolumn{2}{c}{Real occupant 3} \\
\cmidrule(lr){2-3} 
\cmidrule(lr){4-5}
\cmidrule(lr){6-7}
& {Temp. ($\dC{}$)} & {Additional Comment} & {Temp. ($\dC{}$)} & {Additional Comment} &
{Temp. ($\dC{}$)} & {Additional Comment}\\
\midrule

1  & 21  & - &
21.5 & - &
22 & -\\

2  & 23.5  & - &
24  & - &
24.5  & - \\

3 & 26  & - &
23 & - &
23.5  & -  \\

4  & 24.5 & - &
22.5  & - &
23  & - \\

5 & 24 & ``I would prefer these conditions'' &
22 & - &
22.5 & -\\

6 & 24 & ``this is good'' &
22.5 & ``temperature is comfortable for now'' &
23.5& -\\

7 & 24 & - &
22.5 & - &
23 & ``temp is okay''\\

8 & - & - &
- & - &
23 & ``no need to change''\\
\bottomrule
\end{tabular}
}
\caption{Occupants' response to ``Please write down any comment you have regarding the thermal condition (optional)'' query.}
\label{table:additional_comments_1}
\end{table}

\begin{table}[H]
\centering
\resizebox{\textwidth}{!}{
\begin{tabular}{ccccccc}
\toprule
\multirow{2}{*}{Query \#} &
\multicolumn{2}{c}{Real occupant 4} &
\multicolumn{2}{c}{Real occupant 5} &
\multicolumn{2}{c}{Real occupant 6} \\
\cmidrule(lr){2-3} 
\cmidrule(lr){4-5}
\cmidrule(lr){6-7}
& {Temp. ($\dC{}$)} & {Additional Comment} & {Temp. ($\dC{}$)} & {Additional Comment} &
{Temp. ($\dC{}$)} & {Additional Comment}\\
\midrule

1  & 22  & ``temperature is good for me now'' &
21 & - &
21.5 & - \\

2  & 24 & - &
23.5 & - &
23.5 & ``I like the temperature'' \\

3 & 21 & - &
23 & - &
26 & ``I like this temperature'' \\

4 & 23 & - &
21 & - &
26.5 & - \\

5 & 21.5 & - &
20 & - &
28 & ``slightly warm on forehead'' \\

6 & 22.5 & - &
22 & ``good'' &
26 & - \\

7 & 23 & ``temperature is good'' &
22.5 & - &
27 & - \\

8 & 23.5 & ``good'' &
21.5 & ``very good!'' &
26.5 & ``I like this temperature'' \\

9 & 23.5 & - &
22.5 & ``good'' &
24.5 & ``temp. is good'' \\

10 & 23 & ``best of all'' &
22.5  & - &
25 & ``temp. is good'' \\

\bottomrule
\end{tabular}
}
\caption{Occupants' response to ``Please write down any comment you have regarding the thermal condition (optional)'' query.}
\label{table:additional_comments_2}
\end{table}

\subsection{Limitations of Unimodal GP-PE Framework and Future Work}
\label{subsec:limitations}
Despite the promising results presented in the paper (on the topic of inferring the maximally preferred indoor air temperature values based on limited data), there are some unresolved issues that need further attention.

\subsubsection{Limitations imposed due to strict unimodality constraints}\label{subsubsec:limitation_unimodality}
In this work, we have imposed strict unimodality constraints on the structure of utility functions.
That is, we strongly believe (before seeing any data) that there exists a unique maximum (indoor air temperature value) that the occupant will prefer over all the other temperatures. 
However, this might not always be the case (as is evident in case of real occupants 5 and 6). In this study, occupants 5 and 6 were not very sensitive to change in indoor air temperature values. That is, in these cases, the utility functions governing the preferences of occupants had a relatively flat structure.
As is evident from \Crefrange{fig:conv_mean_ro5}{fig:conv_mean_ro6}, convergence of PE runs in such cases to a unique maximally preferred indoor air temperature value is not guaranteed. We propose few possible ways in which  such limitations can be addressed:
\begin{enumerate}
    \item Relaxing the strict unimodality constraints placed on GP prior by tuning hyperparameter $\nu_{\tilde{y}}$ (see \cref{sec:unimodality}) of the proposed unimodal-GP model.
  
    \item Use of conventional GP prior (without any constraints (see \Cref{fig:appen_mon_dec_gp_fit} and \cref{fig:appen_unimodal_normal_gp_fit}). Implementation of derivative based GP preference learning model (without any constraints) is provided in GPPrefElicit package.

    \item Use of pairwise comparison data (``do you prefer the current temperature or do you prefer previous temperature?')' instead of derivative based thermal comfort queries (if the thermal setup and pilot study results allows for it). Support for training Gaussian process (with and without unimodality constraints) preference learning models based on pairwise comparison data is also provided in GPPrefElicit package. 
\end{enumerate}

\subsubsection{Limitations imposed due to the use of only one environmental feature : indoor air temperature}\label{subsubsec:limitation_unimodality}
In this work, we have focused on development of GP-PE framework based on indoor air temperature values. The contributions of other variables (e.g. relative humidity, air velocity) for inferring preferences were found to be considerably smaller than than indoor air temperature values in our experimental setup (revealed through ANOVA on pilot study data and further supported in studies by \cite{ghahramani2015online, daum2011personalized}). However, this might not always be the case. In cases where readers want to include other environmental features into their modeling framework, we propose extending the Unimodal-GP preference learning  model to higher dimensions. This model can naturally be extended to $M$ dimensions by introducing $M$ under the hood GPs, where the $j$'th GP, $g_{j} : \mathbb{R}^{M} \rightarrow \mathbb{R},$ controls the sign of the $j$'th partial derivative of $u$. This proposed approach will also provide additional flexibility to only impose unimodality with respect to a subset of dimensions, if desired. Support for higher dimensions Unimodal-GP surrogate will be provided in GPPrefElicit package.

\subsubsection{Limitations imposed due to stationary nature of Unimodal-GP model}\label{subsubsec:limitations_hiddenfactors}
As evidenced in prior studies \cite{jendritzky2009adaptation, brager1998thermal}, many dynamic environmental and human related features are seen to affect thermal satisfaction of occupants.
As such, one of the limitations of our framework is the failure to capture evolution of occupants’ preferences over time/seasons.
We acknowledge this issue and when it comes to addressing such dynamic variations in preferences (across seasons), we suggest to run the PE framework for each of the seasons and quantify thermal preferences for each season separately. 
Extrapolating our work further and to ensure more precise modeling, our future work entails modeling the evolution of occupants' preferences over time for personalized recommendation (time-aware collaborative filtering approach).  

\subsubsection{Limitations imposed due to failure to model latent hidden variables}\label{subsubsec:limitations_stationary}
In this paper, we have focused on the development of a simple, robust, low-cost, easy to compute and easy to implement personalized PE framework which focuses on indoor air temperature as its most important feature and as such have not included other latent features affecting preferences (such as metabolic rate, clothing level etc.).
Our future work involves addressing these limitations by extending our PE framework to incorporate these temporal effects \cite{liu2015modeling, gao2017collaborative} into our GP-PE based latent factor model.

\subsubsection{Limitations imposed due to failure to capture shared structure in occupants' utility}\label{subsubsec:limitations_shared_structure}
In this work, utility functions governing the preferences of each occupant were modeled independently (independence assumption). However, when thermal preference query data from many occupants is available, our framework, in its present form fails to leverage similarities in preferences across occupants. In future, we want to extend our Unimodal-GP PE model to exploit collaborative information from shared structure in occupant preferences (collaborative GPs for preference elicitation) \cite{wilson2010generalised}.

\section{Conclusion}\label{sec:conclusion}
To provide a satisfactory thermal environment in office buildings, we need to understand the unique  preference requirements of individual occupants.
Therefore, in this paper, we presented a new elicitation framework for learning personalized thermal preferences. 
We addressed the question of ``how can we learn a particular occupant's maximally preferred indoor air temperature without requiring unnecessary or excessive survey data from them?''
More specifically, we constructed a unimodal GP-based model for characterizing our beliefs about the utility function governing the thermal preferences of individual occupants.
We conditioned this unimodal GP based on the thermal preference data to update our state of knowledge about the utility function (posterior predictive distribution).
This newly inferred posterior predictive distribution was then used to drive the elicitation process for selecting a new indoor air temperature to query.
We kept on repeating these steps until we converged towards the maximally preferred indoor air temperature value for each of the individual occupants.
We verified the proposed methodology using synthetic occupants and designed practical/ real-time experiments to infer the maximally preferred indoor air temperatures for six new real occupants working in private offices. We showed that occupants' maximally preferred indoor air temperature can be inferred with only 5 to 10 preference queries to them. 
In this paper, we have focused on the development of simple, robust, low-cost, easy to compute and easy to implement personalized PE framework which focuses on indoor air temperature as its most important feature.
This framework is an important step towards the development of intelligent HVAC systems which would be able to respond to individual occupants' personalized thermal comfort needs.

\section*{Acknowledgements}
This work was funded by the National Science Foundation under Grant No. 1539527 and it was supported by the Center of High Performance Buildings at Purdue University. 
Any opinions, findings, and conclusions or recommendations expressed in this material are those of the authors and do not necessarily reflect the views of the National Science Foundation.

\appendix
\section{Utility theory}
\label{apen_sec: Utility theory}
Classical decision theory deals with making optimal decisions taking into account uncertainty in the outcomes \cite{berger2013statistical}.
It is based on the assumption that complete user preferences are known to us.
Therefore, in order to take optimal decisions, it is crucial to understand user preferences and uncertainty associated with it in the context of the problem.
Due to the challenges and difficulties associated with quantifying and understanding human preferences, a utility function is introduced over it.
This utility function maps each instance/state or item to a real value and compares different instances/states/items based on their utility function values \cite{guo2011bayesian}. 
Von Neumann and Morgenstern \cite{morgenstern1953theory} showed that: if one has preferences defined over outcomes that have uncertain consequences, there is a utility function that assigns a utility to every outcome that represent these preferences. 
Therefore, utility functions therefore allow us to directly compare arbitrary states/items and reduces the complexity of preference learning framework to $\mathcal{O}(n)$). In order to formally define preferences and several axioms/theorems related to it, we follow the review examples given in \cite{Sanders}.

\subsection{Preference Relations}
\label{apen_subsec:Preference Relations}
In our problem, preference relations are an intuitive way of thinking about how people rank different thermal conditions inside an office. We talk about preferences by mainly using three binary relations: $\succ$ (strictly preferred to), $\sim$ (indifferent between) and $\succeq$ (preferred to or indifferent between). Let us now review Von Neumann and Morgenstern’s axioms related to preference relations:
\begin{enumerate}
\item \textbf{Completeness:} For our example/problem, it implies that given two states of thermal conditions in the room $(m,n)$, our preference relation can compare and rank them as either $m \succeq n$ or $m \preceq n$.
\item \textbf{Transitivity:} If a user prefers $m$ to $n$, and prefers $n$ to $p$ in the meantime, 
then it holds that the user prefers $m$ to $p$, i.e.  $m \succeq p$.
\item \textbf{Continuity:} 	Given strict preferences over any three states, i.e. $m \succ n \succ p$, 
there exists a linear combination of the most and least preferred states such that $am+(1-a)p \succ n$ and $p \succ bm+(1-b)n$ where $a,b \in [0,1]$.
\end{enumerate}
Provided that the an occupant's preference for different thermal states satisfies these four axioms, 
Von Neumann and Morgenstern’s main existence theorem states that there exists a utility function that represents preferences over different thermal states of room in our problem.

\subsection{Utility Functions and Monotonic Transformations}
\label{apen_subsec:Utility Functions and Monotonic Transformations}
In most cases, preference relations can be sufficiently described with the help of utility functions.
A utility function assigns scalar values to all states so that, if we have a preference relation between $m$ and $n$ state as $m \succ n$ then we have $u(m)>u(n)$.
For drawing inferences from the utility function values, all that matters is ordinal ranking, and not cardinal ranking. 
In Economics, an ordinal ranking is a function representing the preferences of a consumer on an ordinal scale.
The ordinal utility theory claims that it is only meaningful to ask which option is better than the other but it is meaningless to ask how much better it is. 
Put simply, the scalar numerical utility function values we obtain only matter in the context that we can say one utility level is higher than the other, but the actual values do not mean much.
For example, if $u(m)=100$ and $u(n)=300$, we cannot say that $m$ is three times as good as $n$ (cardinal statement). We can only say that $m$ is preferred to $n$. 
Utility function $u$ is affine i.e. there can be multiple utility functions that describe the same set of preferences. One property of utility functions is that if $u(m)$ is a valid utility and $f$ is a monotonically increasing transformation (e.g. exponential functions), then $f(u(m))$ is also a valid utility function.

In summary, it is useful to operate on preference relations with the use of utility function. 
It assigns a scalar value to a state as a representation of its utility. Consequently, utility functions allow a mathematical framework for preference learning. 
In addition, one can handle uncertainty associated with preferences by introducing Bayesian models for practical problems \cite{guo2011bayesian}.

\section{Gaussian Process Regression (with and without constraints)}
\label{apen_sec:Gaussian Processes (with and without constraints)}
In a classic regression problem, consider a continuous valued input/output dataset $D = \{(x_{i}, y_{i})| i = 1, ..., N\}$ where $y_{i}$ is a noisy observation of function $f$, which is given as: $y_{i} = f(x_{i}) + \epsilon_{i}$. The likelihood \cite{bishop2006pattern} of observing $\mathbf{y}$ given function $f$ and feature value $\mathbf{x}$ is then given as:
\begin{equation}
\label{appendix eq : gp regression likelihood}
p(\mathbf{y}| \mathbf{f}, \sigma) = \mathcal{N}(\mathbf{y}| \mathbf{f}, \sigma^2\mathbf{I}),
\end{equation}
where $\mathbf{y} = \{y_{i}| i = 1,..., N\}$, $\mathbf{f} = \{f(x_{i})| i = 1,..., N\}$ and noise standard deviation $\sigma > 0$. The results of fitting GP (with and without constraints) in such regression settings is discussed in the following sub-sections.

\subsection{GP regression results with and without monotonicity constraints}
\label{appen_subsec: GPR with and without monotonicity}
\begin{figure}[H]
  \centering
  \begin{subfigure}[b]{0.48\linewidth}
    \includegraphics[width=\linewidth]{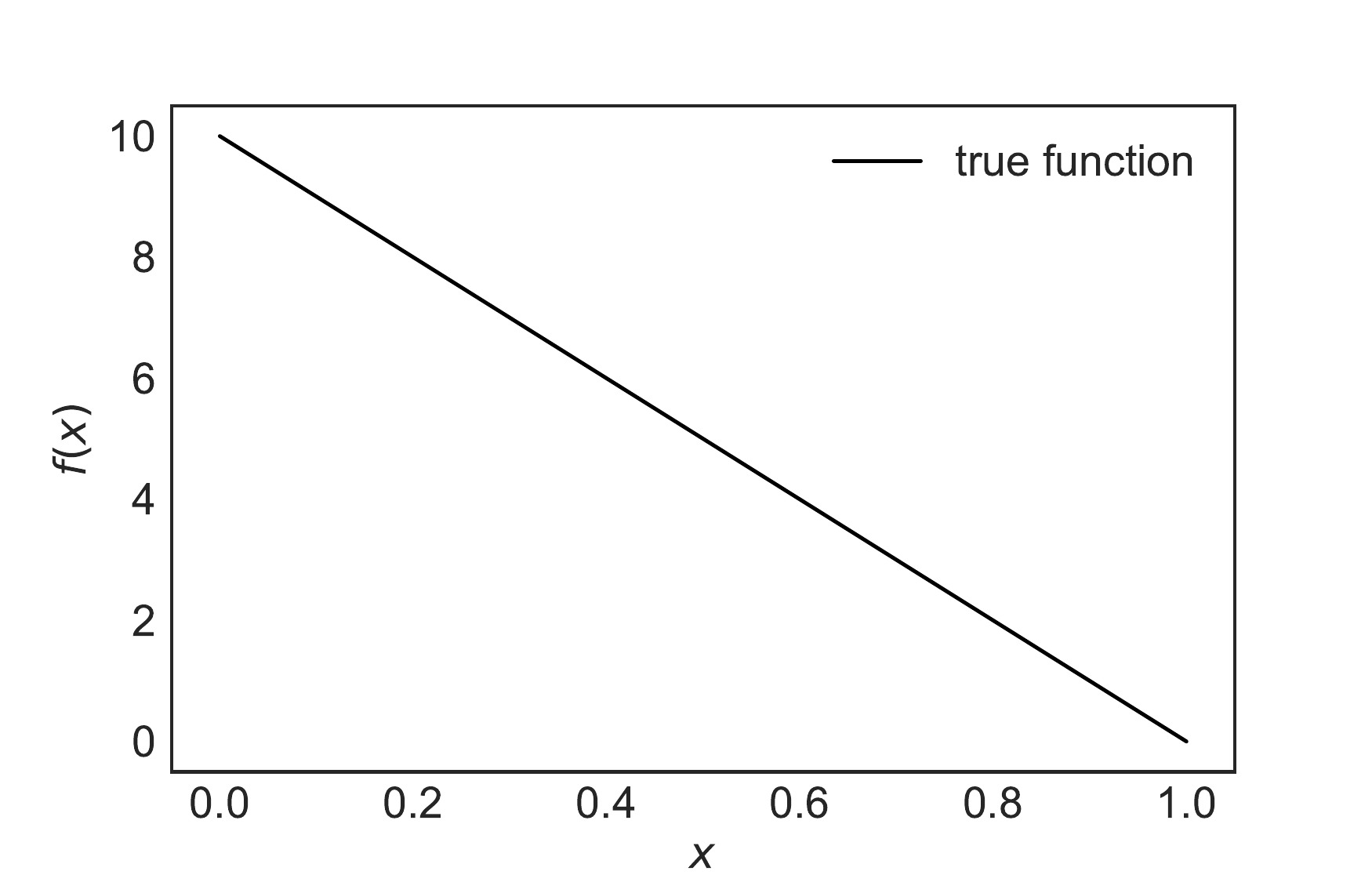}
     \caption{Monotonically decreasing function}
     \label{fig:appen_true_func}
  \end{subfigure}
  \begin{subfigure}[b]{0.48\linewidth}
    \includegraphics[width=\linewidth]{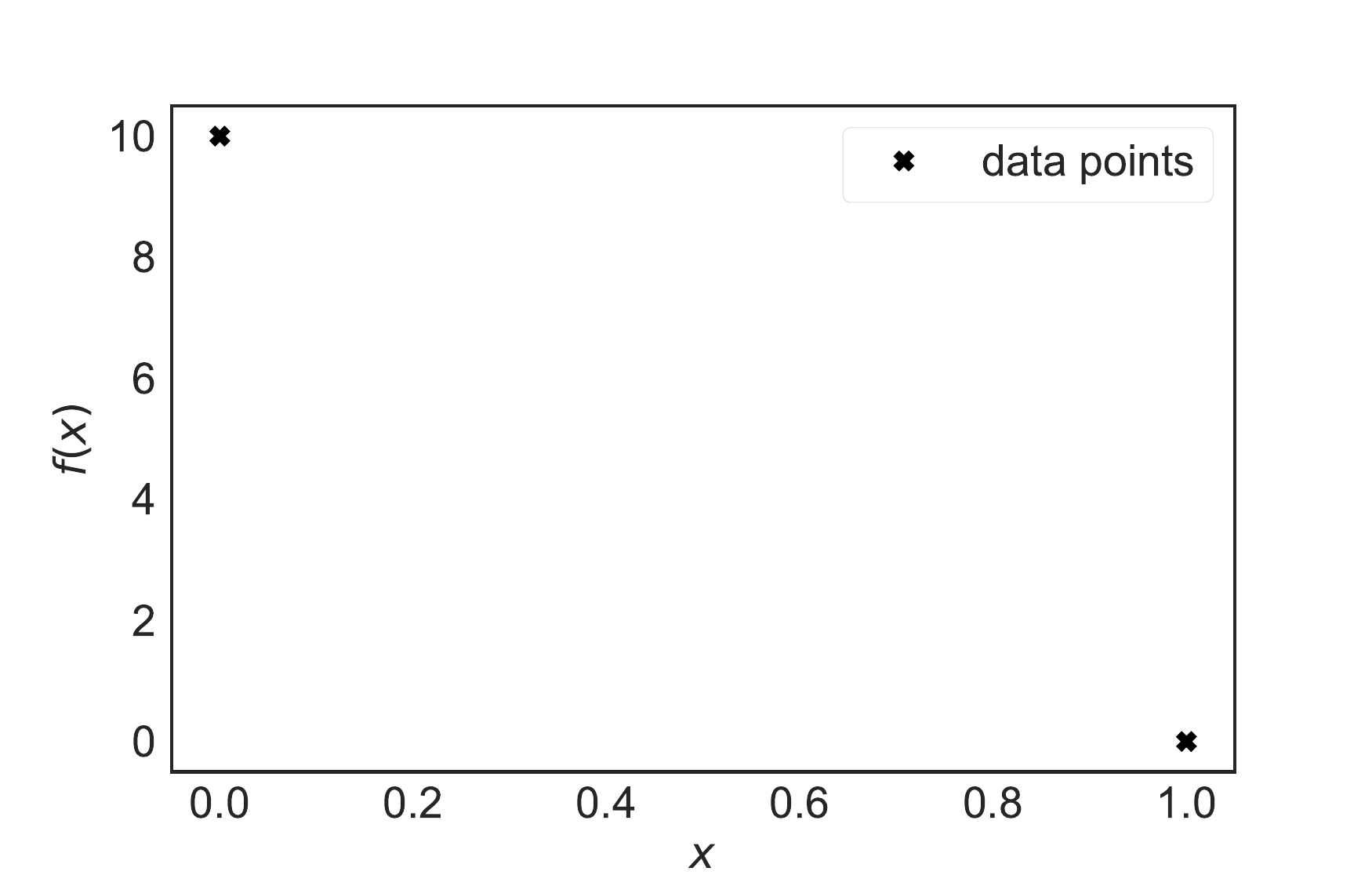}
    \caption{Observed Datapoints $N = 2$}
    \label{fig:appen_obs_data}
  \end{subfigure}
\caption{True monotonically decreasing function $f(x) = (10 - 10x)$ along with associated observed dataset.}
\end{figure}

\begin{figure}[H]
  \centering
  \begin{subfigure}[b]{0.48\linewidth}
    \includegraphics[width=\linewidth]{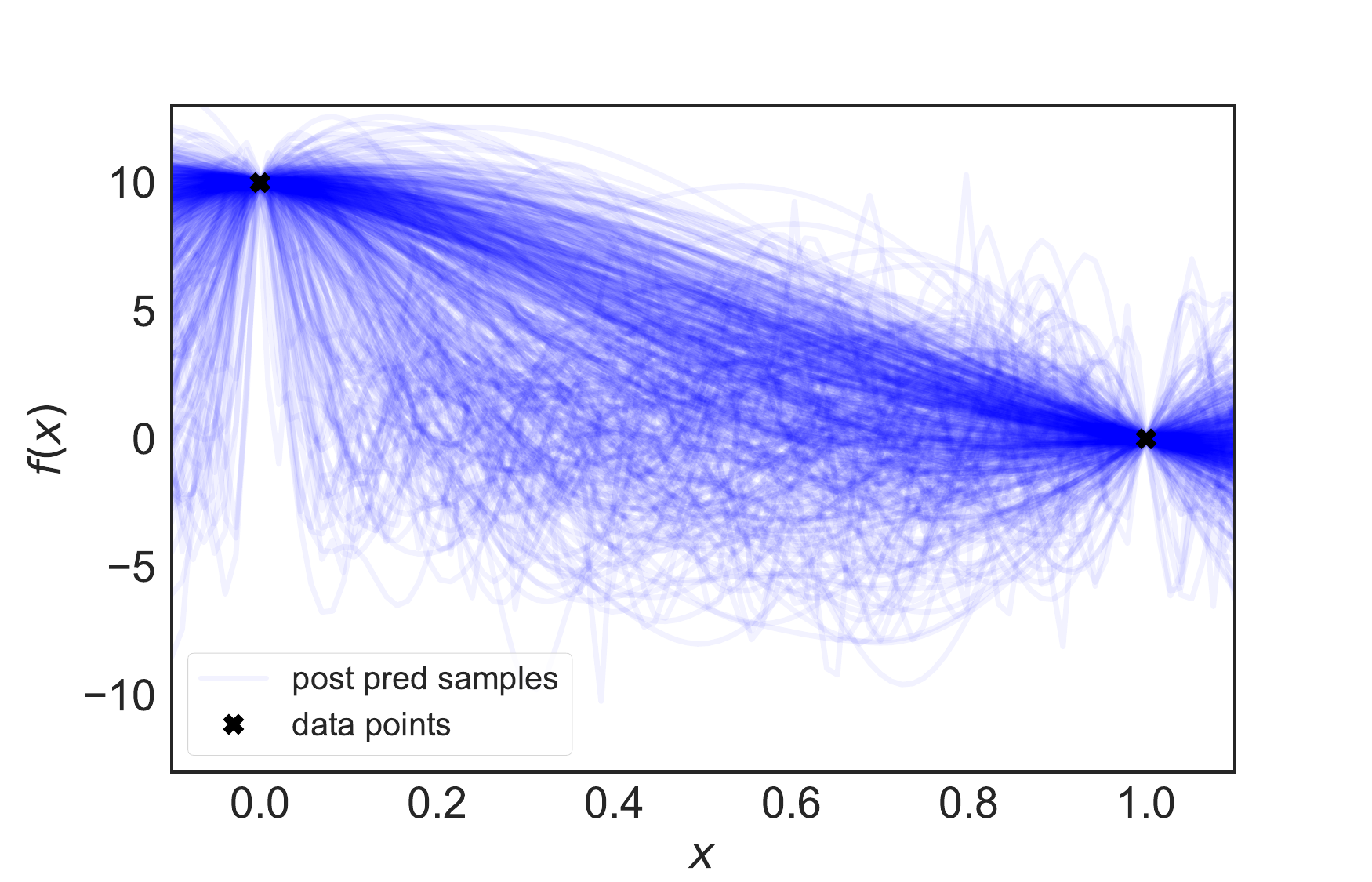}
     \caption{GP fit (without any constraints)}
     \label{fig:appen_mon_dec_gp_fit}
  \end{subfigure}
  \begin{subfigure}[b]{0.48\linewidth}
    \includegraphics[width=\linewidth]{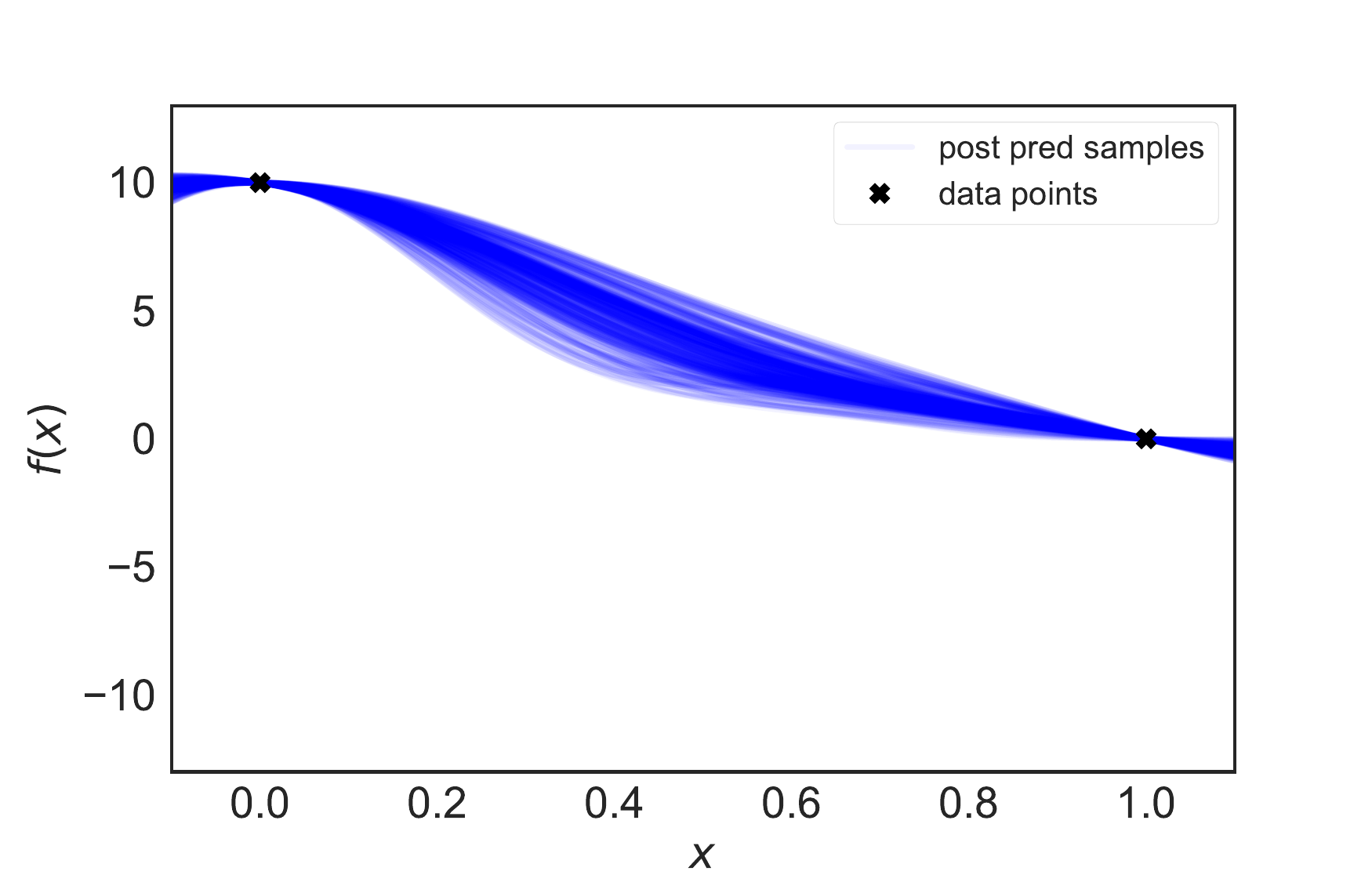}
    \caption{Monotonic GP fit (with monotonic constraints)}
    \label{fig:appen_mon_dec_monotonic_gp_fit}
  \end{subfigure}
\caption{Posterior predictive samples (\textbf{left:} unconstrained GP fit ; \textbf{right:} monotonic GP fit).}
\end{figure}

In this sub-section, we demonstrate how the monotonic GP model we developed in sub-section \ref{sec:unimodality} can be applied to regression problems.
Assume that the true function we want to approximate is given as: $f(x) = 10 - 10x$.
This function is shown in \Cref{fig:appen_true_func}.
Furthermore, assume that we have observed $N = 2$ datapoints associated with function $f$, which are given as $D_{1} = \{(0,10), (1,0)\}$.
These observed points are shown in the \Cref{fig:appen_obs_data}.
We fit two models (1) regular GP without constraints and (2) GP with decreasing monotonicity constraints by conditioning these models on the observed dataset $D_{1}$.
Samples from posterior predictive distribution, fitted using a regular GP model without constraints are shown in \Cref{fig:appen_mon_dec_gp_fit} and from a GP  with decreasing monotonicity constraints are shown in \Cref{fig:appen_mon_dec_monotonic_gp_fit}.

\subsection{GP regression results with and without unimodality constraints}
\label{appen_subsec: GPR with and without unimodality}
\begin{figure}[H]
  \centering
  \begin{subfigure}[b]{0.48\linewidth}
    \includegraphics[width=\linewidth]{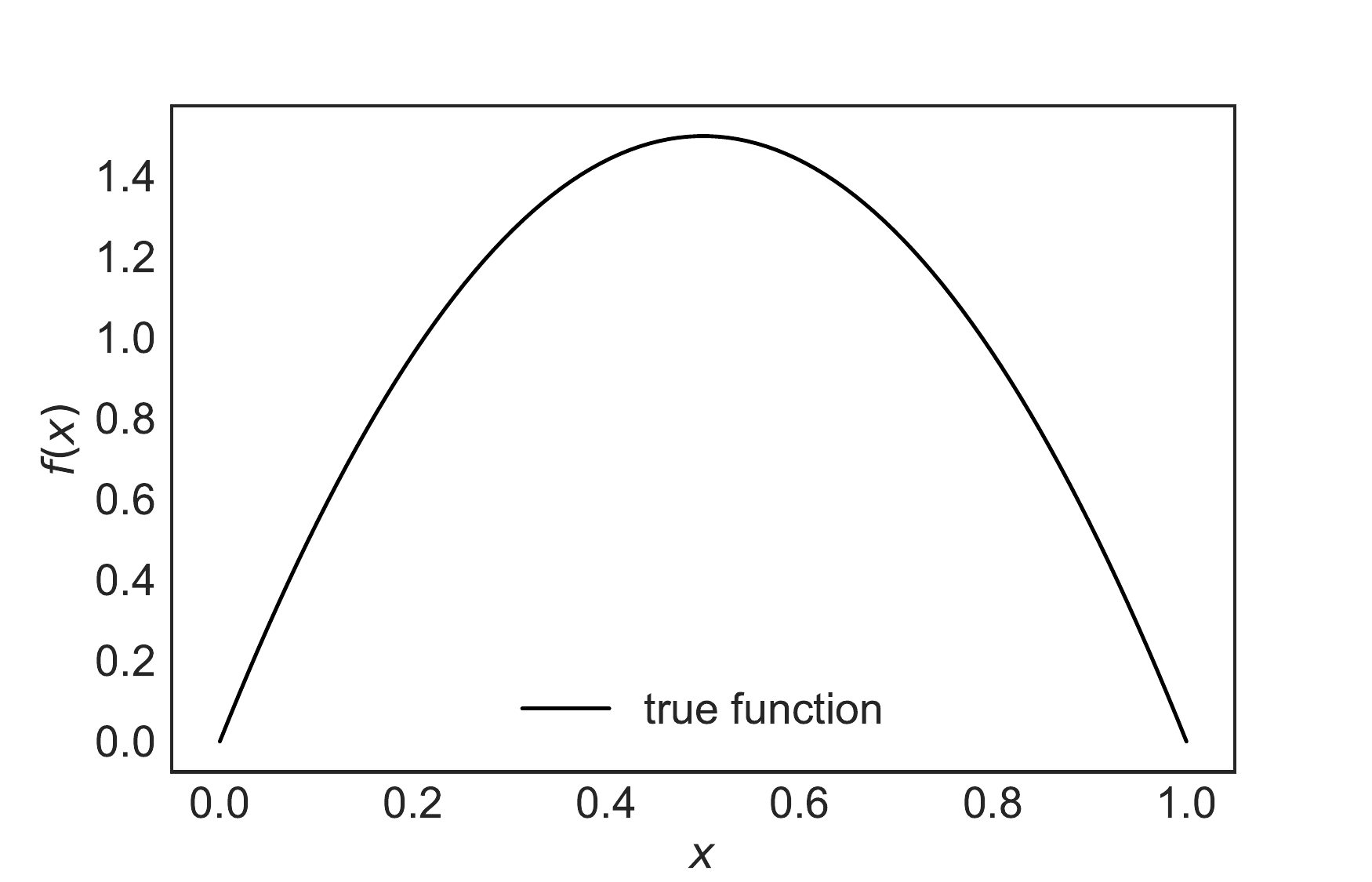}
     \caption{True unimodal function}
     \label{fig:appen_true_unimodal_func}
  \end{subfigure}
  \begin{subfigure}[b]{0.48\linewidth}
    \includegraphics[width=\linewidth]{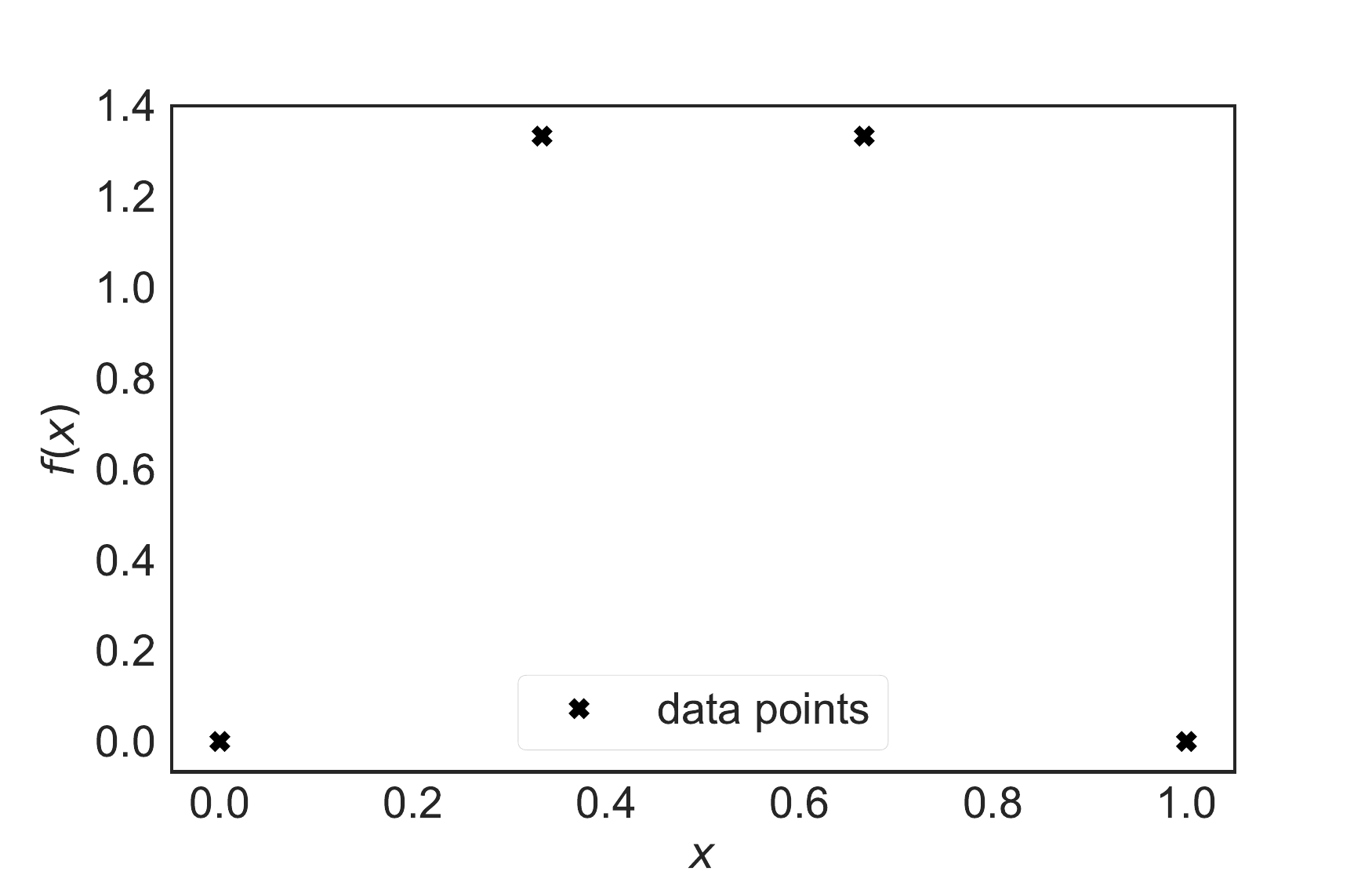}
    \caption{Observed Datapoints $N = 4$}
    \label{fig:appen_unimodal_obs_data}
  \end{subfigure}
\caption{True unimodal function $f(x) = \frac{x(1-x)}{B(1,1)}$ along with associated observed dataset.}
\end{figure}

\begin{figure}[H]
  \centering
  \begin{subfigure}[b]{0.48\linewidth}
    \includegraphics[width=\linewidth]{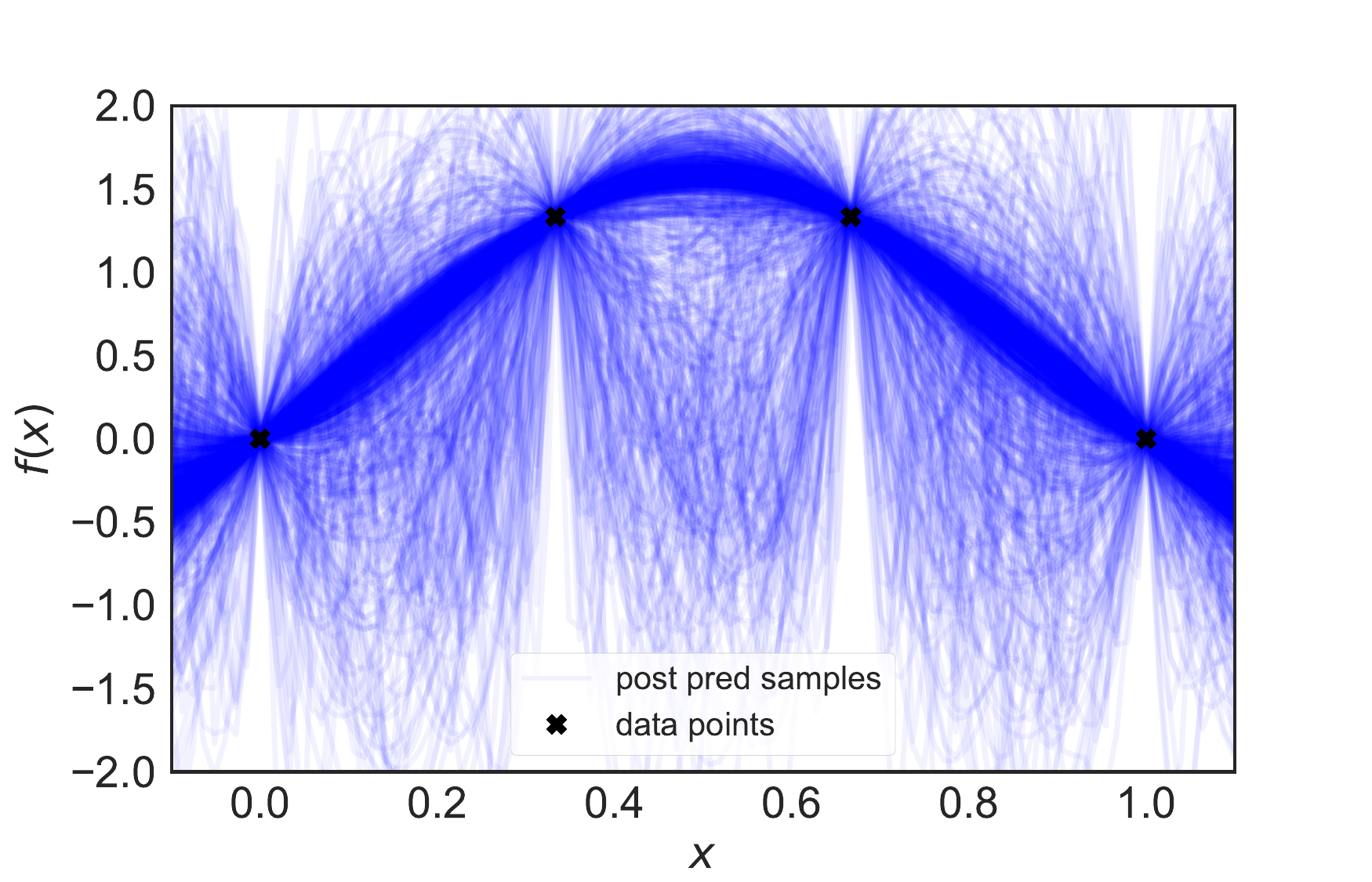}
     \caption{GP fit (without any constraints)}
     \label{fig:appen_unimodal_normal_gp_fit}
  \end{subfigure}
  \begin{subfigure}[b]{0.48\linewidth}
    \includegraphics[width=\linewidth]{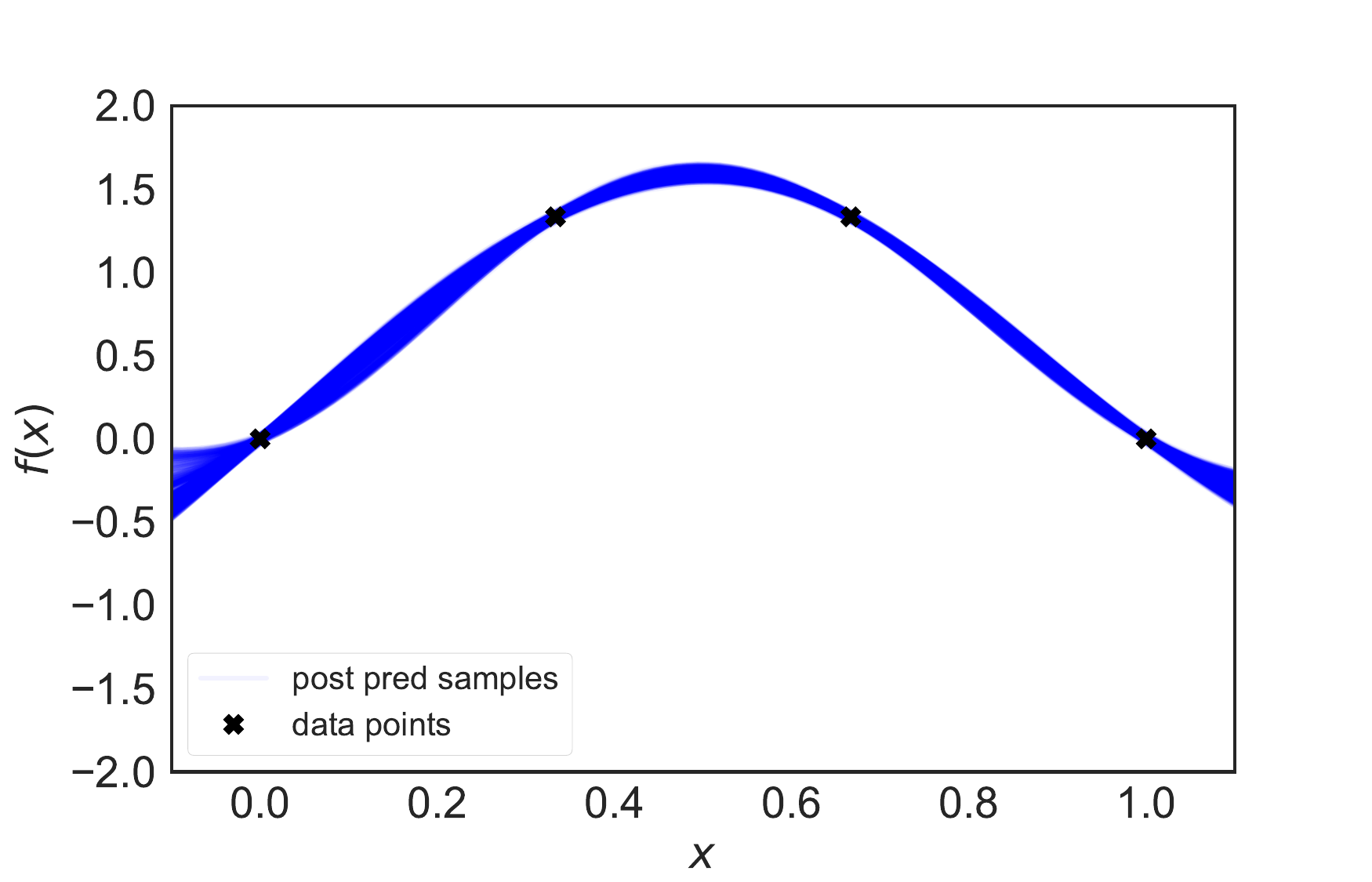}
    \caption{Unimodal GP fit (with unimodality constraints)}
    \label{fig:appen_unimodal_unimodal_gp_fit}
  \end{subfigure}
\caption{Posterior predictive samples (\textbf{left:} unconstrained GP fit ; \textbf{right:} unimodal GP fit).}
\end{figure}

\begin{figure}[H]
  \centering
  \begin{subfigure}[b]{0.48\linewidth}
    \includegraphics[width=\linewidth]{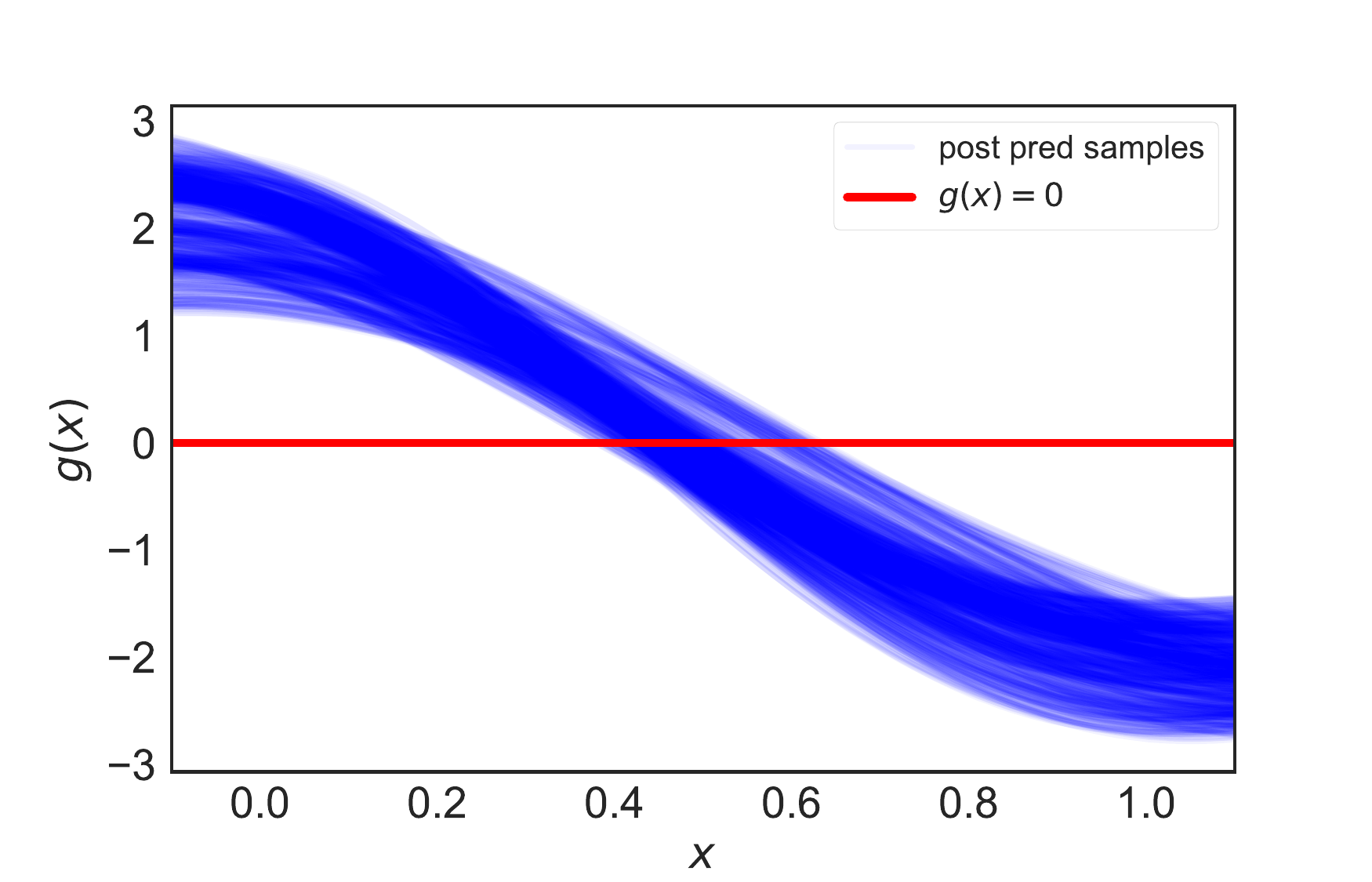}
     \caption{Latent GP $g(x)$ fit}
     \label{fig:appen_unimodal_g}
  \end{subfigure}
  \begin{subfigure}[b]{0.48\linewidth}
    \includegraphics[width=\linewidth]{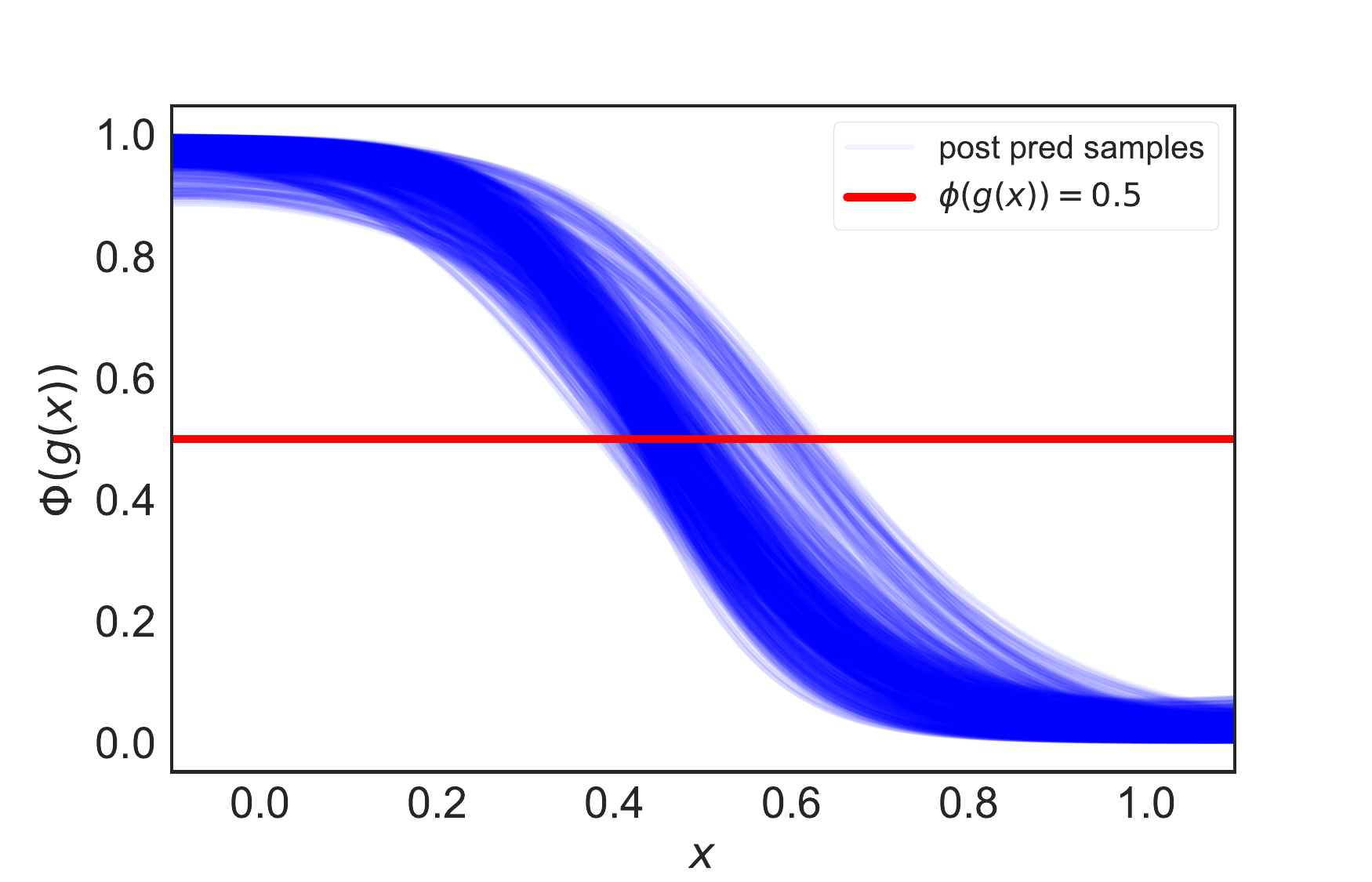}
    \caption{CDF of latent GP $\Phi(g(x))$ fit}
    \label{fig:appen_unimodal_phi_g}
  \end{subfigure}
\caption{Posterior predictive samples (\textbf{left:} latent GP $g(x)$ ; \textbf{right:} CDF of latent GP $\Phi(g(x))$.}
\end{figure}

In this sub-section, we demonstrate how the unimodal GP model we developed in sub-section \ref{sec:unimodality} can be applied to regression problems.
Assume that the true function we want to approximate is given as: $f(x) = \frac{x(1-x)}{B(1,1)}$, where $B(a,b)$ is a Beta function characterized by parameters $a$ and $b$. 
This function is shown in \cref{fig:appen_true_unimodal_func}. Furthermore, assume that we have observed $N = 4$ datapoints associated with function $f$, which are given as $D_{2}=\{(0,0),(0.33,1.33),(0.66,1.33),(1,0)\}$.
These observed points are also shown in \cref{fig:appen_unimodal_obs_data}.
We fit two models (1) regular GP without unimodality constraints and (2) GP with unimodal constraints by conditioning these models on the observed dataset $D_{2}$.
Samples from posterior predictive distribution, fitted using a regular GP model without constraints is shown in \Cref{fig:appen_unimodal_normal_gp_fit} and from a GP  with  unimodality constraints are shown in \Cref{fig:appen_unimodal_unimodal_gp_fit}.
As seen from these figures, using the newly developed unimodal GP model, we are to constrain GP to be unimodal. 
We are also able to locate the maxima $c_0=0.5$ of the fitted GP surrogate.
Maxima is where sign of the derivative changes from positive to negative (i.e. when latent function $g(x)=0$ (see \cref{fig:appen_unimodal_g}) and $\Phi(g(x))=0.5$ (see \cref{fig:appen_unimodal_phi_g})).

\section*{References}
\bibliography{actual_paper.bib}
\end{document}